\documentclass[journal]{IEEEtai}
\usepackage{times}
\usepackage{epsfig}
\usepackage{graphicx}
\usepackage{float}
\usepackage{amsmath}
\usepackage{amssymb}
\usepackage{subcaption}
\usepackage{caption}
\usepackage{amsmath}
\usepackage{amssymb}
\usepackage{amsfonts}
\usepackage{multirow}
\usepackage{rotating}
\usepackage{tabularx}
\usepackage{mathrsfs}
\usepackage{amsthm}
\usepackage{booktabs}
\usepackage[english]{babel}
\usepackage{blindtext}
\usepackage{tikz}
\usepackage{verbatim}
\usepackage{array}
\usepackage{capt-of}
\usepackage{xr}
\usepackage{tgpagella}
\usepackage{xcolor,colortbl}
\usepackage{hyperref}
\hypersetup{
     colorlinks=true,   
     linkcolor=blue,
     filecolor=magenta,      
     urlcolor=blue,
     citecolor=blue,
 }
\usepackage{todonotes}
\usepackage[compress]{cite}
\usepackage{balance}
\usepackage{algorithm}
\usepackage{algpseudocode}

\begin{document}
\title{Non-Uniform Illumination Attack for Fooling Convolutional Neural Networks}

\author{Akshay Jain, Shiv Ram Dubey, \IEEEmembership{Senior Member,~IEEE}, Satish Kumar Singh, \IEEEmembership{Senior Member,~IEEE}, 
\\ KC Santosh, \IEEEmembership{Senior Member,~IEEE}, 
Bidyut Baran Chaudhuri, \IEEEmembership{Life Fellow,~IEEE}
\thanks{A. Jain, S.R. Dubey and S.K. Singh are with the Computer Vision and Biometrics Lab, Department of Information Technology, Indian Institute of Information Technology Allahabad, Prayagraj, Uttar Pradesh-211015, India (e-mail: jainvakshay97@gmail.com, srdubey@iiita.ac.in, sk.singh@iiita.ac.in).}
\thanks{KC Santosh is with the AI Research Lab, Department of Computer Science, University of South Dakota, Vermillion, SD 57069 USA (e-mail:
santosh.kc@usd.edu).}
\thanks{B.B. Chaudhuri was with the Computer Vision and Pattern Recognition Unit at Indian Statistical Institute, Kolkata-700108, India (e-mail: bidyutbaranchaudhuri@gmail.com).}
}

\markboth{CNN Robustness under Non-Uniform Illumination}
 {Jain \MakeLowercase{\textit{et al.}}}

\maketitle

\begin{abstract}
Convolutional Neural Networks (CNNs) have made remarkable strides; however, they remain susceptible to vulnerabilities, particularly in the face of minor image perturbations that humans can easily recognize. This weakness, often termed as `attacks,' underscores the limited robustness of CNNs and the need for research into fortifying their resistance against such manipulations. This study introduces a novel Non-Uniform Illumination (NUI) attack technique, where images are subtly altered using varying NUI masks. Extensive experiments are conducted on widely-accepted datasets including CIFAR10, TinyImageNet, and CalTech256, focusing on image classification with 12 different NUI attack models. The resilience of VGG, ResNet, MobilenetV3-small and InceptionV3 models against NUI attacks are evaluated. Our results show a substantial decline in the CNN models' classification accuracy when subjected to NUI attacks, indicating their vulnerability under non-uniform illumination. To mitigate this, a defense strategy is proposed, including NUI-attacked images, generated through the new NUI transformation, into the training set. The results demonstrate a significant enhancement in CNN model performance when confronted with perturbed images affected by NUI attacks. This strategy seeks to bolster CNN models' resilience against NUI attacks. 
\footnote{The code is available at \url{https://github.com/Akshayjain97/Non-Uniform_Illumination}}
\end{abstract}

\begin{IEEEImpStatement}
While CNN models demonstrate strong performance on controlled data, their susceptibility to manipulation raises significant concerns about their robustness and suitability for real-world applications, as they can potentially fooled by data perturbation. In this context, we explore non-uniform illumination (NUI) masks that manipulate images to deceive CNN models while preserving their semantic content. Additionally, we introduce a key defense strategy involving NUI augmentation during training to enhance CNN model robustness. Given the prevalence of illumination variations in practical computer vision applications, our NUI masks offer a crucial means of bolstering model resilience.
\end{IEEEImpStatement}


\section{Introduction}
Deep learning, a subfield of artificial intelligence, known for neural networks with multiple interconnected layers, enables the automated extraction of progressively abstract features from input data \cite{lecun2015deep}.
Its resurgence in the 2010s was catalyzed by ample data availability, enhanced computational resources, and novel architectures such as convolutional and recurrent networks. Ongoing research in optimization, interpretability, and robustness continues to refine deep learning's efficacy and broaden its applicability across intricate real-world problem domains. The convolutional and recurrent networks made significant advancements in diverse domains including computer vision \cite{ioannidou2017deep, guo2016deep}, natural language processing \cite{young2018recent}, health informatics \cite{ravi2016deep}, and sentiment analysis \cite{glorot2011domain}
The Convolutional Neural Networks (CNNs) are utilized for computer vision applications \cite{li2021survey}, such as image recognition \cite{dai2021random, dubey2022adainject}, COVID-19 grading \cite{de2021automated}, image quality assessment \cite{pan2022no}, image super-resolution \cite{esmaeilzehi2022ultralight} and human action recognition \cite{ahmad2021graph}. CNN models employ backpropagation to learn the weights  \cite{kingma2014adam, dubey2023adanorm}. However, if a CNN model is more complex than the dataset and appropriate regularization techniques are not utilized, they are susceptible to overfitting the training data. Common regularization approaches include Dropout \cite{srivastava2014dropout}, Batch Normalization \cite{ioffe2015batch}, and Data Augmentation \cite{cubuk2019autoaugment}.

\begin{figure}[tbp]
    \centering
    \includegraphics[width=1.0 \linewidth]{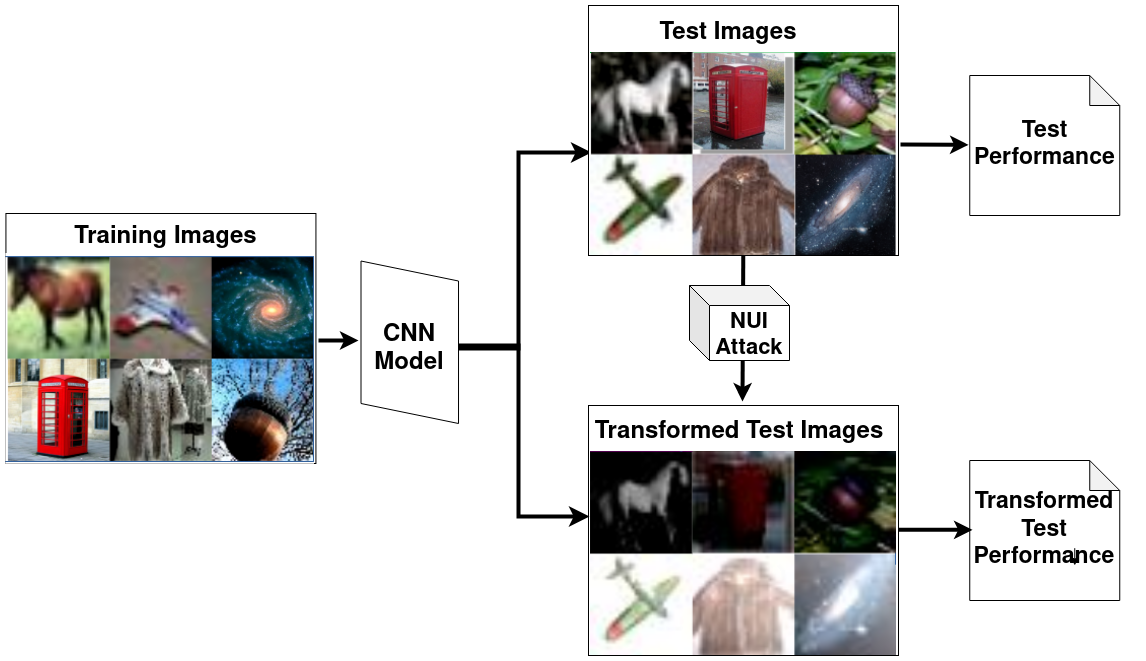}
    \caption{An overview of the NUI attack to fool the CNN models (image classification): test images are transformed through NUI attacks and their corresponding performance.}
    \label{fig:my_label}
\end{figure}

Recent studies uncovered that the CNN models can be deceived via data perturbation in multiple different ways \cite{goodfellow2014explaining, moosavi2016deepfool, elsayed2018adversarial, su2019one, deb2020advfaces, singh2022powerful}. To address this issue, many defense methods and network robustness aspects were studied \cite{wang2021fakespotter, ren2022perturbation, akhtar2018defense}. However, none of them studied the robustness of CNN models against non-uniform illumination. In this paper, we propose mask-based non-uniform illumination (NUI) variations as depicted in \autoref{fig:my_label} to fool the CNN models. Existing methods for adversarial attacks and defense techniques depend on data and the model's gradient. The proposed NUI attack is data-independent and utilizes varying weights of brightness and darkness. 

The majority of the techniques to perturb test images have a few drawbacks: prior knowledge of the model and dataset limits their applications in unfamiliar scenarios, and the inability to add non-uniform illumination variations in the brightness of the images, whereas the NUI attack technique adds non-uniform brightness to the image while keeping the semantic meaning intact. 
The following are the contributions of this paper:
\begin{itemize}
    \item The proposed NUI attack produces the attacked images by combining the input image with a NUI mask. Specifically, 12 NUI attack masks are presented.
    \item The NUI attack mask is created using several non-linear transformations generating non-uniform variations of brightness and darkness exploiting the spatial structure of the image. 
    \item We analyze the robustness of the CNN models including VGG, ResNet, MobilenetV3 and InceptionV3 over the proposed NUI attack on various benchmark datasets, including CIFAR10, CalTech256, and TinyImageNet. 
    \item We also train the CNN models on the NUI-attacked images to evaluate the robustness of the models when the NUI attack is used as a data augmentation technique. 
\end{itemize}

The remaining paper is structured as follows: \autoref{sec:Rel_wrk} describes the related work; \autoref{sec:NUI} describes the proposed NUI attack; \autoref{sec:Exp_set} describes the experimental settings, datasets, and training settings used; \autoref{sec:Res} illustrates the experimental results with observations; 
and \autoref{sec:Conclusion} concludes the paper.

\section{Related work}
\label{sec:Rel_wrk}
This section briefs about the adversarial attacks using brightness and defense mechanisms to such attacks.

\subsection{Adversarial Attacks Using Brightness}
Several works have focused on attacking the neural network models by perturbing the intensity values of the image pixels.
Nguyen et al. \cite{nguyen2020adversarial} have explored the possibility and practicality of performing real-time physical attacks on face recognition systems using adversarial light projections.
Singh et al.  \cite{singh2021brightness} have generated adversarial examples using Curriculum Learning.
The natural adversarial lighting conditions are generated by utilizing a physical lighting model proposed by Zhang et al. \cite{zhang2021adversarial} for conducting an adversarial relighting attack.
Given an image, Yang et al. \cite{yang2022random} have generated the adversarial examples by applying a brightness transformation to an image and feeding it into a CNN. 
Hsiung et al. \cite{hsiung2023towards} have utilized the component-wise projected gradient descent and automatic attack-order scheduling to find the optimal attack composition for creating the composite adversarial examples. 

Most existing methods require a neural network to generate adversarial examples. The colour channel perturbation (CCP) attack, perturbs the channels of images to generate the mixed colour channels randomly \cite{ccp}. The impact of colour is also studied in \cite{de2021impact} on the robustness of deep learning models. The paper aims to judge the robustness of CNN models against various non-uniform illumination variations generated through different masks. The proposed method is data-independent, does not require any neural network and gives a high attack success rate.

\subsection{defense Against Brightness Attacks}
The primary defense mechanism employed by most methods includes the attacked samples in the training set through data augmentation and retrains the model. A survey of defense strategies is presented in \cite{zhang2021adversarial_survey}. 
Agarwal et al. \cite{agarwal2020image} have exploited the image transformations, including Discrete Wavelet Transform and Discrete Sine Transform, against adversarial perturbation using deep models.
The performance of CNN models on CCP-attacked images greatly improved when the models were trained on the training set containing the CCP-attacked samples \cite{ccp}. 
The adversarial examples generated in \cite{singh2021brightness} are designed to be resilient against variations in real-world brightness conditions. Agarwal et al. \cite{agarwal2021damad} have developed an adversarial perturbation detector agnostic to databases, attacks, and models. Adversarial visual reconstruction is used against DeepFakes in \cite{he2022defeating}.
Hsiung et al. \cite{hsiung2023towards} have performed the generalized adversarial training (GAT) to enhance the robustness of the model against composite semantic perturbations, including combinations of Hue, Saturation, Brightness, Contrast, and Rotation. Recently, a self-supervised defense mechanism has been utilized in \cite{deb2023faceguard} against adversarial face images.
Premakumara et al. \cite{premakumara2023enhancing} have systematically investigated the amount of artificial perturbation needed to enhance the models' generalization by augmenting the data for object detection using neural networks.
We propose a primary defense mechanism against the NUI attack by employing data augmentation through NUI attack in the training set and retraining the CNN models for the image classification task. The proposed defense technique can be useful in common use cases where the input image gets distorted due to exposure to sunlight or part of the image becomes relatively darker because of reflection.

\begin{table*}[!t]
\caption{List of all masks and their region of perturbation. $x$ and $y$ are the horizontal and vertical coordinate axis variables, respectively, and $u$ and $v$ represent the image size which remained constant (i.e., $32$) throughout the experiment. $a$ is the amount of brightness to be added to the input image.}
{\renewcommand{\arraystretch}{1.5}%
\resizebox{\textwidth}{!}{%
\begin{tabular}{|p{0.06\textwidth}|p{0.53\textwidth}|p{0.3\textwidth}|}    
\hline
\textbf{Mask ID} & \textbf{Mask} & \textbf{Region of perturbation}\\
\hline
Mask $1$ & $a = ((u - x) \times \frac{30}{u}) + ((v - y) \times \frac{30}{v}) + ((u - y) \times \frac{20}{u}) + ((v - y) \times \frac{20}{v})$  & Focused more on the left side \\
\hline
Mask $2$ & $a = (x \times \frac{30}{u}) + ((v - y) \times \frac{30}{v}) + (y \times \frac{20}{u}) + ((v - x) \times \frac{20}{v})$
 & Distributed throughout \\
\hline
Mask $3$ & $a = ((u - x) \times \frac{30}{u}) + (y \times \frac{30}{v}) + ((u - y) \times \frac{20}{u}) + (y \times \frac{20}{v})$
 & Focused on the top right corner \\
\hline
Mask $4$ & $a = (x \times \frac{30}{u}) + (y \times \frac{30}{v}) + (x \times \frac{20}{u}) + (y \times \frac{20}{v})$
 & Focused on the bottom right corner \\
\hline
Mask $5$ & $a = \text{abs}(16-x) \times \text{abs}(16-y)$
 & The curved diamond shape \\
\hline
Mask $6$ \newline Mask $7$ \newline Mask $8$ & $a = 144 - \text{abs}(16-x) \times \text{abs}(16-y)$ \newline $a = 100 - \text{abs}(16-x) \times \text{abs}(16-y)$ \newline $a = 50 - \text{abs}(16-x) \times \text{abs}(16-y)$
 & Circular perturbation with different radius at centre \\
\hline
Mask $9$ & \(\textbf{if} (0\leq y \leq 5 \ or \ 10\leq y\leq 15 \ or \ 20\leq y\leq 25 \ or \ 30\leq y\leq 32):
   a = \text{Mask } 1
  \newline  \textbf{else}: a = -\text{Mask } 2\) & A pattern of vertical lines\\
\hline
Mask $10$ & \(\textbf{if} (0\leq x \leq 5 \ or \ 10\leq x\leq 15 \ or \ 20\leq x\leq 25 \ or \ 30\leq x\leq 32): 
    a =  \text{Mask } 1
    \newline \textbf{else}: a = -\text{Mask } 2\) & A pattern of horizontal lines \\
\hline
Mask $11$ & $\textbf{if}(x\leq 16 \ and \ y\leq 16):a = \text{Mask } 1$ 
\newline$\textbf{if}(x\leq 16 \ and \ y > 16):a = \text{Mask } 2$ \newline$\textbf{if}(x>16 \ and \ y\leq 16):a = \text{Mask } 3$ \newline$\textbf{if}(x>16 \ and \ y>16):a = \text{Mask } 4$  & Differs for different quadrants of the image \\
\hline
Mask $12$ & $\textbf{if}(x\leq 16 \ and \ y\leq 16):   a = \text{Mask } 1$
    \newline $\textbf{if}(x\leq 16 \ and \ y>16):a = \text{Mask } 2$
    \newline$\textbf{if}(x>16 \ and \ y\leq 16):a = -\text{Mask } 3$
    \newline$\textbf{if}(x>16 \ and \ y>16):a = -\text{Mask } 4$ & Differs for different quadrants of the image and produces a pattern effect of vertical lines\\
\hline
\end{tabular}}}
\label{tab:mask}
\end{table*}

\section{Proposed Non-Uniform Illumination Attack}
\label{sec:NUI}
In recent years, various attack methods have been investigated to judge the robustness of CNN models. However, the conventional attack methods do not take advantage of creating non-uniform illumination variations with different brightness and darkness levels.

\subsection{Proposed NUI Attacks}
We propose a simple yet effective non-uniform illumination (NUI) attack on test image data. The rationale behind developing this attack technique stemmed from a desire to investigate perturbation methods applicable to convolutional neural network (CNN) models which can give a high attack success rate and do not require any Neural Network to model such attack. Specifically, the aim is to explore how illumination variations could be utilized to attack these models. In the earlier stages of the experiments we considered only Mask $1$ to Mask $4$, but later to experiment with the region of attack, we added Mask $5$ to Mask $12$ given in \autoref{tab:mask}. The proposed NUI attack brightens or darkens the image pixels non-uniformly to generate the synthesized test images to fool the CNN models. The core of the proposed attack is the weight of image brightness and darkness. The weight ($k$) value controls the brightness or darkness added to the test image based on certain patterns.
The proposed attack technique uses several masking strategies to generate different masks ($a$) for the images of size $h\times w$, where $h$ and $w$ are image height and width, respectively. The created masks are applied to the test images to generate the synthesized test images to fool the CNN models.
In this paper, we experiment using $12$ different masks. We analyzed the robustness of CNN models on the Attacks caused by different NUI masks. The formulas utilized to create these masks ($a$) are given in \autoref{tab:mask} with its region of perturbation in the image. 
There are a total of $23$ different weight values $k$ used in this paper, ranging from $-2.2$ to $+2.2$ with a gap of $0.2$. It leads to $23 \times 12 = 276$ experiments for a given model on any dataset.

The masking function, Mask $1$, is considered from \cite{dubey2015multi}. Mask $2$, $3$, and $4$ are the variations of Mask $1$ and are formulated by considering the exploitation of spatial locality. Mask $5$ perturbs the image centre up to the centres of each side in the shape of a curved diamond. The effect of Mask $6$, $7$ and $8$ is similar, but with different severity. These masks create a circular perturbation effect in the images. The amount of perturbation is highest for Mask $6$ and lowest for Mask $8$. Mask $9$ and $10$ use Mask $1$ and negative of the Mask $2$ in specific conditions leading to perturbation of the pattern of vertical and horizontal lines, respectively. Mask $11$ adds perturbations of Mask $1$ $2$, $3$, and $4$ in different quadrants. The effect of the Mask $12$ is similar to Mask $11$, except for the right part of the image which becomes darker instead of brighter.

\begin{algorithm}[!t]
\caption{Proposed NUI Attack Algorithm} \label{alg:nui_attack}
\textbf{Input: }Image data $I\in\mathbb{R}^{u \times v}$ as input, Attack Mask ID $i \in \{1, 2, \ldots, 12\}$, and perturbation weight value $k \in \{-2.2, -2.0, \ldots, 2.0, 2.2\}$\\
\textbf{Output: }NUI Attacked Image $I_{M_i,k}$. 
\begin{algorithmic}[1]
\State Generate the $i^{th}$ Mask using Table 1 as 
\Statex $M_i = a(x,y),  \forall{x\in\{1,u\}} \text{ and } \forall{y\in\{1,v\}}$.
\State Perform the mask weighting with weight $k$ as 
\Statex $M_{i,k} = M_i \times k$.
\State Attack the image ($I$) to generate the perturbed image ($I_{M_i,k}$) as 
\Statex $I_{M_i,k} = I + M_{i,k}$.
\end{algorithmic}
\end{algorithm}

The algorithm for the proposed NUI attack is illustrated in Algorithm \ref{alg:nui_attack}. The input image ($I$) is attacked to purturbed image ($I_{M_i,k}$) using the $i^{th}$ Mask and weight value $k$. As shown in Table \ref{tab:mask}, 12 NUI Masks are used in this paper. Based on the chosen Mask and weight, the final Mask is computed and added in the input image to generate the attacked image.

\begin{figure*}[!t]
    \centering
    \newcommand\width{1.285cm}
    \includegraphics[width=\width]{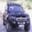}
    \includegraphics[width=\width]{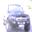}
    \includegraphics[width=\width]{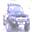}
    \includegraphics[width=\width]{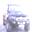}
    \includegraphics[width=\width]{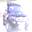}
    \includegraphics[width=\width]{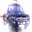}
    \includegraphics[width=\width]{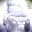}
    \includegraphics[width=\width]{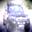}
    \includegraphics[width=\width]{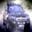}
    \includegraphics[width=\width]{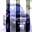}
    \includegraphics[width=\width]{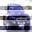}
    \includegraphics[width=\width]{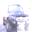}
    \includegraphics[width=\width]{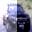}
    \\
    \includegraphics[width=\width]{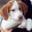}
    \includegraphics[width=\width]{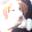}
    \includegraphics[width=\width]{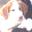}
    \includegraphics[width=\width]{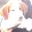}
    \includegraphics[width=\width]{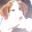}
    \includegraphics[width=\width]{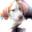}
    \includegraphics[width=\width]{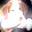}
    \includegraphics[width=\width]{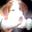}
    \includegraphics[width=\width]{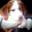}
    \includegraphics[width=\width]{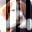}
    \includegraphics[width=\width]{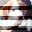}
    \includegraphics[width=\width]{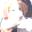}
    \includegraphics[width=\width]{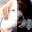}
    \\
    \includegraphics[width=\width]{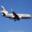}
    \includegraphics[width=\width]{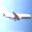}
    \includegraphics[width=\width]{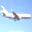}
    \includegraphics[width=\width]{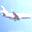}
    \includegraphics[width=\width]{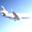}
    \includegraphics[width=\width]{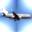}
    \includegraphics[width=\width]{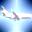}
    \includegraphics[width=\width]{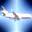}
    \includegraphics[width=\width]{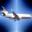}
    \includegraphics[width=\width]{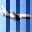}
    \includegraphics[width=\width]{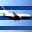}
    \includegraphics[width=\width]{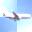}
    \includegraphics[width=\width]{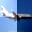}
    \\
    \includegraphics[width=\width]{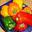}
    \includegraphics[width=\width]{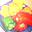}
    \includegraphics[width=\width]{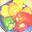}
    \includegraphics[width=\width]{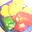}
    \includegraphics[width=\width]{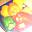}
    \includegraphics[width=\width]{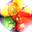}
    \includegraphics[width=\width]{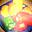}
    \includegraphics[width=\width]{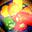}
    \includegraphics[width=\width]{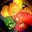}
    \includegraphics[width=\width]{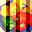}
    \includegraphics[width=\width]{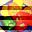}
    \includegraphics[width=\width]{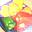}
    \includegraphics[width=\width]{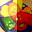}
    \\
    \includegraphics[width=\width]{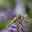}
    \includegraphics[width=\width]{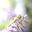}
    \includegraphics[width=\width]{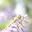}
    \includegraphics[width=\width]{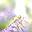}
    \includegraphics[width=\width]{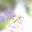}
    \includegraphics[width=\width]{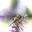}
    \includegraphics[width=\width]{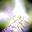}
    \includegraphics[width=\width]{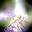}
    \includegraphics[width=\width]{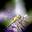}
    \includegraphics[width=\width]{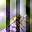}
    \includegraphics[width=\width]{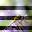}
    \includegraphics[width=\width]{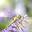}
    \includegraphics[width=\width]{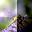}
    \\
\caption{$1^{st}$ column in the figure contains original images. The $2^{nd}$ to $13^{th}$ columns contain the images perturbed using mask $1^{st}$ to $12^{th}$ , respectively. The images are taken from CIFAR10, TinyImageNet, and CalTech256 datasets.}
\label{fig:effect}
\end{figure*}

\subsection{Effect of NUI Attacks}
The effect of different NUI attacks is illustrated in \autoref{fig:effect} using the sample images as to how the brightness, colour, details, appearance, etc. change after applying different NUI masks. Here, the perturbation weight ($k$) value is different for all columns and is positive, because of which all the images look brighter than their original form.  The $1^{st}$ column contains the original sample images. The $2^{nd}$ to $13^{th}$ columns correspond to the images generated using Mask $1$ to $12$, respectively. As mentioned, the perturbed image is brighter on the left side and the perturbation drops when it goes to the right for Mask $1$. The image is bright in general for Mask $2$. The images appear bright in the top right corner for Mask $3$. The perturbations are focused more in the bottom right corner for the masking function $4$. These masking functions are simple and do not change the underlying semantic meaning of the input image, but can provide a good attack success rate. 
The effect of a curved diamond can be observed for the Mask $5$. The perturbations for Mask function $6$, $7$, and  $8$, respectively, produce samples like the reverse of the Mask $5$. The images produced using Mask $6$ are perturbed with higher intensity values. However, the amount of perturbation is reduced for Mask $7$ which is further reduced for Mask $8$. Moreover, the attack success increases for Mask $8$ without losing the visual perceptibility of the image. 
The perturbations caused by Mask $9$ and $10$ respectively have vertical and horizontal patterns of alternate brightness and darkness. Masks $11$ and $12$ perturb the images using different masks in different quadrants. Mask $11$ adds mask value in each quadrant, while mask $12$ adds mask value in the left side quadrants and subtracts in the right side quadrants. We also show the effect on the histogram in Supplementary.

\begin{figure*}[!t]
    \centering
    \includegraphics[trim={0.2cm 0 0.5cm 0},clip, width=\linewidth, height=7cm]{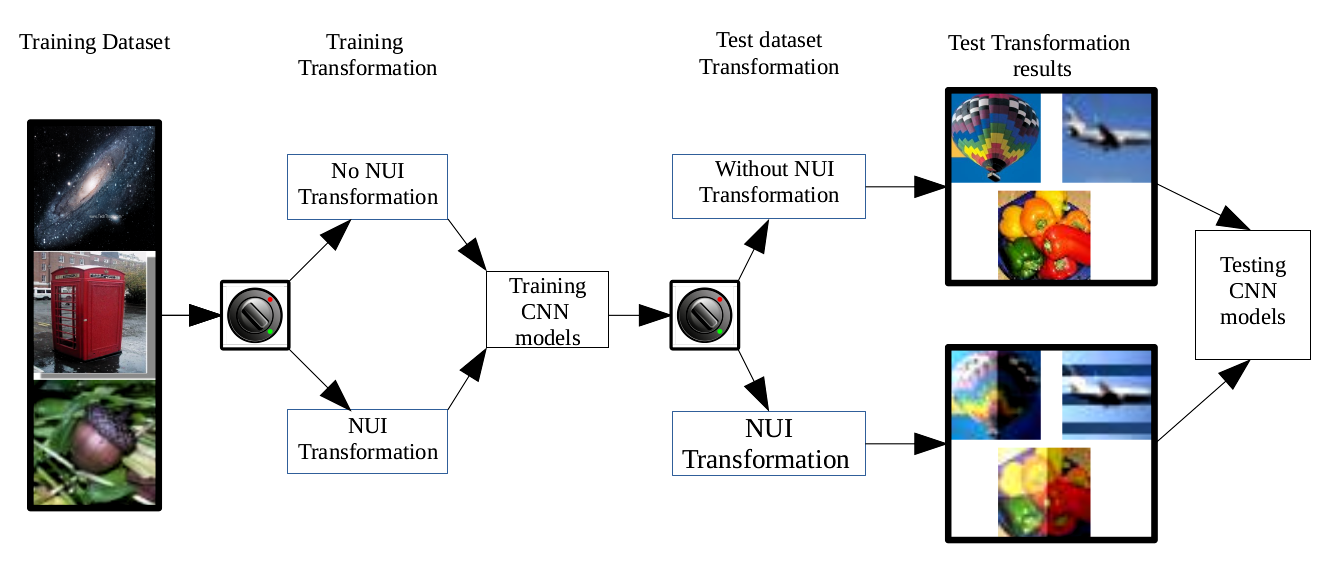}
    \caption{The workflow of the proposed method and the experimental settings used for the training and testing of the CNN models using NUI attack.}
    \label{fig:method}
\end{figure*}

\subsection{Proposed Workflow using NUI Attacks}
The workflow of the proposed method is illustrated in \autoref{fig:method}. To analyse the robustness of the CNN models against NUI attacks, we trained models on the original datasets and tested them for all NUI masks for all values of ($k$). Further to analyse the defense capability, the CNN models are trained on the NUI-attacked datasets and again tested.

For training models on perturbed datasets, the NUI perturbation is added to $80\%$ of the training set. We limit the weight factor ($k$) in the training part to $12$ different settings to avoid high bias in the training set towards severe perturbation, i.e., from $-1.2$ to $+1.2$ with a gap of $0.2$ excluding $0.0$ as it is already included in the $20\%$ part of the training set. The number of masks for perturbation during training is reduced to $10$ only, excluding Mask $6$ and Mask $7$ as these are similar to Mask $8$. Mask $12$ is replaced with the following mask for training:
\newline
$\textbf{if}(x\leq16 \ and \ y\leq16):a = +$Mask 1
\newline $\textbf{if}(x\leq16 \ and \ y>16):a = -$Mask 2
\newline $\textbf{if}(x>16 \ and \ y\leq16):a = +$Mask 3
\newline $\textbf{if}(x>16 \ and \ y>16):a = -$Mask 4 
\newline
which subtracts Mask $2$ and Mask $4$ in the leading diagonal quadrants, respectively, and adds Mask $1$ and Mask $3$ in the other two quadrants, respectively. This represents the general case for quadrant perturbation. 
After being trained on perturbed images, the CNN models not only preserved the original accuracy on unperturbed data but also became robust to NUI attacks.

\section{Experimental Settings}
\label{sec:Exp_set}

\subsection{Datasets}
To examine the impact of the proposed NUI attacks, we conduct the image classification experiments on three benchmark datasets, including CIFAR10 \cite{cifar}, CalTech256 \cite{griffin2007CalTech}, and TinyImageNet \cite{tinyimagenet}. 
The $60,000$ images in the CIFAR10 dataset are equally divided into $10$ different categories. Out of $60,000$ images $10,000$ images are marked as the test set and the rest as the training set.
The $30,607$ images in the CalTech256 dataset represent $257$ different object categories. $20\%$ of the CalTech256 dataset is utilized for testing, while the rest for training. The CalTech256 dataset exhibits a high level of complexity due to several categories and more instances within each category, it also exhibits high inter-class similarity. 
The training set of the TinyImageNet dataset contains $100,000$ images and the validation set consists of $10,000$ images. The dataset comprises $200$ categories which have $500$ training images and $50$ validation images for each category.  It consists of a subset of images from ImageNet, specifically curated for small-scale experiments.

\subsection{CNN Architectures Used}
We used VGG \cite{simonyan2014very}, ResNet \cite{he2016deep}, MobilenetV3 \cite{howard2019searching} and InceptionV3 \cite{szegedy2015rethinking} to demonstrate the effects of the proposed non-uniform illumination attack.
The VGG network is a deep CNN model containing $16$ or $19$ trainable layers. The principal thought behind the VGG network is to utilize a series of convolutional layers with small filter sizes (3$\times$3) and stack them together to create a deeper network. 
For experiments on the CIFAR10 and TinyImageNet datasets, VGG16 is used and for experiments on the CalTech256 dataset, VGG19 is used.
The ResNet model includes the residual connections that allow the flow of gradients during backpropagation effectively. 
Deep CNNs utilizing the residual model demonstrate improved convergence, leading to enhanced performance. The ResNet18 model is used with all the datasets for experiments.
MobileNetV3 is a convolutional neural network specifically optimized for mobile phone CPUs through a combination of hardware-aware network architecture search (NAS).
This network has been further refined through several innovative architectural improvements, including integrating complementary search methodologies, developing new efficient nonlinearities suitable for mobile environments and creating efficient network design tailored for mobile applications.

Inception-v3 represents an advanced convolutional neural network architecture within the Inception series, incorporating several enhancements. These include Label Smoothing, factorized $7\times7$ convolutions, and the integration of an auxiliary classifier to propagate label information to earlier network layers with the implementation of batch normalization within the auxiliary head layers.
Cifar10 dataset has been used for experimentation with MobilenetV3-small and InceptionV3.

\begin{figure*}[!t]
  \centering
  \begin{minipage}[t]{0.13\textwidth}
    \centering
    \includegraphics[width=\textwidth]{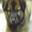}
    Prediction: \textcolor{blue}{\textbf{dog}}\\
    Prob: 0.99999 \\ Test Sample
  \end{minipage}%
  \hfill
  \begin{minipage}[t]{0.13\textwidth}
    \centering
    \includegraphics[width=\textwidth]{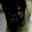}
    Prediction: \textcolor{blue}{\textbf{dog}}\\
    Prob: 0.93482\\
    $\text{M}1$, $k=-1.6$
  \end{minipage}%
  \hfill
  \begin{minipage}[t]{0.13\textwidth}
    \centering
    \includegraphics[width=\textwidth]{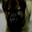}
    Prediction: \\ \textcolor{red}{\textbf{cat}}\\
    Prob: 0.83713\\
    $\text{M}2$, $k=-1.6$
  \end{minipage}
  \hfill
  \begin{minipage}[t]{0.13\textwidth}
    \centering
    \includegraphics[width=\textwidth]{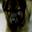}
    Prediction: \\ \textcolor{red}{\textbf{cat}}\\
    Prob: 0.99993\\
    $\text{M}3$, $k=-1.6$
  \end{minipage}
  \hfill
  \begin{minipage}[t]{0.13\textwidth}
    \centering
    \includegraphics[width=\textwidth]{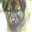}
    Prediction: \textcolor{blue}{\textbf{dog}}\\
    Prob: 0.73697\\
    $\text{M}4$, $k=1.2$
  \end{minipage}
  \hfill
  \begin{minipage}[t]{0.13\textwidth}
    \centering
    \includegraphics[width=\textwidth]{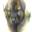}
    Prediction: \textcolor{blue}{\textbf{dog}}\\
    Prob: 0.63013\\
    $\text{M}5$, $k=1.2$
  \end{minipage}
  \hfill
  \begin{minipage}[t]{0.13\textwidth}
    \centering
    \includegraphics[width=\textwidth]{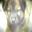}
    Prediction: \textcolor{blue}{\textbf{dog}}\\
    Prob: 0.93180\\
    $\text{M}6$, $k=0.8$
  \end{minipage}
  \hfill
  \vspace{0.1cm}
  \begin{minipage}[t]{0.13\textwidth}
    \centering
    \includegraphics[width=\textwidth]{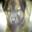}
    Prediction: \textcolor{red}{\textbf{deer}}\\
    Prob: 0.89235\\
    $\text{M}7$, $k=0.8$
  \end{minipage}
  \hfill
  \begin{minipage}[t]{0.13\textwidth}
    \centering
    \includegraphics[width=\textwidth]{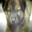}
    Prediction: \textcolor{red}{\textbf{deer}}\\
    Prob: 0.83318\\
    $\text{M}8$, $k=0.8$
  \end{minipage}
  \hfill
  \begin{minipage}[t]{0.13\textwidth}
    \centering
    \includegraphics[width=\textwidth]{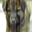}
    Prediction: \textcolor{blue}{\textbf{dog}}\\
    Prob: 0.75995\\
    $\text{M}9$, $k=0.4$
  \end{minipage}
  \hfill
  \begin{minipage}[t]{0.13\textwidth}
    \centering
    \includegraphics[width=\textwidth]{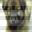}
    Prediction: \textcolor{red}{\textbf{ship}}\\
    Prob: 0.94852\\
    $\text{M}10$, $k=0.4$
  \end{minipage}
  \hfill
  \begin{minipage}[t]{0.13\textwidth}
    \centering
    \includegraphics[width=\textwidth]{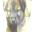}
    Prediction: \textcolor{blue}{\textbf{dog}}\\
    Prob: 0.56874\\
    $\text{M}11$, $k=1.6$
  \end{minipage}
  \hfill
  \begin{minipage}[t]{0.13\textwidth}
    \centering
    \includegraphics[width=\textwidth]{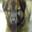}
    Prediction: \textcolor{red}{\textbf{ship}}\\
    Prob: 0.98438\\
    $\text{M}12$, $k=0.4$
  \end{minipage}

  \caption{Predictions of ResNet18 for an original test image and NUI attacked images using different masks with varying weight ($k$). `Prob' refers to probability and M$i$ refers to $i^{th}$ mask.}
   \label{fig:prediction}
\end{figure*}

\subsection{Training Settings}
All the experiments are performed using the PyTorch framework \cite{paszke2019pytorch}. The batch size of $64$ is used for VGG and ResNet models, $256$ for MobileNet model and $128$ for Inception model. Using the Adam optimizer, the models are trained for $100$ epochs. For the first $80$ epochs, the learning rate is set at $10^{-3}$ for CIFAR10 and TinyImageNet and $10^{-4}$ for the CalTech256 dataset, and for the final $20$ epochs, it is reduced by a factor of $10$. The categorical cross-entropy loss function is used as an objective function to measure the dissimilarity between predicted and actual class labels. Batch normalization is used for regularization. 
The following data augmentation is used during training: random cropping of size $32$, random horizontal flipping, and normalization to zero mean and unit standard deviation. The images are also resized to $32 \times 32$ resolution for VGG and ResNet models, whereas the MobileNet and Inception models accept images of size $224\times224$ and $299\times299$, respectively.

\section{Experimental Results and Analysis}
\label{sec:Res}
In this section, the qualitative and quantitative results are presented for image classification using VGG and ResNet models on CIFAR10, TinyImageNet and CalTech256 datasets as well as MobileNet and InceptionV3 models on CIFAR10 dataset.

\begin{figure*}[!t]
    \centering
    \newcommand\wide{2.93cm}
    \includegraphics[width=\wide]{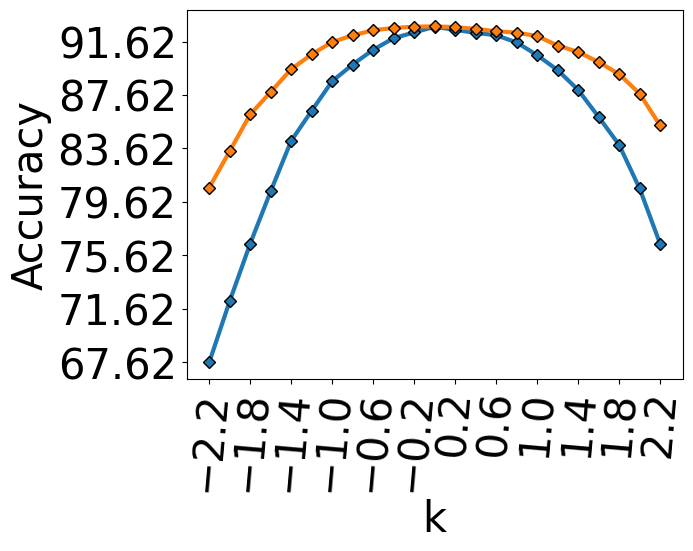}
    \includegraphics[width=\wide]{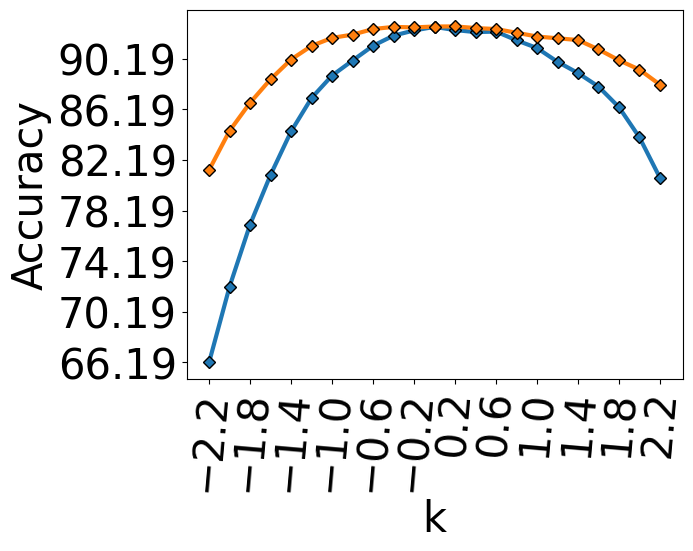}
    \includegraphics[width=\wide]{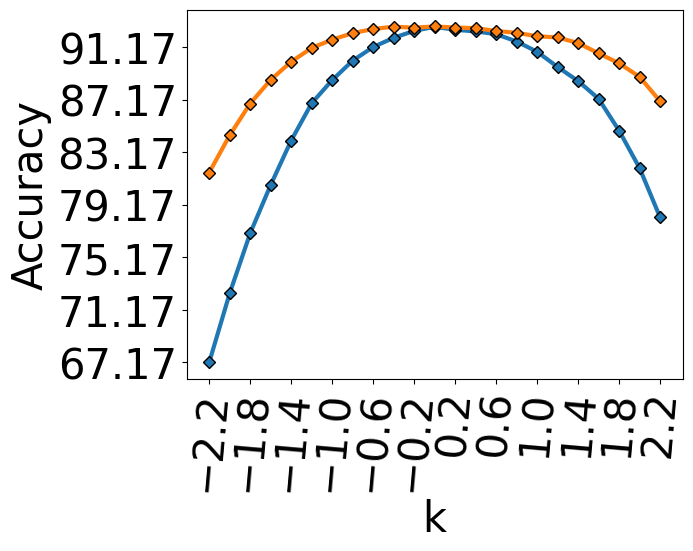}
    \includegraphics[width=\wide]{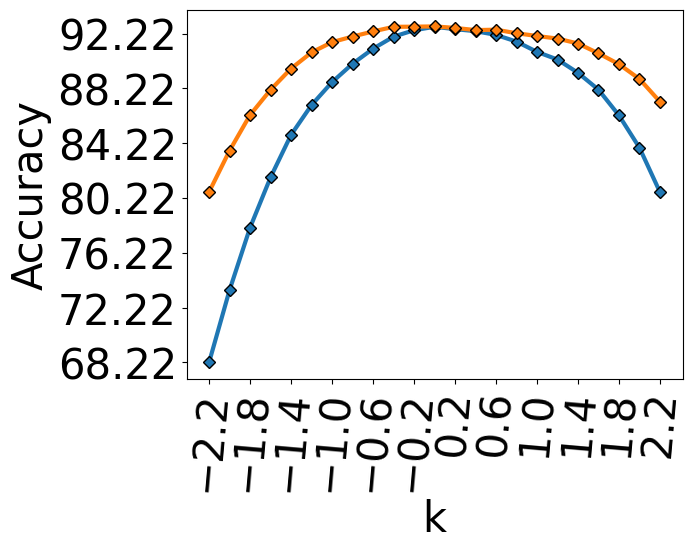}
    \includegraphics[width=\wide]{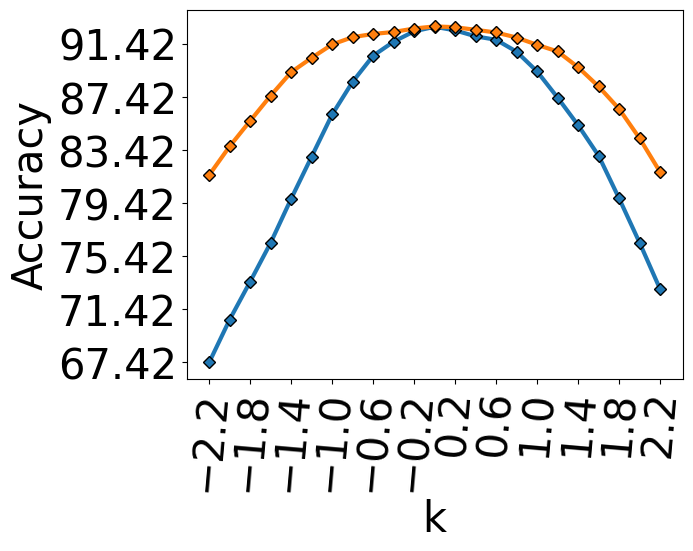}
    \includegraphics[width=\wide]{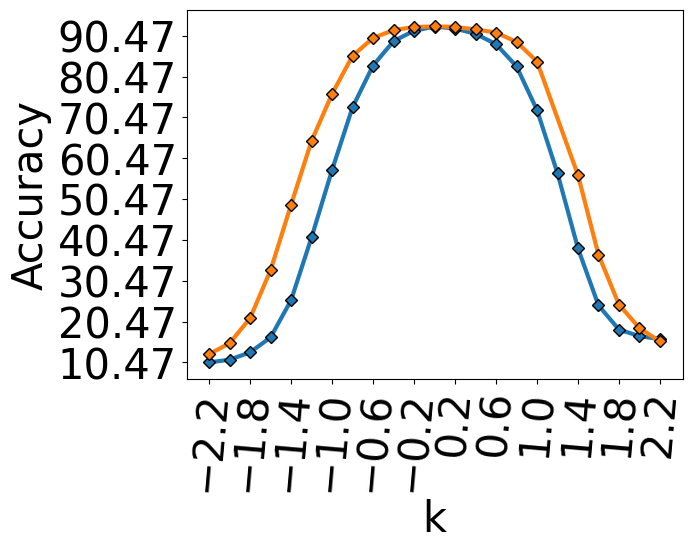}
    \includegraphics[width=\wide]{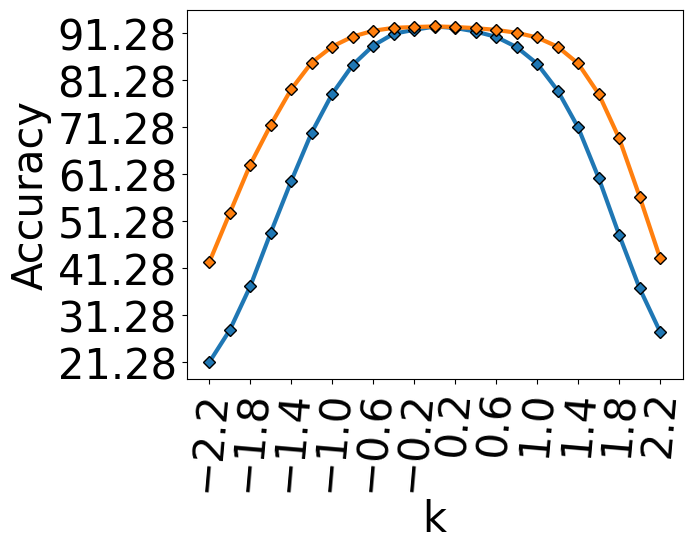}
    \includegraphics[width=\wide]{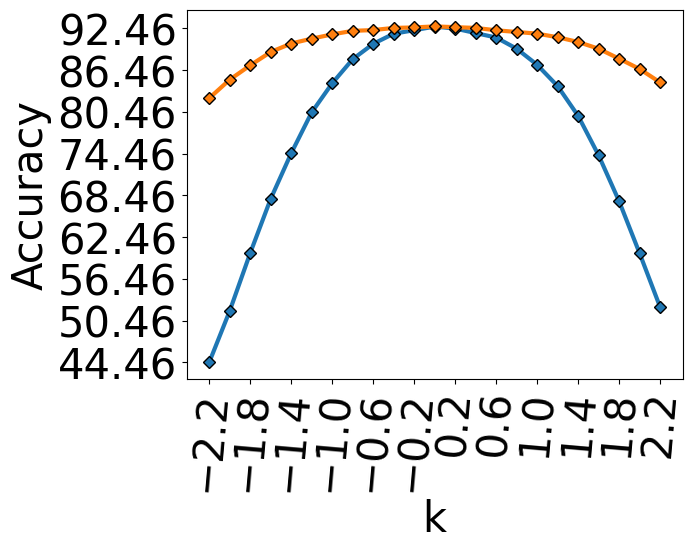}
    \includegraphics[width=\wide]{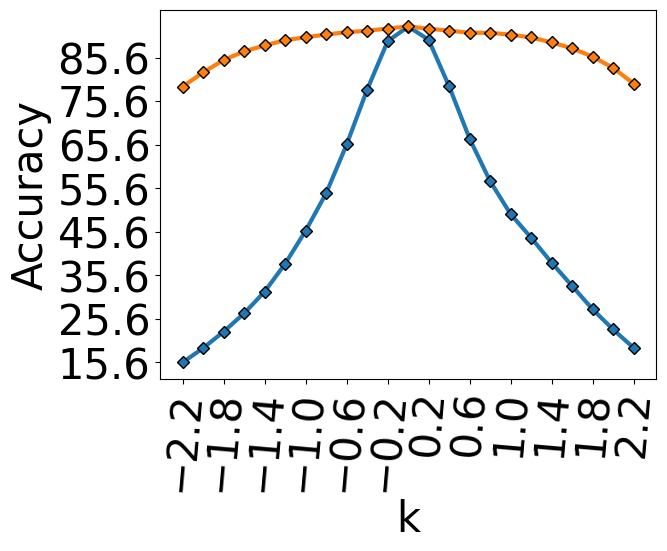}
    \includegraphics[width=\wide]{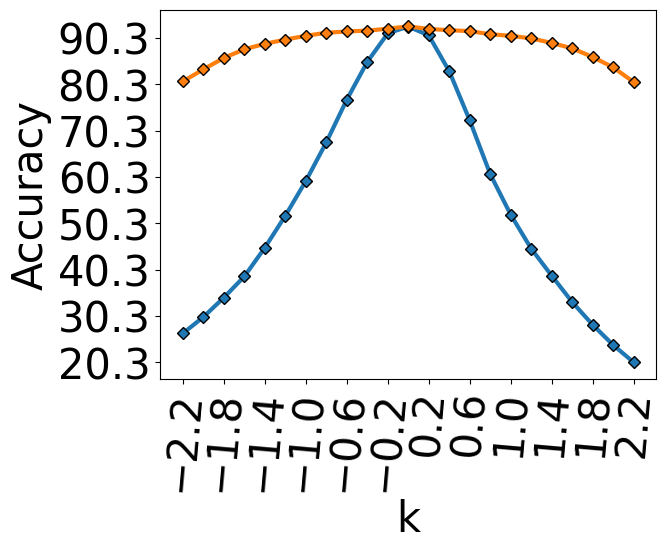}
    \includegraphics[width=\wide]{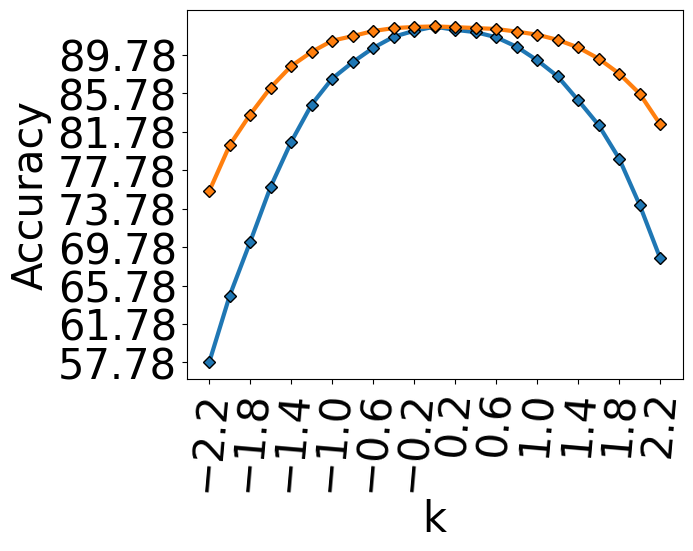}
    \includegraphics[width=\wide]{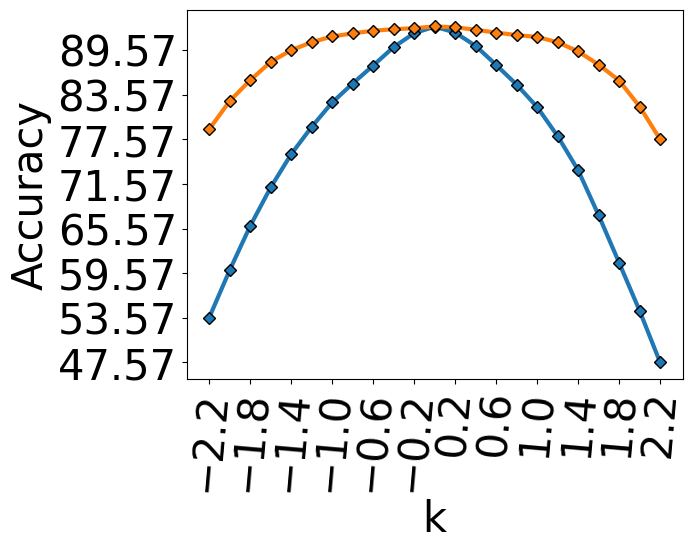}
    \caption{Results of VGG16 model on CIFAR10 dataset under different NUI attacks (test set). \textcolor{blue}{Blue} and \textcolor{orange}{orange} curves show the performance of the model trained on the original training set and the NUI perturbed training set, respectively.} 
    \label{fig:cifar_vgg}
\end{figure*}

\begin{figure*}[!t]
    \centering
    \newcommand\wide{2.93cm}
    \includegraphics[width=\wide]{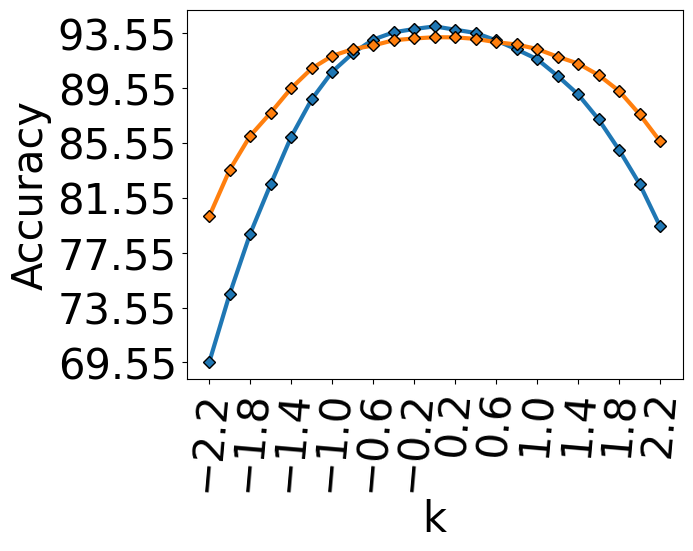}
    \includegraphics[width=\wide]{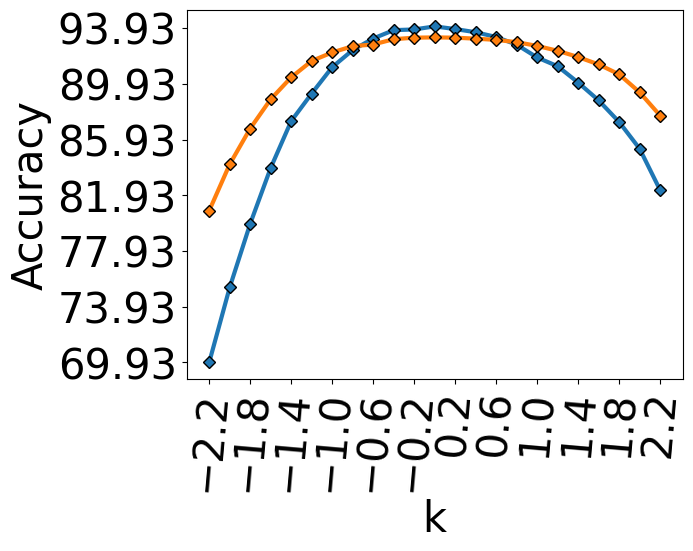}
    \includegraphics[width=\wide]{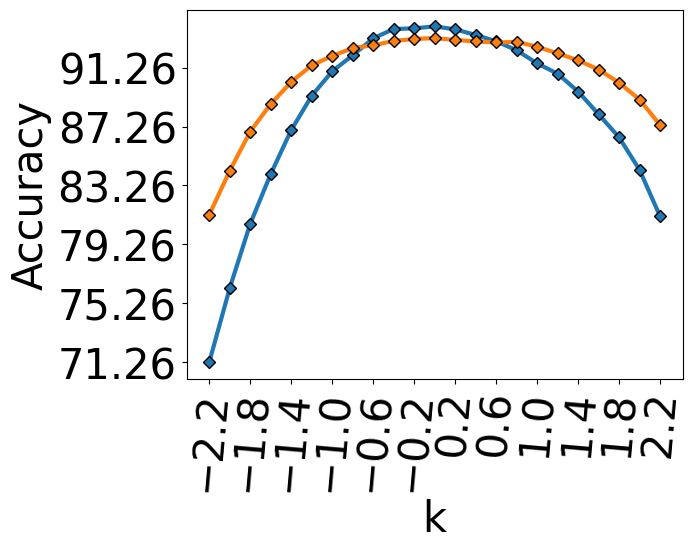}
    \includegraphics[width=\wide]{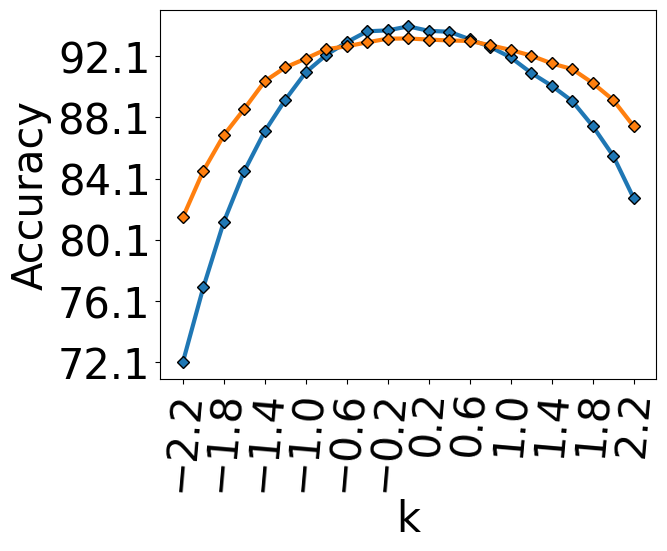}
    \includegraphics[width=\wide]{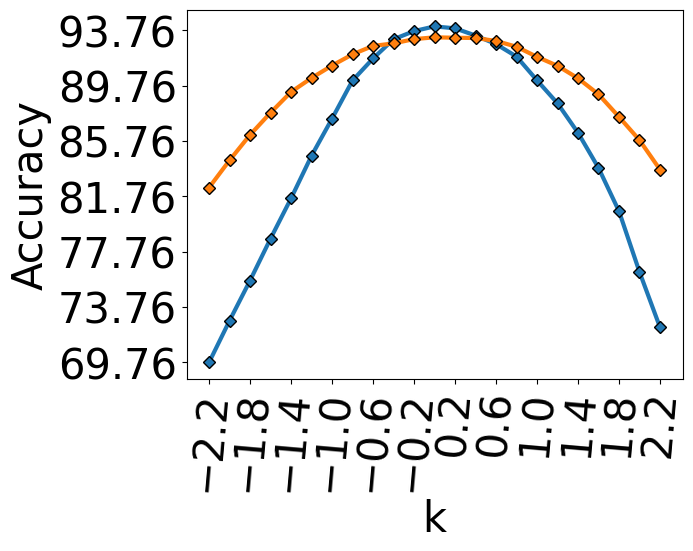}
    \includegraphics[width=\wide]{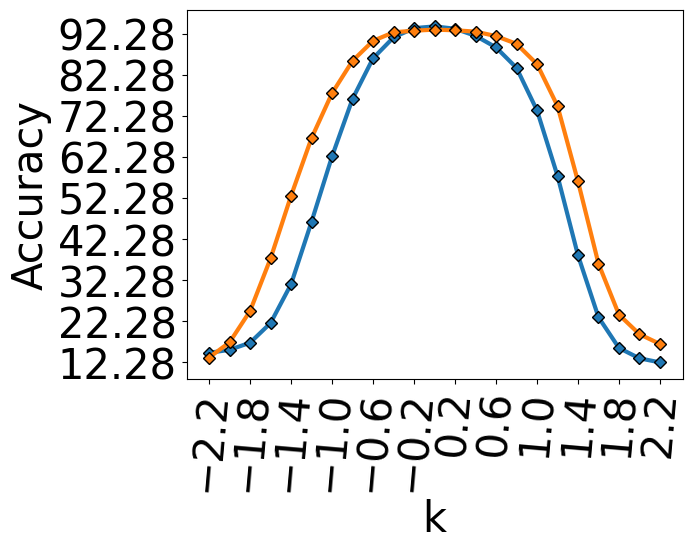}
    \includegraphics[width=\wide]{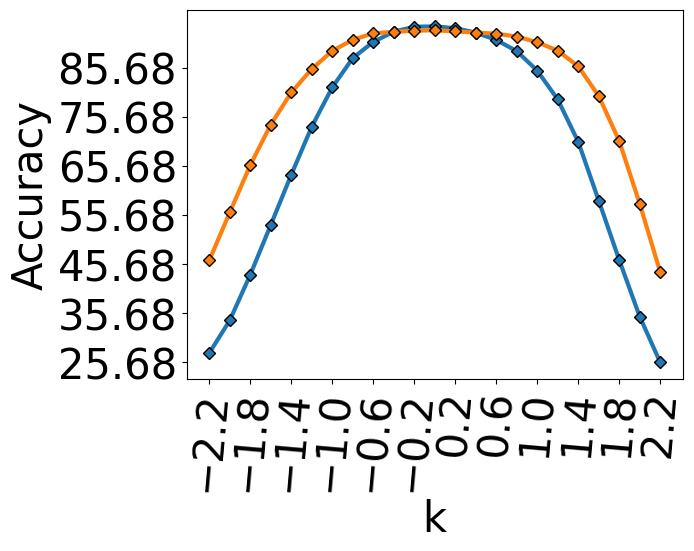}
    \includegraphics[width=\wide]{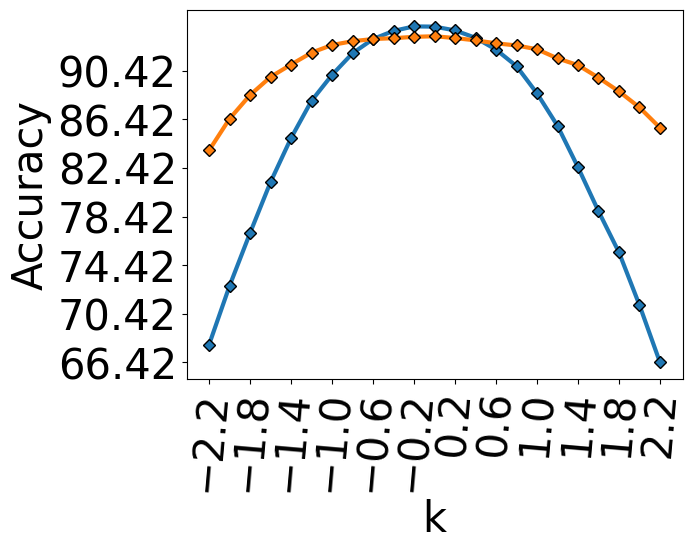}
    \includegraphics[width=\wide]{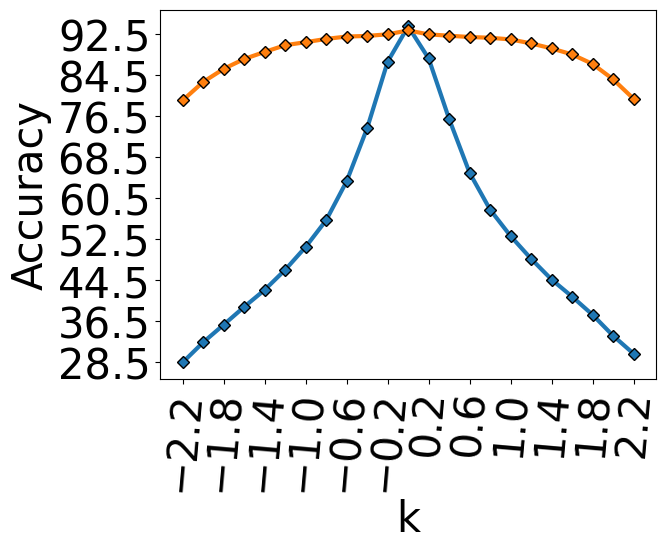}
    \includegraphics[width=\wide]{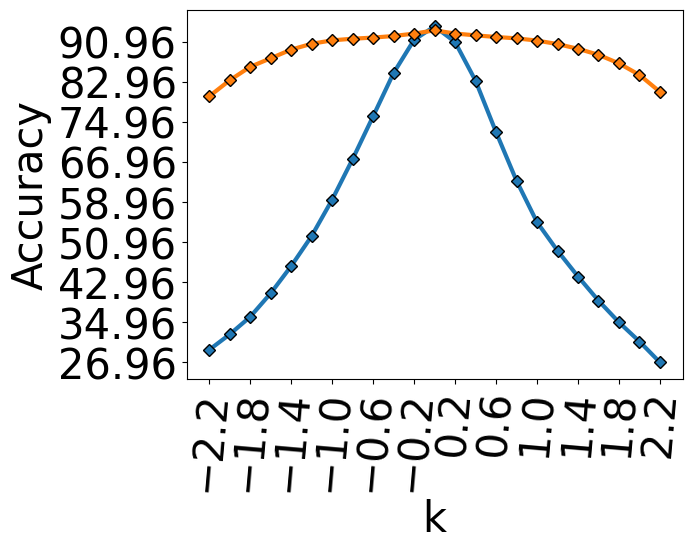}
    \includegraphics[width=\wide]{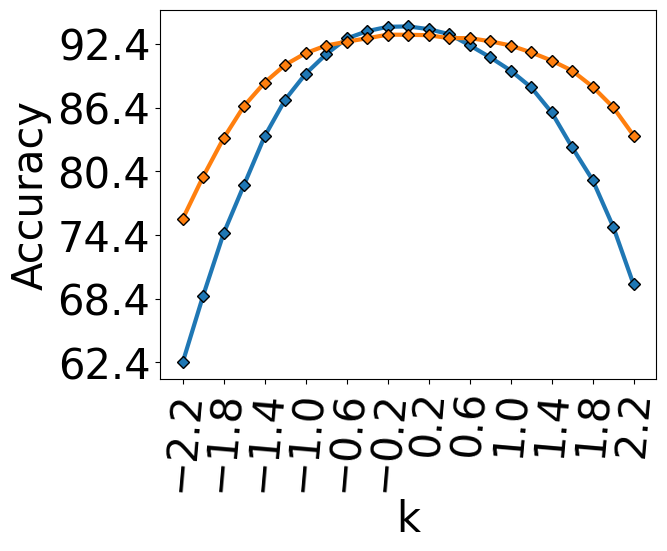}
    \includegraphics[width=\wide]{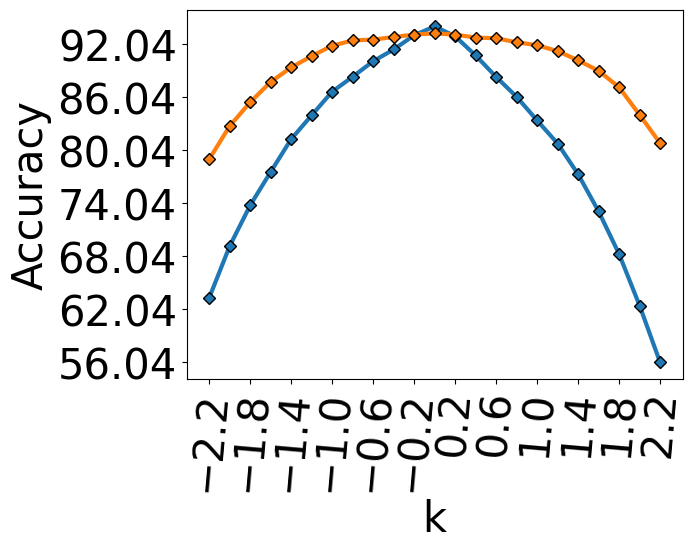}
    \caption{Results of ResNet18 model on CIFAR10 dataset under different NUI attacks (test set). \textcolor{blue}{Blue} and \textcolor{orange}{orange} curves show the performance of the model trained on the original training set and the NUI perturbed training set, respectively.}
    \label{fig:cifar_res}
\end{figure*}

\begin{figure*}[!t]
    \centering
    \newcommand\wide{2.93cm}
    \includegraphics[width=\wide]{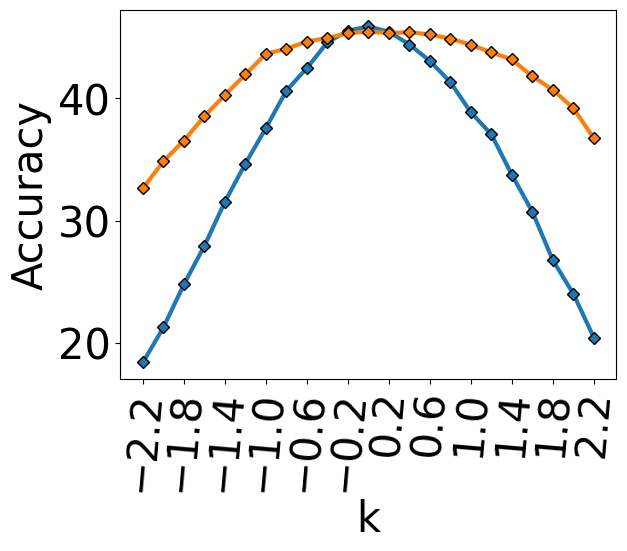}
    \includegraphics[width=\wide]{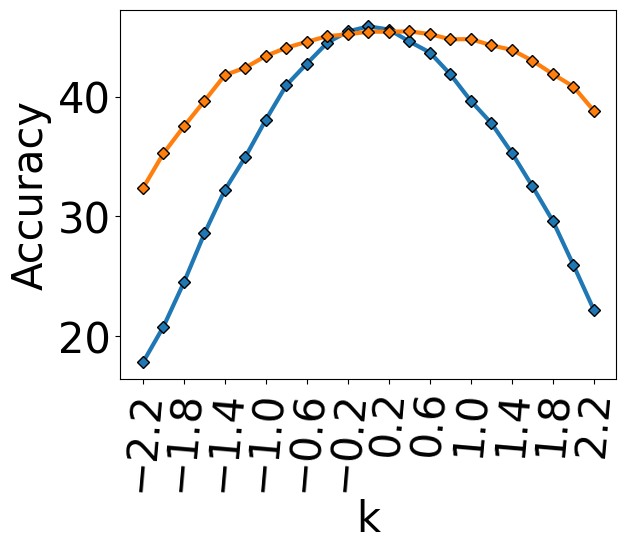}
    \includegraphics[width=\wide]{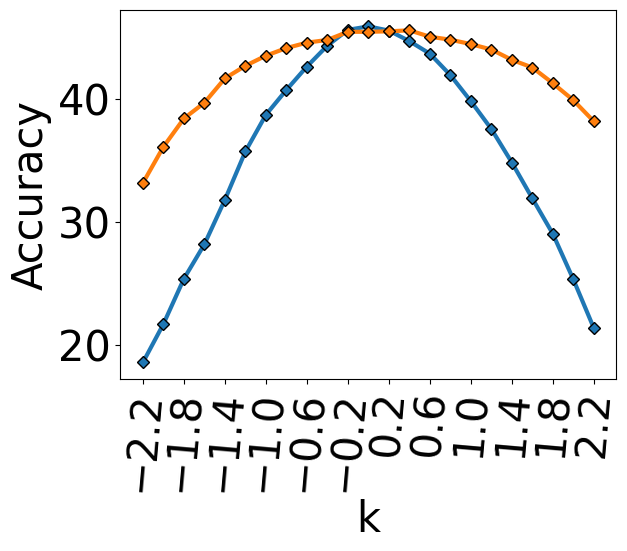}
    \includegraphics[width=\wide]{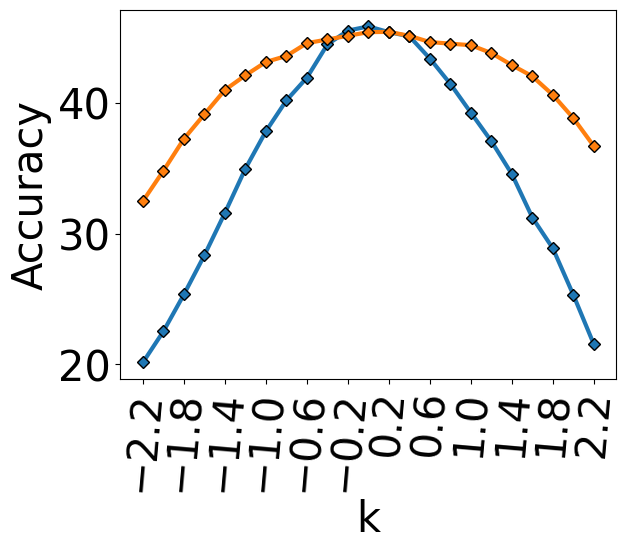}
    \includegraphics[width=\wide]{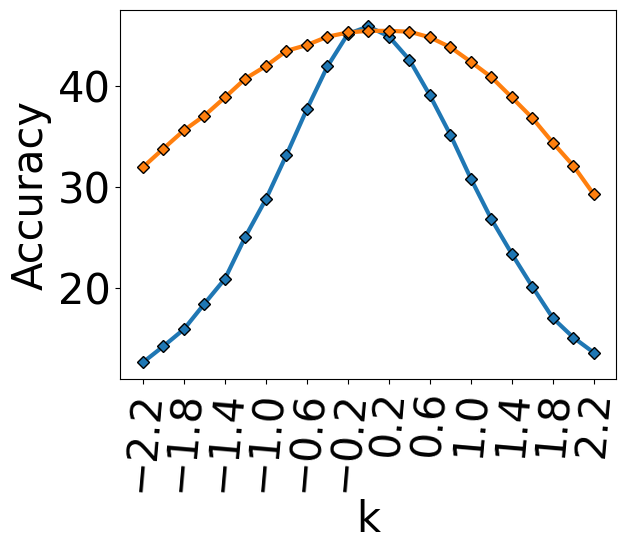}
    \includegraphics[width=\wide]{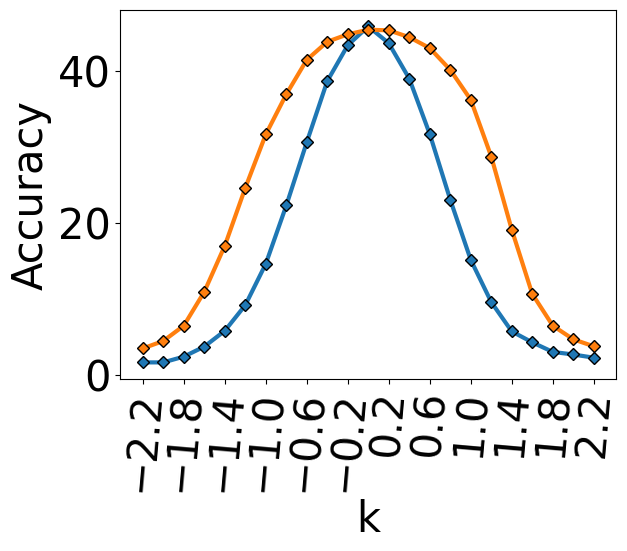}
    \includegraphics[width=\wide]{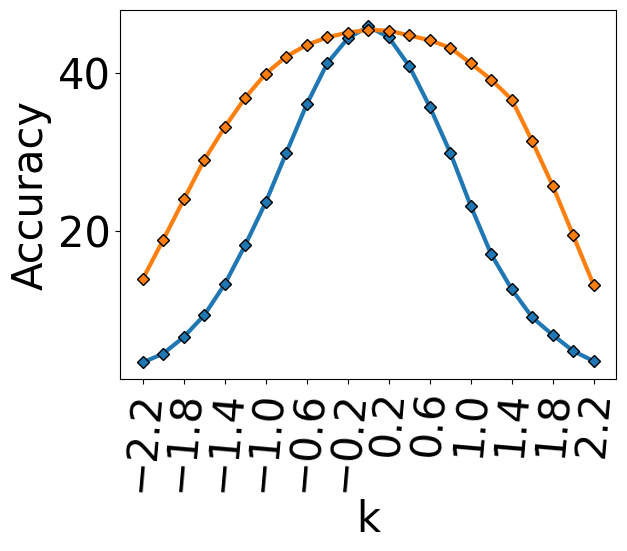}
    \includegraphics[width=\wide]{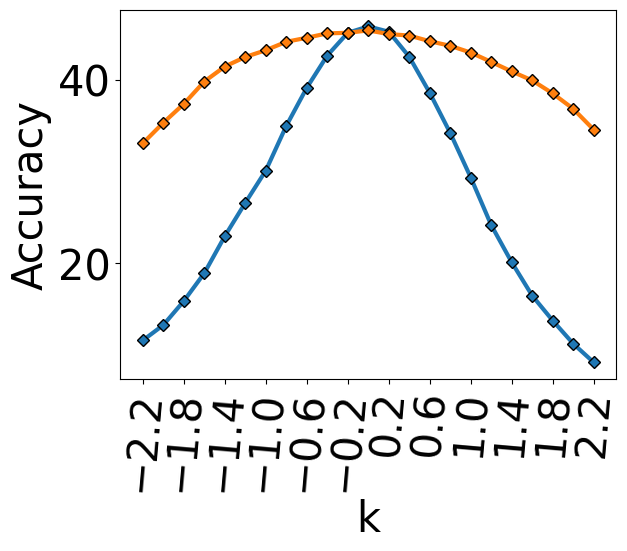}
    \includegraphics[width=\wide]{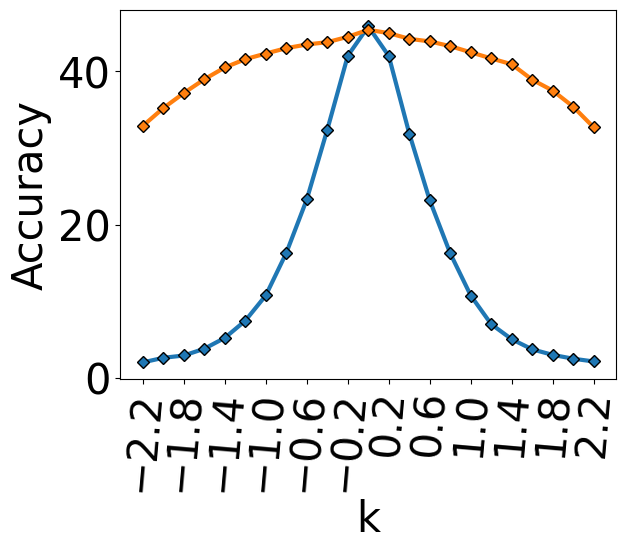}
    \includegraphics[width=\wide]{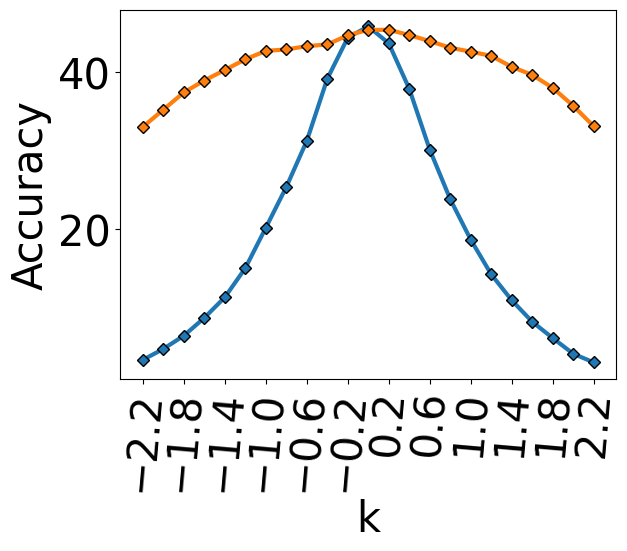}
    \includegraphics[width=\wide]{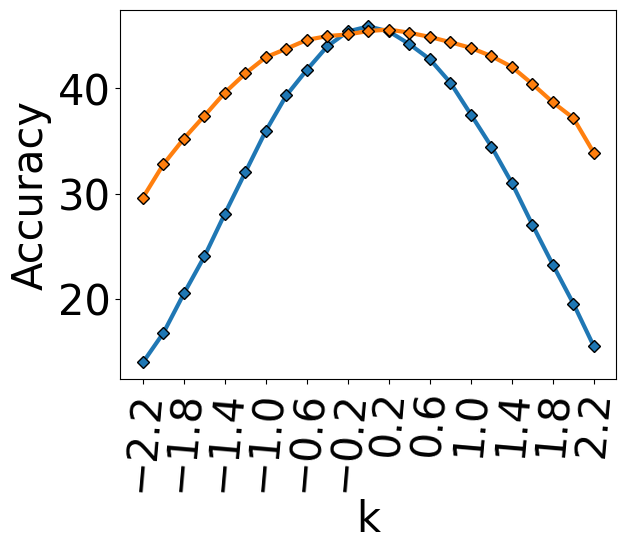}
    \includegraphics[width=\wide]{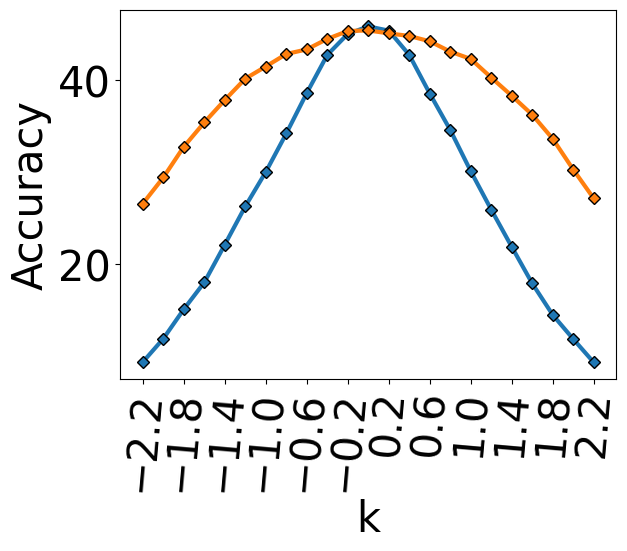}
    \caption{Results of VGG16 model on TinyImageNet dataset under different NUI attacks (test set). \textcolor{blue}{Blue} and \textcolor{orange}{orange} curves show the results of the model trained on the original training set and the NUI perturbed training set, respectively.} 
    \label{fig:tiny_vgg}
\end{figure*}

\begin{figure*}[!t]
    \centering
    \newcommand\wide{2.93cm}
    \includegraphics[width=\wide]{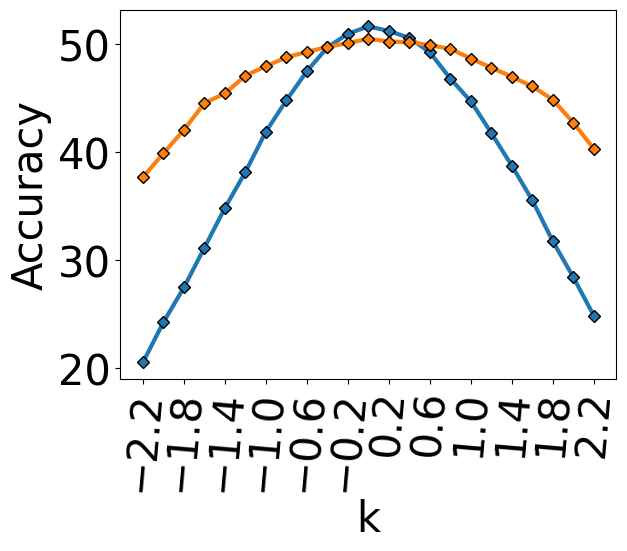}
    \includegraphics[width=\wide]{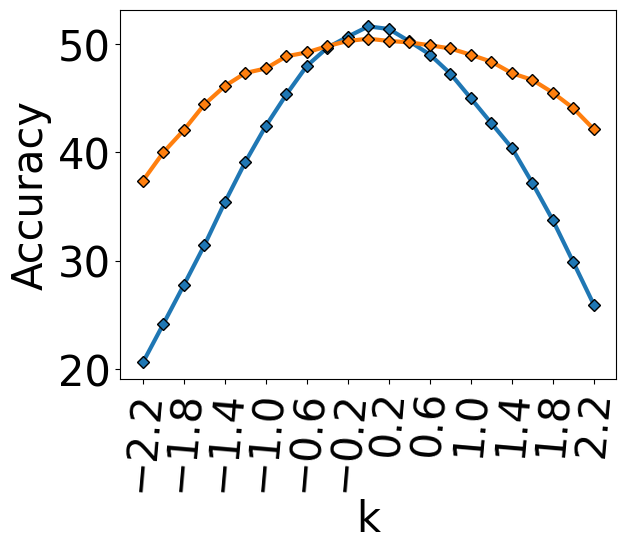}
    \includegraphics[width=\wide]{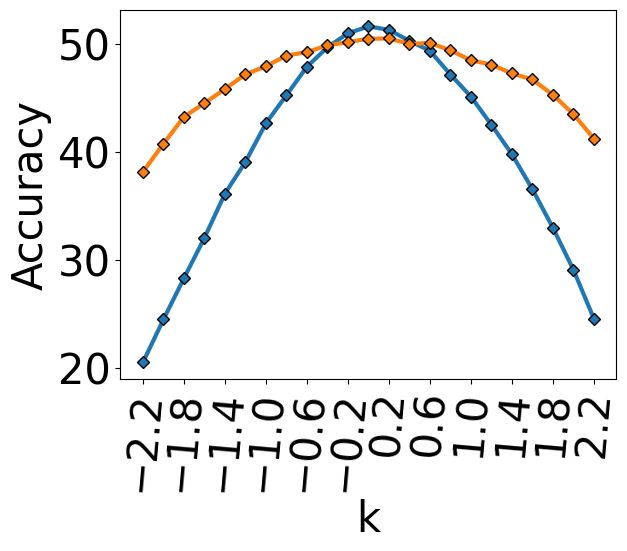}
    \includegraphics[width=\wide]{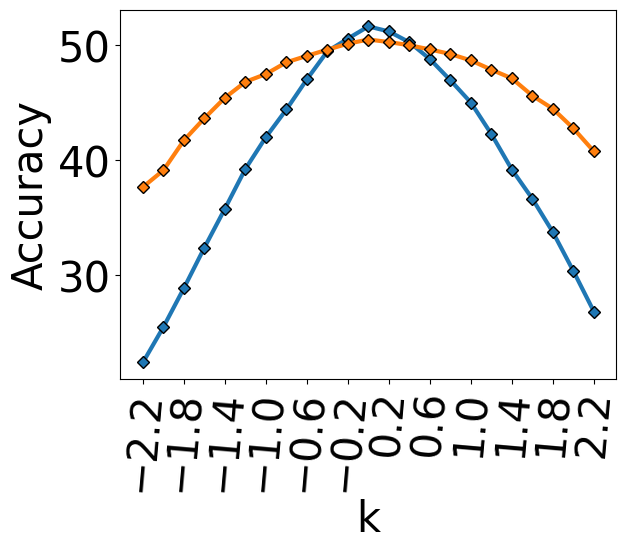}
    \includegraphics[width=\wide]{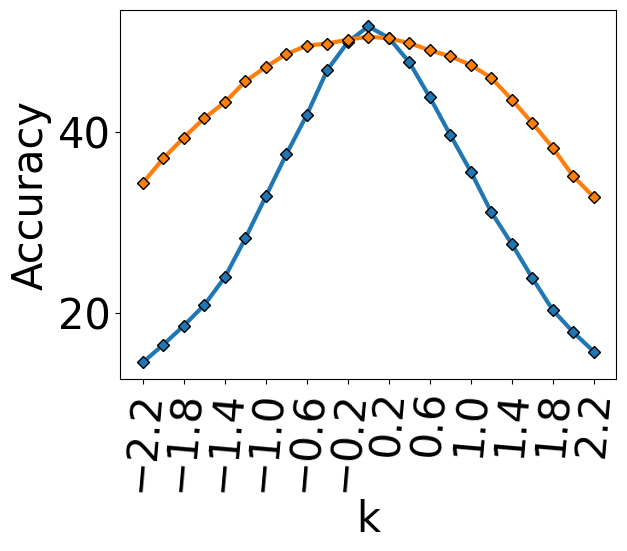}
    \includegraphics[width=\wide]{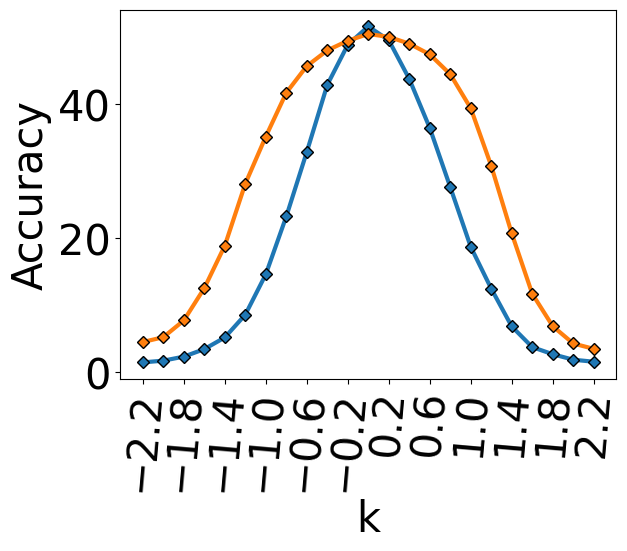}
    \includegraphics[width=\wide]{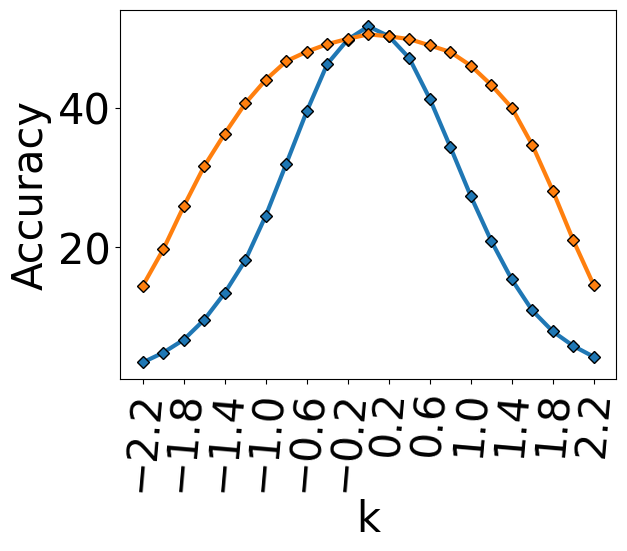}
    \includegraphics[width=\wide]{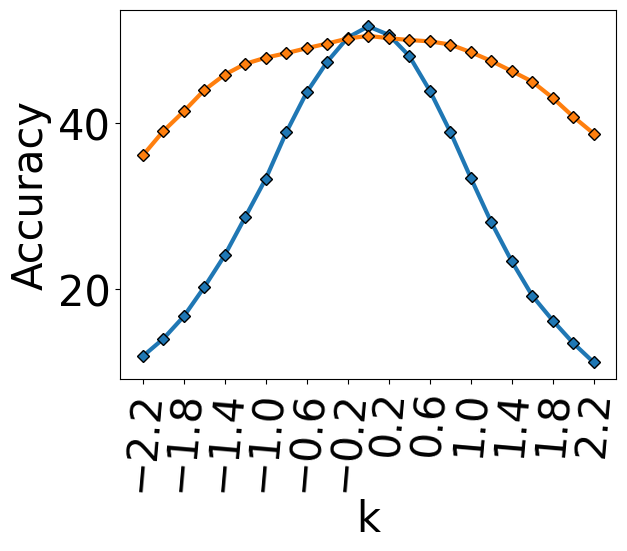}
    \includegraphics[width=\wide]{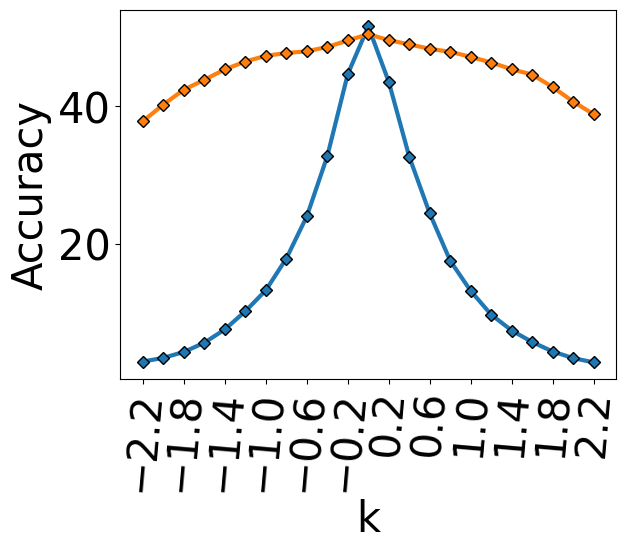}
    \includegraphics[width=\wide]{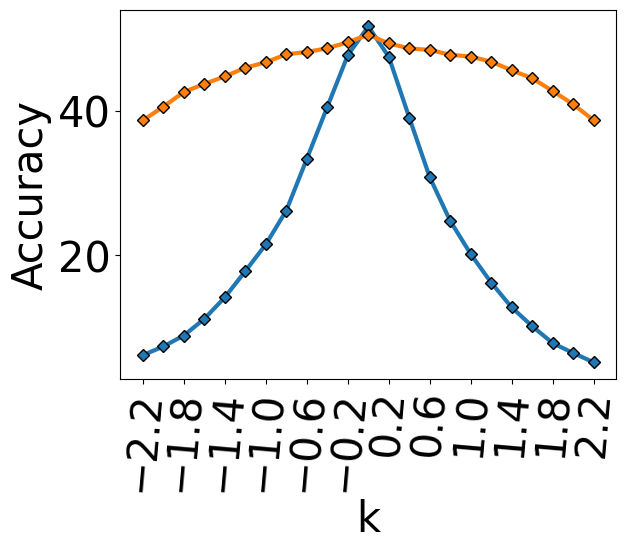}
    \includegraphics[width=\wide]{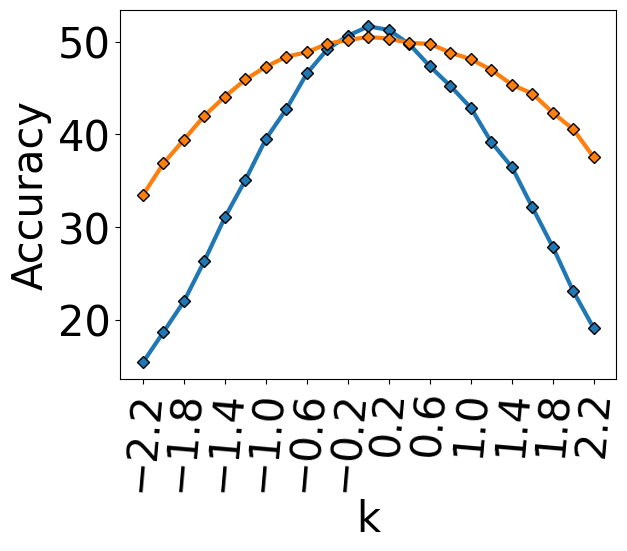}
    \includegraphics[width=\wide]{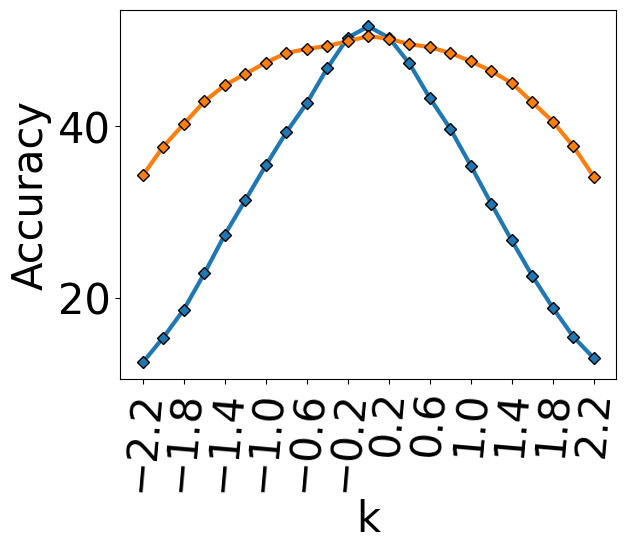}
    \caption{Results of ResNet18 model on TinyImageNet dataset under different NUI attacks (test set). \textcolor{blue}{Blue} and \textcolor{orange}{orange} curves show the results of the model trained on the original training set and the NUI perturbed training set, respectively.}
    \label{fig:tiny_res}
\end{figure*}

\begin{figure*}[!t]
    \centering
    \newcommand\wide{2.93cm}
    \includegraphics[width=\wide]{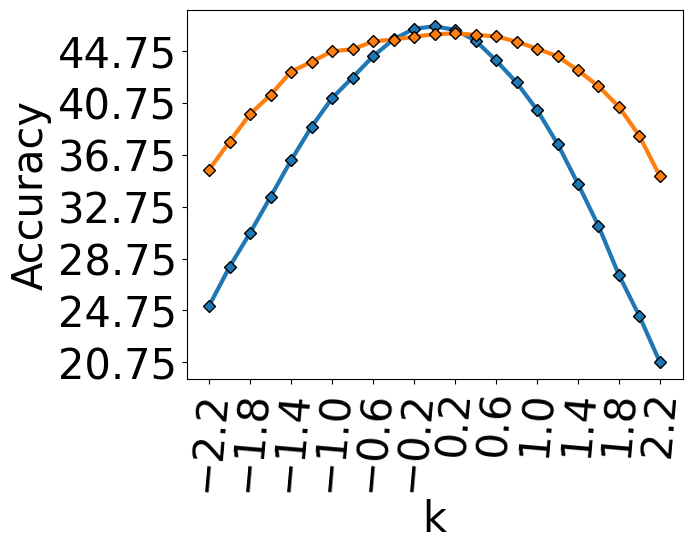}
    \includegraphics[width=\wide]{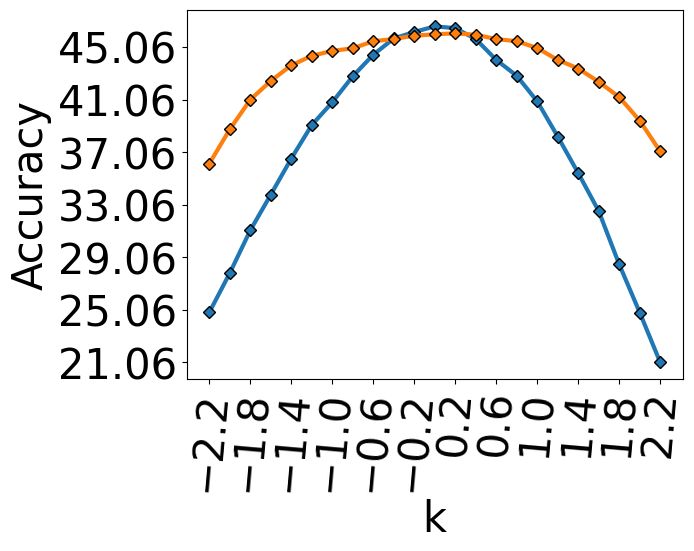}
    \includegraphics[width=\wide]{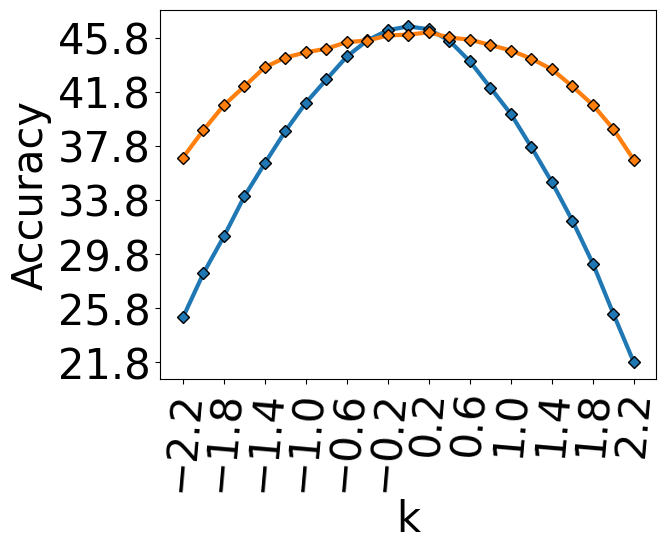}
    \includegraphics[width=\wide]{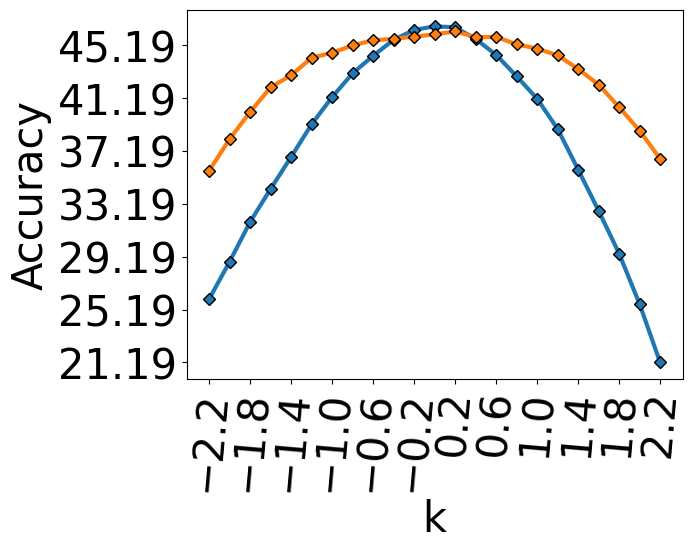}
    \includegraphics[width=\wide]{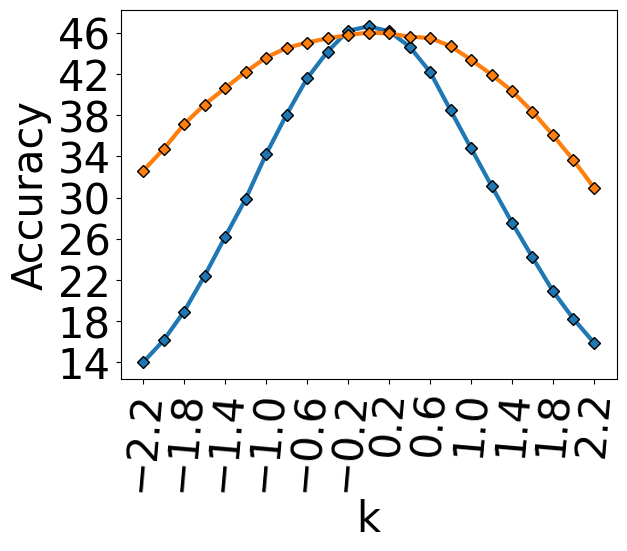}
    \includegraphics[width=\wide]{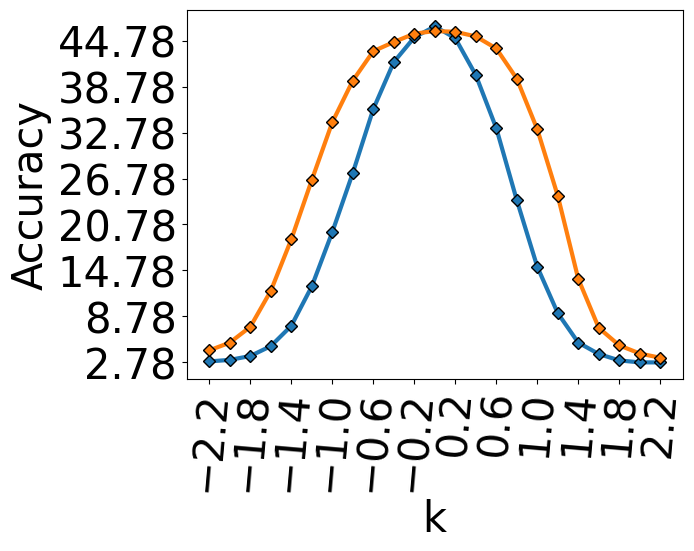}
    \includegraphics[width=\wide]{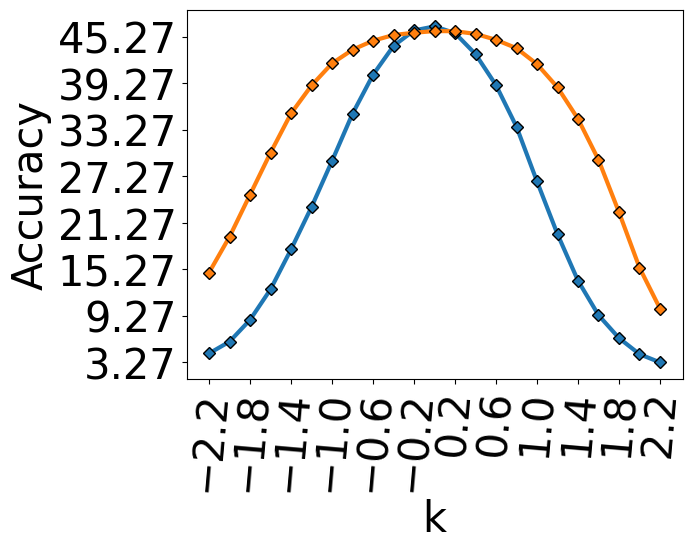}
    \includegraphics[width=\wide]{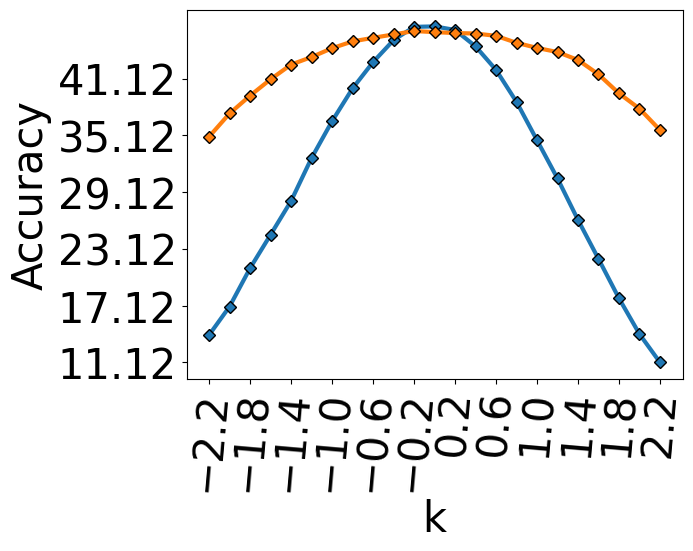}
    \includegraphics[width=\wide]{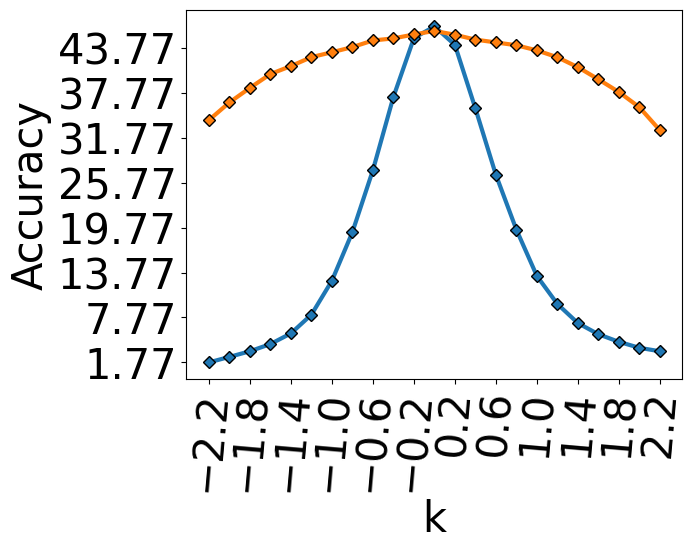}
    \includegraphics[width=\wide]{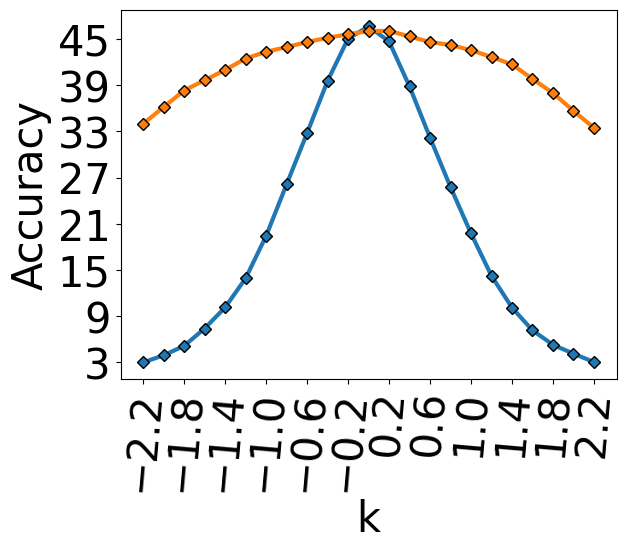}
    \includegraphics[width=\wide]{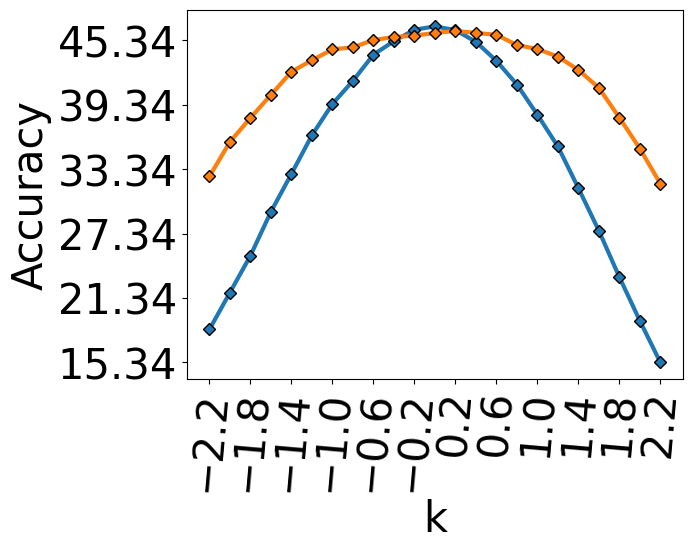}
    \includegraphics[width=\wide]{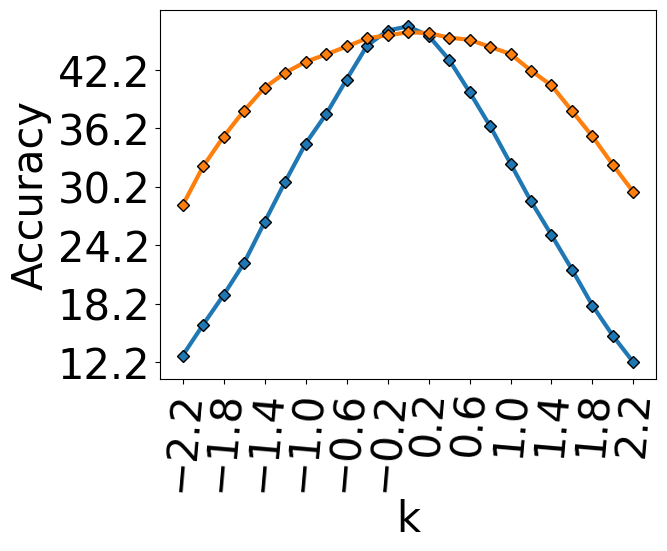}
    \caption{Results of VGG19 model on CalTech256 dataset under different NUI attacks (test set). \textcolor{blue}{Blue} and \textcolor{orange}{orange} curves show the results of the model trained on the original training set and the NUI perturbed training set, respectively.} 
    \label{fig:cal_vgg}
\end{figure*}

\begin{figure*}[!t]
    \centering
    \newcommand\wide{2.93cm}
    \includegraphics[width=\wide]{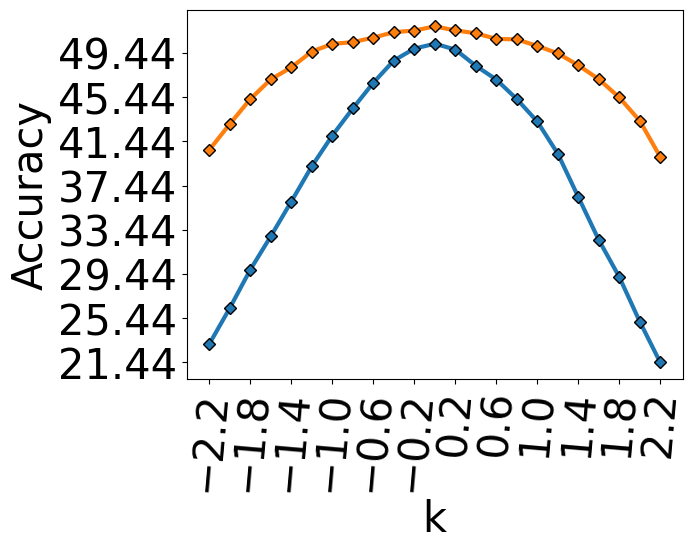}
    \includegraphics[width=\wide]{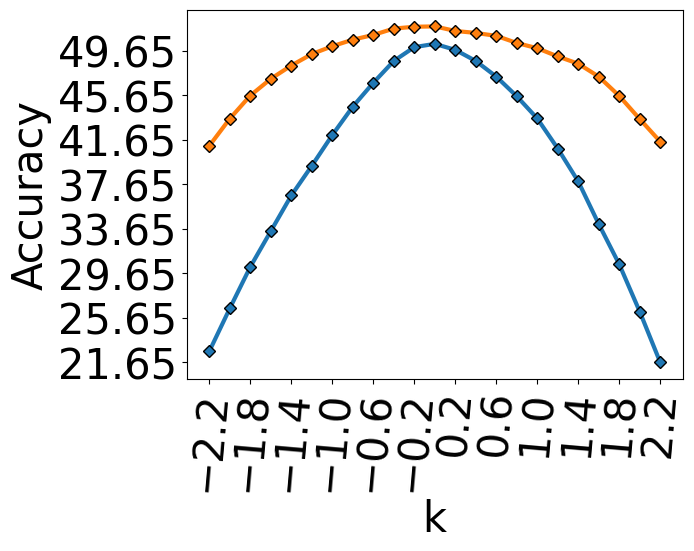}
    \includegraphics[width=\wide]{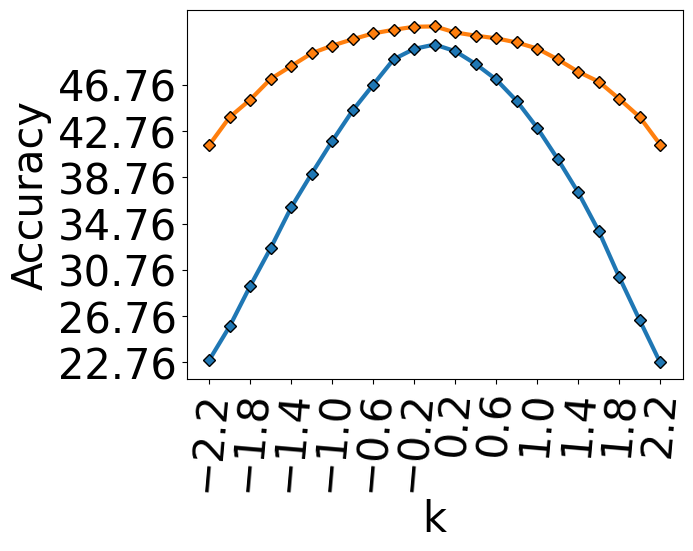}
    \includegraphics[width=\wide]{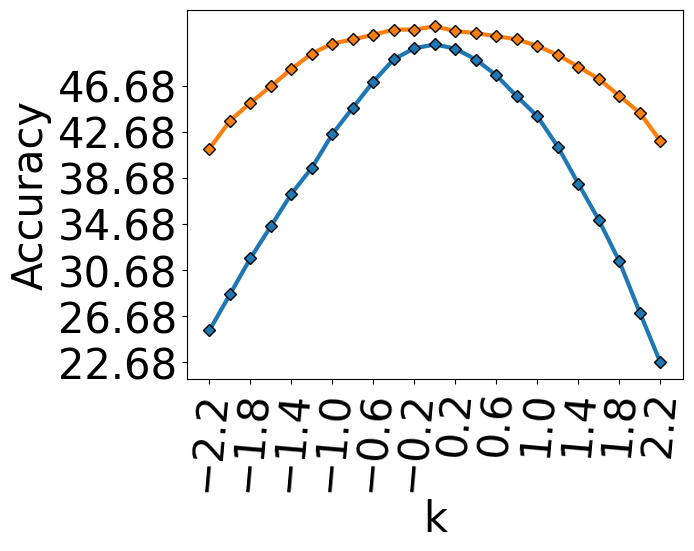}
    \includegraphics[width=\wide]{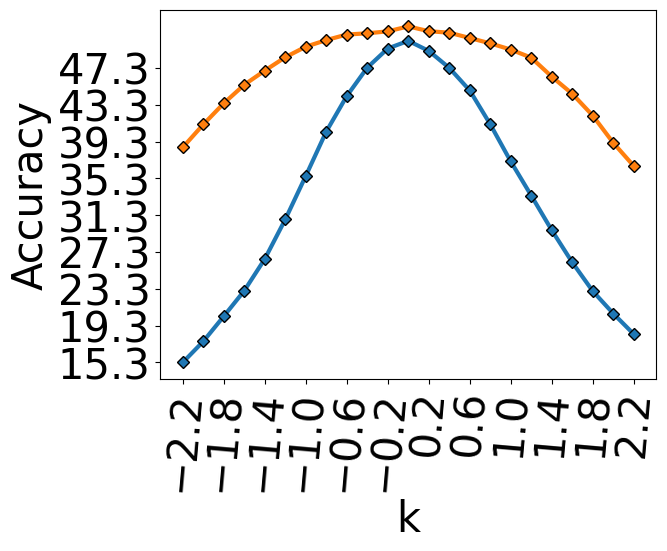}
    \includegraphics[width=\wide]{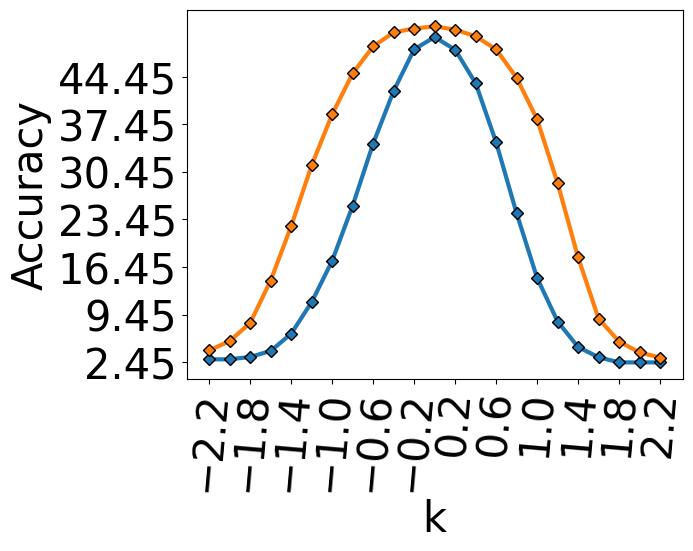}
    \includegraphics[width=\wide]{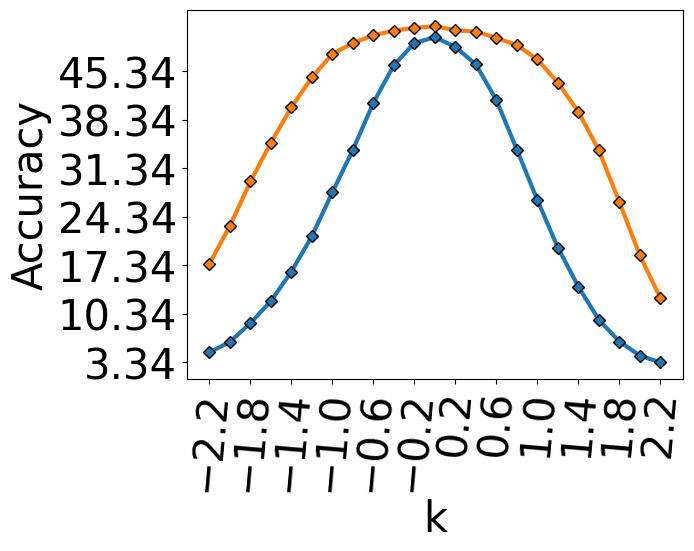}
    \includegraphics[width=\wide]{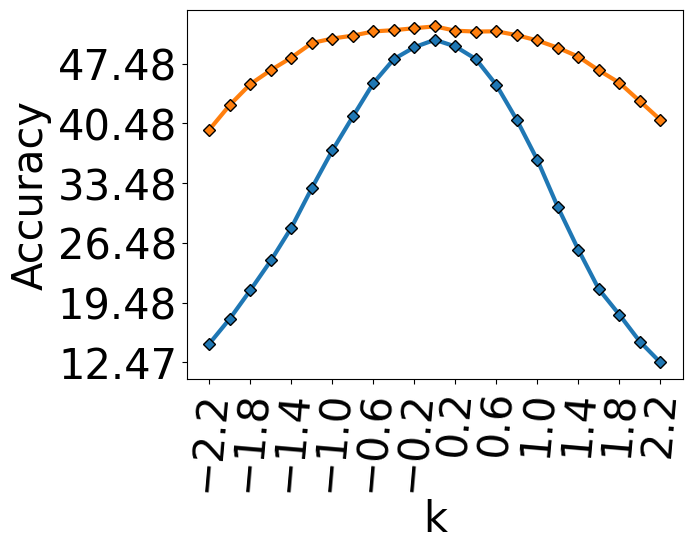}
    \includegraphics[width=\wide]{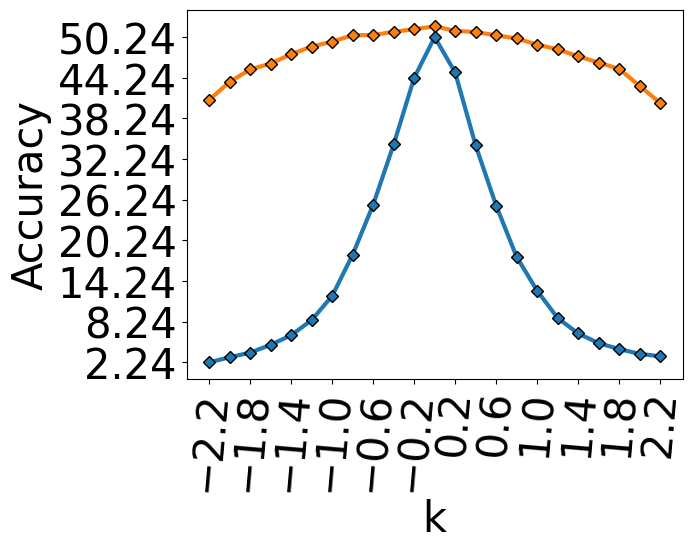}
    \includegraphics[width=\wide]{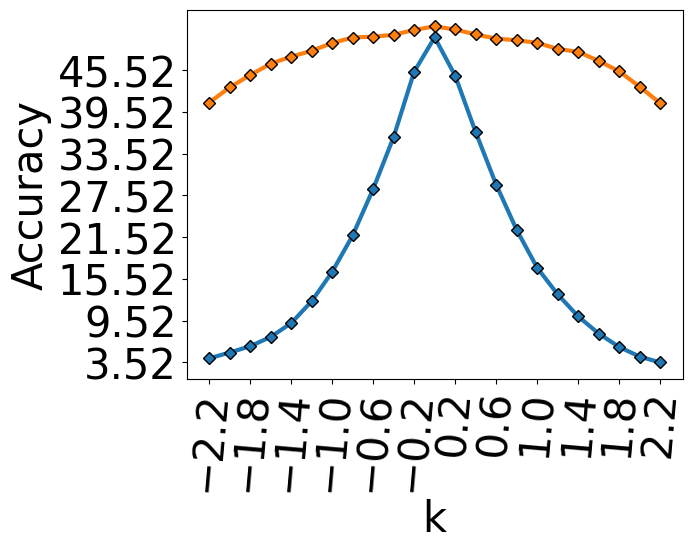}
    \includegraphics[width=\wide]{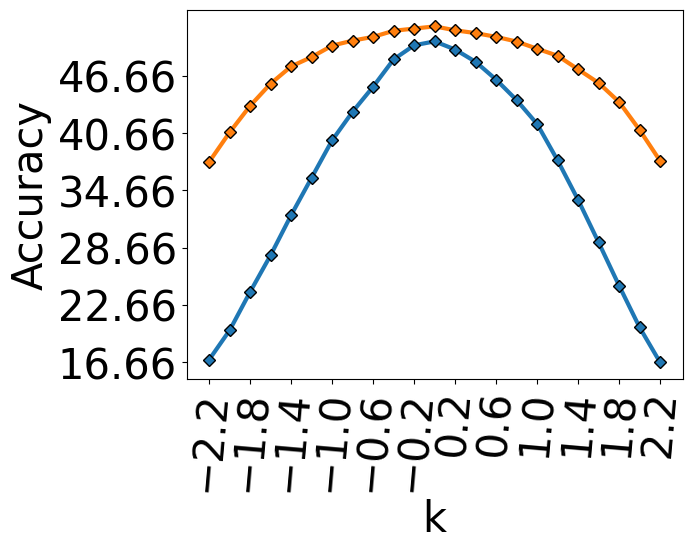}
    \includegraphics[width=\wide]{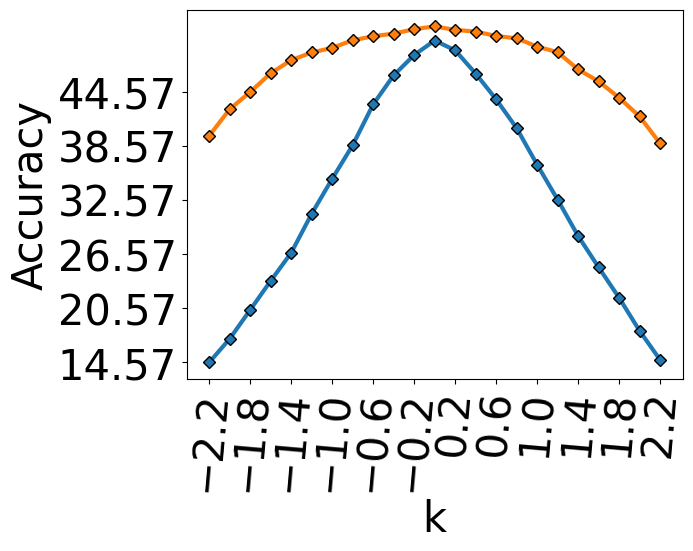}
    \caption{The results of the ResNet18 model on the CalTech256 dataset under different NUI attacks (test set). The \textcolor{blue}{blue} and \textcolor{orange}{orange} curves show the results of the model trained on the original training set and the NUI perturbed training set, respectively.}
    \label{fig:cal_res}
\end{figure*}

\begin{figure*}[!t]
    \centering
    \newcommand\wide{2.93cm}
    \includegraphics[width=\wide]{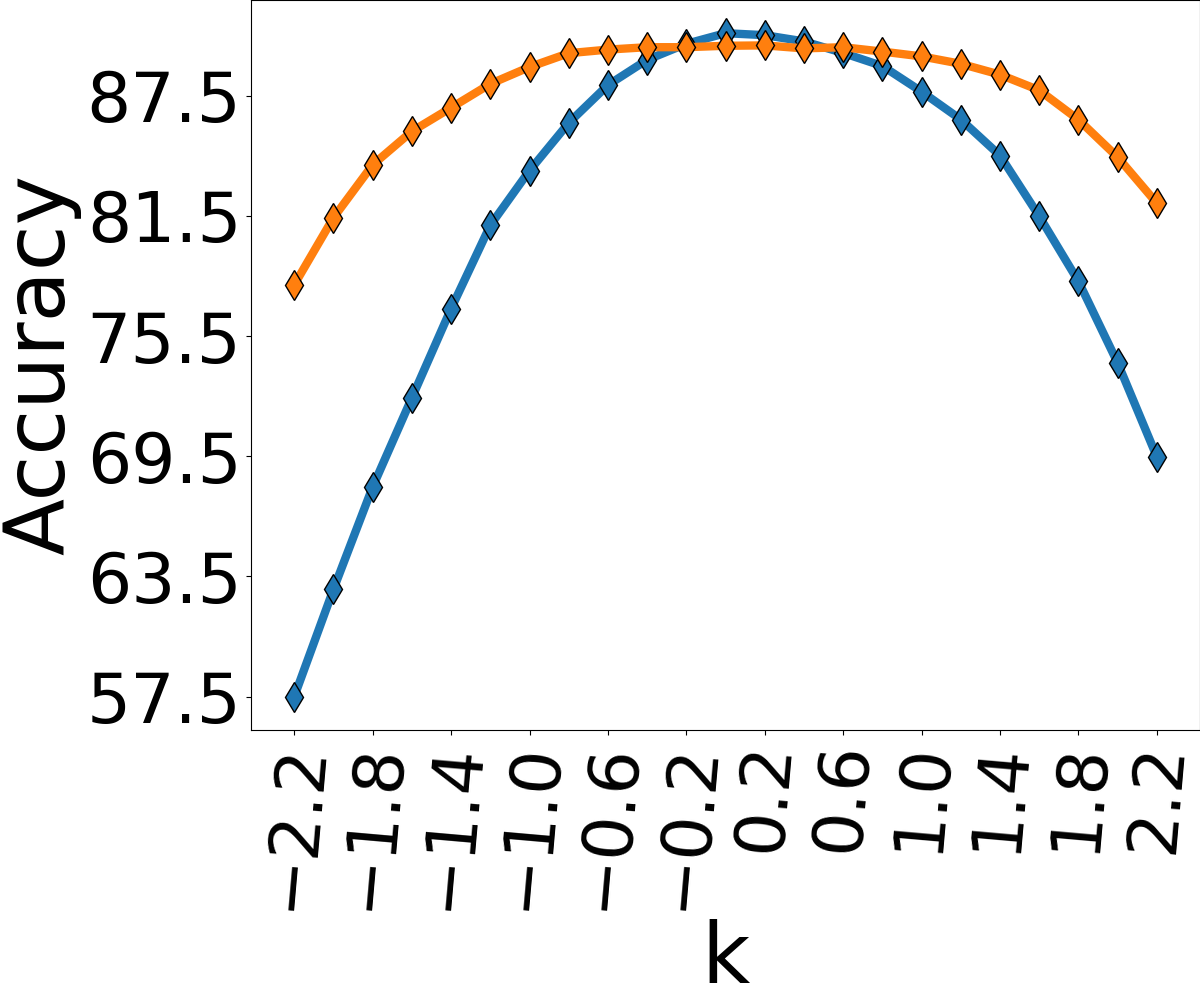}
    \includegraphics[width=\wide]{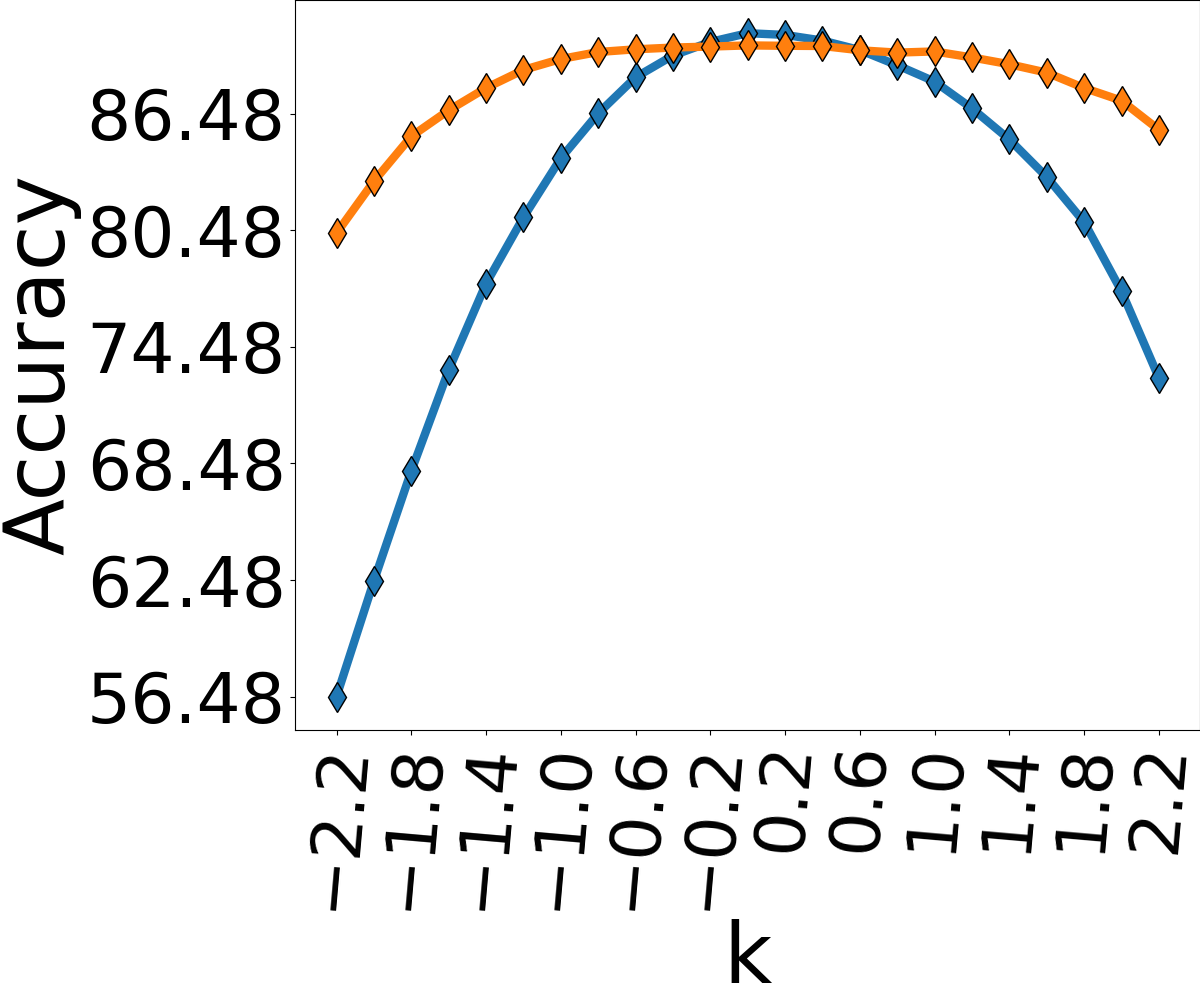}
    \includegraphics[width=\wide]{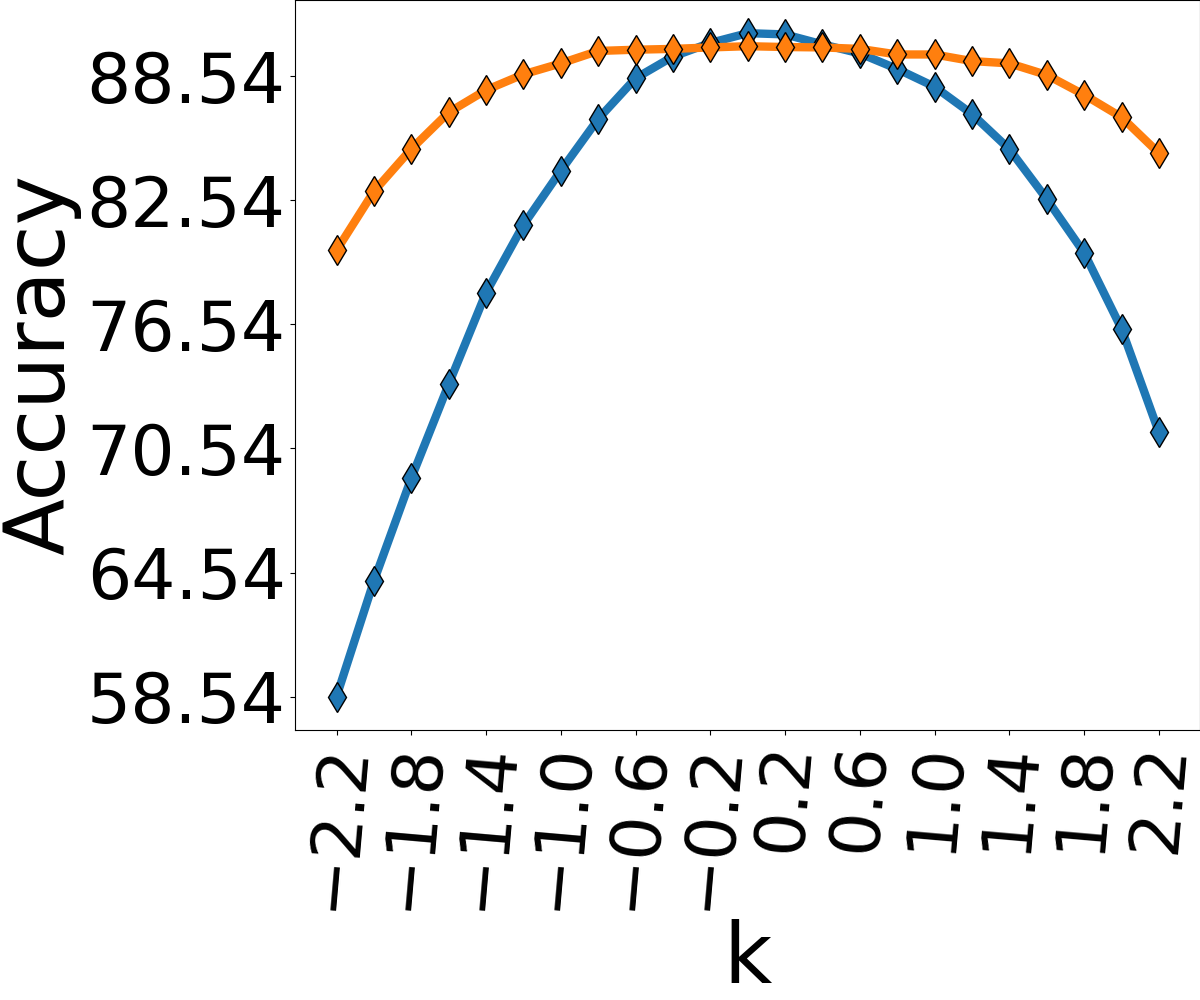}
    \includegraphics[width=\wide]{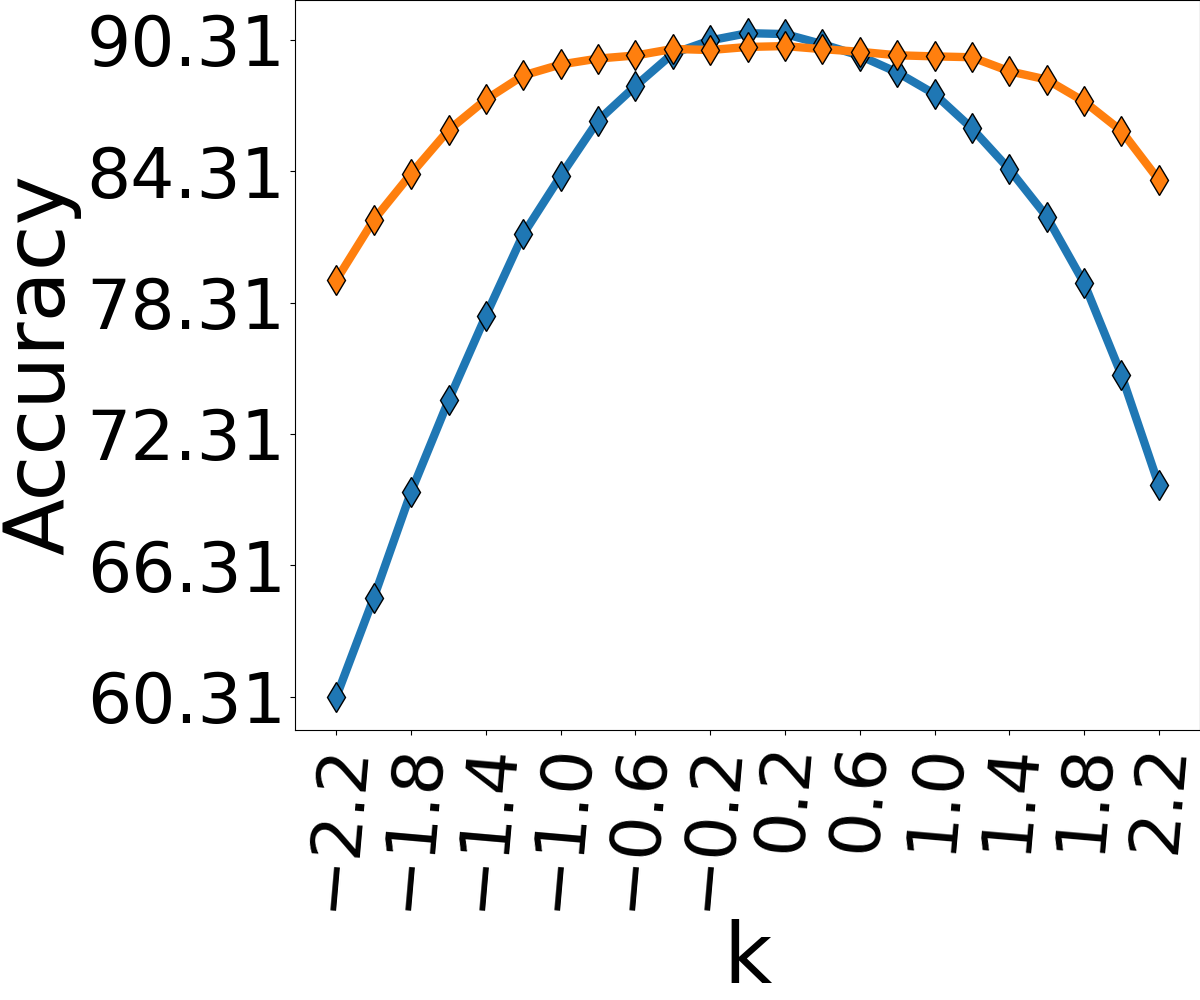}
    \includegraphics[width=\wide]{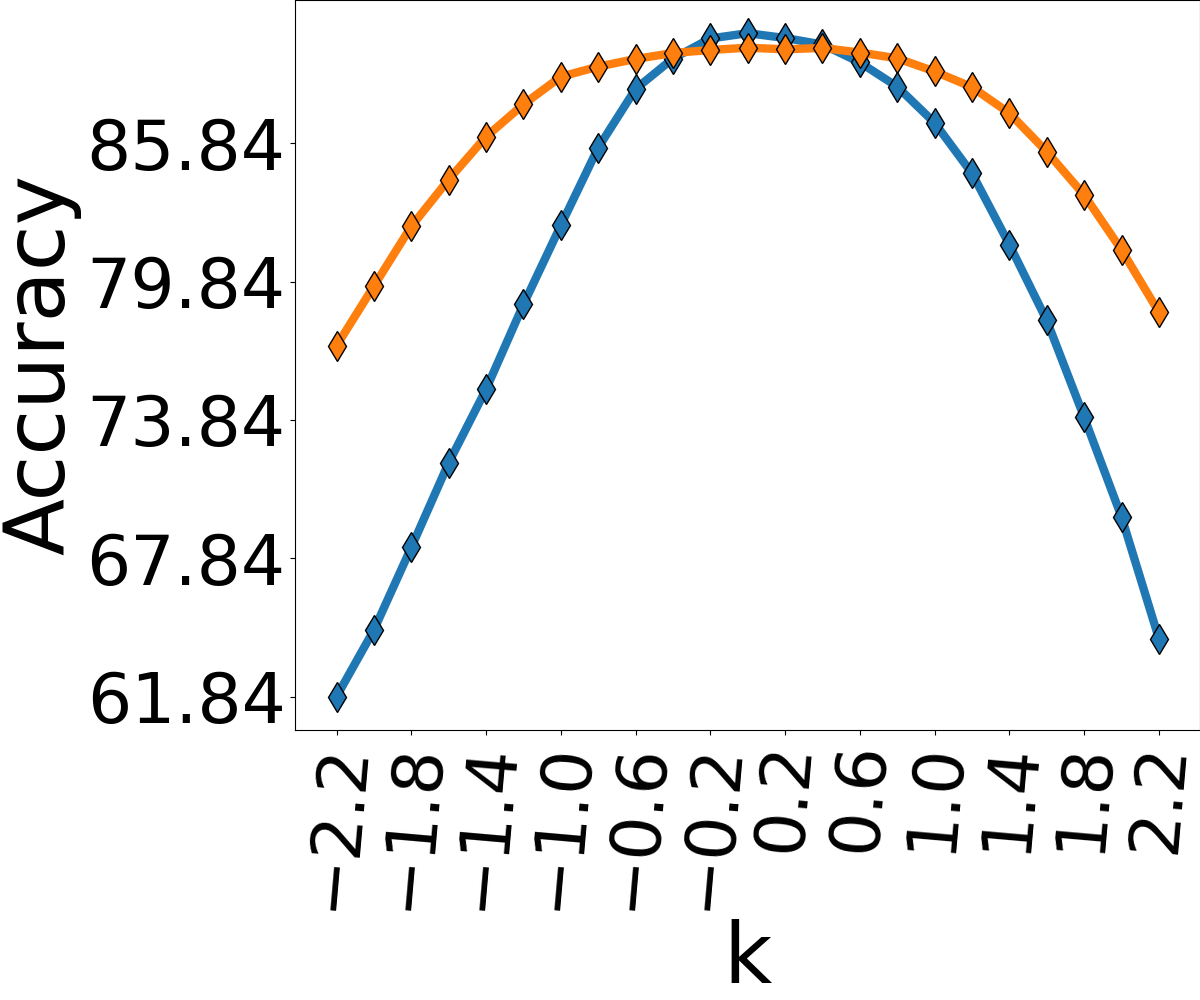}
    \includegraphics[width=\wide]{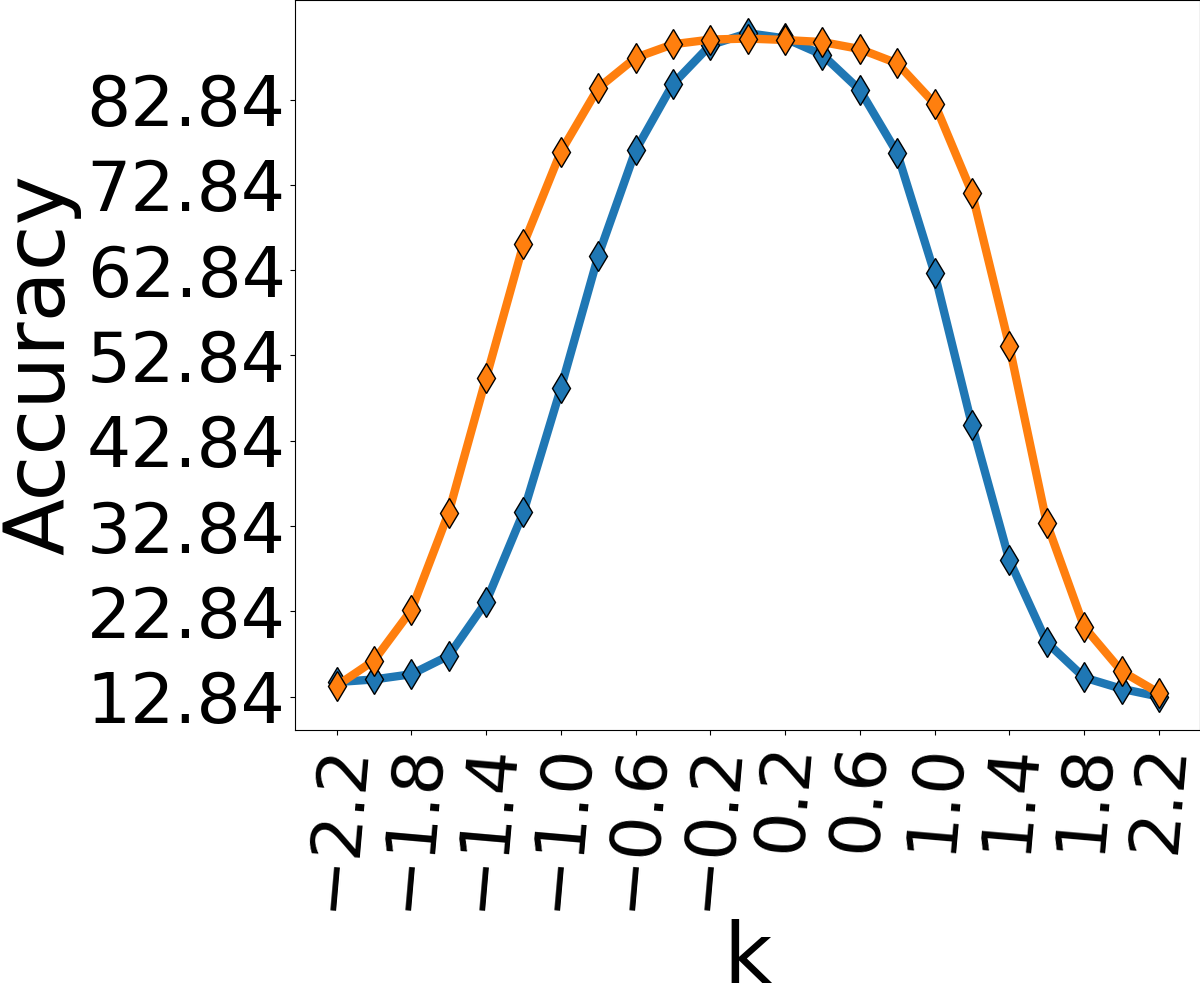}
    \includegraphics[width=\wide]{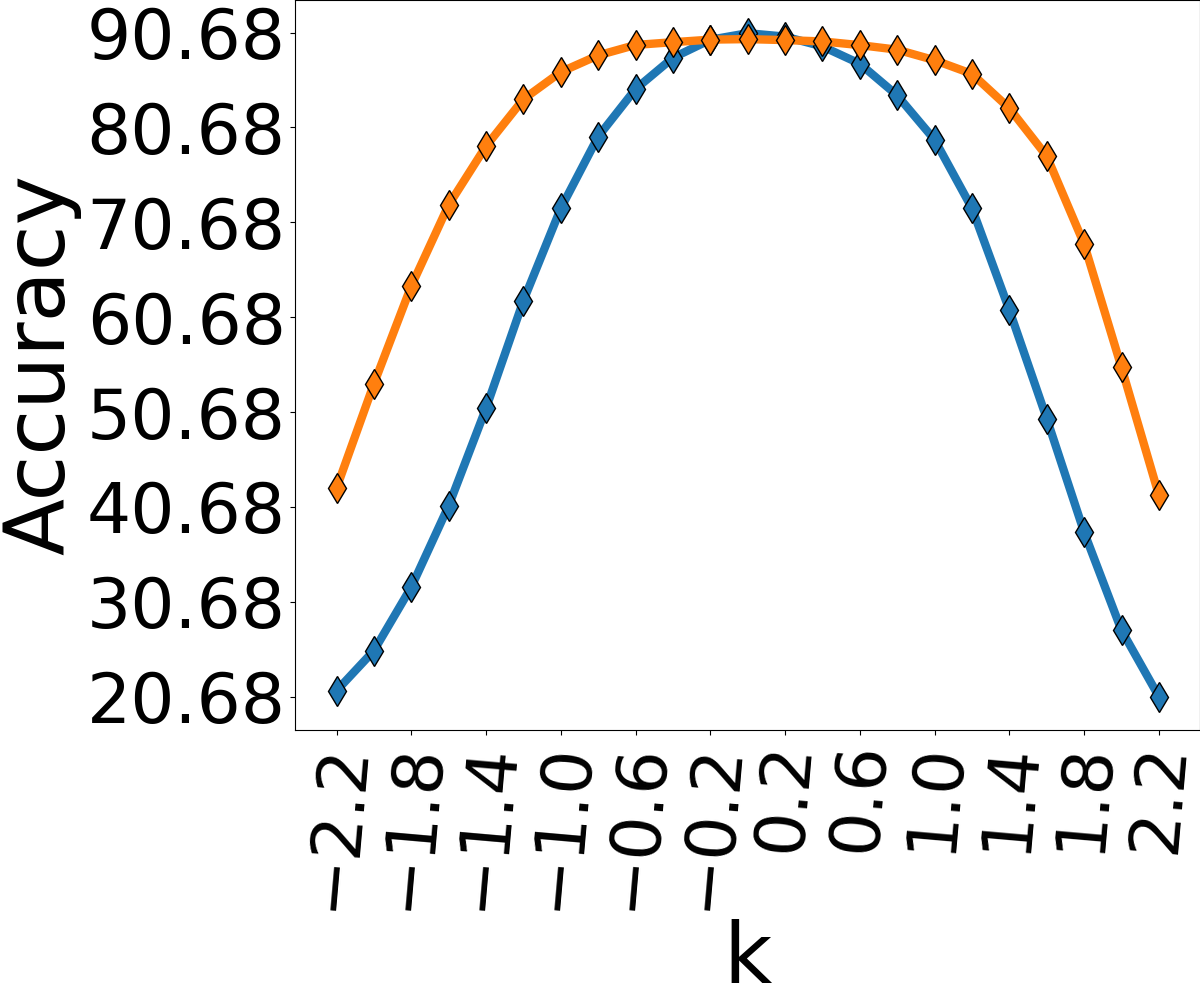}
    \includegraphics[width=\wide]{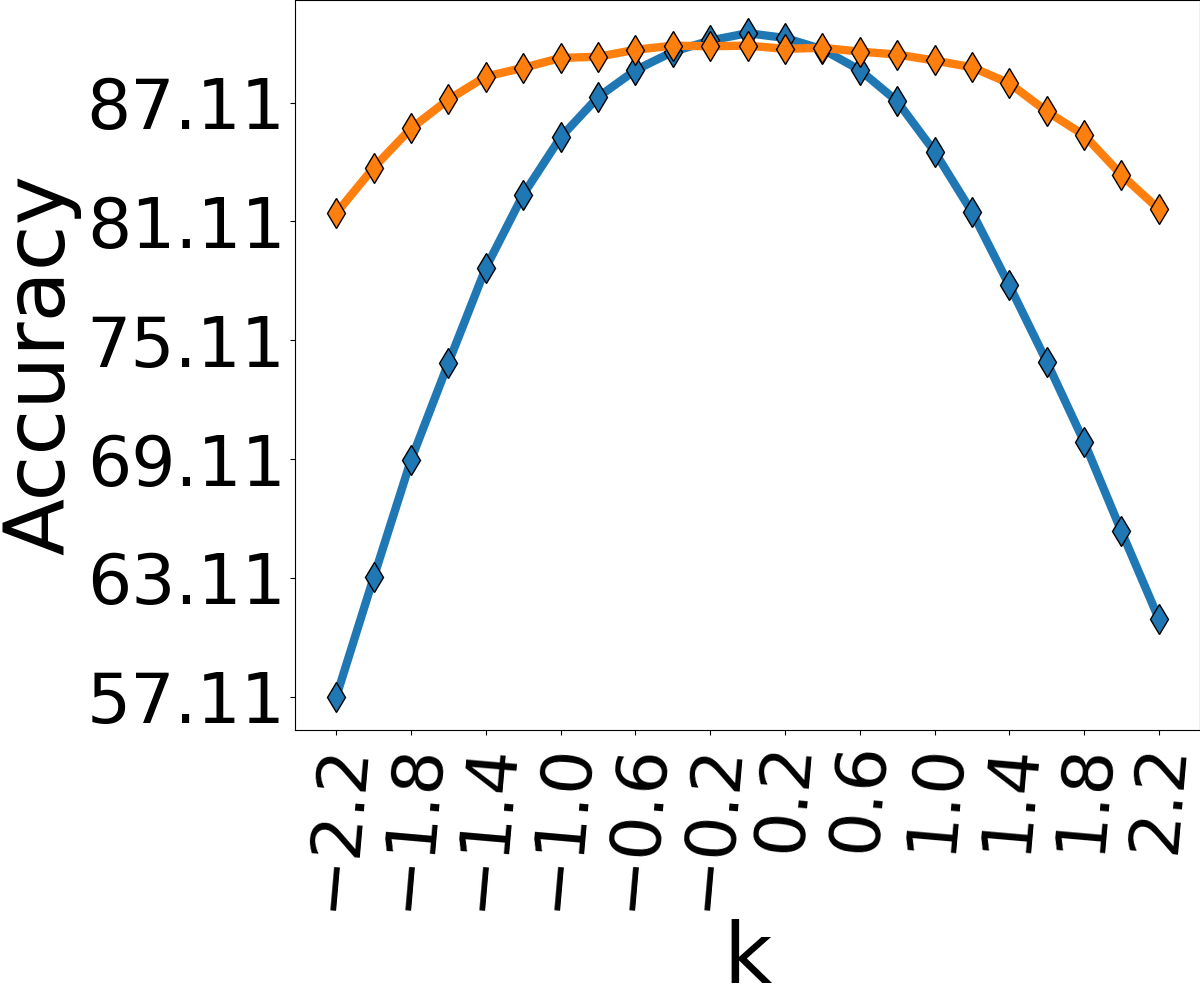}
    \includegraphics[width=\wide]{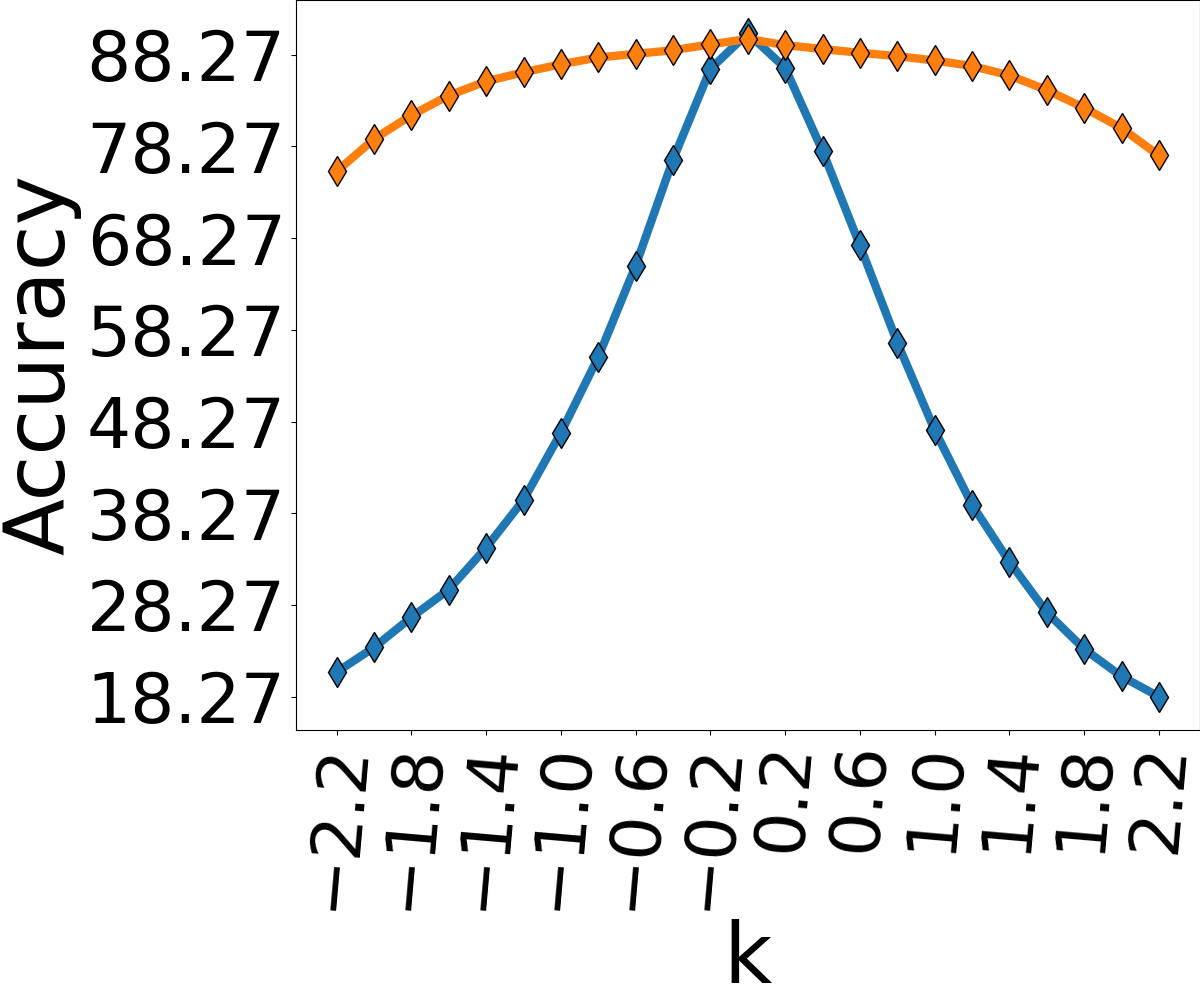}
    \includegraphics[width=\wide]{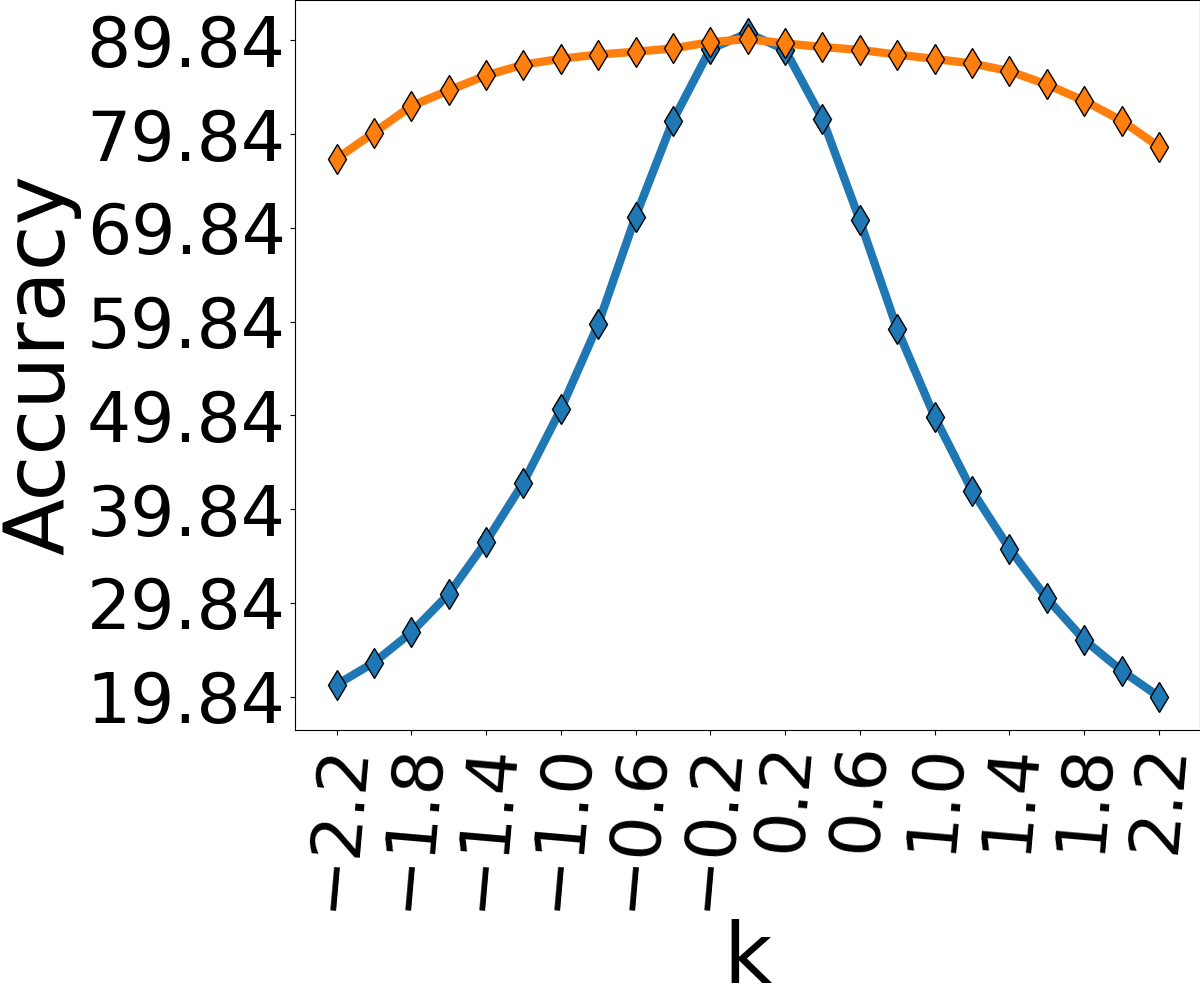}
    \includegraphics[width=\wide]{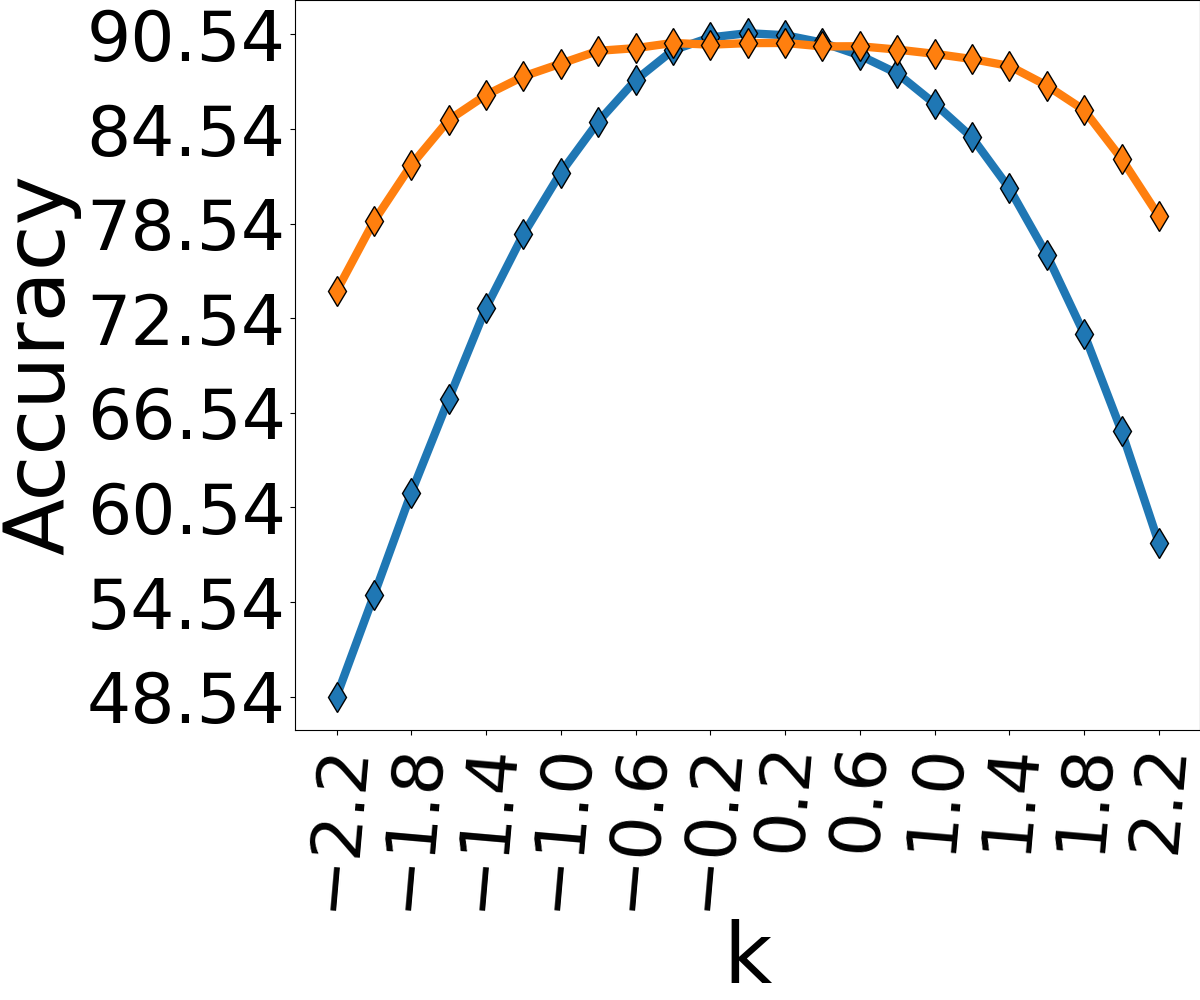}
    \includegraphics[width=\wide]{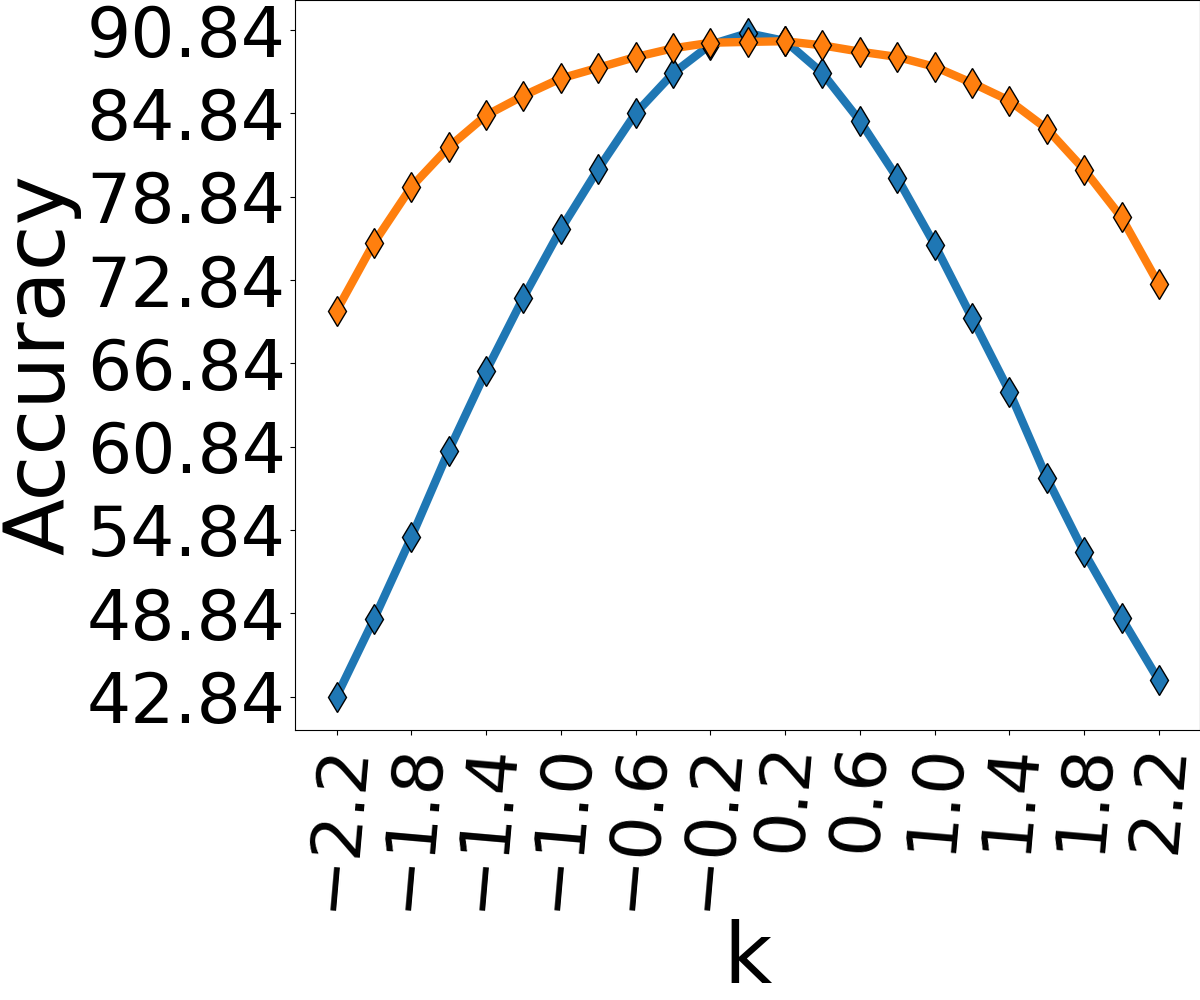}
    \caption{The results of the MobilenetV3-small model on the CIFAR10 dataset under different NUI attacks (test set). The \textcolor{blue}{blue} and \textcolor{orange}{orange} curves show the results of the model trained on the original training set and the NUI perturbed training set, respectively.}
    \label{fig:cifar_mob}
\end{figure*}

\begin{figure*}[!t]
    \centering
    \newcommand\wide{2.93cm}
    \includegraphics[width=\wide]{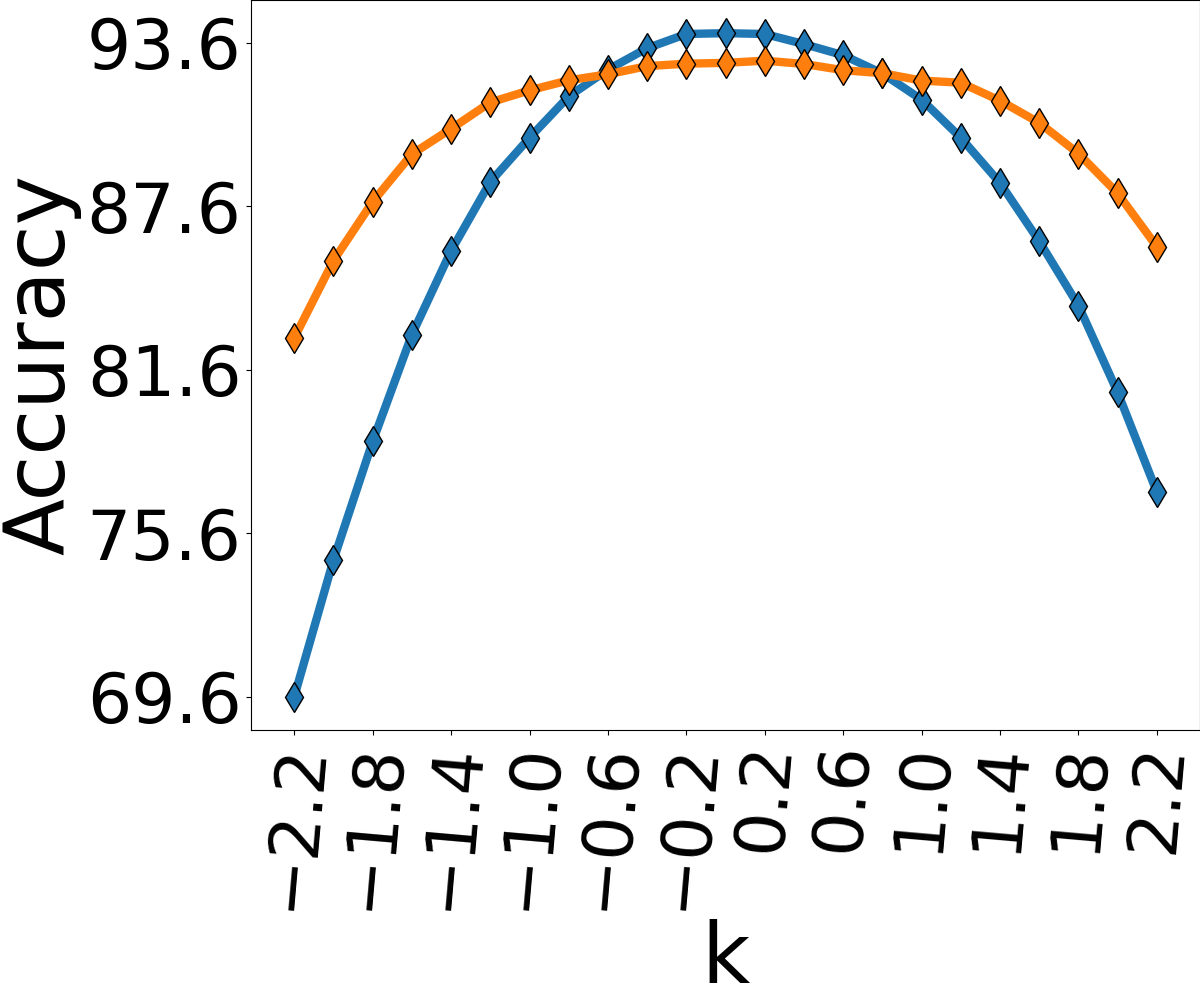}
    \includegraphics[width=\wide]{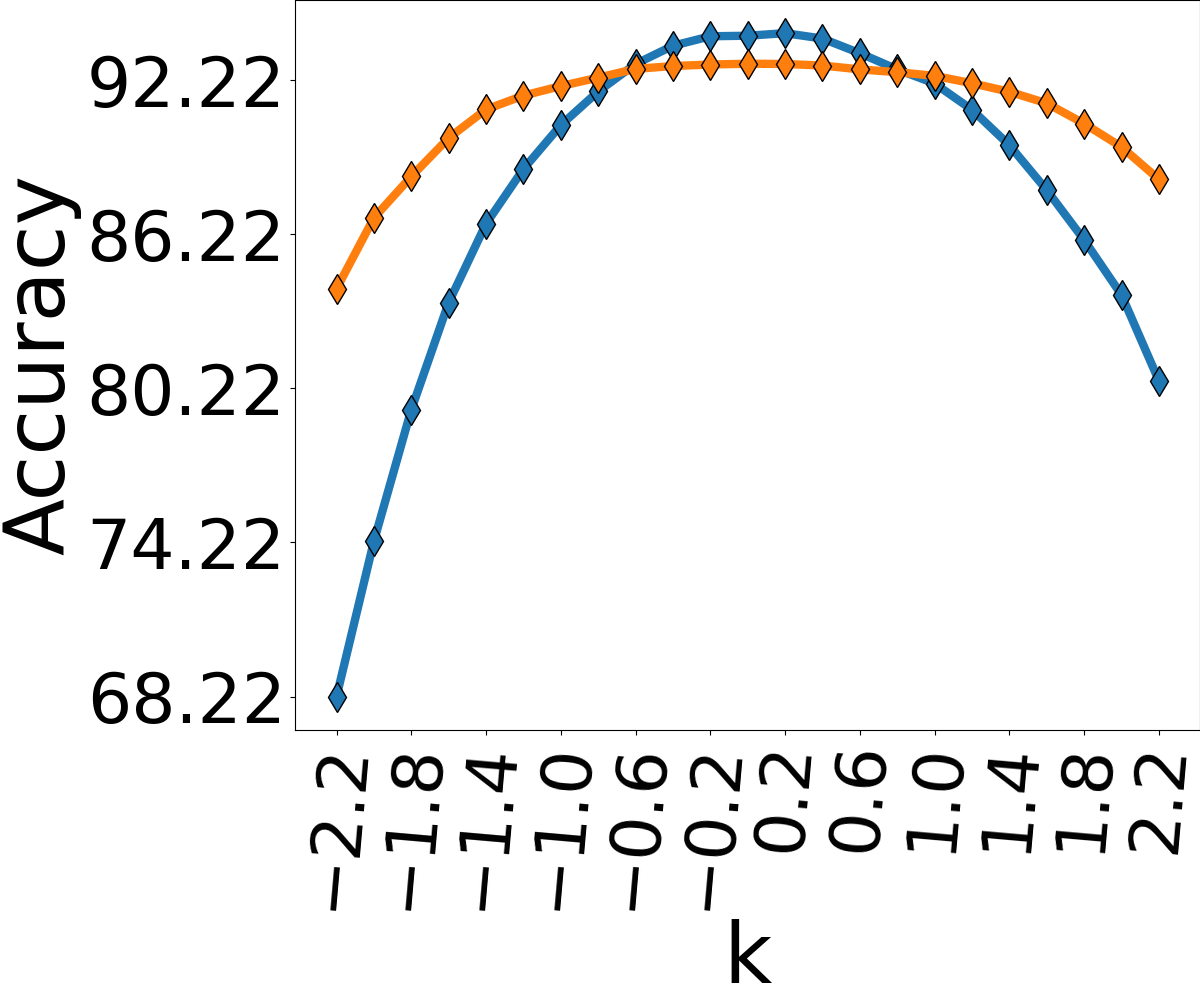}
    \includegraphics[width=\wide]{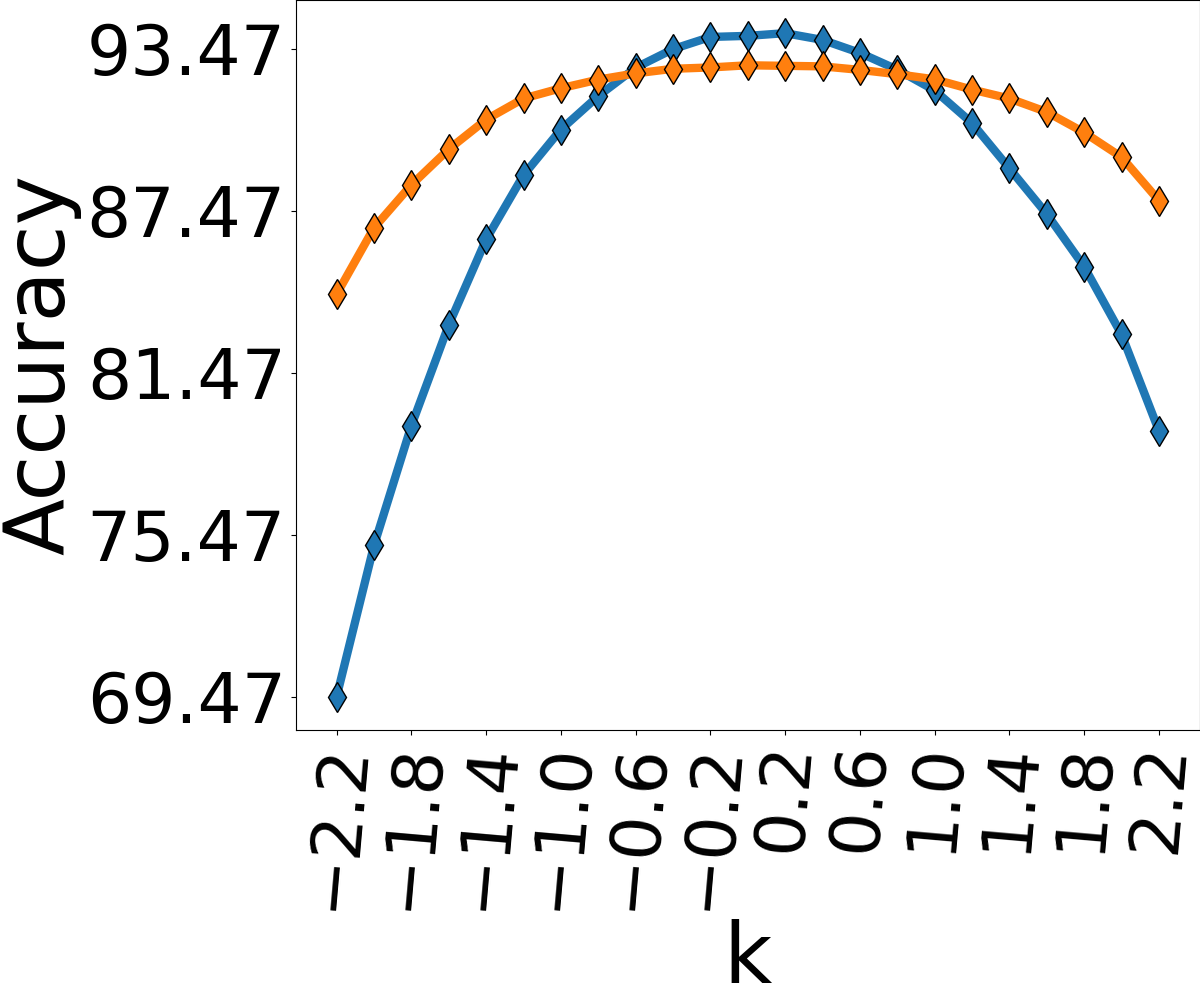}
    \includegraphics[width=\wide]{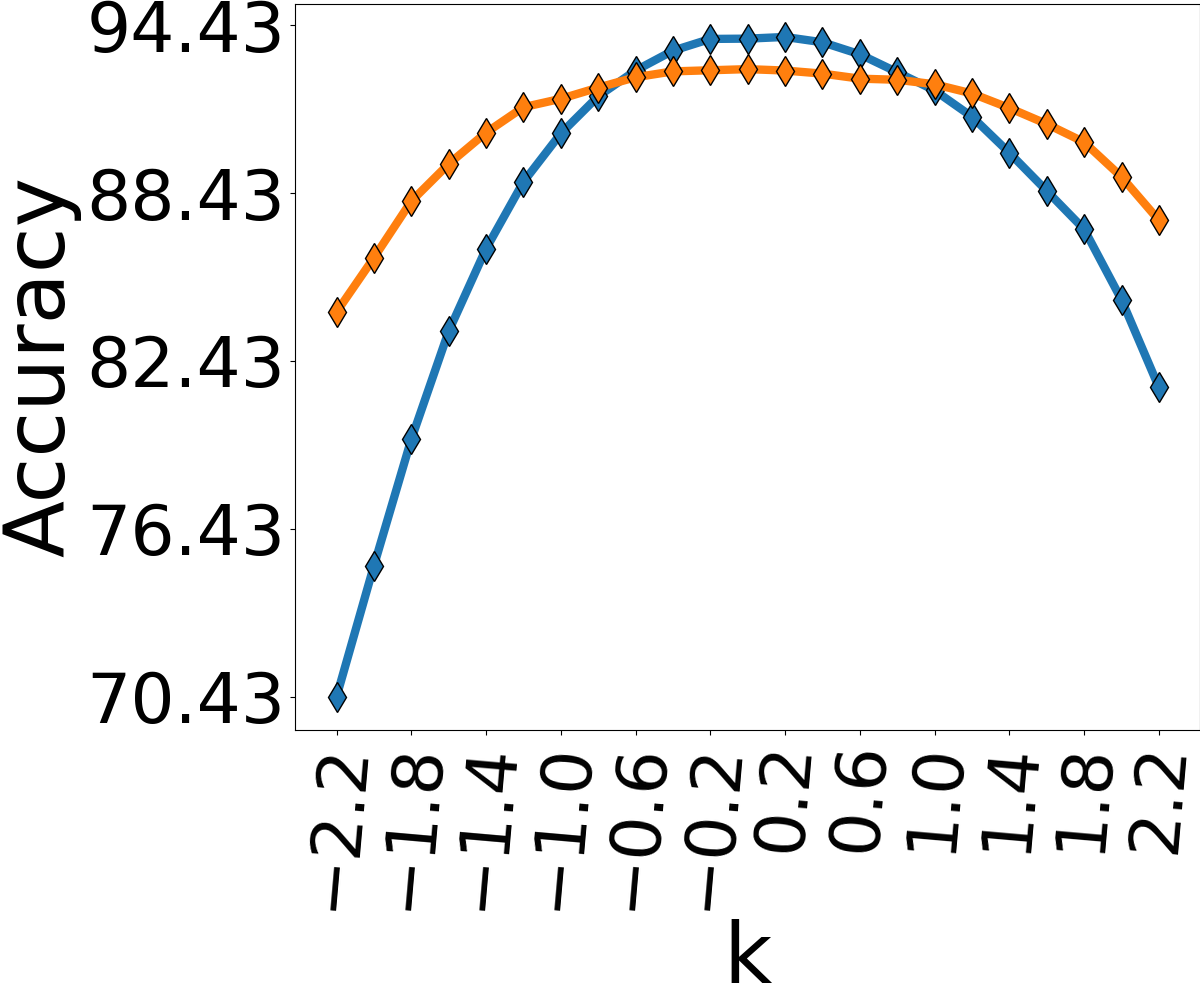}
    \includegraphics[width=\wide]{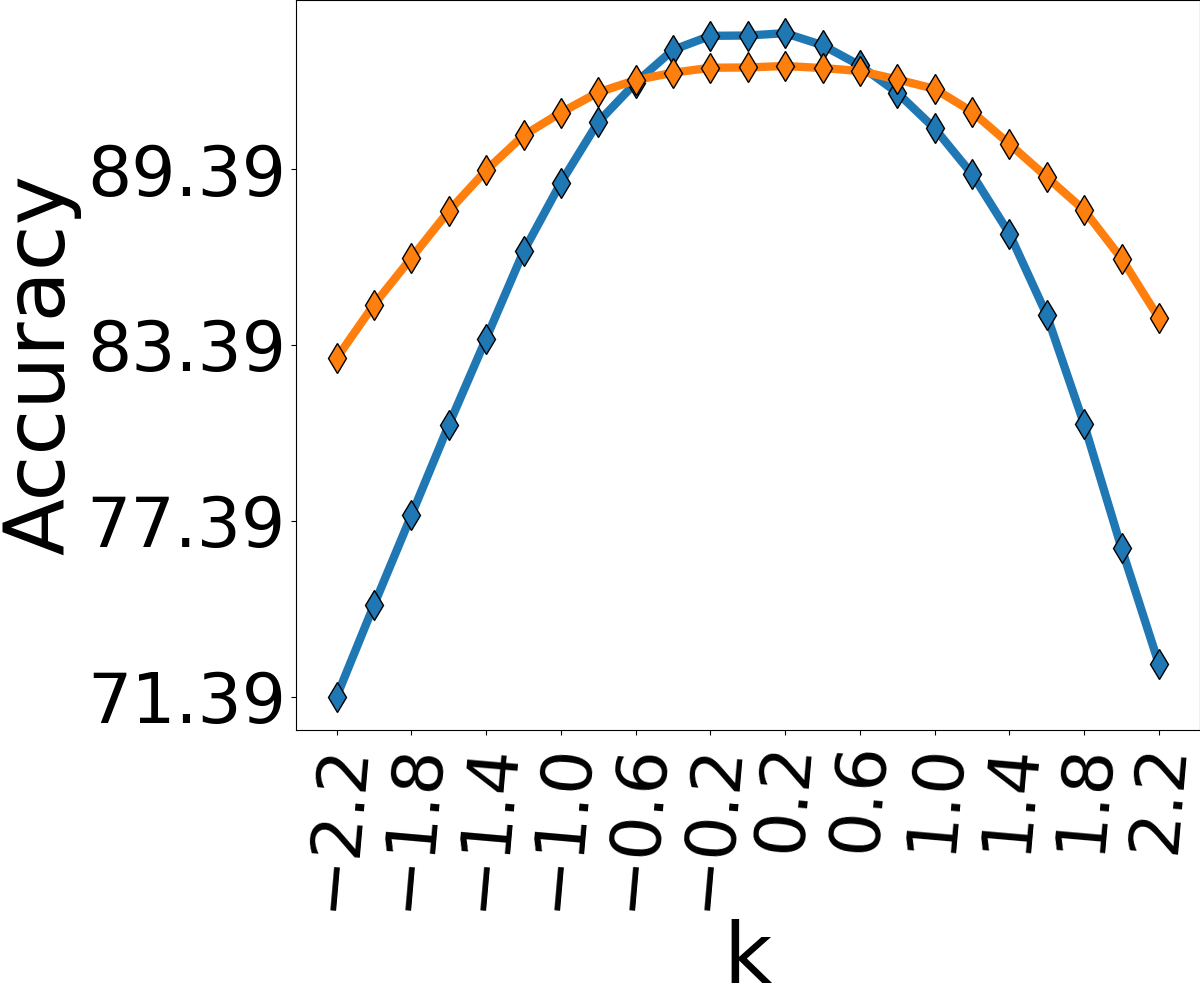}
    \includegraphics[width=\wide]{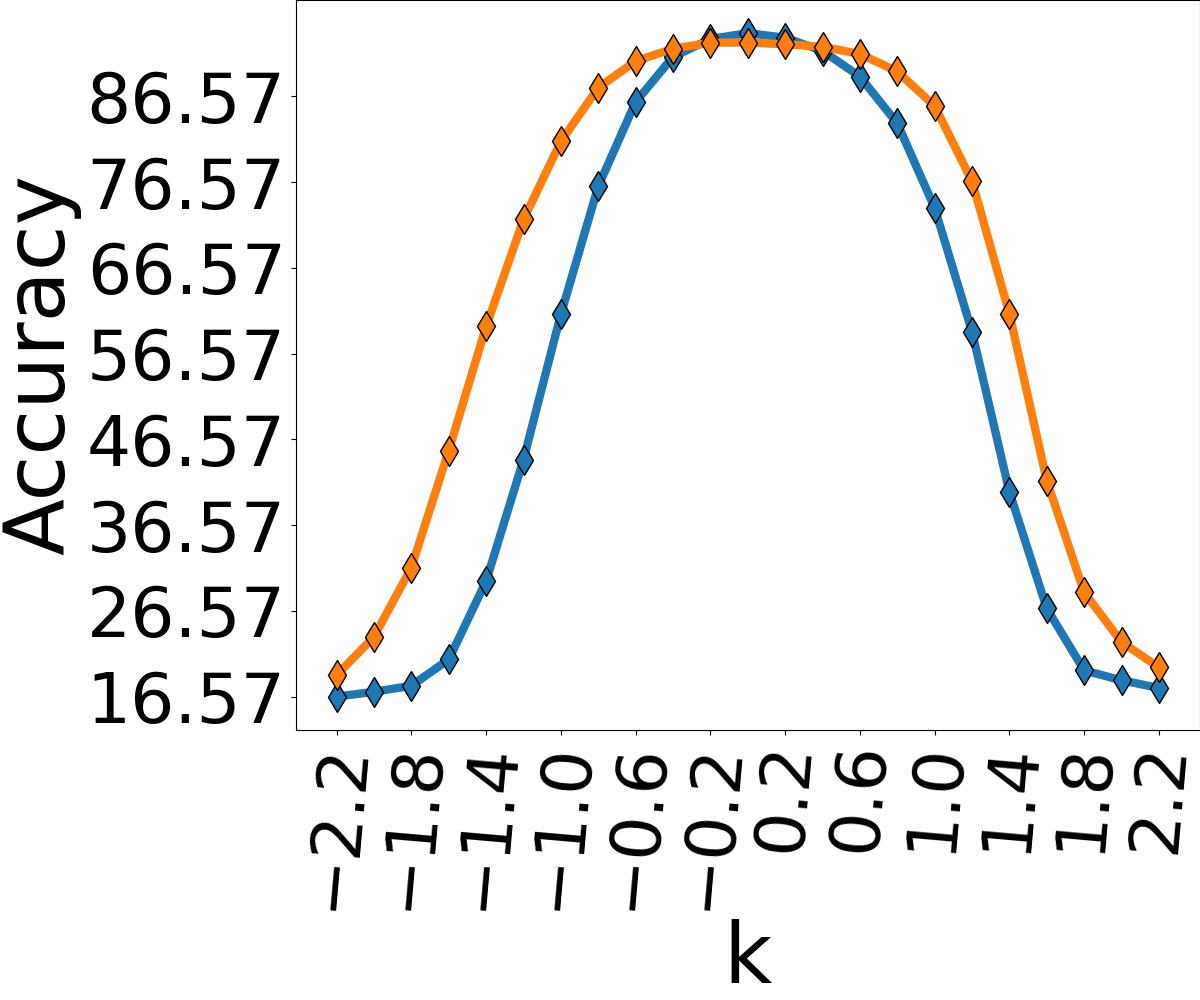}
    \includegraphics[width=\wide]{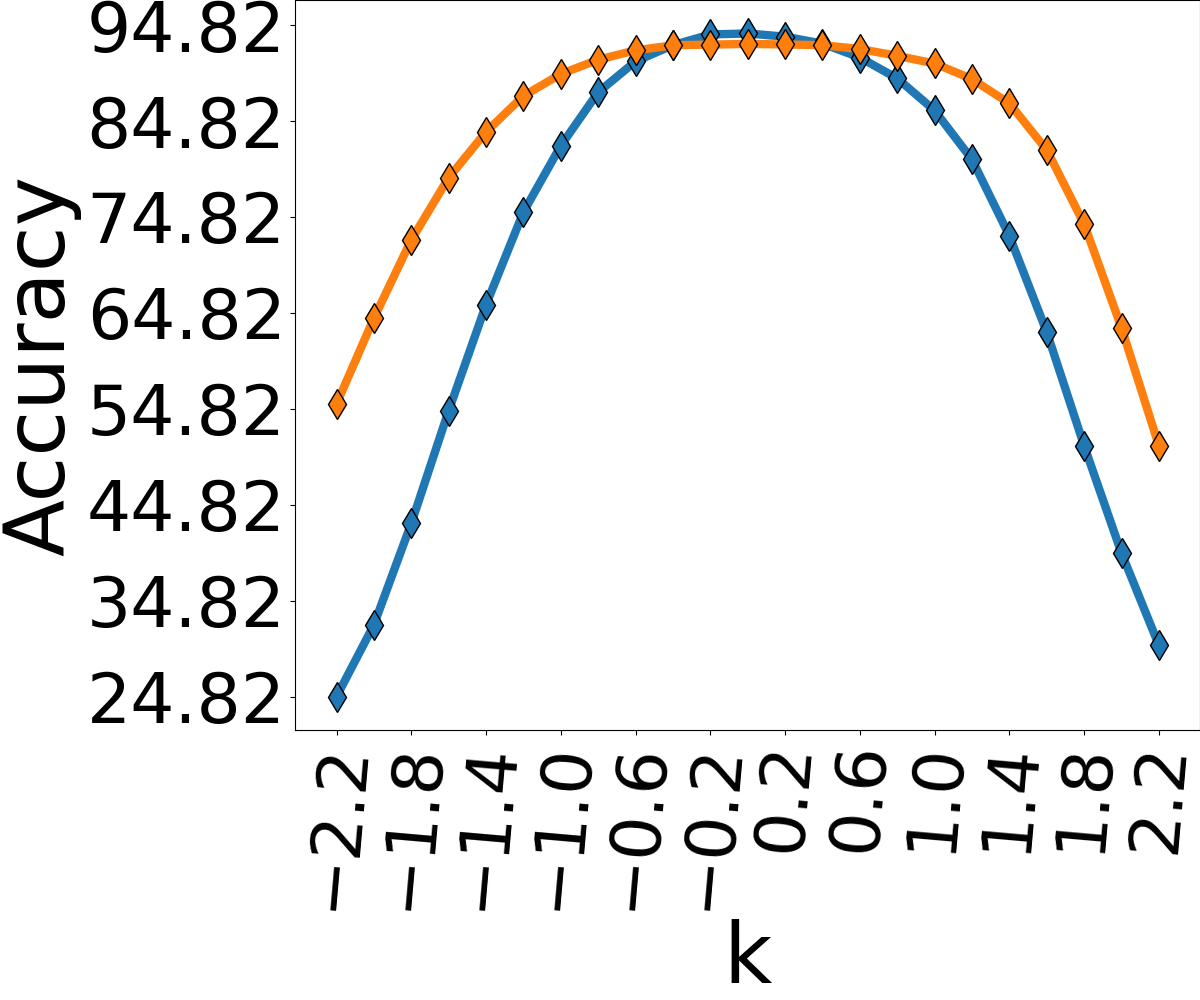}
    \includegraphics[width=\wide]{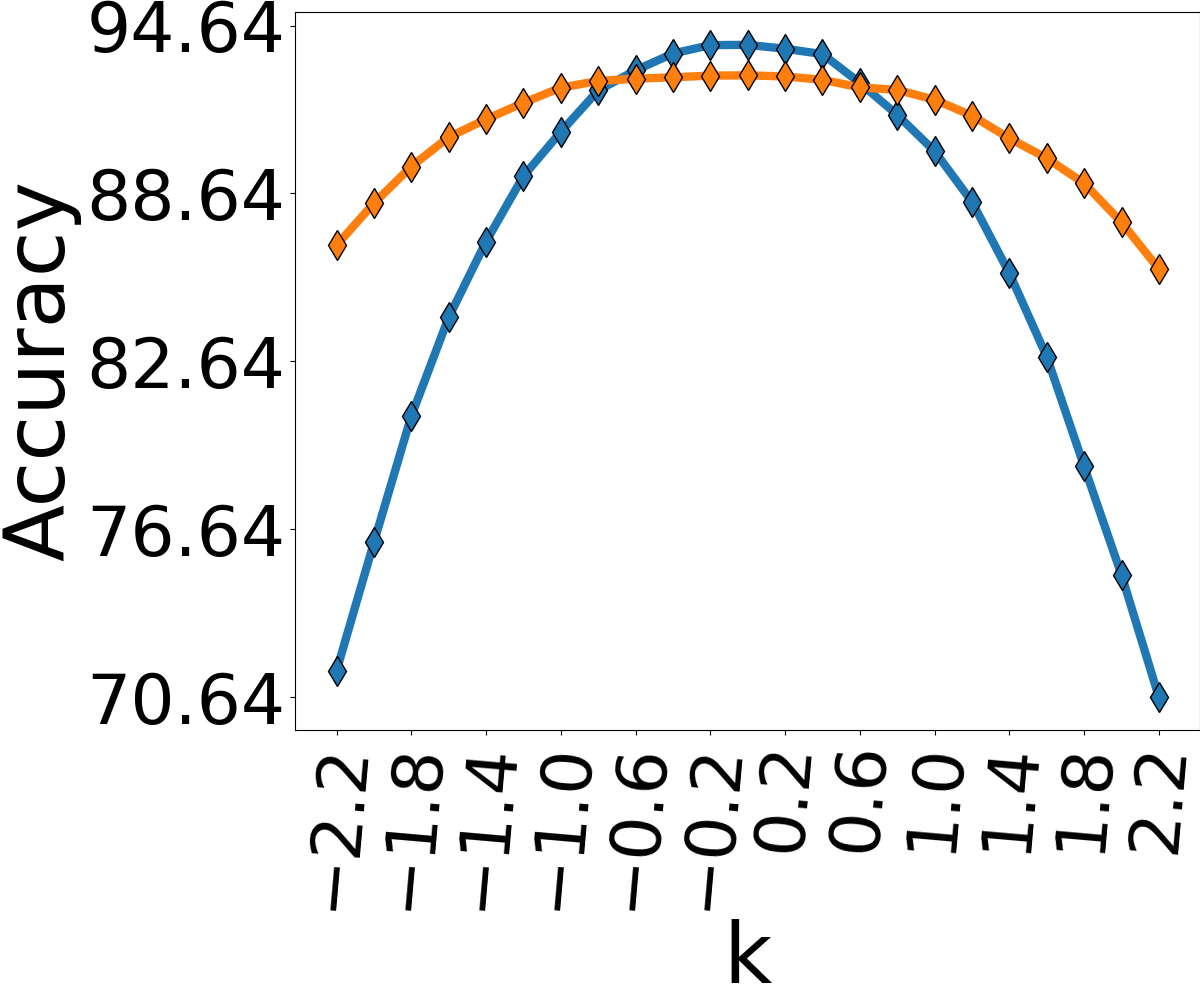}
    \includegraphics[width=\wide]{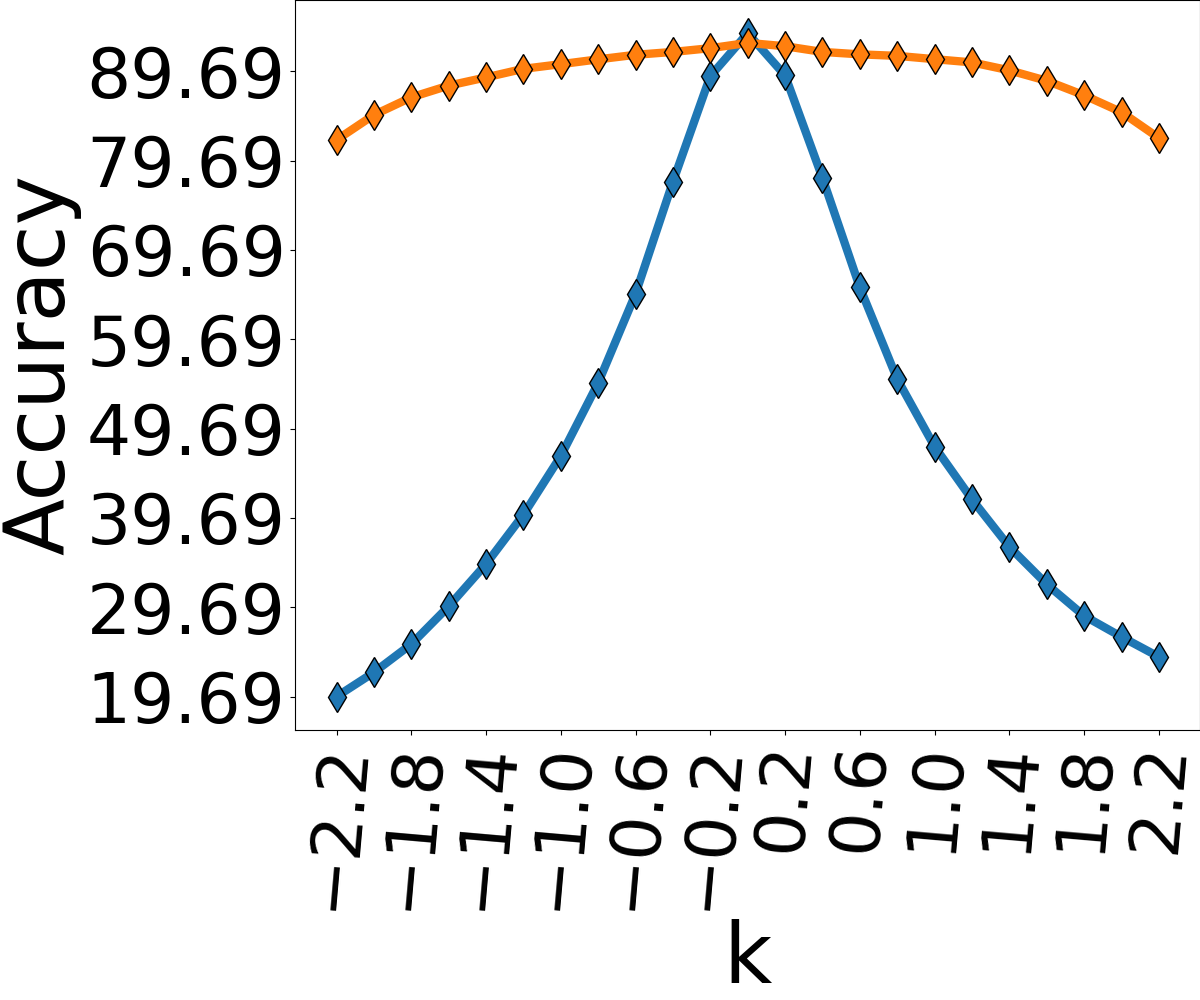}
    \includegraphics[width=\wide]{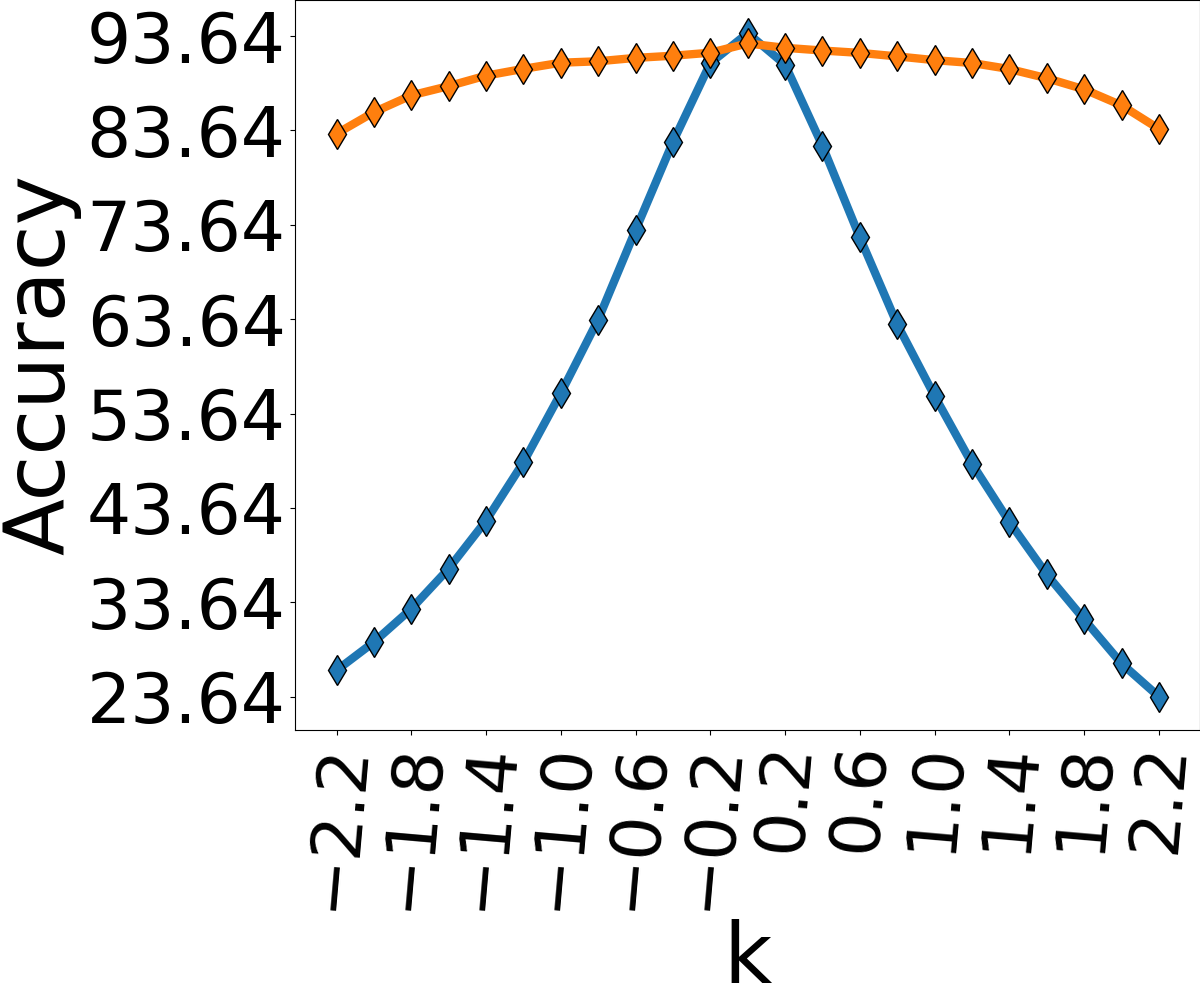}
    \includegraphics[width=\wide]{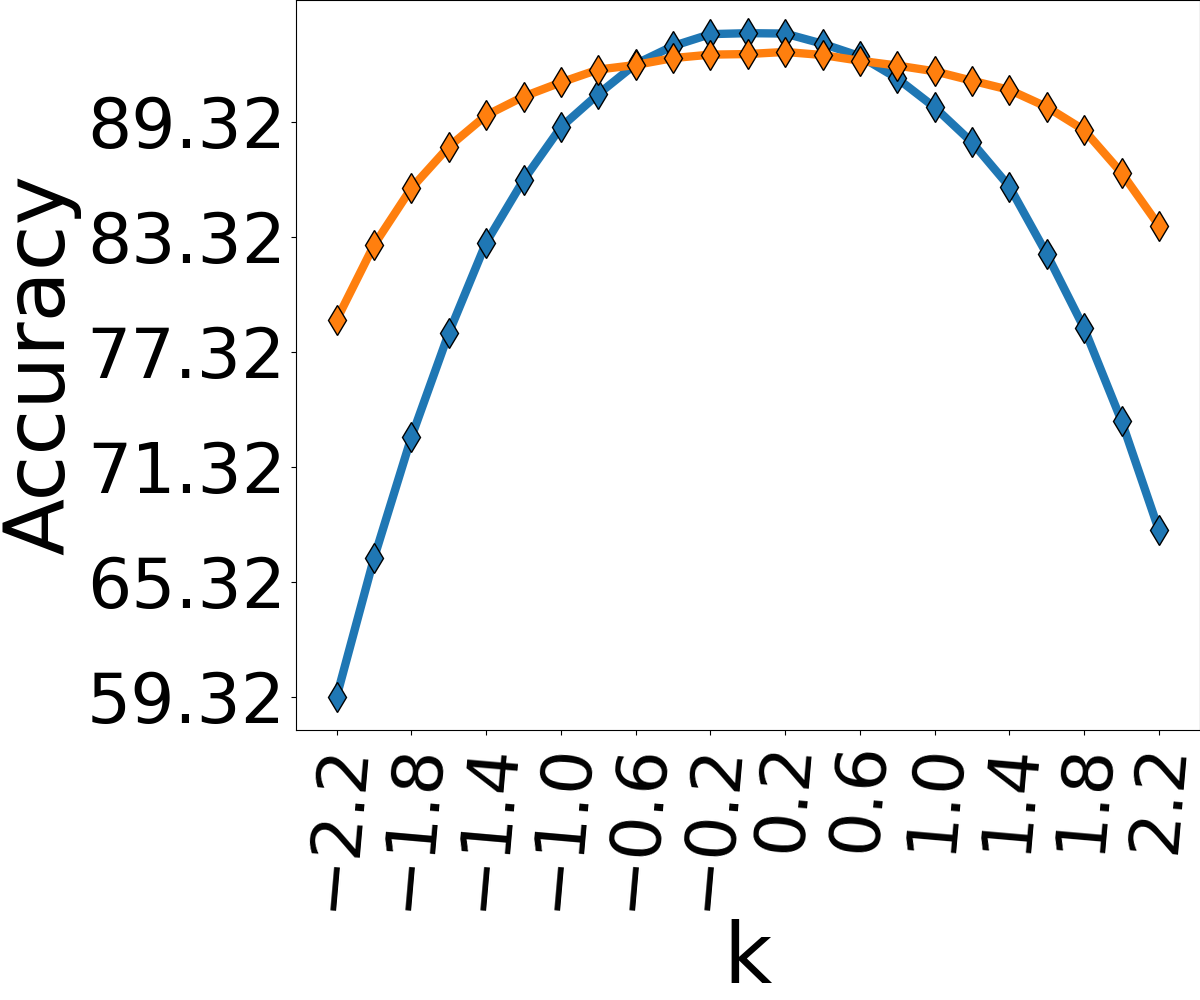}
    \includegraphics[width=\wide]{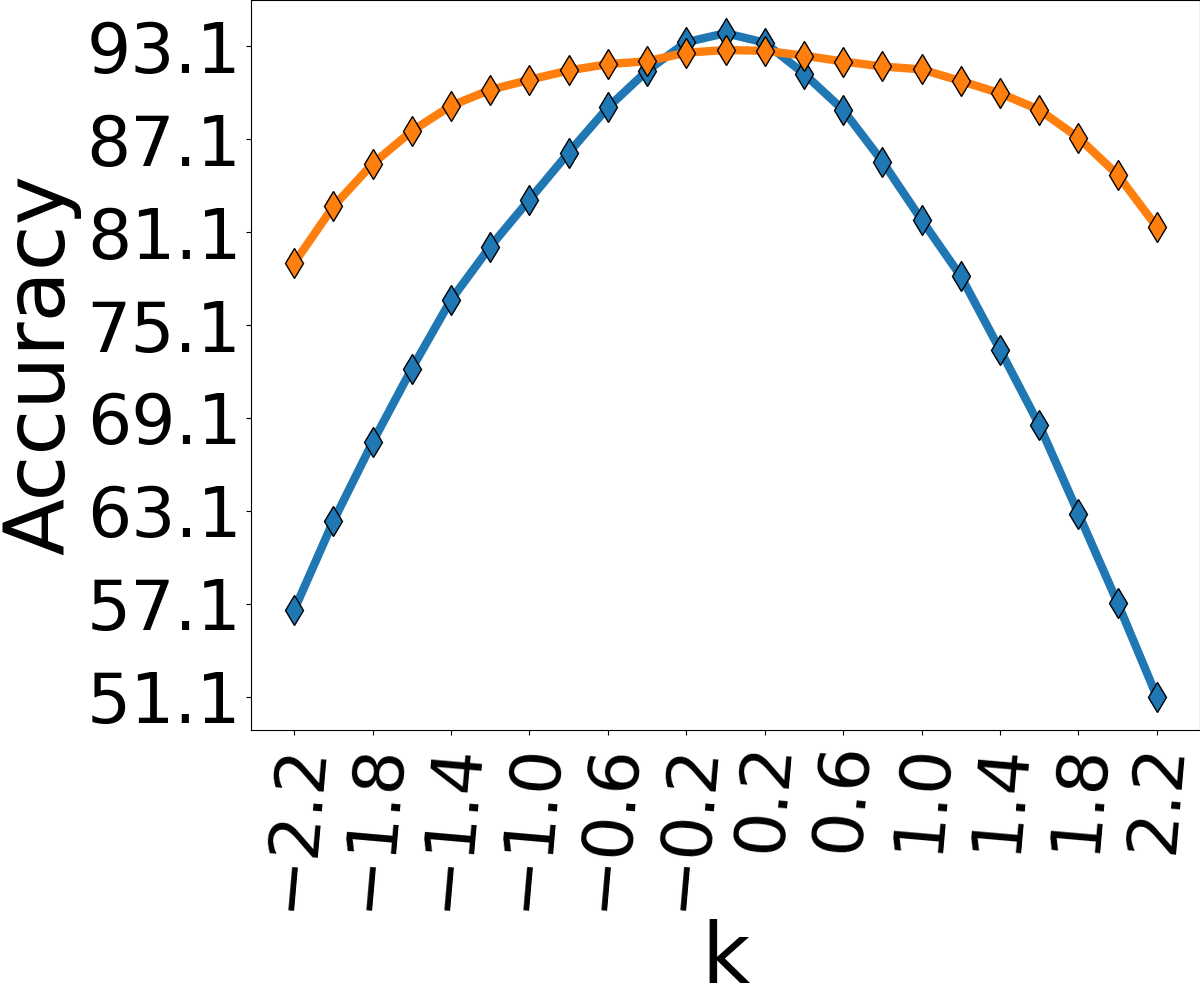}
    \caption{The results of the InceptionV3 model on the CIFAR10 dataset under different NUI attacks (test set). The \textcolor{blue}{blue} and \textcolor{orange}{orange} curves show the results of the model trained on the original training set and the NUI perturbed training set, respectively.}
    \label{fig:Cifar_inc}
\end{figure*}

\subsection{Qualitative Results}
The visual results for a sample image from the CIFAR10 dataset under different NUI attacks are shown using the ResNet18 model in \autoref{fig:prediction} following the predicted category with the probability of classification. The $1^{st}$ image in the $1^{st}$ row is an original dog image taken from the CIFAR10 dataset, and the model predicts it as a dog with very high probability. The $2^{nd}$ to $7^{th}$ images in the $1^{st}$ row and the $1^{st}$ to $6^{th}$ images in the $2^{nd}$ row represent samples generated using the $1^{st}$ to $12^{th}$ mask in the same order along with its predicted category with probability. Different values of NUI weight ($k$) are used with different masks. Note that when for negative $k$, the resultant image becomes darker and vice-versa. The images are misclassified with high probability under NUI attacks with $2^{nd}$, $3^{rd}$, $7^{th}$, $8^{th}$, $10^{th}$ and $12^{th}$ masks. Whereas, the probability of classification to correct class is decreased under other NUI attacks. It is evident from these results that almost all the images are visually perceptible to the original image with some amount of brightness or darkness, however, these images are either misclassified by a trained CNN model or confidence of classification decreases. We refer to the Supplementary materials to observe the impact of the NUI attack on image pixel value distributions.

\begin{table*}[!t]
    \caption{The \% decrease in the accuracy of models on various test sets under NUI attack. Results are reported for $k = -1.4$}
    \centering
    \newcommand\wide{0.041\textwidth}
    \begin{tabular}{|p{0.068\textwidth}|p{0.105\textwidth}|    p{\wide}|p{\wide}|p{\wide}|p{\wide}|p{\wide}|p{\wide}|p{\wide}|p{\wide}|p{\wide}|p{\wide}|p{\wide}|p{\wide}|p{\wide}|}
    \hline
    Model & Dataset & M1 & M2 & M3 & M4 & M5 & M6 & M7 & M8 & M9 & M10 & M11 & M12 \\
    \hline
    VGG16 & CIFAR10 & $9.19$ & $8.91$ & $9.33$ & $8.5$ & $13.9$ & $72.29$ & $35.47$ & $19.63$ & $65.63$ & $51.44$ & $12.93$ & $18.49$ \\
    \hline
    VGG16 & TinyImageNet & $31.22$ & $29.74$ & $30.7$ & $31.09$ & $54.4$ & $87.3$ & $71.17$ & $49.9$ & $88.61$ & $75.31$ & $38.72$ & $51.72$ \\
    \hline
    VGG19 & CalTech256 & $22.12$ & $21.56$ & $21.66$ & $21.17$ & $43.86$ & $83.8$ & $61.71$ & $39.58$ & $87.9$ & $78.26$ & $29.45$ & $43.02$ \\
    \hline
    ResNet18 & CIFAR10 & $8.55$ & $7.23$ & $7.4$ & $7.2$ & $13.21$ & $66.7$ & $32.07$ & $9.75$ & $54.65$ & $50.93$ & $11$ & $13.55$ \\
    \hline
    ResNet18 & TinyImageNet & $32.53$ & $31.4$ & $30.01$ & $30.67$ & $53.67$ & $89.89$ & $73.89$ & $53.29$ & $85.43$ & $72.57$ & $39.08$ & $46.97$ \\
    \hline
    ResNet18 & CalTech256 & $28.47$ & $27.09$ & $28.04$ & $25.76$ & $47.17$ & $86.72$ & $67.27$ & $43.75$ & $87.57$ & $81.78$ & $36.17$ & $46.84$ \\
    \hline
    Mobilenet & CIFAR10 & $15.10$ & $14.22$ & $13.84$ & $14.23$ & $17.04$ & $73.59$ & $43.63$ & $13.07$ & $61.91$ & $59.87$ & $19.21$ & $26.87$ \\
    \hline
    Inception & CIFAR10 & $8.45$ & $7.79$ & $8.02$ & $7.99$ & $11.00$ & $68.02$ & $30.10$ & $7.48$ & $63.24$ & $54.98$ & $11.63$ & $18.32$ \\
    \hline
    \end{tabular}
    \label{tab:perc1}
\end{table*}

\begin{figure*}[!t]
  \centering
  \begin{minipage}[t]{0.13\textwidth}
    \centering
    \includegraphics[width=\textwidth]{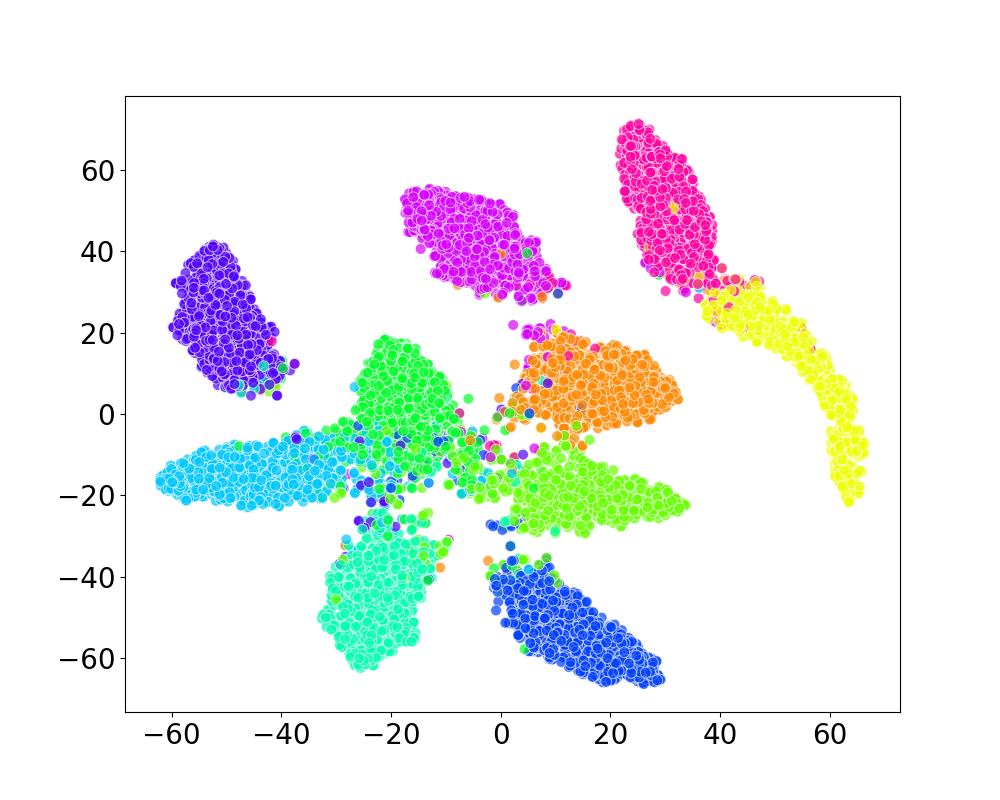}
      Original Test Set
  \end{minipage}%
  \hfill
  \begin{minipage}[t]{0.13\textwidth}
    \centering
    \includegraphics[width=\textwidth]{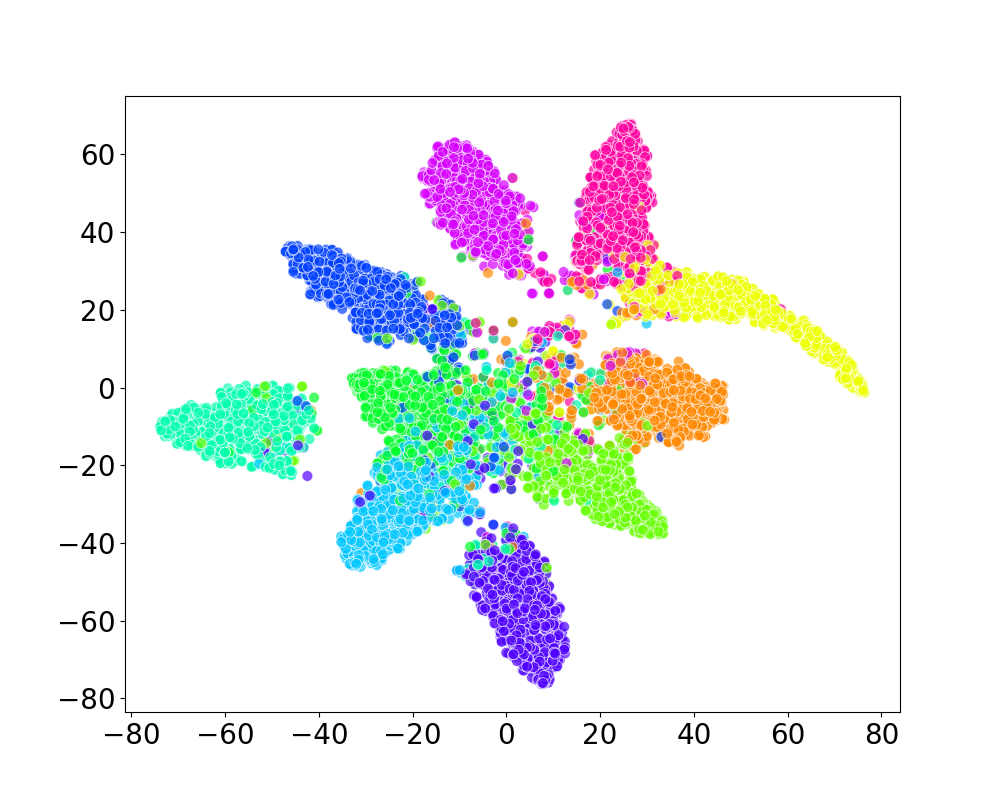}
    $\text{M}1$, $k=-1.2$
  \end{minipage}%
  \hfill
  \begin{minipage}[t]{0.13\textwidth}
    \centering
    \includegraphics[width=\textwidth]{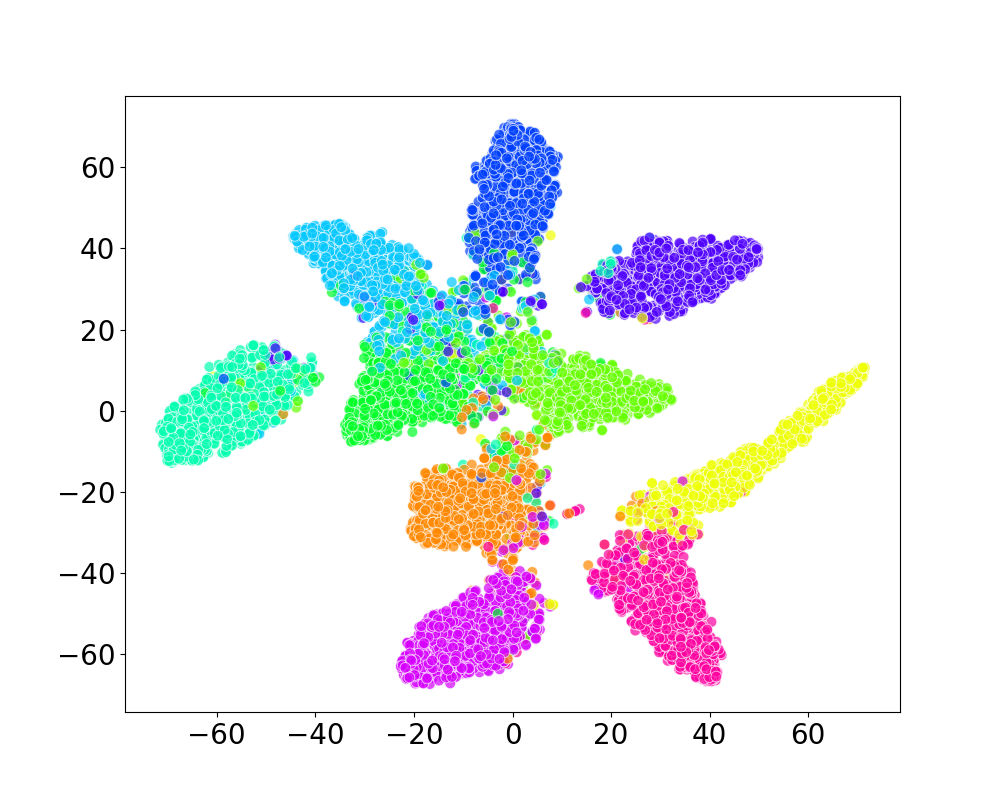}
    $\text{M}2$, $k=1.2$
  \end{minipage}
  \hfill
  \begin{minipage}[t]{0.13\textwidth}
    \centering
    \includegraphics[width=\textwidth]{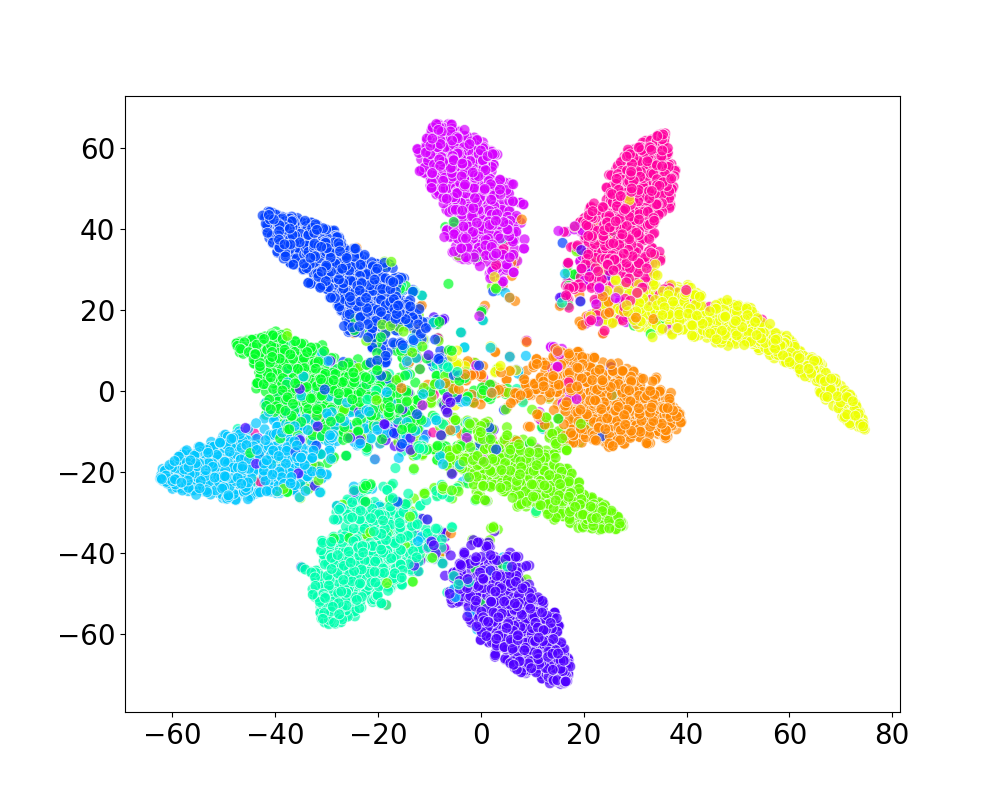}
    $\text{M}3$, $k=-1.2$
  \end{minipage}
  \hfill
  \begin{minipage}[t]{0.13\textwidth}
    \centering
    \includegraphics[width=\textwidth]{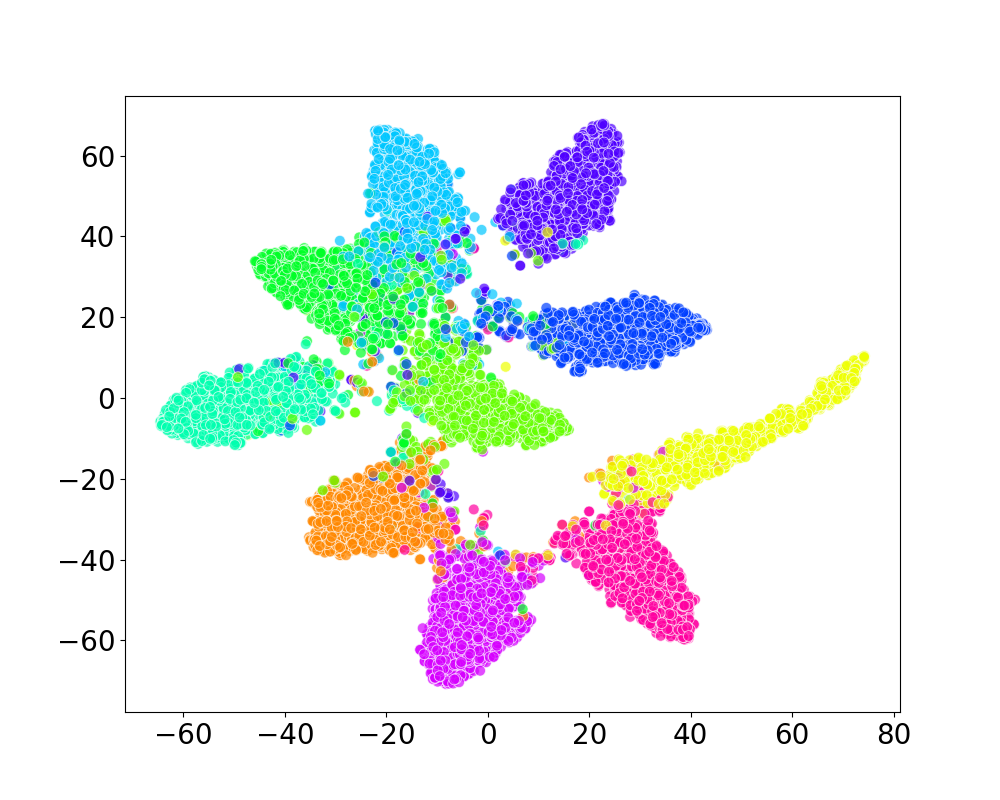}
    $\text{M}4$, $k=1.2$
  \end{minipage}
  \hfill
  \begin{minipage}[t]{0.13\textwidth}
    \centering
    \includegraphics[width=\textwidth]{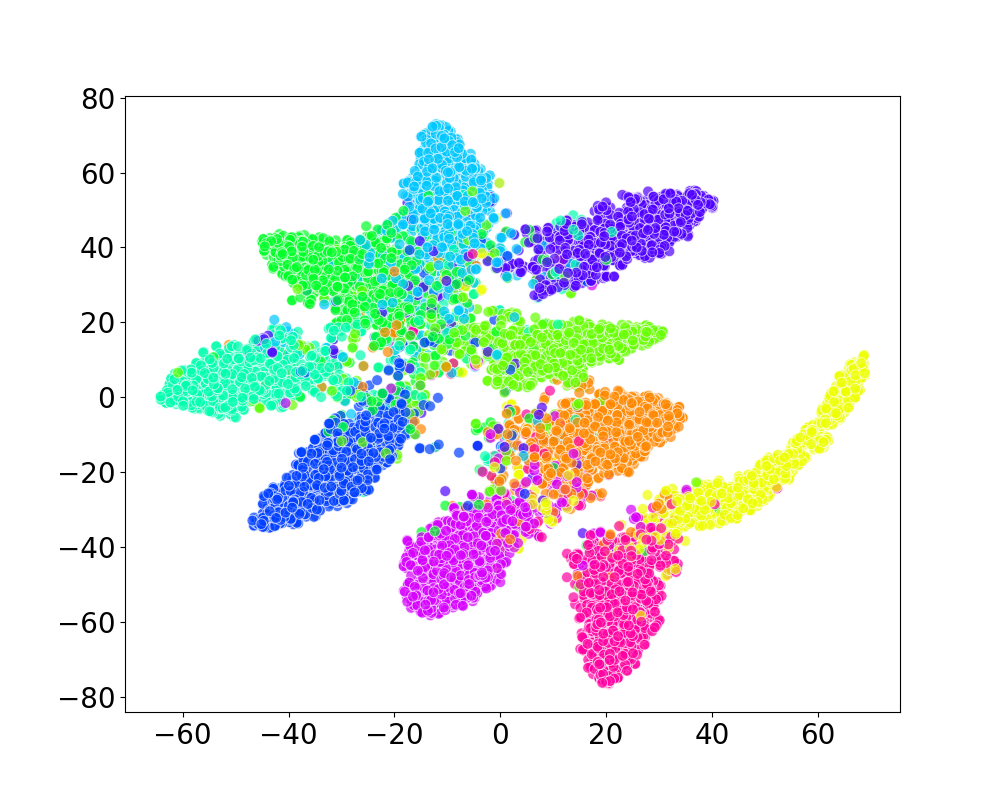}
    $\text{M}5$, $k=-1.2$
  \end{minipage}
  \hfill
  \begin{minipage}[t]{0.13\textwidth}
    \centering
    \includegraphics[width=\textwidth]{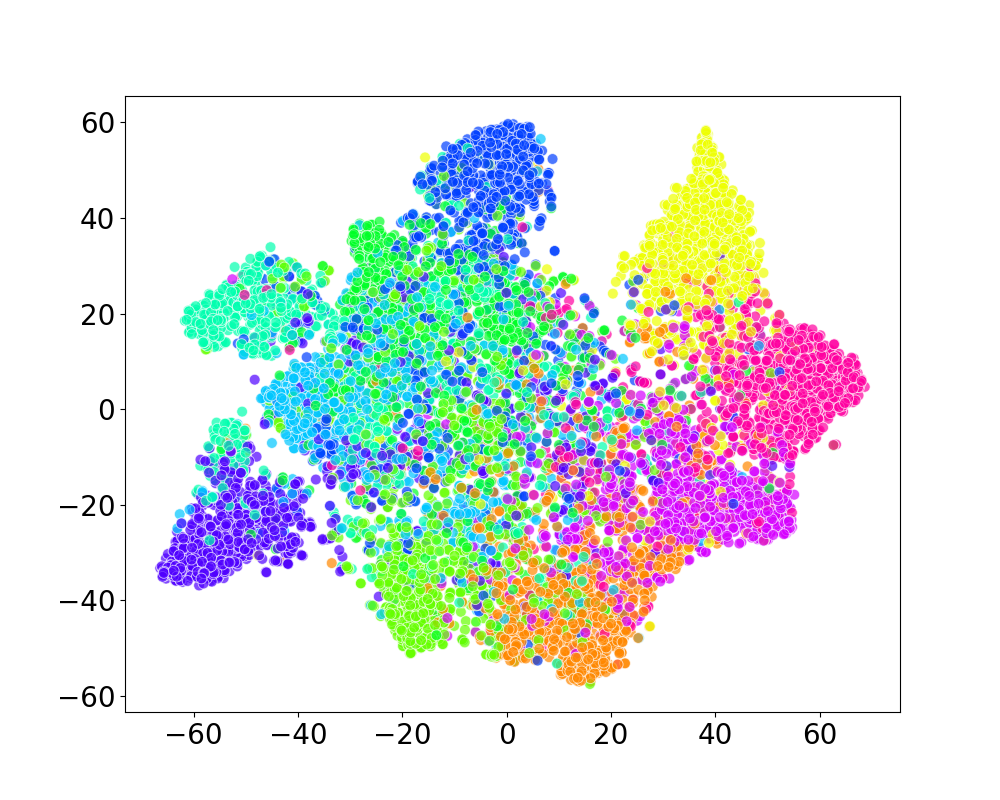}
    $\text{M}6$, $k=1.2$
  \end{minipage}
  \hfill
  \vspace{0.1cm}
  \begin{minipage}[t]{0.13\textwidth}
    \centering
    \includegraphics[width=\textwidth]{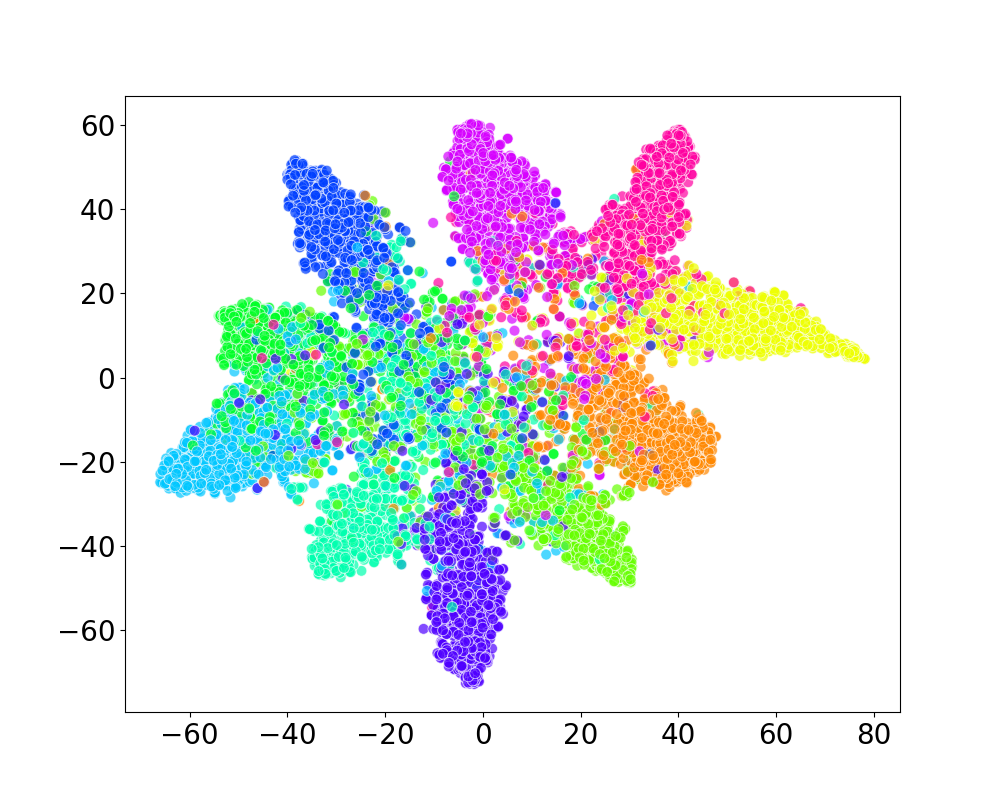}
    $\text{M}7$, $k=-1.2$
  \end{minipage}
  \hfill
  \begin{minipage}[t]{0.13\textwidth}
    \centering
    \includegraphics[width=\textwidth]{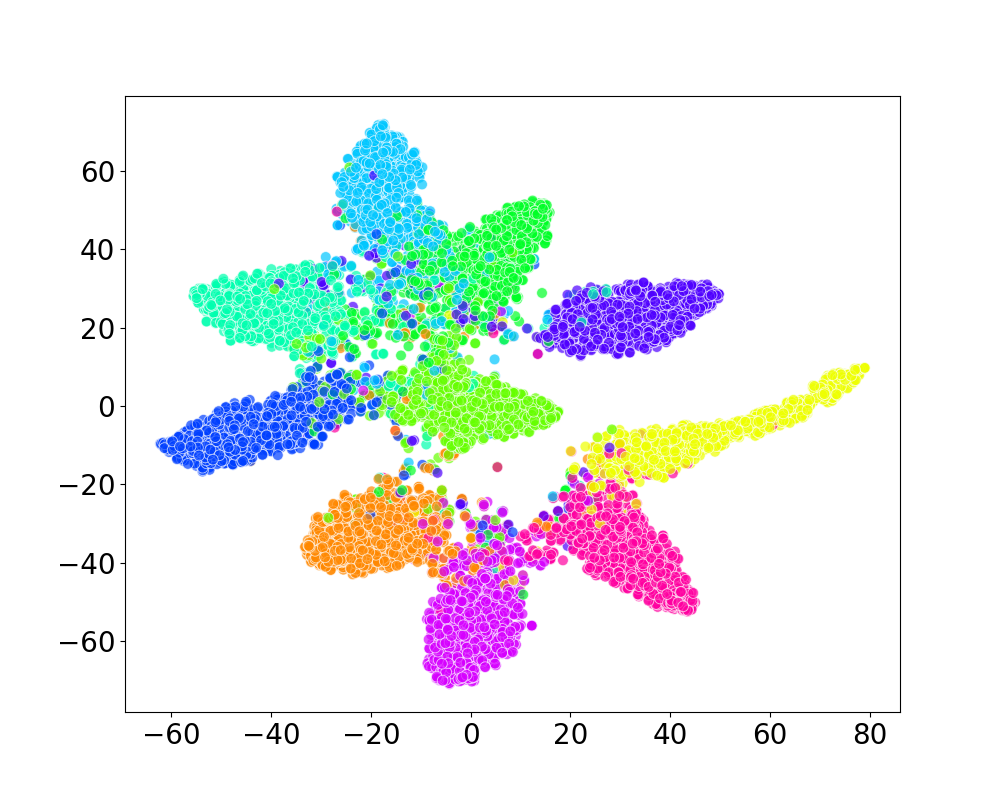}
    $\text{M}8$, $k=1.2$
  \end{minipage}
  \hfill
  \begin{minipage}[t]{0.13\textwidth}
    \centering
    \includegraphics[width=\textwidth]{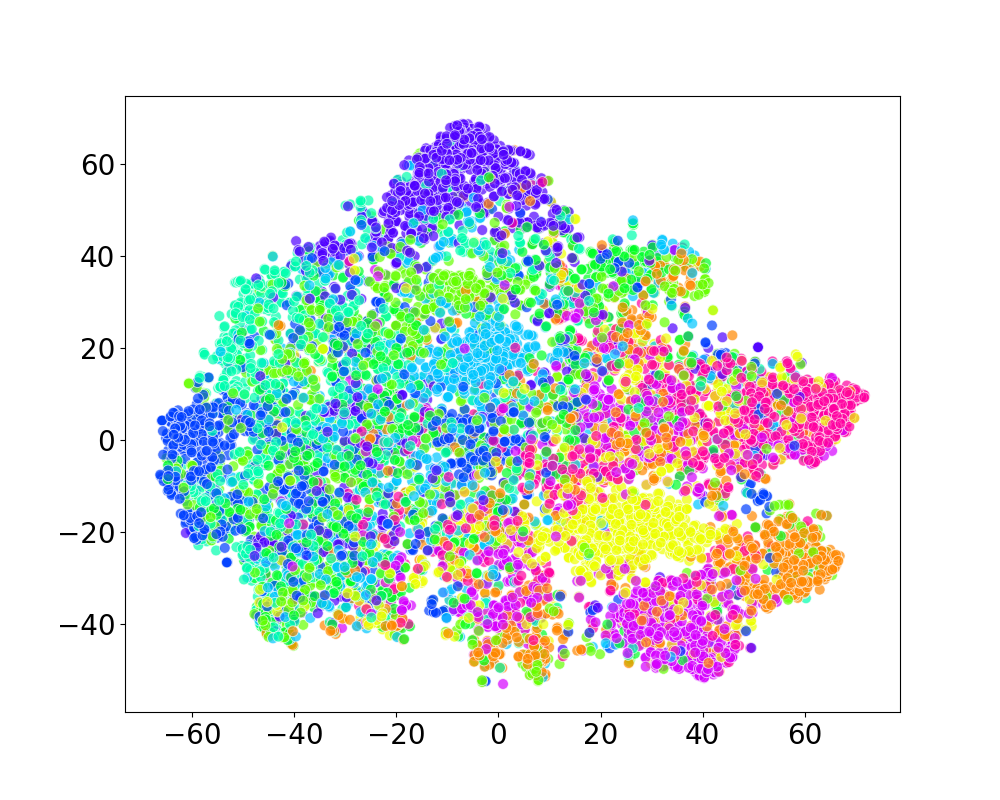}
    $\text{M}9$, $k=-1.2$
  \end{minipage}
  \hfill
  \begin{minipage}[t]{0.13\textwidth}
    \centering
    \includegraphics[width=\textwidth]{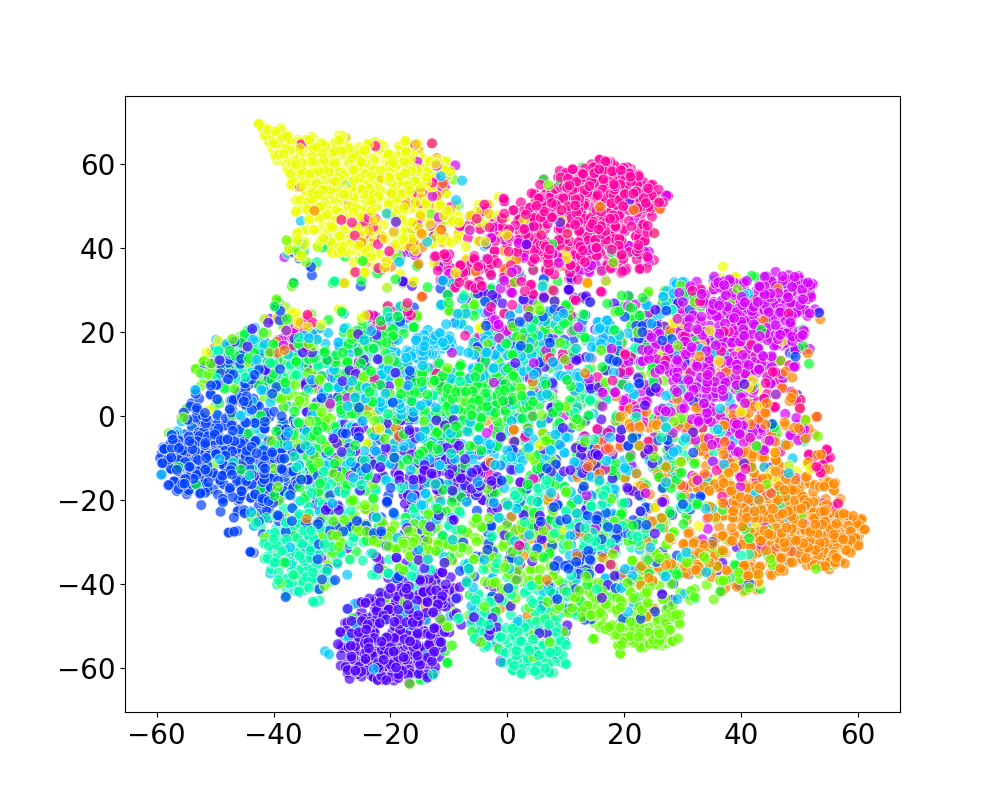}
    $\text{M}10$, $k=1.2$
  \end{minipage}
  \hfill
  \begin{minipage}[t]{0.13\textwidth}
    \centering
    \includegraphics[width=\textwidth]{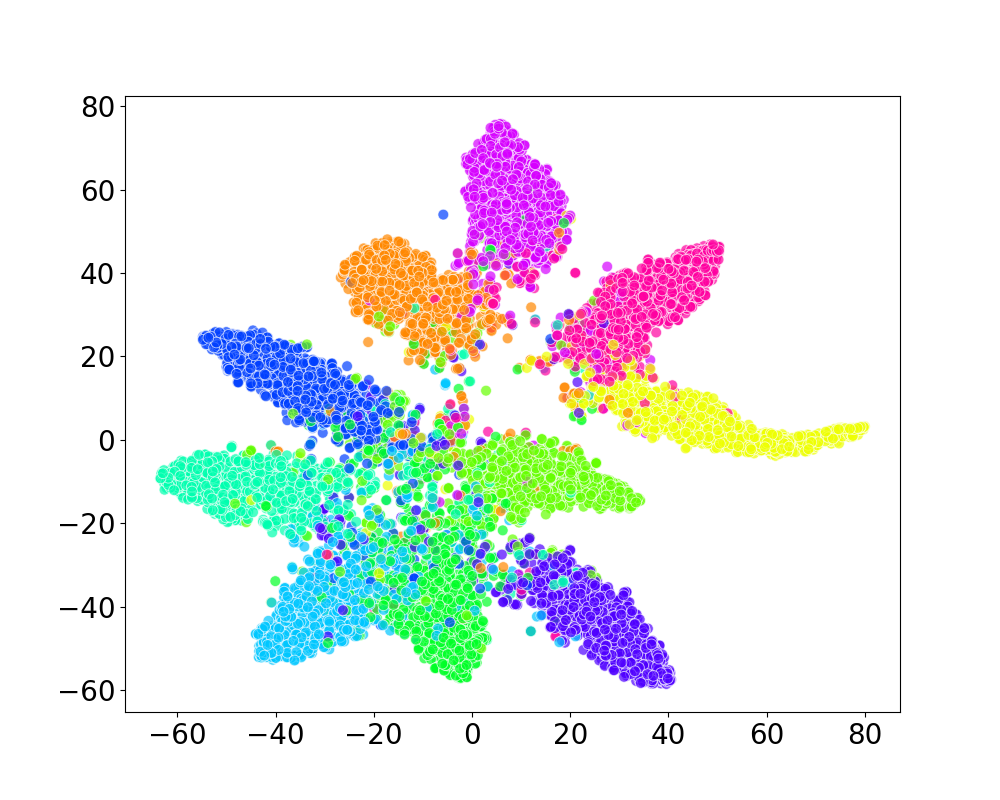}
    $\text{M}11$, $k=-1.2$
  \end{minipage}
  \hfill
  \begin{minipage}[t]{0.13\textwidth}
    \centering
    \includegraphics[width=\textwidth]{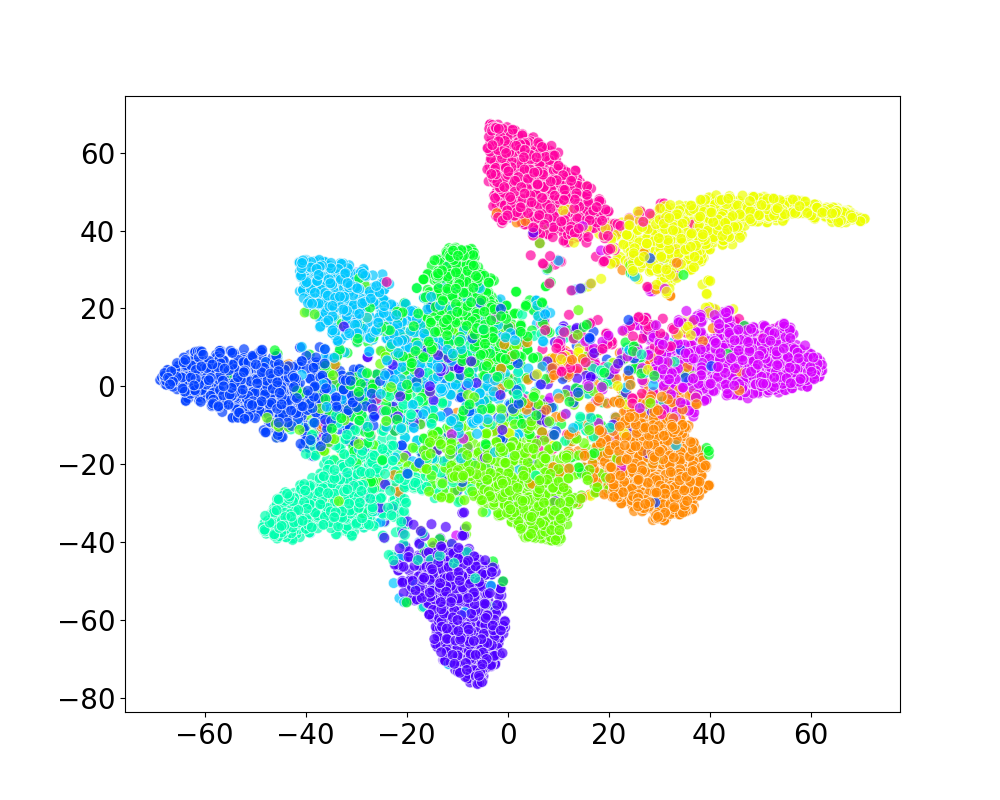}
    $\text{M}12$, $k=1.2$
  \end{minipage}

  \caption{Shows the t-SNE plot on CIFAR10 test set using InceptionV3 model for $12$ masks and $1.2$ and $-1.2$ values of ($k$). Here, $M_i$ refers to $i^{th}$ mask.}
   \label{fig:TSNE-inc}
\end{figure*}

\begin{figure*}[!t]
  \centering
  \begin{minipage}[t]{0.13\textwidth}
    \centering
    \includegraphics[width=\textwidth]{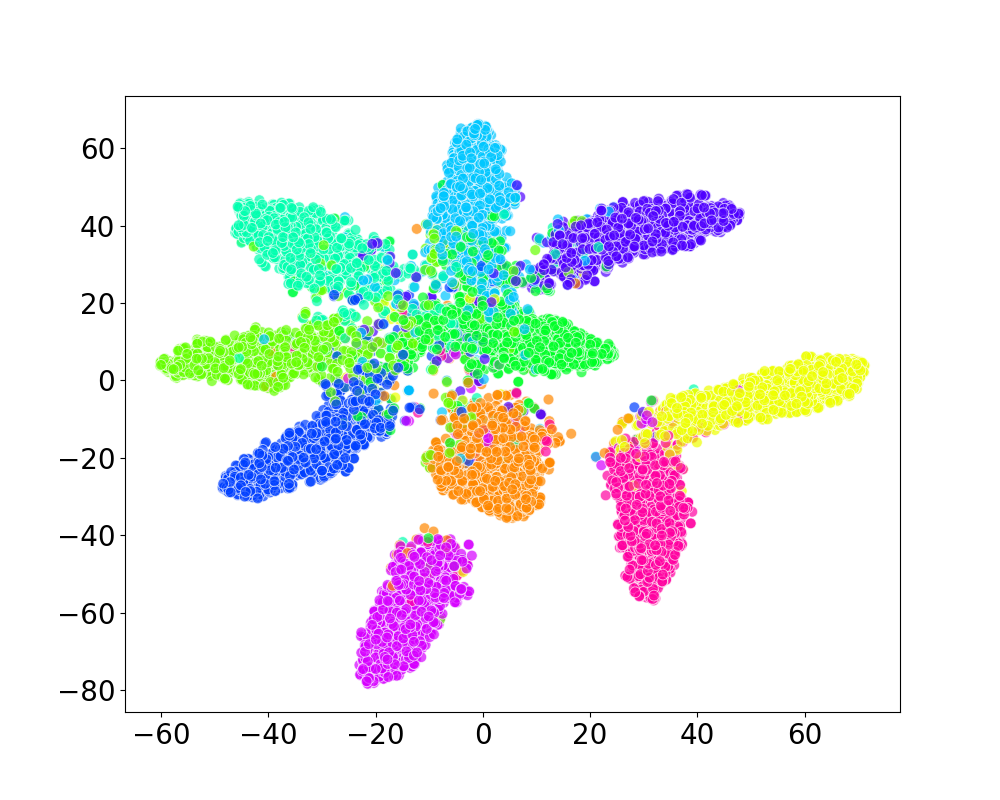}
      Original Test Set
  \end{minipage}%
  \hfill
  \begin{minipage}[t]{0.13\textwidth}
    \centering
    \includegraphics[width=\textwidth]{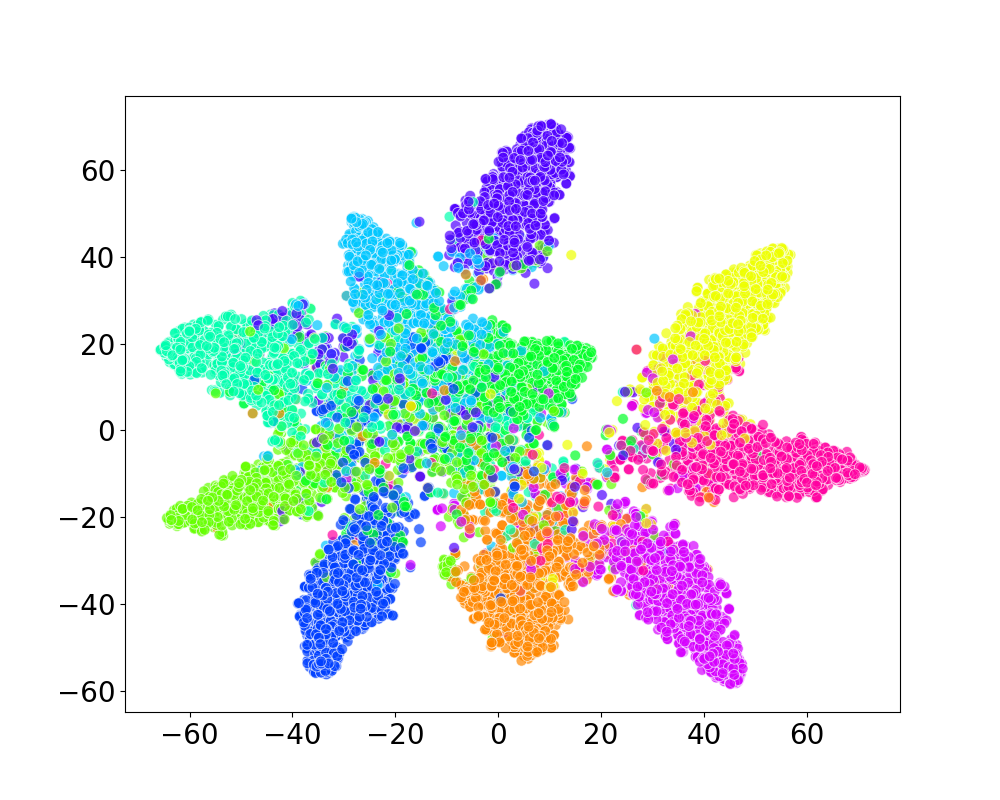}
    $\text{M}1$, $k=-1.2$
  \end{minipage}%
  \hfill
  \begin{minipage}[t]{0.13\textwidth}
    \centering
    \includegraphics[width=\textwidth]{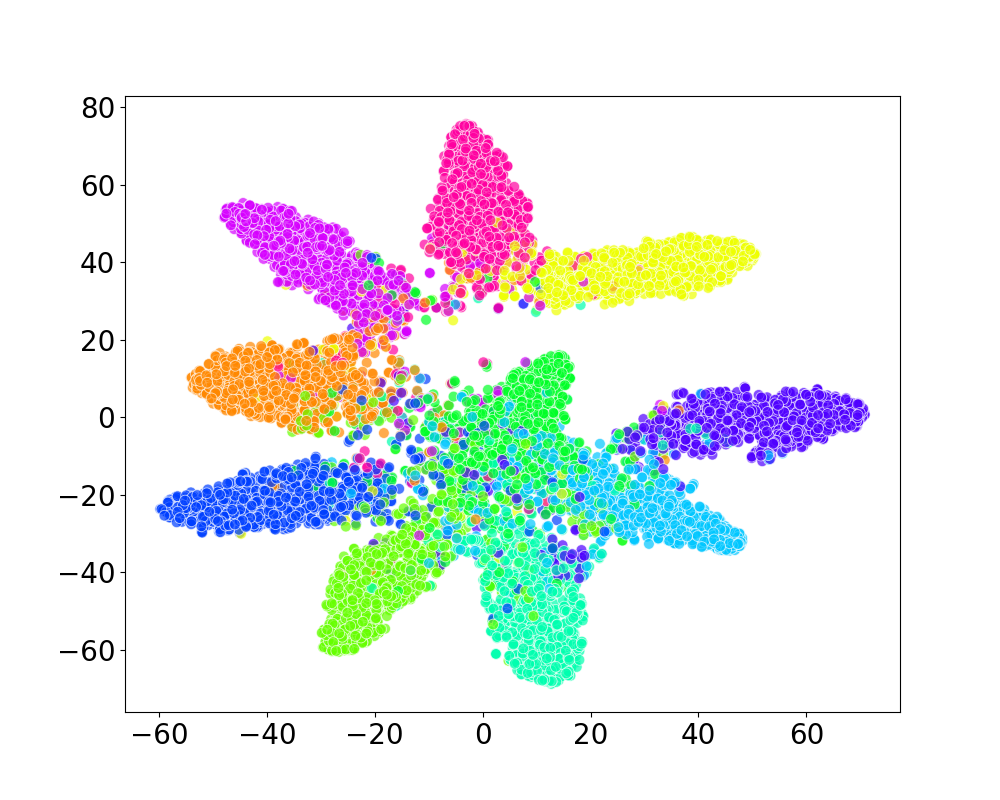}
    $\text{M}2$, $k=1.2$
  \end{minipage}
  \hfill
  \begin{minipage}[t]{0.13\textwidth}
    \centering
    \includegraphics[width=\textwidth]{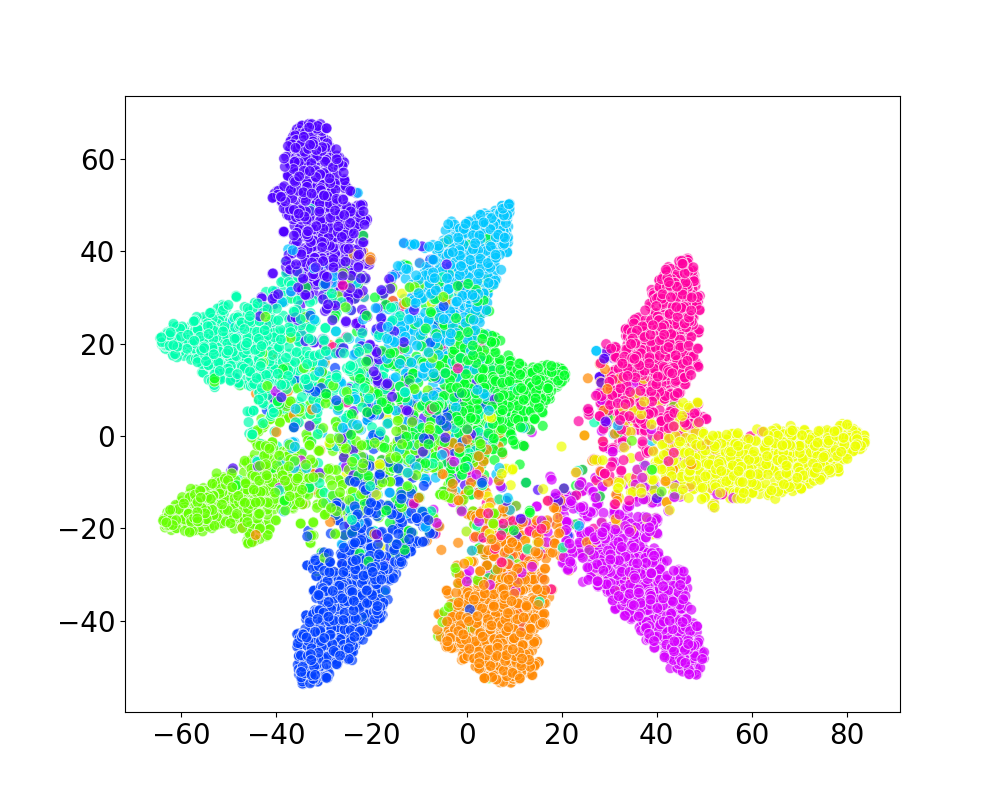}
    $\text{M}3$, $k=-1.2$
  \end{minipage}
  \hfill
  \begin{minipage}[t]{0.13\textwidth}
    \centering
    \includegraphics[width=\textwidth]{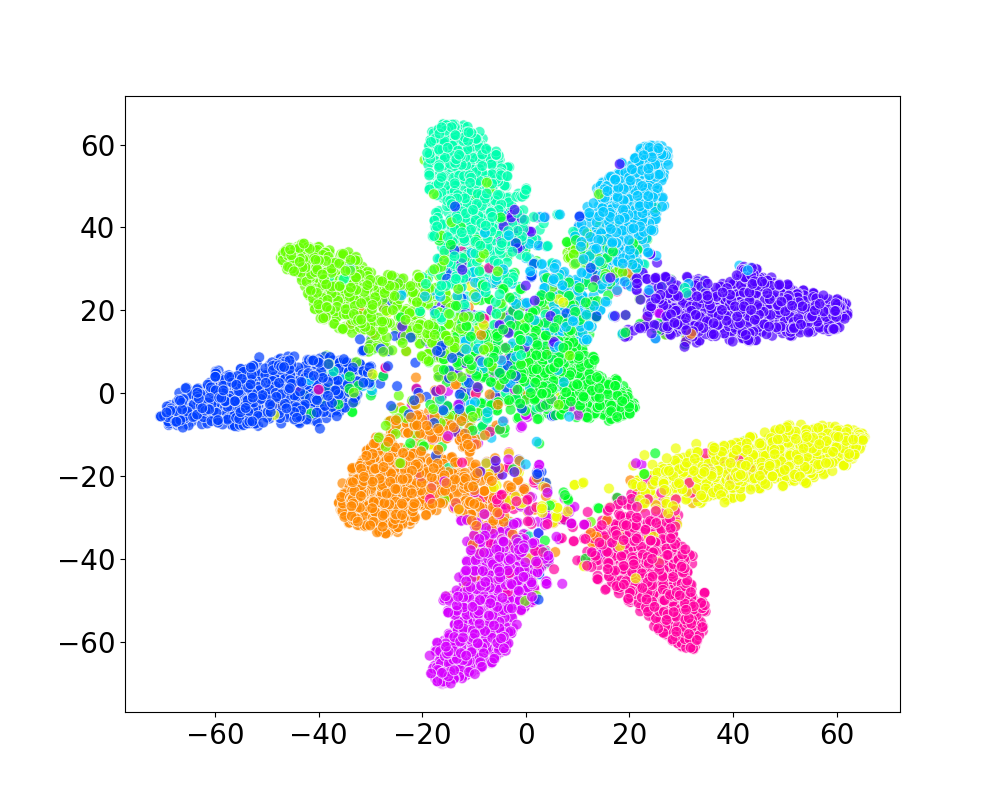}
    $\text{M}4$, $k=1.2$
  \end{minipage}
  \hfill
  \begin{minipage}[t]{0.13\textwidth}
    \centering
    \includegraphics[width=\textwidth]{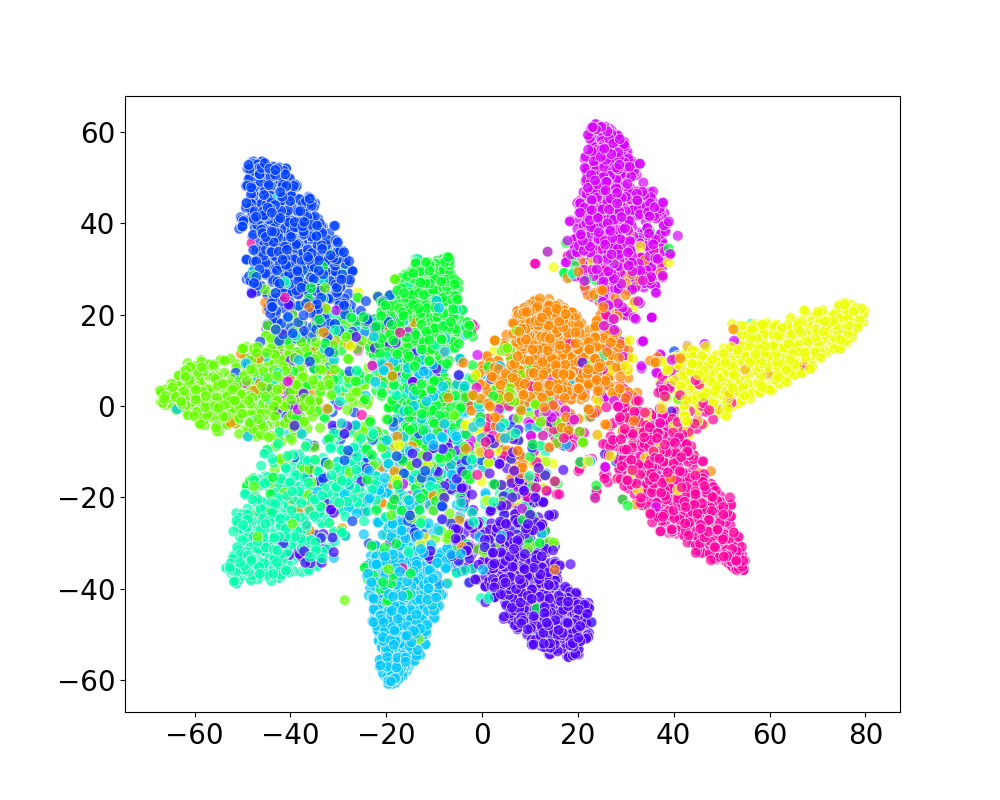}
    $\text{M}5$, $k=-1.2$
  \end{minipage}
  \hfill
  \begin{minipage}[t]{0.13\textwidth}
    \centering
    \includegraphics[width=\textwidth]{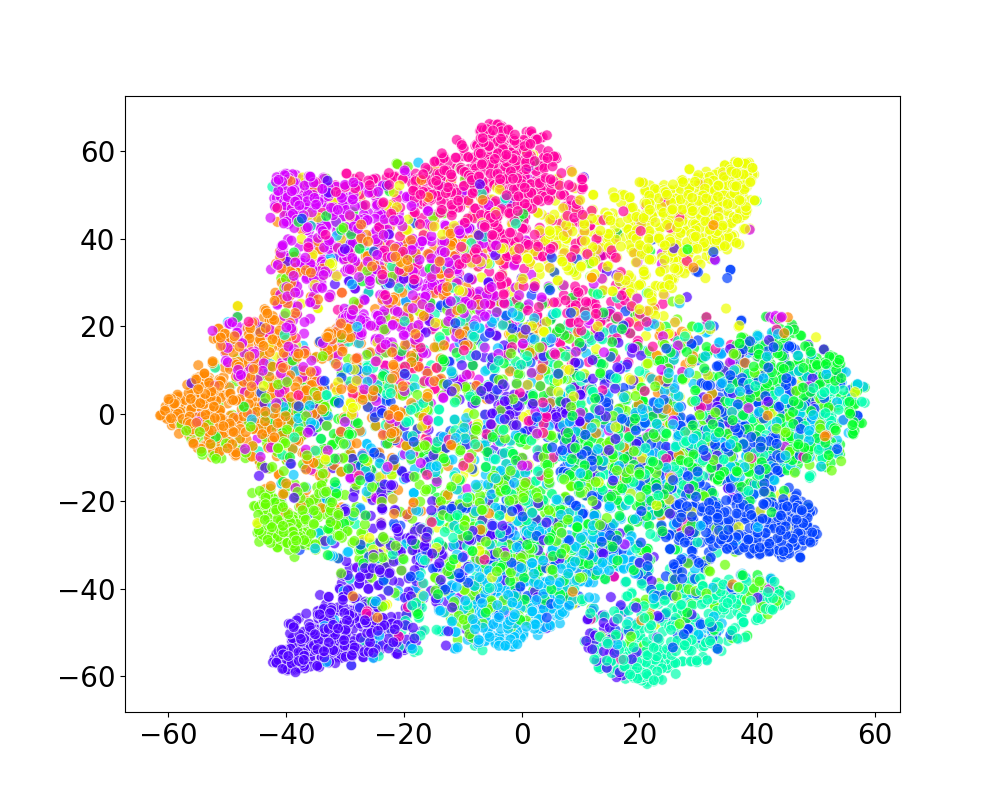}
    $\text{M}6$, $k=1.2$
  \end{minipage}
  \hfill
  \vspace{0.1cm}
  \begin{minipage}[t]{0.13\textwidth}
    \centering
    \includegraphics[width=\textwidth]{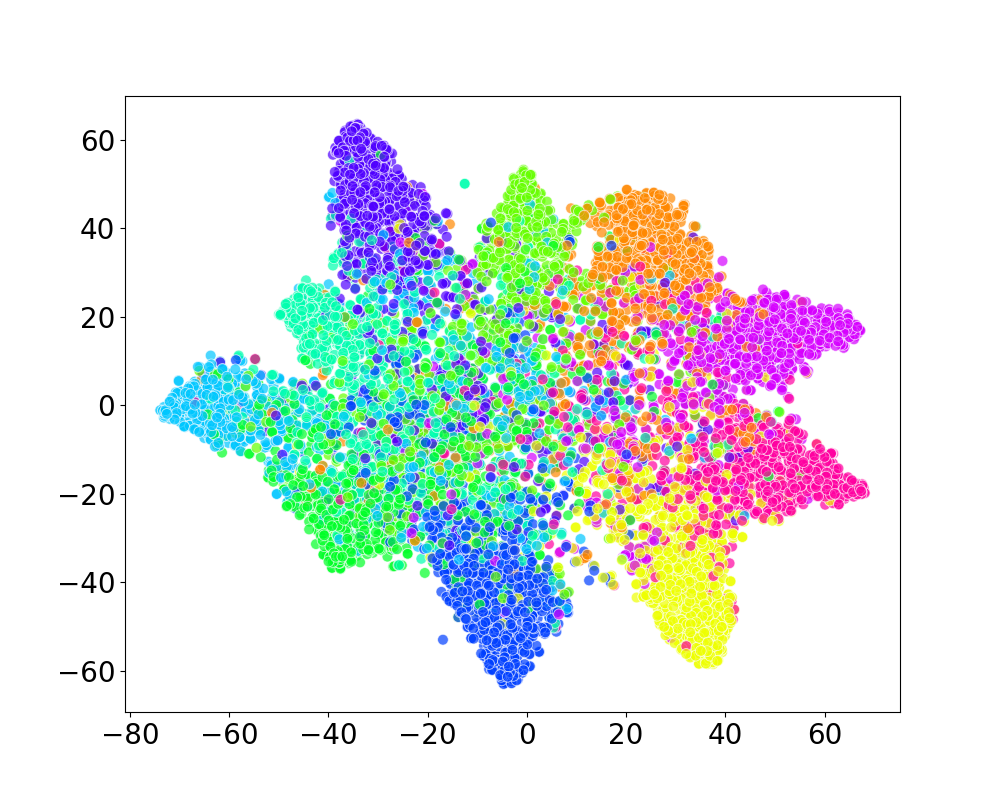}
    $\text{M}7$, $k=-1.2$
  \end{minipage}
  \hfill
  \begin{minipage}[t]{0.13\textwidth}
    \centering
    \includegraphics[width=\textwidth]{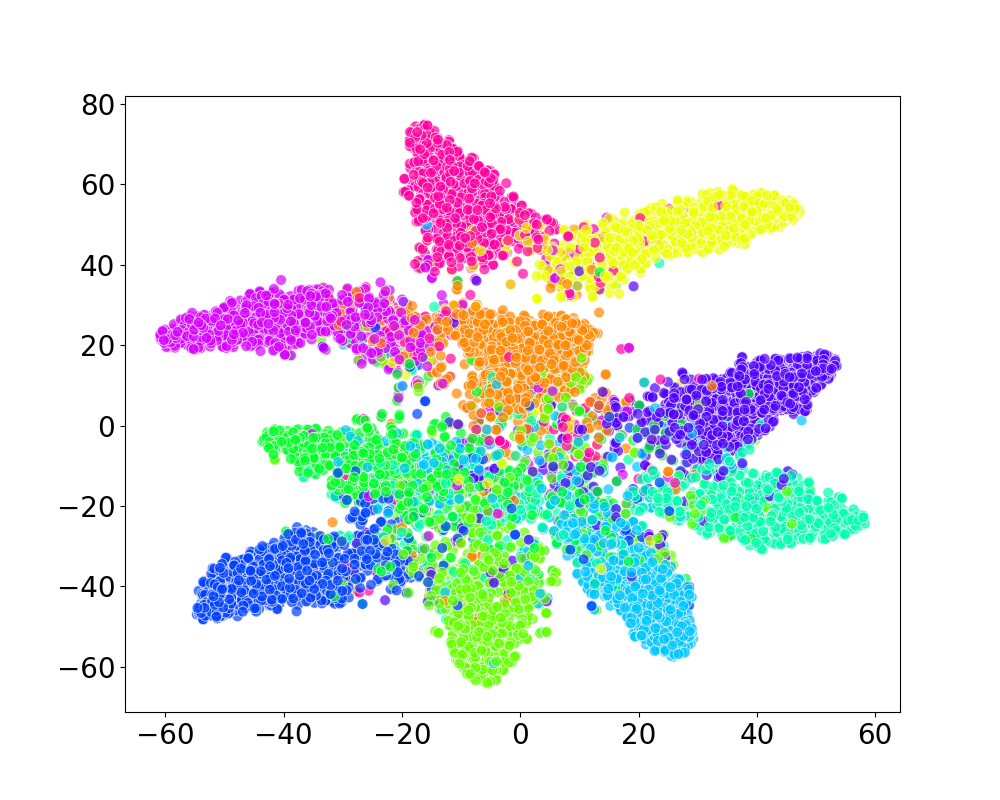}
    $\text{M}8$, $k=1.2$
  \end{minipage}
  \hfill
  \begin{minipage}[t]{0.13\textwidth}
    \centering
    \includegraphics[width=\textwidth]{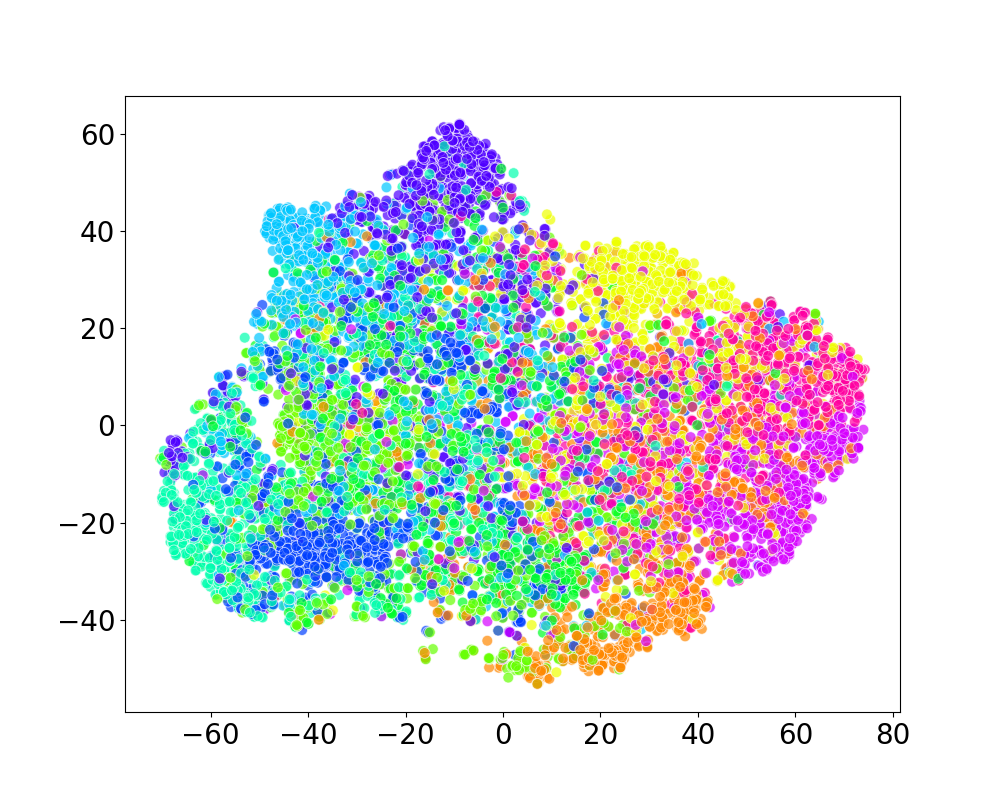}
    $\text{M}9$, $k=-1.2$
  \end{minipage}
  \hfill
  \begin{minipage}[t]{0.13\textwidth}
    \centering
    \includegraphics[width=\textwidth]{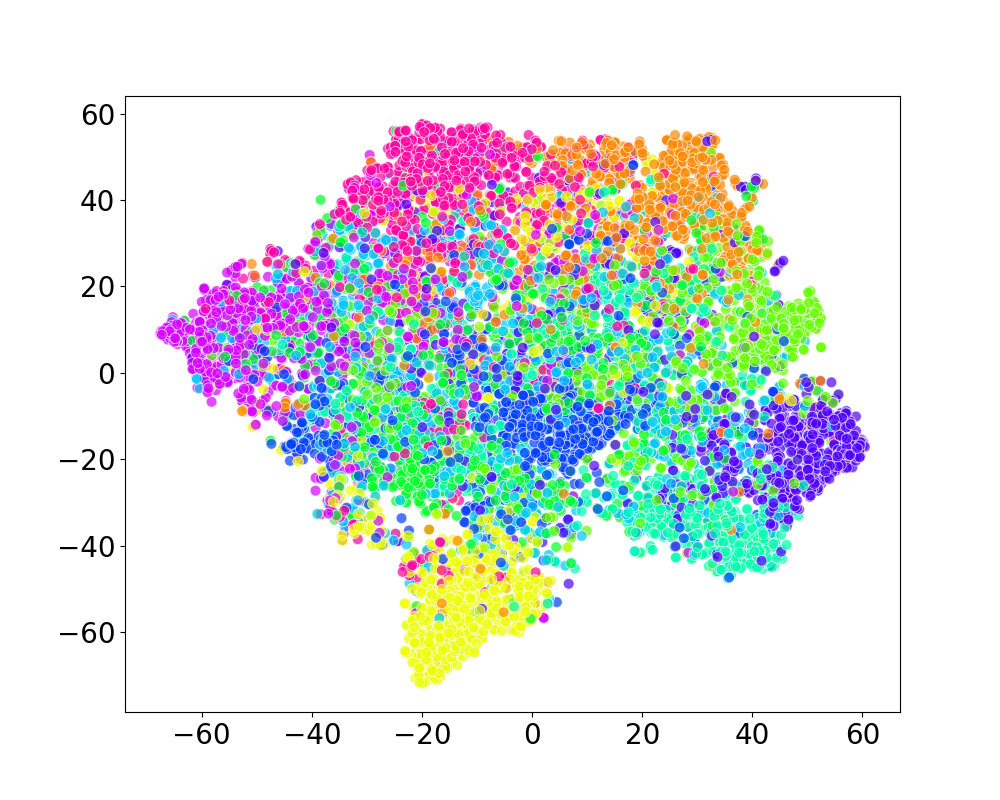}
    $\text{M}10$, $k=1.2$
  \end{minipage}
  \hfill
  \begin{minipage}[t]{0.13\textwidth}
    \centering
    \includegraphics[width=\textwidth]{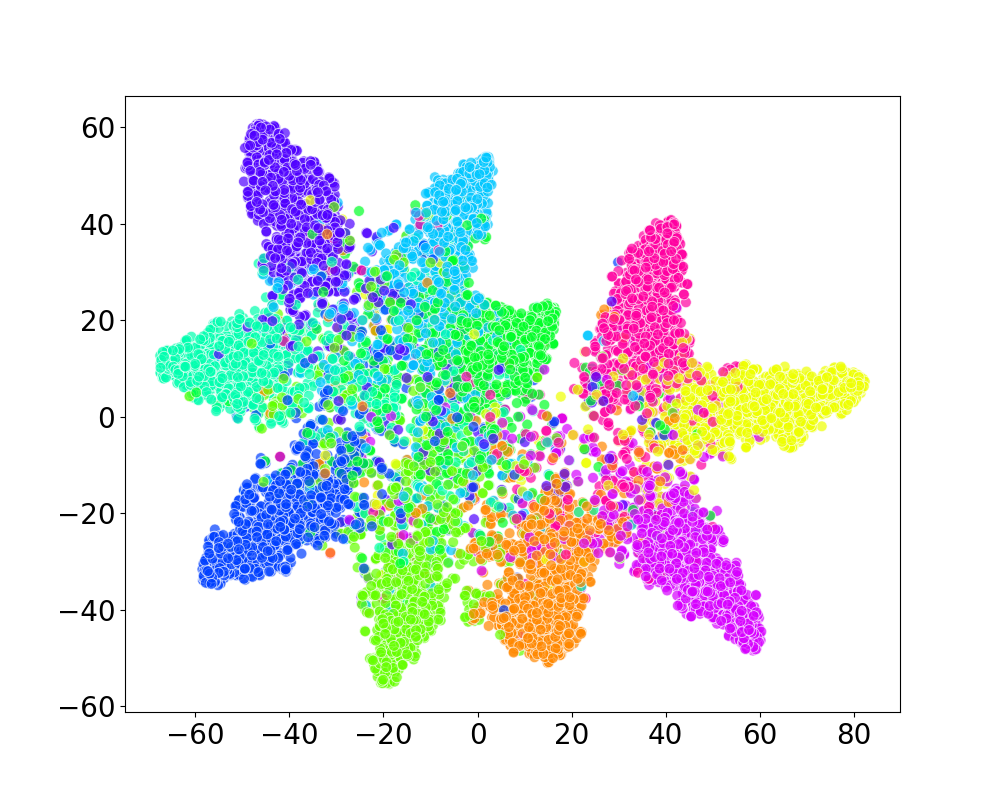}
    $\text{M}11$, $k=-1.2$
  \end{minipage}
  \hfill
  \begin{minipage}[t]{0.13\textwidth}
    \centering
    \includegraphics[width=\textwidth]{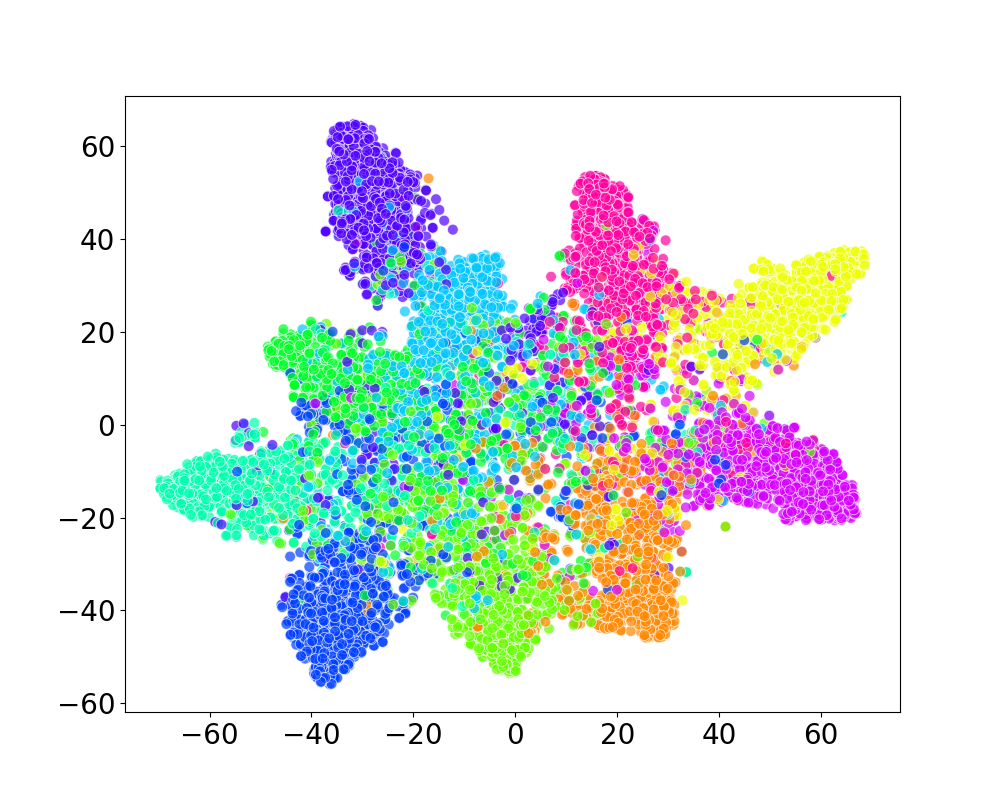}
    $\text{M}12$, $k=1.2$
  \end{minipage}

  \caption{Shows the t-SNE plot on CIFAR10 test set using MoblineV3-small model for $12$ masks and $1.2$ and $-1.2$ values of ($k$). Here, $M_i$ refers to $i^{th}$ mask.}
   \label{fig:TSNE-mob}
\end{figure*}

\subsection{Quantitative Results}
The goal of this study was to evaluate the robustness of CNN models under NUI attacks, on different datasets. After conducting several experiments, we have recorded a substantial drop in the accuracy of CNNs on all datasets. \autoref{fig:cifar_vgg} shows the performance of VGG16 over CIFAR10 under different NUI attacks on the test set, similarly later figures up to \autoref{fig:cal_res}  shows the performance curve for different CNNs over different datasets. The plots are reported in blue colour when the models are trained on the original training set and in orange colour on the augmented training set with NUI transformations. Each Figure contains $12$ Sub-Figures corresponding to NUI attacks with $1^{st}$ to $12^{th}$ masks in the order of $1^{st}$ row from left to right for $1^{st}$ to $6^{th}$ masks and $2^{nd}$ row from left to right for $7^{th}$ to $12^{th}$ masks, respectively. The \textit{x}-axis and \textit{y}-axis represent different NUI weights ($k$) and Accuracy ($\%$), respectively. Note that $k=0$ indicates no attack. From these plots, it is clear that the performance of the CNN models decreases on the NUI-attacked test sets. However, the performance is enhanced by including the NUI attack-based augmentation during training. It is also observed that the accuracy of the CNN models decreases as the weight ($k$) of the NUI attack moves towards extreme positive or negative values. We can observe that the curve for a mask remains similar for a particular dataset irrespective of the model used. It depicts the generalizability of proposed NUI attacks for different CNN models. The performance of a particular mask on a dataset also depends on the number of classes. If the number of classes is less, the probability of correctly classifying a test image is high as compared to a dataset with more classes.

\begin{table*}[!t]
\caption{The \% increase in the accuracy of models on various test sets, after applying the proposed defense technique. Results are reported for $k = -1.4$}
    \centering
    \newcommand\wide{0.041\textwidth}
    \begin{tabular}{|p{0.068\textwidth}|p{0.105\textwidth}|
    p{\wide}|p{\wide}|p{\wide}|p{\wide}|p{\wide}|p{\wide}|p{\wide}|p{\wide}|p{\wide}|p{\wide}|p{\wide}|p{\wide}|p{\wide}|}
    \hline
    Model & Dataset & M1 & M2 & M3 & M4 & M5 & M6 & M7 & M8 & M9 & M10 & M11 & M12 \\
    \hline
    VGG16 & CIFAR10 & $6.36$ & $6.73$ & $7.14$ & $5.7$ & $11.9$ & $91$ & $32.65$ & $21.17$ & $177.3$ & $97.66$ & $9.77$ & $18.4$ \\
    \hline
    VGG16 & TinyImageNet & $27.64$ & $29.73$ & $31$ & $19.74$ & $85.83$ & $193.7$ & $150.5$ & $80.66$ & $675.8$ & $256$ & $40.8$ & $71$ \\
    \hline
    VGG19 & CalTech256 & $18.81$ & $19.4$ & $19.27$ & $16.8$ & $55.21$ & $149.4$ & $98.25$ & $51.18$ & $632.7$ & $304.2$ & $28.92$ & $51.87$ \\
    \hline
    ResNet18 & CIFAR10 & $4.11$ & $3.59$ & $3.75$ & $3.75$ & $9.42$ & $68.7$ & $26.18$ & $7.12$ & $108.8$ & $93.73$ & $6.82$ & $9.97$ \\
    \hline
    ResNet18 & TinyImageNet & $30.4$ & $30.26$ & $26.84$ & $26.92$ & $80.85$ & $262.4$ & $169.2$ & $90$ & $503.4$ & $215.8$ & $41.82$ & $63.62$ \\
    \hline
    ResNet18 & CalTech256 & $33.71$ & $31.85$ & $33.81$ & $29$ & $77$ & $237$ & $144.1$ & $70.44$ & $664.1$ & $418.5$ & $48.5$ & $79.98$ \\
    \hline
    Mobilenet & CIFAR10 & $13.04$ & $12.95$ & $12.55$ & $12.65$ & $14.58$ & $109.65$ & $54.09$ & $12.26$ & $147.37$ & $136.82$ & $18.42$ & $27.81$ \\
    \hline
    Inception & CIFAR10 & $5.21$ & $5.13$ & $7.62$ & $4.8$ & $6.86$ & $98.86$ & $27.44$ & $5.02$ & $157.46$ & $111.51$ & $7.99$ & $16.32$ \\
    \hline
    \end{tabular}
    
    \label{tab:perc2}
\end{table*}

The blue curves show that the CNN models are not robust against the NUI attacks as these models get fooled by the perturbed images. The $6^{th}$, $7^{th}$, $9^{th}$ and $10^{th}$ masks lead to a very high impact on the performance degradation of the CNN models. The poor performance of the models for Mask $6$ is due to severe circular perturbation which leads to the complex generated images. Mask $7$ is similar to Mask $6$ but with reduced complexity. Still, the complexity of images generated by Mask $7$ is very high to fool the CNN models.
Mask $9$ and $10$ add perturbation as a pattern in the horizontal and vertical directions, respectively. Using these NUI attacks, the images after adding the mask are still visually perceptible, however, the performance of CNN models has significantly dropped. Moreover, only a small value of $k$ can produce a powerful NUI attack with high fooling success using Mask $9$ and Mask $10$.
The red curves depict that there has been considerable improvement in the performance of the CNN models after being trained on NUI-augmented training data. We exclude Mask $6$ and Mask $7$ in the training set, hence the improvement after NUI augmentation is low under these attacks on the test set.

\autoref{tab:perc1} summarizes the percentage reduction in the accuracy of the CNN models under different NUI attacks for $k = -1.4$ w.r.t. without attack. A high attack success rate is achieved using Mask $6$, Mask $7$, Mask $9$ and Mask $10$. TinyImageNet images are more prone to heavy perturbation using NUI attacks as depicted by the highest performance drop among all the datasets. The success rate of attack is higher for datasets for which the number of classes is large as the perturbation creates more confusion in class probabilities.
\autoref{tab:perc2} summarizes the percentage increase in the accuracy of the CNN models after being trained on the NUI perturbed dataset. The percentage improvement in the performance is calculated on the model's performance on the NUI attack and the model's performance after being trained on the NUI perturbed dataset. If a model on a particular dataset has a higher percentage reduction in \autoref{tab:perc1} then in most of such cases a higher percentage increase is observed in the model's performance on the same dataset in \autoref{tab:perc2}. Mask $9$ and Mask $10$ lead to the highest increment when trained on the NUI perturbed dataset. The readings also indicate that using NUI transformation as data augmentation is an effective technique and results in considerable performance improvements on NUI-attacked test sets.

\subsection{Analysis}
NUI has given a high attack success rate for all the models (VGG, ResNet, MobileNetV3 and InceptionV3). As mentioned in \autoref{tab:perc1} the classification accuracy of all the models decreased by at least $7\%$ which proves the effectiveness of the attack across various architecture and dataset complexity. 
The t-Distributed Stochastic Neighbor Embedding (t-SNE) plots are shown in \autoref{fig:TSNE-inc} and \autoref{fig:TSNE-mob} on CIFAR10 Test Set using InceptionV and MobileNet-small models, respectively. The t-SNE plots present the effect of different masks on the discriminative ability of the embedding distribution of CNN models leading to lower classification accuracy. 
It can be noticed that the separation between the distribution of the embedding of different classes decreases after applying the NUI attacks leading to mis-classifications. The t-SNE plots of $6^{th}$, $7^{th}$, $9^{th}$, $10^{th}$ and $11^{th}$ masks show heavy degradation of the separation between the distributions which leads to a huge accuracy drop. In addition to the t-SNE plot, we have provided histograms to better understand the change in data distribution in the Supplementary. The accuracy drops are managed via the proposed defense technique effectively. The defense strategy enhances models' performance on perturbed data and preserves the original accuracy. 
\autoref{tab:perc2} shows at least $4\%$ increase in the models' accuracy after applying the defense technique. The metrics Precision, Recall and F1-score, given in Supplementary, also support the above discussion.

Compared to attack approaches that require a neural network, the proposed NUI attack is swift and data-independent. The challenge with this approach is its fixed nature, which may prove ineffective in certain scenarios requiring an attack technique of a dynamic nature. Testing of such scenarios is out of the scope of this paper. We tested the proposed attack extensively through various evaluation metrics which gives a better understanding of how the attack technique works.

\section{Conclusion}
\label{sec:Conclusion}
In this research, we introduce non-uniform illumination (NUI) attacks to study the robustness of the CNN models. The proposed NUI attacks can deceive the CNN models for image classification. The attack is simple and data-independent. It leverages the pixel brightness with spatial information to create the different masks that are included in the original image with a weight factor to generate the perturbed images. The images generated using NUI attacks retain their semantic significance. Through extensive experimentation using VGG and ResNet models on CIFAR10, TinyImageNet, and CalTech256 datasets as well as MobilenetV3-small and InceptionV3 models on CIFAR10 dataset, we observe a significant decline in classification performance across all the NUI-attacked test sets. Notably, several samples that were correctly classified with high confidence in the original test set, were incorrectly classified with high confidence after undergoing the NUI attack. The proposed NUI attack is also utilized as a data augmentation during training as a primary defense mechanism and to make the models resilient against such attacks. We have also observed the effects of the NUI attack on different colour channels through a brief experiment, detailed in Supplementary, which we would like to extend in future as a topic of our next research.  

{\small
\bibliographystyle{IEEEtran}
\bibliography{References}

\begin{thebibliography}{10}
\providecommand{\url}[1]{#1}
\csname url@samestyle\endcsname
\providecommand{\newblock}{\relax}
\providecommand{\bibinfo}[2]{#2}
\providecommand{\BIBentrySTDinterwordspacing}{\spaceskip=0pt\relax}
\providecommand{\BIBentryALTinterwordstretchfactor}{4}
\providecommand{\BIBentryALTinterwordspacing}{\spaceskip=\fontdimen2\font plus
\BIBentryALTinterwordstretchfactor\fontdimen3\font minus
  \fontdimen4\font\relax}
\providecommand{\BIBforeignlanguage}[2]{{%
\expandafter\ifx\csname l@#1\endcsname\relax
\typeout{** WARNING: IEEEtran.bst: No hyphenation pattern has been}%
\typeout{** loaded for the language `#1'. Using the pattern for}%
\typeout{** the default language instead.}%
\else
\language=\csname l@#1\endcsname
\fi
#2}}
\providecommand{\BIBdecl}{\relax}
\BIBdecl

\bibitem{lecun2015deep}
Y.~LeCun, Y.~Bengio, and G.~Hinton, ``Deep learning,'' \emph{Nature}, vol. 521,
  no. 7553, pp. 436--444, 2015.

\bibitem{ioannidou2017deep}
A.~Ioannidou, E.~Chatzilari, S.~Nikolopoulos, and I.~Kompatsiaris, ``Deep
  learning advances in computer vision with 3d data: A survey,'' \emph{ACM
  Computing Surveys}, vol.~50, no.~2, pp. 1--38, 2017.

\bibitem{guo2016deep}
Y.~Guo, Y.~Liu, A.~Oerlemans, S.~Lao, S.~Wu, and M.~S. Lew, ``Deep learning for
  visual understanding: A review,'' \emph{Neurocomputing}, vol. 187, pp.
  27--48, 2016.

\bibitem{young2018recent}
T.~Young, D.~Hazarika, S.~Poria, and E.~Cambria, ``Recent trends in deep
  learning based natural language processing,'' \emph{IEEE Computational
  Intelligence Magazine}, vol.~13, no.~3, pp. 55--75, 2018.

\bibitem{ravi2016deep}
D.~Rav{\`\i}, C.~Wong, F.~Deligianni, M.~Berthelot, J.~Andreu-Perez, B.~Lo, and
  G.-Z. Yang, ``Deep learning for health informatics,'' \emph{IEEE Journal of
  Biomedical and Health Informatics}, vol.~21, no.~1, pp. 4--21, 2016.

\bibitem{glorot2011domain}
X.~Glorot, A.~Bordes, and Y.~Bengio, ``Domain adaptation for large-scale
  sentiment classification: A deep learning approach,'' in \emph{28th
  International Conference on Machine Learning}, 2011, pp. 513--520.

\bibitem{li2021survey}
Z.~Li, F.~Liu, W.~Yang, S.~Peng, and J.~Zhou, ``A survey of convolutional
  neural networks: analysis, applications, and prospects,'' \emph{IEEE
  Transactions on Neural Networks and Learning Systems}, vol.~33, no.~12, pp.
  6999--7019, 2022.

\bibitem{dai2021random}
D.~Dai, Z.~Zhuang, J.~Wei, S.~Xia, Y.~Li, and H.~Zhu, ``Random sharing
  parameters in the global region of convolutional neural network,'' \emph{IEEE
  Transactions on Artificial Intelligence}, vol.~3, no.~5, pp. 738--748, 2021.

\bibitem{dubey2022adainject}
S.~R. Dubey, S.~S. Basha, S.~K. Singh, and B.~B. Chaudhuri, ``Adainject:
  Injection based adaptive gradient descent optimizers for convolutional neural
  networks,'' \emph{IEEE Transactions on Artificial Intelligence}, 2022.

\bibitem{de2021automated}
C.~de~Vente, L.~H. Boulogne, K.~V. Venkadesh, C.~Sital, N.~Lessmann, C.~Jacobs,
  C.~I. S{\'a}nchez, and B.~van Ginneken, ``Automated covid-19 grading with
  convolutional neural networks in computed tomography scans: a systematic
  comparison,'' \emph{IEEE Transactions on Artificial Intelligence}, vol.~3,
  no.~2, pp. 129--138, 2021.

\bibitem{pan2022no}
Z.~Pan, F.~Yuan, X.~Wang, L.~Xu, X.~Shao, and S.~Kwong, ``No-reference image
  quality assessment via multibranch convolutional neural networks,''
  \emph{IEEE Transactions on Artificial Intelligence}, vol.~4, no.~1, pp.
  148--160, 2022.

\bibitem{esmaeilzehi2022ultralight}
A.~Esmaeilzehi, M.~O. Ahmad, and M.~Swamy, ``Ultralight-weight three-prior
  convolutional neural network for single image super resolution,'' \emph{IEEE
  Transactions on Artificial Intelligence}, 2022.

\bibitem{ahmad2021graph}
T.~Ahmad, L.~Jin, X.~Zhang, S.~Lai, G.~Tang, and L.~Lin, ``Graph convolutional
  neural network for human action recognition: A comprehensive survey,''
  \emph{IEEE Transactions on Artificial Intelligence}, vol.~2, no.~2, pp.
  128--145, 2021.

\bibitem{kingma2014adam}
D.~P. Kingma and J.~Ba, ``Adam: a method for stochastic optimization,'' in
  \emph{International Conference on Learning Representations}, 2014.

\bibitem{dubey2023adanorm}
S.~R. Dubey, S.~K. Singh, and B.~B. Chaudhuri, ``Adanorm: Adaptive gradient
  norm correction based optimizer for cnns,'' in \emph{IEEE/CVF Winter
  Conference on Applications of Computer Vision}, 2023, pp. 5284--5293.

\bibitem{srivastava2014dropout}
N.~Srivastava, G.~Hinton, A.~Krizhevsky, I.~Sutskever, and R.~Salakhutdinov,
  ``Dropout: a simple way to prevent neural networks from overfitting,''
  \emph{The Journal of Machine Learning Research}, vol.~15, no.~1, pp.
  1929--1958, 2014.

\bibitem{ioffe2015batch}
S.~Ioffe and C.~Szegedy, ``Batch normalization: Accelerating deep network
  training by reducing internal covariate shift,'' in \emph{International
  Conference on Machine Learning}, 2015, pp. 448--456.

\bibitem{cubuk2019autoaugment}
E.~D. Cubuk, B.~Zoph, D.~Mane, V.~Vasudevan, and Q.~V. Le, ``Autoaugment:
  Learning augmentation strategies from data,'' in \emph{IEEE/CVF Conference on
  Computer Vision and Pattern Recognition}, 2019, pp. 113--123.

\bibitem{goodfellow2014explaining}
I.~J. Goodfellow, J.~Shlens, and C.~Szegedy, ``Explaining and harnessing
  adversarial examples,'' in \emph{International Conference on Learning
  Representations}, 2015.

\bibitem{moosavi2016deepfool}
S.-M. Moosavi-Dezfooli, A.~Fawzi, and P.~Frossard, ``Deepfool: a simple and
  accurate method to fool deep neural networks,'' in \emph{IEEE Conference on
  Computer Vision and Pattern Recognition}, 2016, pp. 2574--2582.

\bibitem{elsayed2018adversarial}
G.~Elsayed, S.~Shankar, B.~Cheung, N.~Papernot, A.~Kurakin, I.~Goodfellow, and
  J.~Sohl-Dickstein, ``Adversarial examples that fool both computer vision and
  time-limited humans,'' in \emph{Advances in Neural Information Processing
  Systems}, vol.~31, 2018.

\bibitem{su2019one}
J.~Su, D.~V. Vargas, and K.~Sakurai, ``One pixel attack for fooling deep neural
  networks,'' \emph{IEEE Transactions on Evolutionary Computation}, vol.~23,
  no.~5, pp. 828--841, 2019.

\bibitem{deb2020advfaces}
D.~Deb, J.~Zhang, and A.~K. Jain, ``Advfaces: Adversarial face synthesis,'' in
  \emph{IEEE International Joint Conference on Biometrics}, 2020, pp. 1--10.

\bibitem{singh2022powerful}
I.~Singh, T.~Araki, and K.~Kakizaki, ``Powerful physical adversarial examples
  against practical face recognition systems,'' in \emph{IEEE/CVF Winter
  Conference on Applications of Computer Vision}, 2022, pp. 301--310.

\bibitem{wang2021fakespotter}
R.~Wang, F.~Juefei-Xu, L.~Ma, X.~Xie, Y.~Huang, J.~Wang, and Y.~Liu,
  ``Fakespotter: a simple yet robust baseline for spotting ai-synthesized fake
  faces,'' in \emph{Twenty-Ninth International Joint Conference on Artificial
  Intelligence}, 2021, pp. 3444--3451.

\bibitem{ren2022perturbation}
M.~Ren, Y.~Zhu, Y.~Wang, and Z.~Sun, ``Perturbation inactivation based
  adversarial defense for face recognition,'' \emph{IEEE Transactions on
  Information Forensics and Security}, vol.~17, pp. 2947--2962, 2022.

\bibitem{akhtar2018defense}
N.~Akhtar, J.~Liu, and A.~Mian, ``Defense against universal adversarial
  perturbations,'' in \emph{IEEE Conference on Computer Vision and Pattern
  Recognition}, 2018, pp. 3389--3398.

\bibitem{nguyen2020adversarial}
D.-L. Nguyen, S.~S. Arora, Y.~Wu, and H.~Yang, ``Adversarial light projection
  attacks on face recognition systems: A feasibility study,'' in \emph{IEEE/CVF
  Conference on Computer Vision and Pattern Recognition Workshops}, 2020, pp.
  814--815.

\bibitem{singh2021brightness}
I.~Singh, S.~Momiyama, K.~Kakizaki, and T.~Araki, ``On brightness agnostic
  adversarial examples against face recognition systems,'' in
  \emph{International Conference of the Biometrics Special Interest
  Group}.\hskip 1em plus 0.5em minus 0.4em\relax IEEE, 2021, pp. 1--5.

\bibitem{zhang2021adversarial}
Q.~Zhang, Q.~Guo, R.~Gao, F.~Juefei-Xu, H.~Yu, and W.~Feng, ``Adversarial
  relighting against face recognition,'' \emph{arXiv preprint
  arXiv:2108.07920}, 2021.

\bibitem{yang2022random}
B.~Yang, K.~Xu, H.~Wang, and H.~Zhang, ``Random transformation of image
  brightness for adversarial attack,'' \emph{Journal of Intelligent \& Fuzzy
  Systems}, vol.~42, no.~3, pp. 1693--1704, 2022.

\bibitem{hsiung2023towards}
L.~Hsiung, Y.-Y. Tsai, P.-Y. Chen, and T.-Y. Ho, ``Towards compositional
  adversarial robustness: Generalizing adversarial training to composite
  semantic perturbations,'' in \emph{IEEE/CVF Conference on Computer Vision and
  Pattern Recognition}, 2023, pp. 24\,658--24\,667.

\bibitem{ccp}
J.~Kantipudi, S.~R. Dubey, and S.~Chakraborty, ``Color channel perturbation
  attacks for fooling convolutional neural networks and a defense against such
  attacks,'' \emph{IEEE Transactions on Artificial Intelligence}, vol.~1,
  no.~2, pp. 181--191, 2020.

\bibitem{de2021impact}
K.~De and M.~Pedersen, ``Impact of colour on robustness of deep neural
  networks,'' in \emph{Proceedings of the IEEE/CVF international conference on
  computer vision}, 2021, pp. 21--30.

\bibitem{zhang2021adversarial_survey}
X.~Zhang, X.~Zheng, and W.~Mao, ``Adversarial perturbation defense on deep
  neural networks,'' \emph{ACM Computing Surveys (CSUR)}, vol.~54, no.~8, pp.
  1--36, 2021.

\bibitem{agarwal2020image}
A.~Agarwal, R.~Singh, M.~Vatsa, and N.~Ratha, ``Image transformation-based
  defense against adversarial perturbation on deep learning models,''
  \emph{IEEE Transactions on Dependable and Secure Computing}, vol.~18, no.~5,
  pp. 2106--2121, 2020.

\bibitem{agarwal2021damad}
A.~Agarwal, G.~Goswami, M.~Vatsa, R.~Singh, and N.~K. Ratha, ``Damad: Database,
  attack, and model agnostic adversarial perturbation detector,'' \emph{IEEE
  Transactions on Neural Networks and Learning Systems}, vol.~33, no.~8, pp.
  3277--3289, 2021.

\bibitem{he2022defeating}
Z.~He, W.~Wang, W.~Guan, J.~Dong, and T.~Tan, ``Defeating deepfakes via
  adversarial visual reconstruction,'' in \emph{Proceedings of the 30th ACM
  International Conference on Multimedia}, 2022, pp. 2464--2472.

\bibitem{deb2023faceguard}
D.~Deb, X.~Liu, and A.~K. Jain, ``Faceguard: A self-supervised defense against
  adversarial face images,'' in \emph{2023 IEEE 17th International Conference
  on Automatic Face and Gesture Recognition (FG)}.\hskip 1em plus 0.5em minus
  0.4em\relax IEEE, 2023, pp. 1--8.

\bibitem{premakumara2023enhancing}
N.~Premakumara, B.~Jalaian, N.~Suri, and H.~Samani, ``Enhancing object
  detection robustness: A synthetic and natural perturbation approach,''
  \emph{arXiv preprint arXiv:2304.10622}, 2023.

\bibitem{dubey2015multi}
S.~R. Dubey, S.~K. Singh, and R.~K. Singh, ``A multi-channel based illumination
  compensation mechanism for brightness invariant image retrieval,''
  \emph{Multimedia Tools and Applications}, vol.~74, pp. 11\,223--11\,253,
  2015.

\bibitem{cifar}
A.~Krizhevsky, ``Learning multiple layers of features from tiny images,''
  \emph{Master's thesis, University of Tront}, 2009.

\bibitem{griffin2007CalTech}
G.~Griffin, A.~Holub, and P.~Perona, ``Caltech-256 object category dataset,''
  \emph{California Institute of Technology}, 2007.

\bibitem{tinyimagenet}
Y.~Le and X.~Yang, ``Tiny imagenet visual recognition challenge,'' \emph{CS
  231N}, vol.~7, p.~7, 2015.

\bibitem{simonyan2014very}
K.~Simonyan and A.~Zisserman, ``Very deep convolutional networks for
  large-scale image recognition,'' in \emph{International Conference on
  Learning Representations}, 2015.

\bibitem{he2016deep}
K.~He, X.~Zhang, S.~Ren, and J.~Sun, ``Deep residual learning for image
  recognition,'' in \emph{IEEE Conference on Computer Vision and Pattern
  Recognition}, 2016, pp. 770--778.

\bibitem{howard2019searching}
A.~Howard, M.~Sandler, G.~Chu, L.-C. Chen, B.~Chen, M.~Tan, W.~Wang, Y.~Zhu,
  R.~Pang, V.~Vasudevan \emph{et~al.}, ``Searching for mobilenetv3,'' in
  \emph{Proceedings of the IEEE/CVF international conference on computer
  vision}, 2019, pp. 1314--1324.

\bibitem{szegedy2015rethinking}
C.~Szegedy, V.~Vanhoucke, S.~Ioffe, J.~Shlens, and Z.~Wojna, ``Rethinking the
  inception architecture for computer vision,'' in \emph{Proceedings of the
  IEEE conference on computer vision and pattern recognition}, 2016, pp.
  2818--2826.

\bibitem{paszke2019pytorch}
A.~Paszke, S.~Gross, F.~Massa, A.~Lerer, J.~Bradbury, G.~Chanan, T.~Killeen,
  Z.~Lin, N.~Gimelshein, L.~Antiga \emph{et~al.}, ``Pytorch: An imperative
  style, high-performance deep learning library,'' \emph{Advances in Neural
  Information Processing Systems}, vol.~32, 2019.

\end{thebibliography}
}

\begin{IEEEbiography}[{\includegraphics[width=1in,height=1.25in,clip,keepaspectratio]{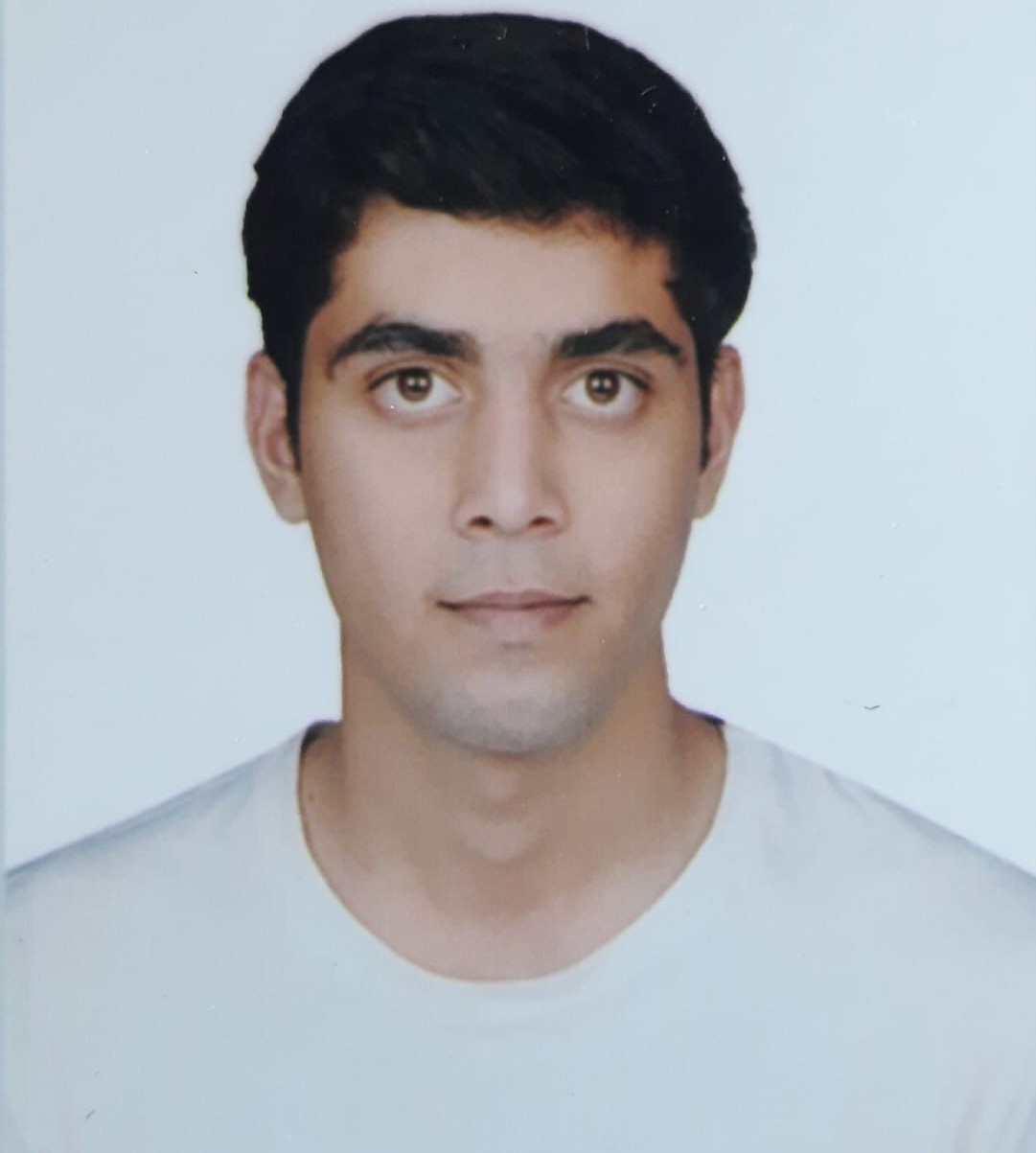}}]{Akshay Jain} was born in Indore, Madhya Pradesh, India in 1997. He completed his Bachelor of Engineering from Jabalpur Engineering College, Jabalpur in Information Technology in  2020. He completed his Master of Technology from the Indian Institute of Information Technology, Allahabad (IIIT-A) in Information Technology in 2023. He worked as a teaching assistant in IIIT-A from 2021 to 2023. He is currently working as a Junior Engineer at Netweb Technologies and he is interested in the field of computer vision.
\end{IEEEbiography}

\begin{IEEEbiography}[{\includegraphics[width=1in,height=1.25in,clip,keepaspectratio]{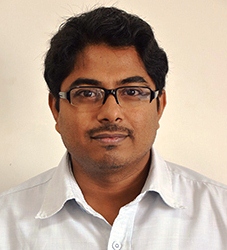}}]{Shiv Ram Dubey} is with the Indian Institute of Information Technology (IIIT), Allahabad since July 2021, where he is currently the Assistant Professor of Information Technology. He was with IIIT Sri City as Assistant Professor from Dec 2016 to July 2021 and Research Scientist from June 2016 to Dec 2016. He received the PhD degree from IIIT Allahabad in 2016. Before that, from 2012 to 2013, he was a Project Officer at Indian Institute of Technology (IIT), Madras. He was a recipient of several awards including the Best PhD Award in PhD Symposium at IEEE-CICT2017. Dr. Dubey is serving as the Secretary of IEEE Signal Processing Society Uttar Pradesh Chapter. His research interest includes Computer Vision and Deep Learning.
\end{IEEEbiography}

\begin{IEEEbiography}[{\includegraphics[width=1in,height=1.25in,clip,keepaspectratio]{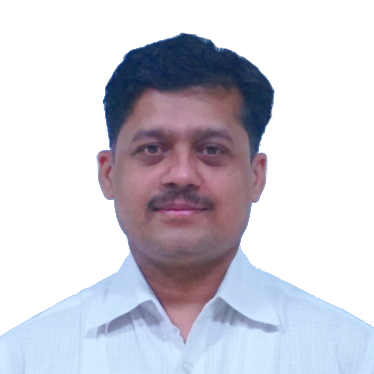}}]{Satish Kumar Singh} is serving at Indian Institute of Information Technology, Allahabad from 2013, and presently working as an Associate Professor in the Department of Information Technology. Dr. Singh is heading the Computer Vision and Biometrics Lab (CVBL) at IIIT Allahabad. His areas of interest include Image Processing, Computer Vision, Biometrics, Deep Learning, and Pattern Recognition. Dr. Singh was the Section Chair IEEE Uttar Pradesh Section (2021-2023) and a member of IEEE India Council (2021). He also served as the Vice-Chair, Operations, Outreach and Strategic Planning of IEEE India Council (2020-2024). Dr. Singh is also the technical committee affiliate of IEEE SPS IVMSP and MMSP. Currently, Dr. Singh is the Chair of IEEE Signal Processing Society Chapter of Uttar Pradesh Section and Associate Editor of IEEE Signal Processing Letters.
\end{IEEEbiography}

\begin{IEEEbiography}[{\includegraphics[width=1in,height=1.25in,clip,keepaspectratio]{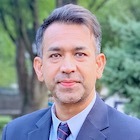}}]{KC Santosh,} a highly accomplished AI expert, is the chair of the Department of Computer Science, University of South Dakota. He served the National Institutes of Health as a research fellow. Before that, he worked as a postdoctoral research scientist at the LORIA research centre, Universitè de Lorraine in direct collaboration with industrial partner, ITESOFT, France. He earned his PhD in Computer Science - Artificial Intelligence from INRIA Nancy Grand East Research Centre (France). With funding of over \$1.3 million, including a \$1 million grant from DEPSCOR (2023) for AI/ML capacity building at USD, he has authored 10 books and published over 240 peer-reviewed research articles. He is an associate editor of multiple prestigious journals such as IEEE Transactions on AI, Int. J of Machine Learning \& Cybernetics, and Int. J of Pattern Recognition \& Artificial Intelligence. To name a few, Prof. Santosh is the proud recipient of the Cutler Award for Teaching and Research Excellence (USD, 2021), the President's Research Excellence Award (USD, 2019) and the Ignite Award from the U.S. Department of Health \& Human Services (HHS, 2014). As the founder of AI programs at USD, he has taken significant strides to increase enrolment in the graduate program, resulting in over 3,000\% growth in just three years. His leadership has helped build multiple inter-disciplinary AI/Data Science related academic programs, including collaborations with Biology, Physics, Biomedical Engineering, Sustainability and Business Analytics departments. Prof. Santosh is highly motivated in academic leadership, and his contributions have established USD as a pioneer in AI programs within the state of SD. More info. \href{https://kc-santosh.org/}{https://kc-santosh.org/.}
\end{IEEEbiography}

\begin{IEEEbiography}[{\includegraphics[width=1in,height=1.25in,clip,keepaspectratio]{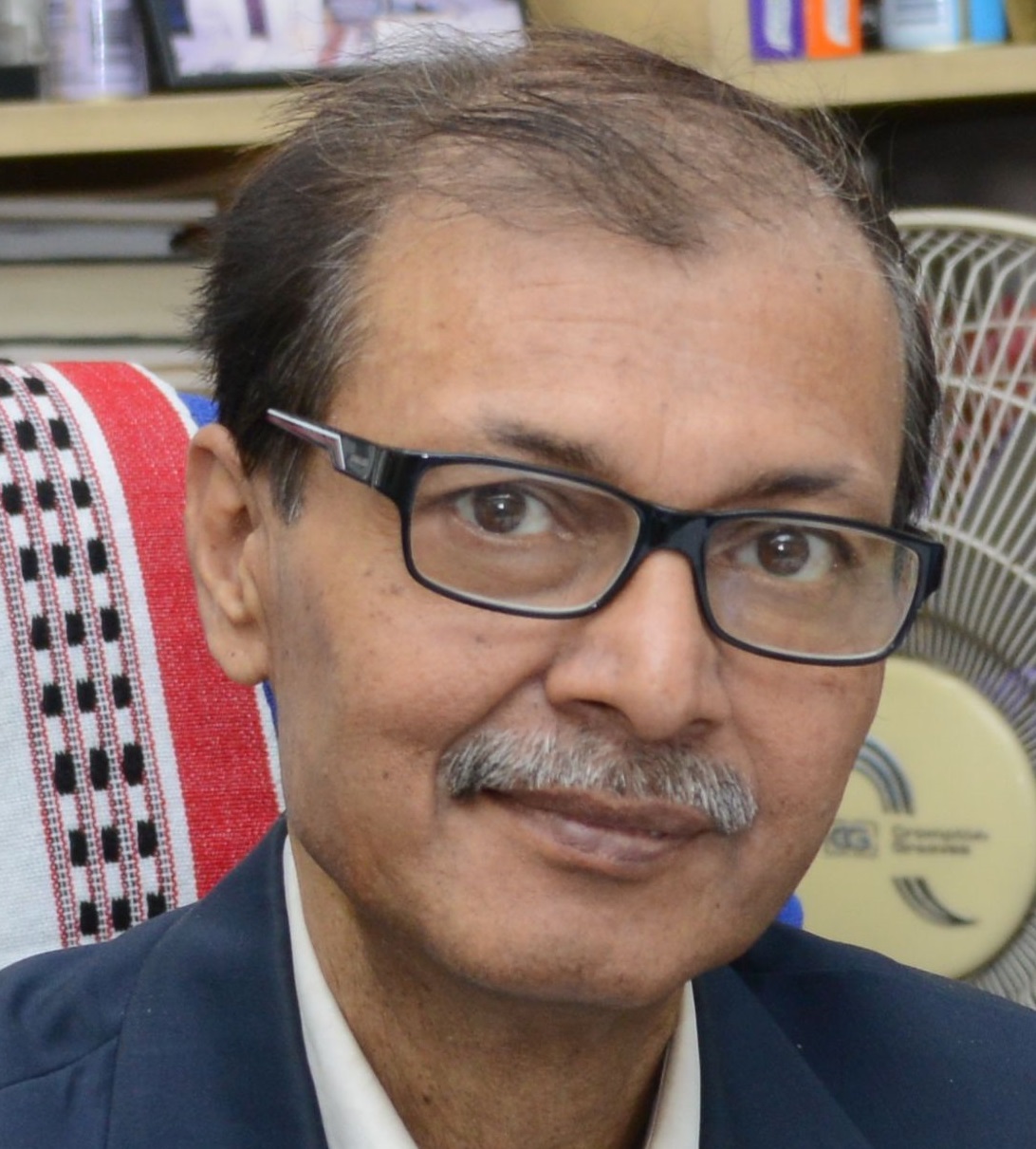}}]{Bidyut Baran Chaudhuri}
received the Ph.D. degree from IIT Kanpur, in 1980. He was a Leverhulme Postdoctoral Fellow with Queen’s University, U.K., from 1981 to 1982. He joined the Indian Statistical Institute, in 1978, where he worked as an INAE Distinguished Professor and a J C Bose Fellow at Computer Vision and Pattern Recognition Unit of Indian Statistical Institute. He is now affiliated to Techno India University, Kolkata as Pro-Vice Chancellor (Academic). His research interests include Pattern Recognition, Image Processing, Computer Vision, Natural Language Processing (NLP), Signal processing, Digital Document Processing, Deep learning etc. He pioneered the first workable OCR system for printed Indian scripts Bangla, Assamese and Devnagari. He also developed computerized \textit{Bharati Braille system} with speech synthesizer and has done statistical analysis of Indian language.  He has published about 425 research papers in international journals and conference proceedings. Also, he has authored/edited seven books in these fields. Prof. Chaudhuri received Leverhulme fellowship award, Sir J. C. Bose Memorial Award, M. N. Saha Memorial Award, Homi Bhabha Fellowship, Dr. Vikram Sarabhai Research Award, C. Achuta Menon Award, Homi Bhabha Award: Applied Sciences, Ram Lal Wadhwa Gold Medal, Jawaharlal Nehru Fellowship, J C Bose fellowship, Om Prakash Bhasin Award etc. Prof. Chaudhuri is the associate editor of three international journals and a fellow of INSA, NASI, INAE, IAPR, The World Academy of Sciences (TWAS) and life fellow of IEEE (2015). He acted as General Chair and Technical Co-chair at various International Conferences.
\end{IEEEbiography}
\balance

\pagebreak

\section{Supplementary}
\subsection{Effect of NUI Attacks}
\begin{figure*}[p]
    \centering
    \newcommand\width{0.1462\columnwidth}
    \includegraphics[width=\width]{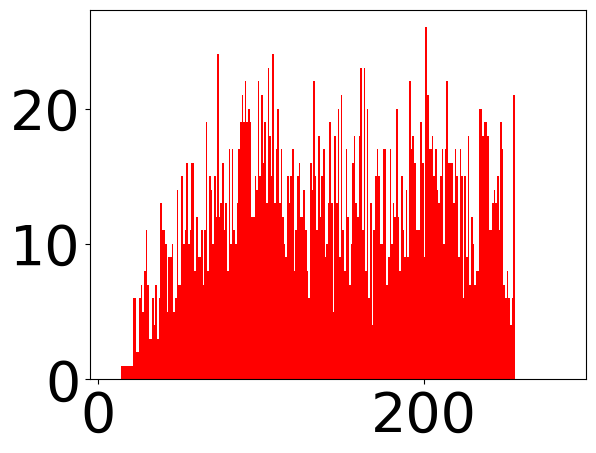}
    \includegraphics[width=\width]{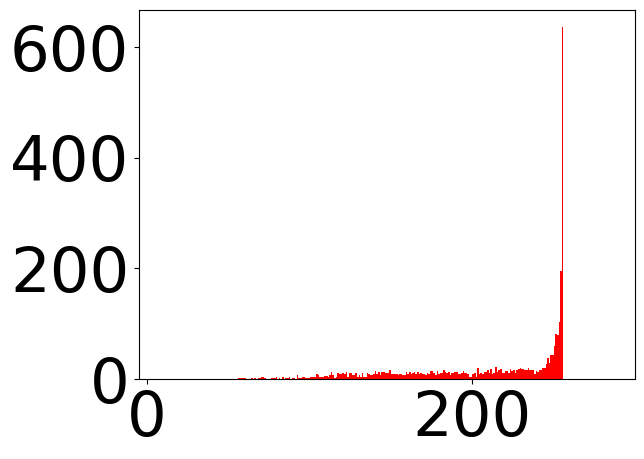}
    \includegraphics[width=\width]{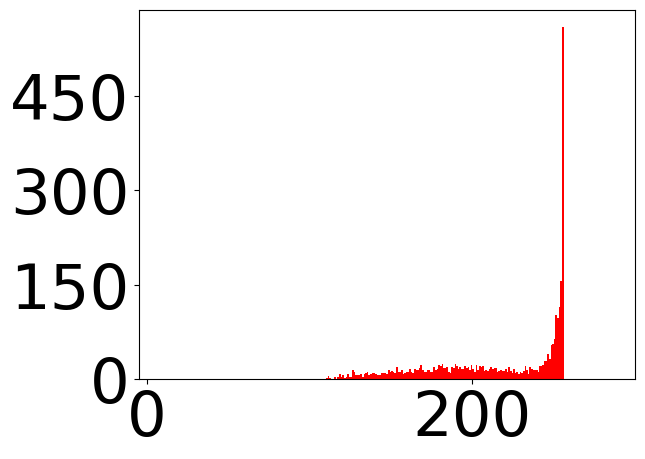}
    \includegraphics[width=\width]{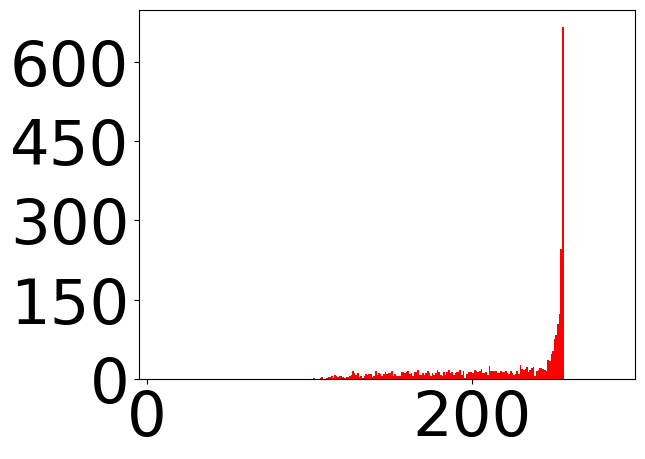}
    \includegraphics[width=\width]{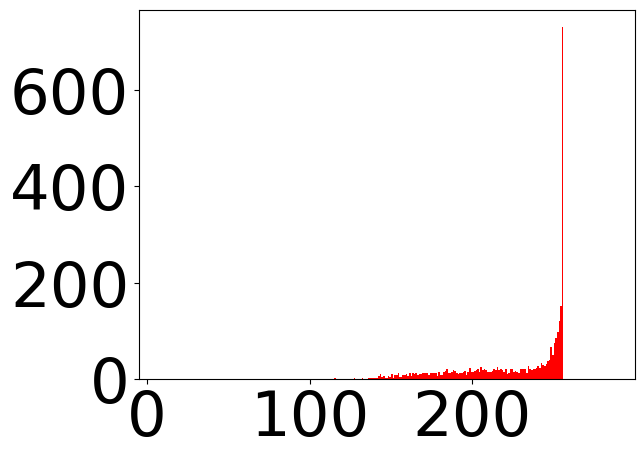}
    \includegraphics[width=\width]{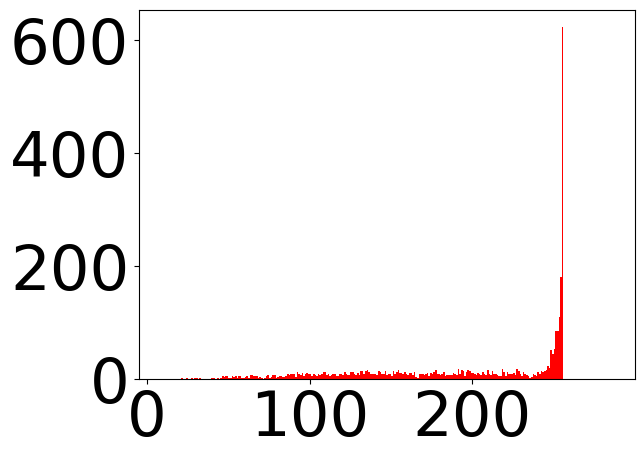}
    \includegraphics[width=\width]{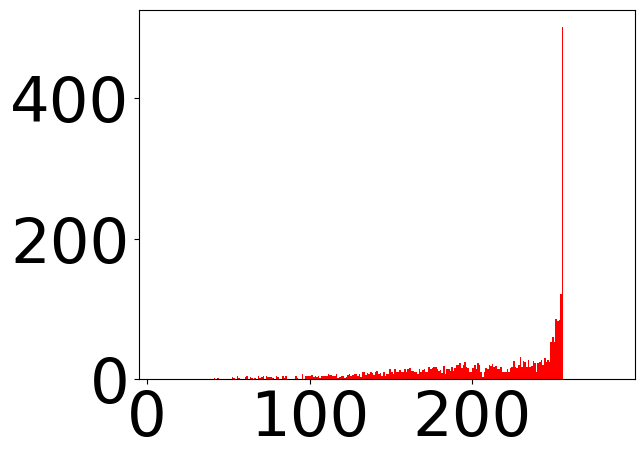}
    \includegraphics[width=\width]{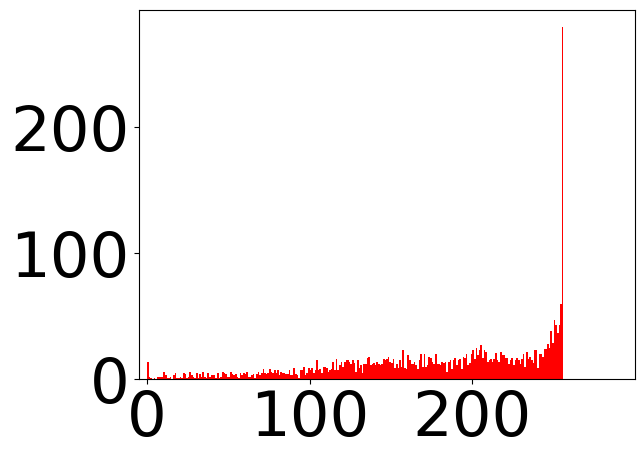}
    \includegraphics[width=\width]{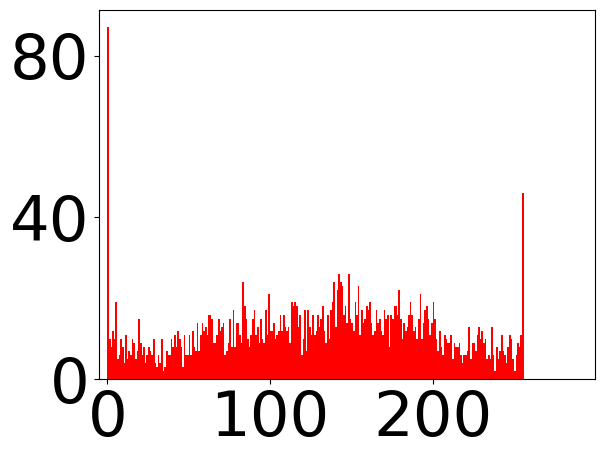}
    \includegraphics[width=\width]{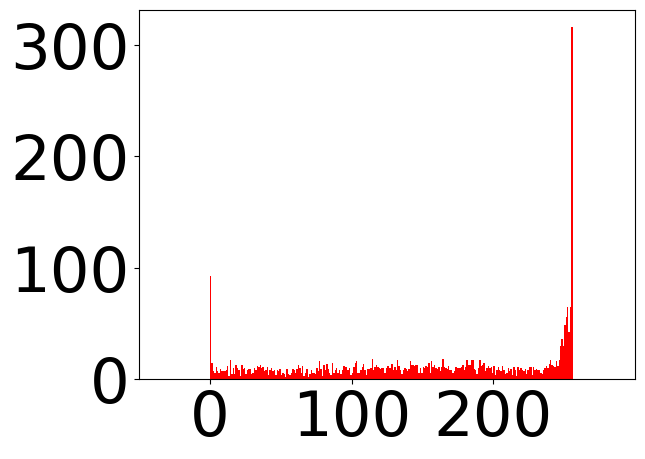}
    \includegraphics[width=\width]{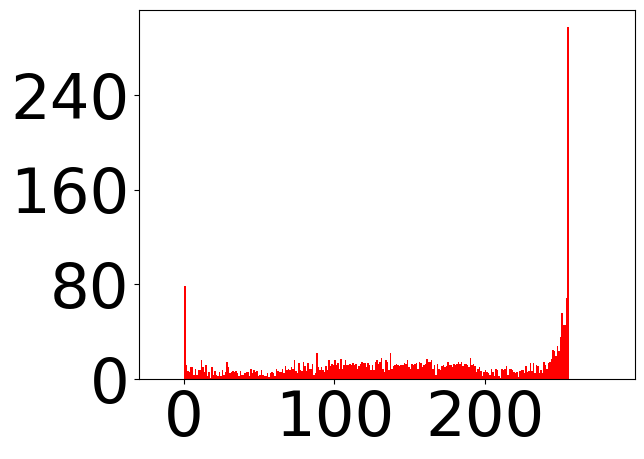}
    \includegraphics[width=\width]{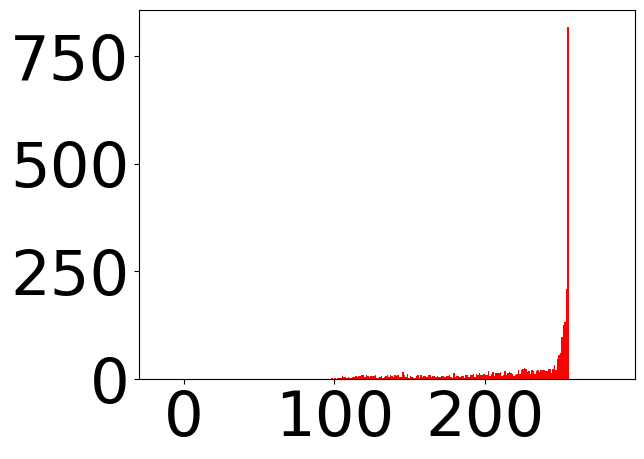}
    \includegraphics[width=\width]{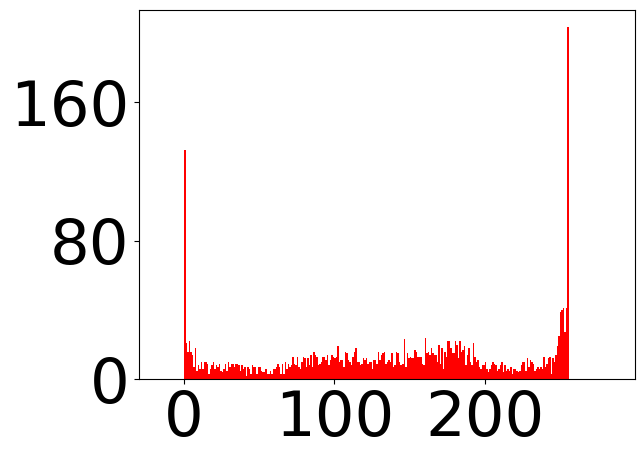}
    \\
    \includegraphics[width=\width]{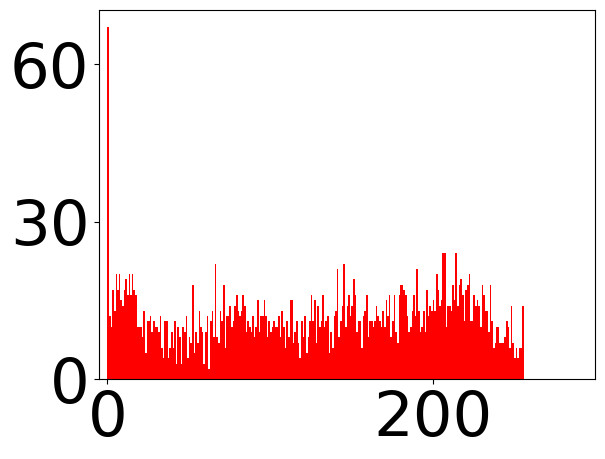}
    \includegraphics[width=\width]{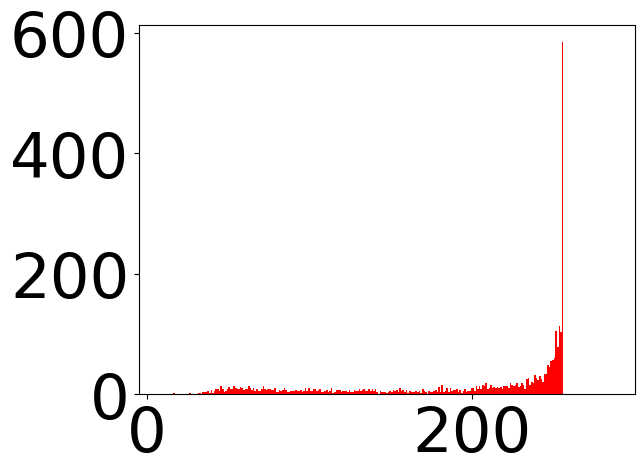}
    \includegraphics[width=\width]{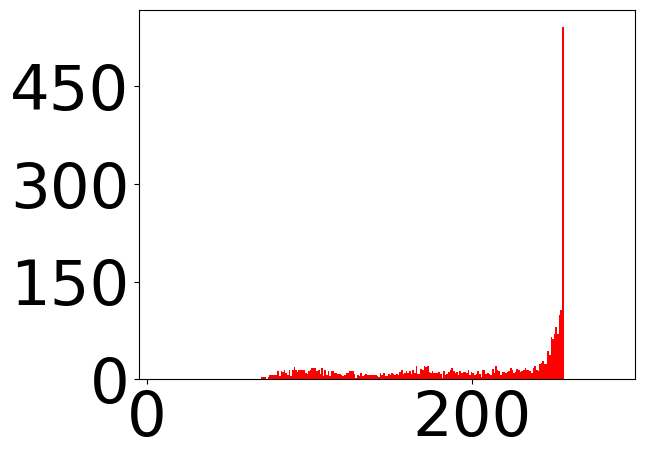}
    \includegraphics[width=\width]{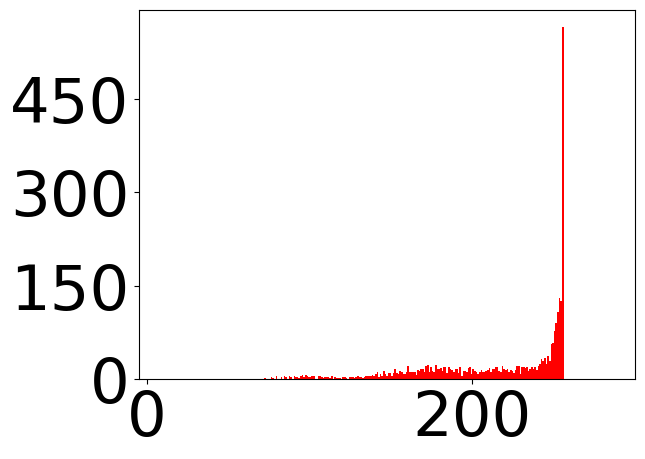}
    \includegraphics[width=\width]{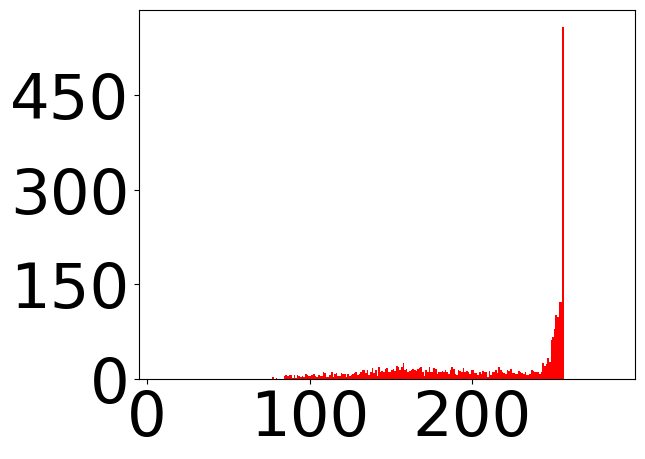}
    \includegraphics[width=\width]{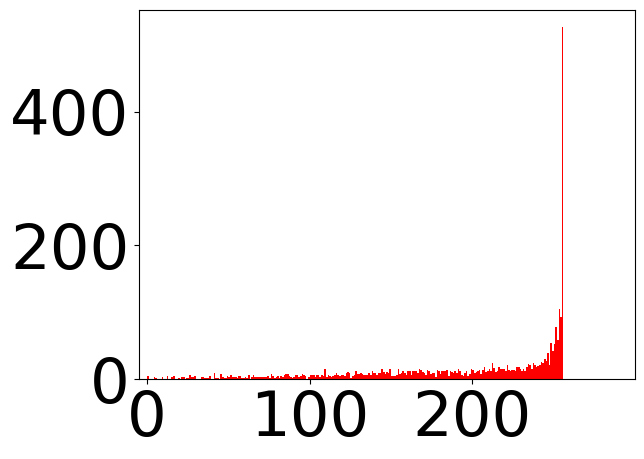}
    \includegraphics[width=\width]{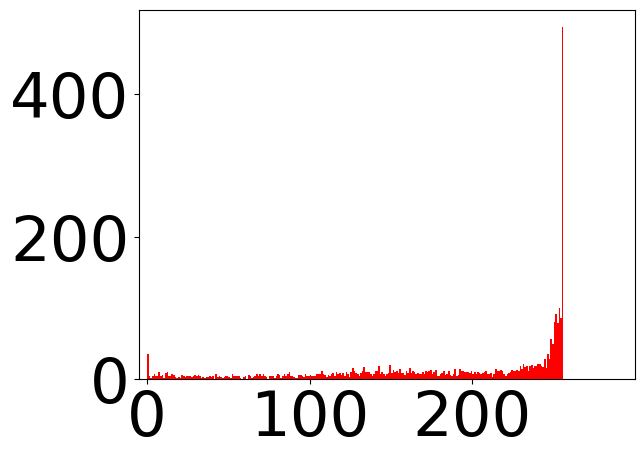}
    \includegraphics[width=\width]{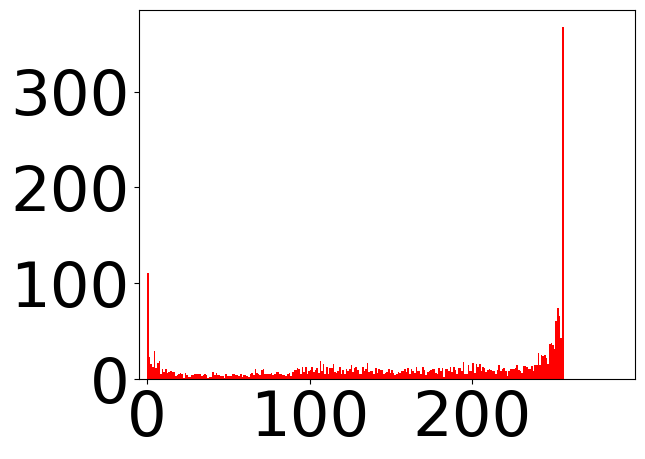}
    \includegraphics[width=\width]{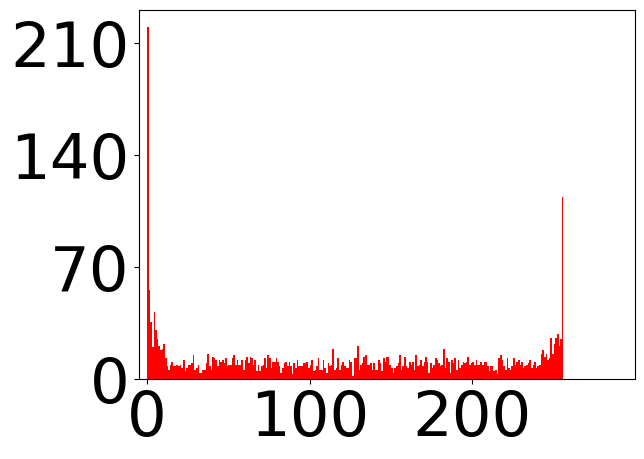}
    \includegraphics[width=\width]{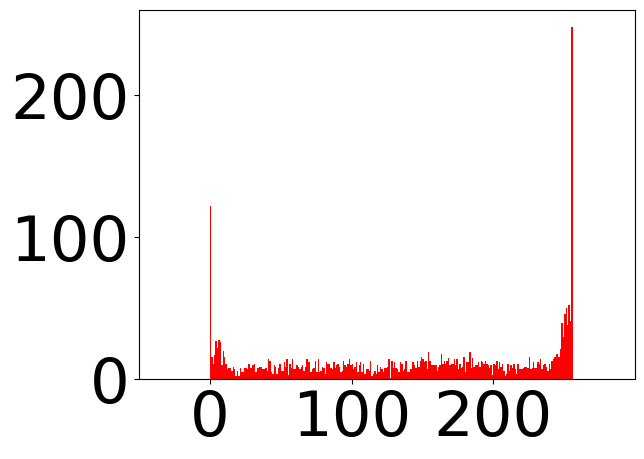}
    \includegraphics[width=\width]{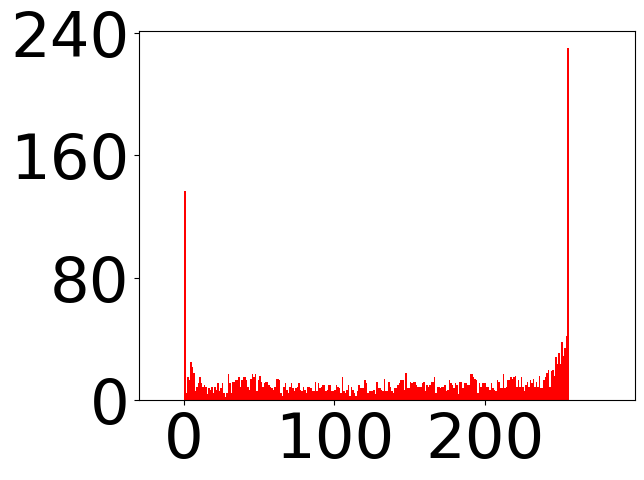}
    \includegraphics[width=\width]{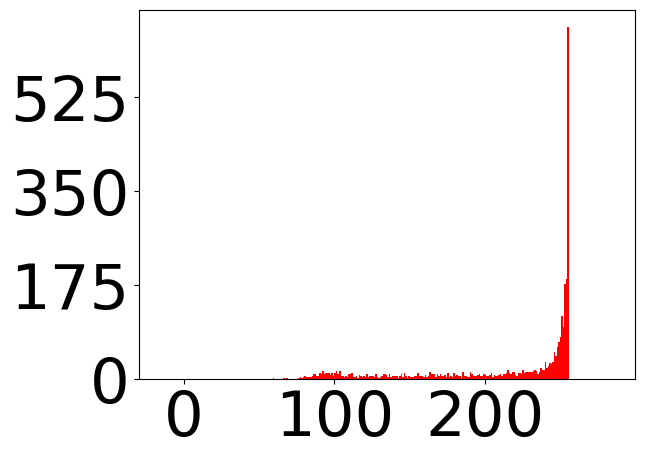}
    \includegraphics[width=\width]{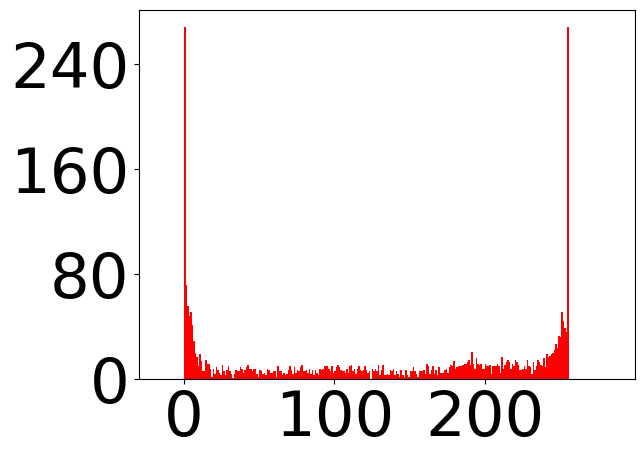}
    \\
    \includegraphics[width=\width]{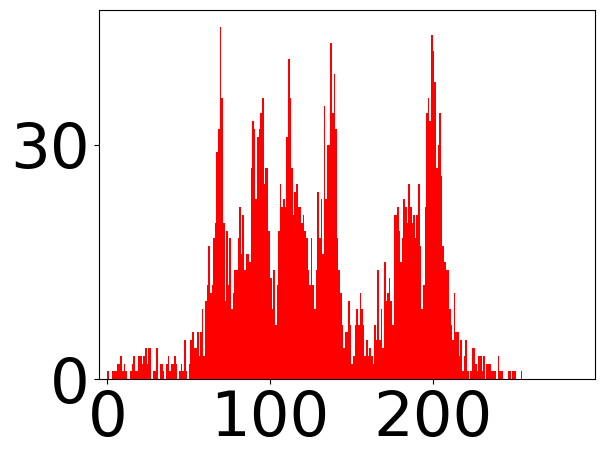}
    \includegraphics[width=\width]{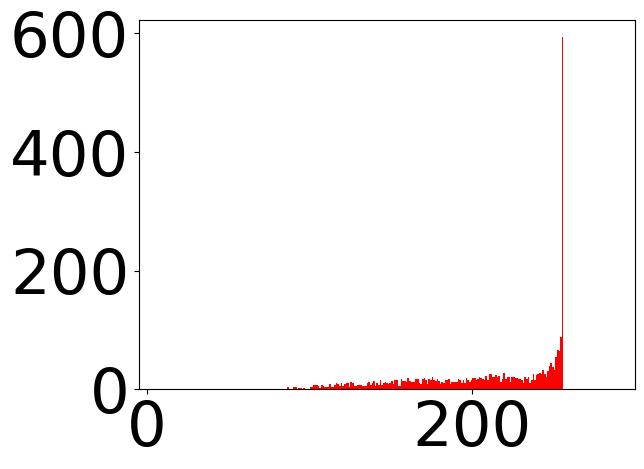}
    \includegraphics[width=\width]{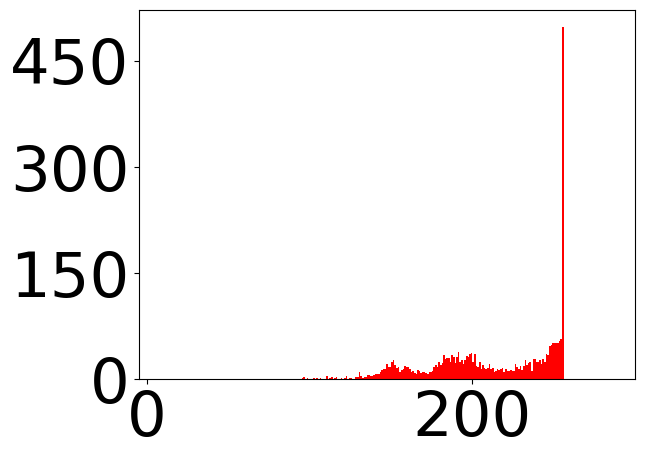}
    \includegraphics[width=\width]{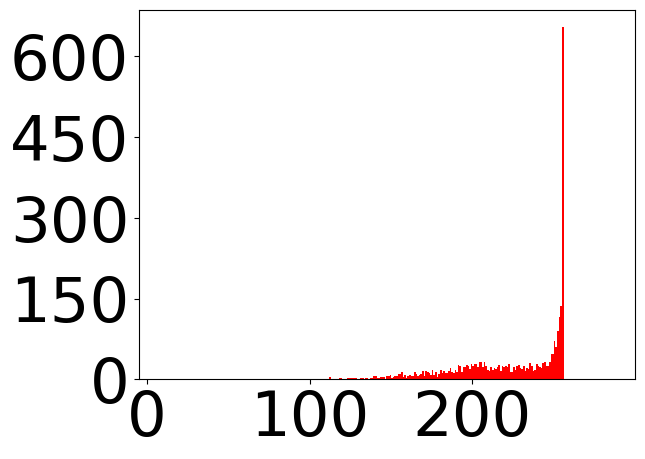}
    \includegraphics[width=\width]{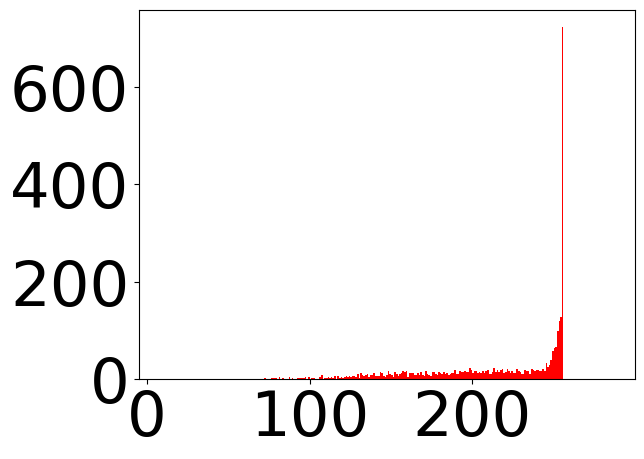}
    \includegraphics[width=\width]{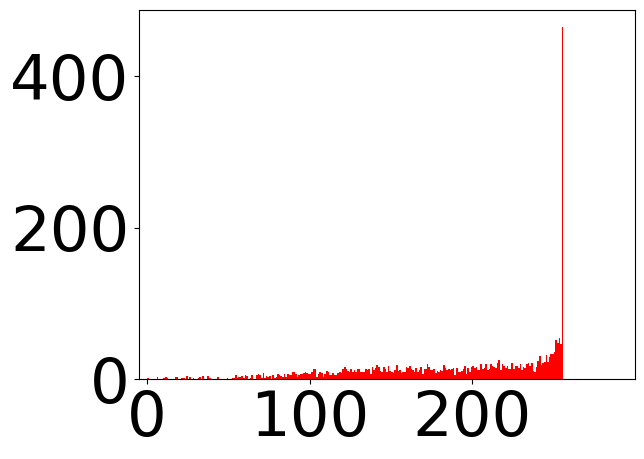}
    \includegraphics[width=\width]{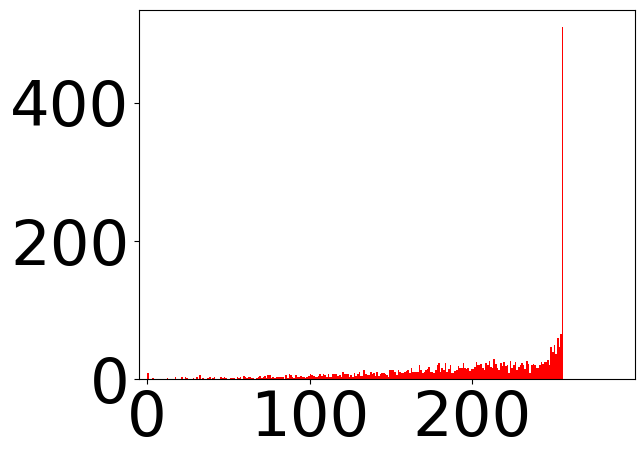}
    \includegraphics[width=\width]{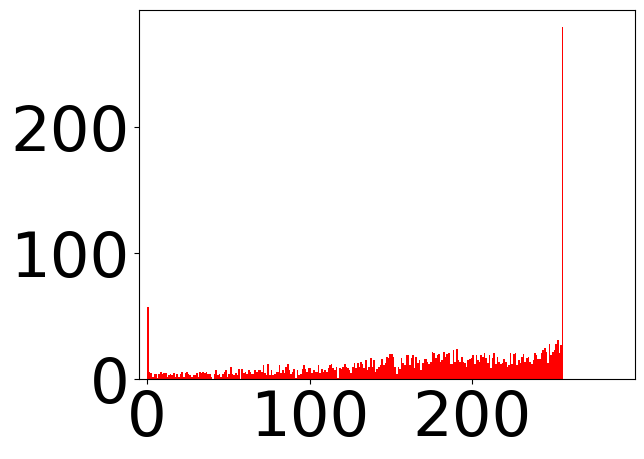}
    \includegraphics[width=\width]{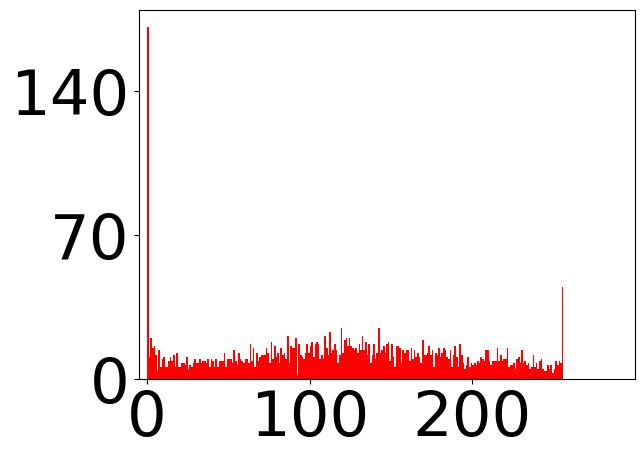}
    \includegraphics[width=\width]{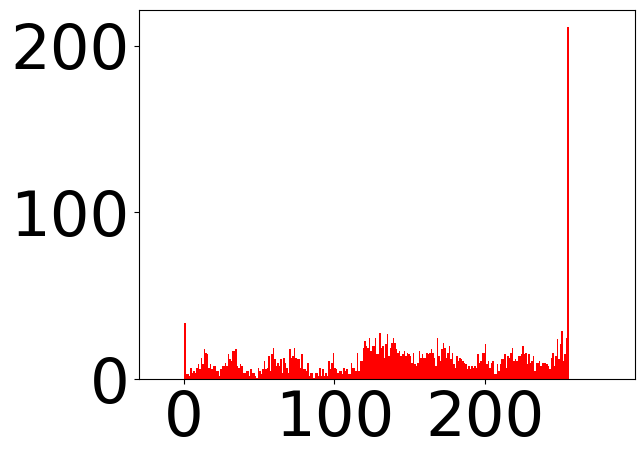}
    \includegraphics[width=\width]{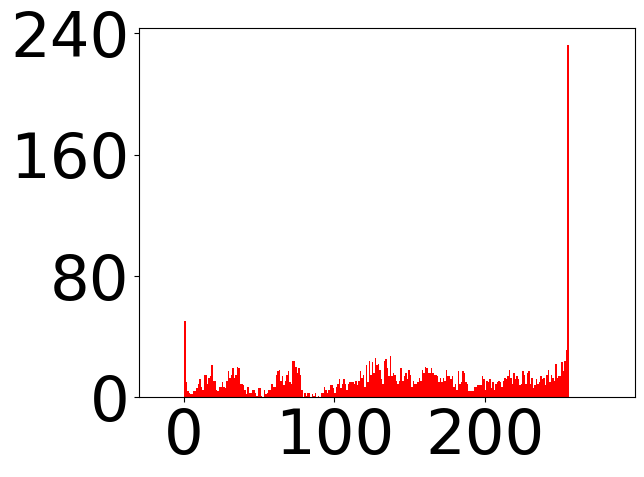}
    \includegraphics[width=\width]{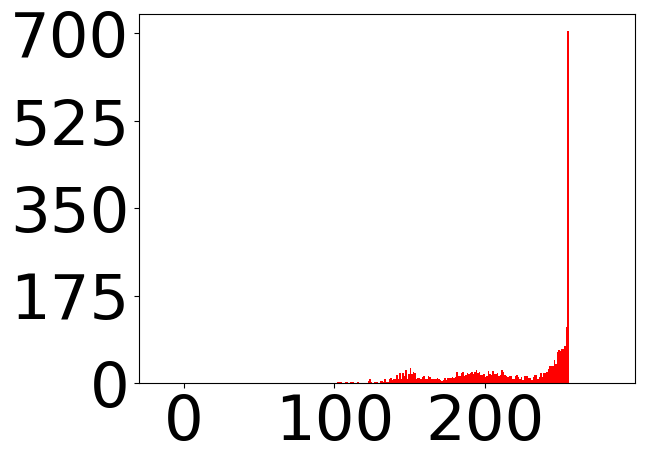}
    \includegraphics[width=\width]{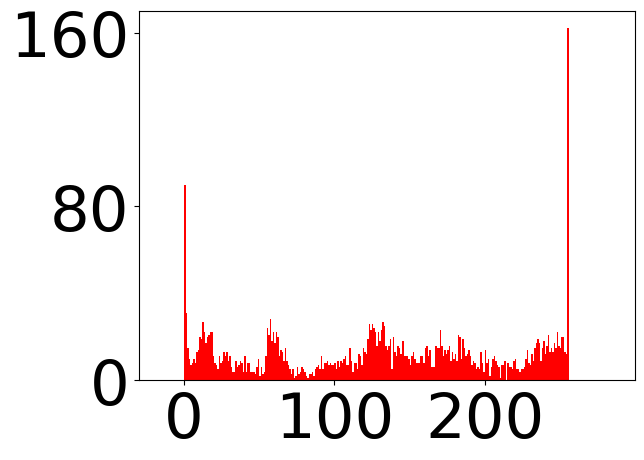}
    \\
    \includegraphics[width=\width]{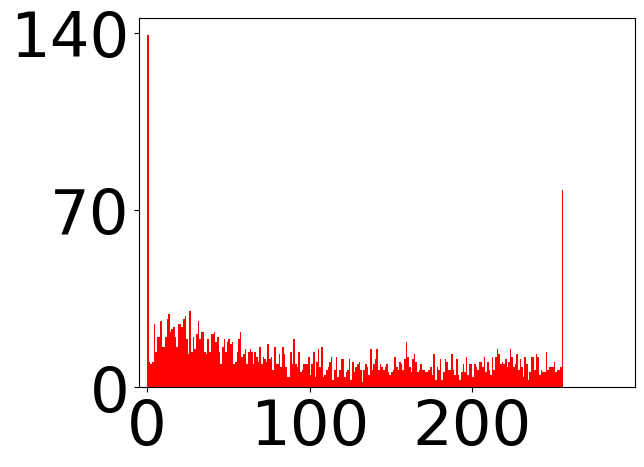}
    \includegraphics[width=\width]{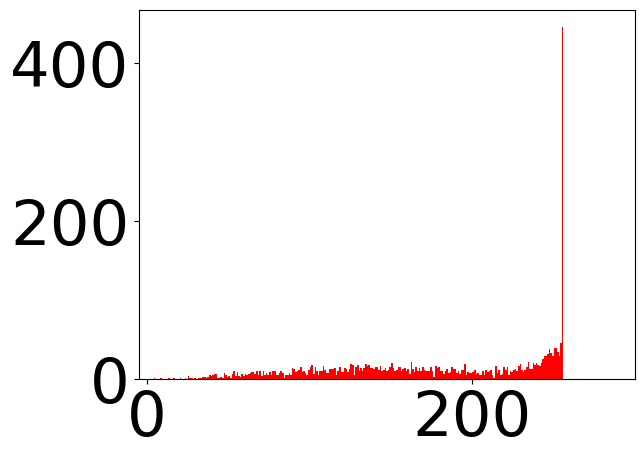}
    \includegraphics[width=\width]{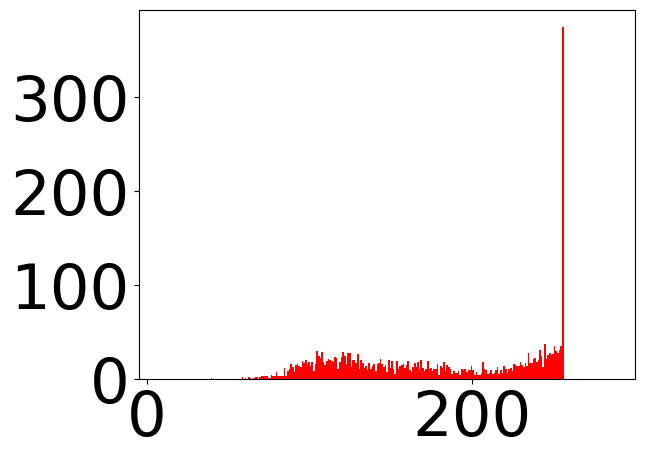}
    \includegraphics[width=\width]{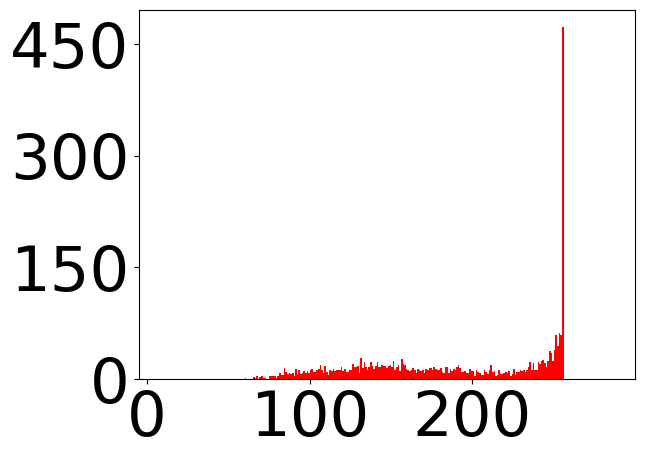}
    \includegraphics[width=\width]{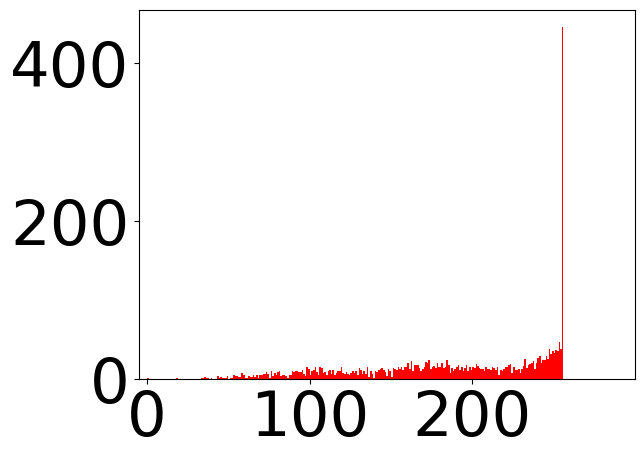}
    \includegraphics[width=\width]{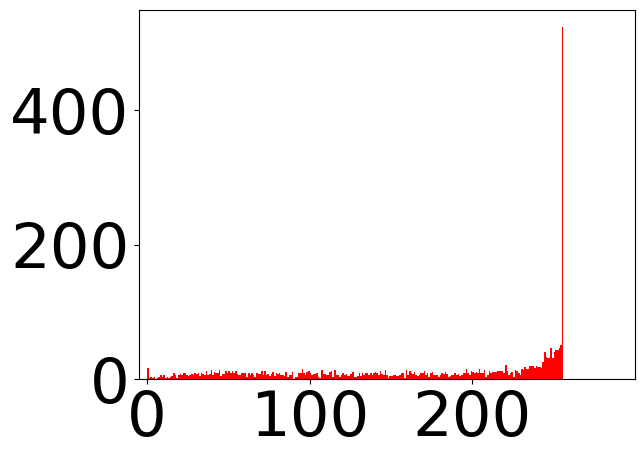}
    \includegraphics[width=\width]{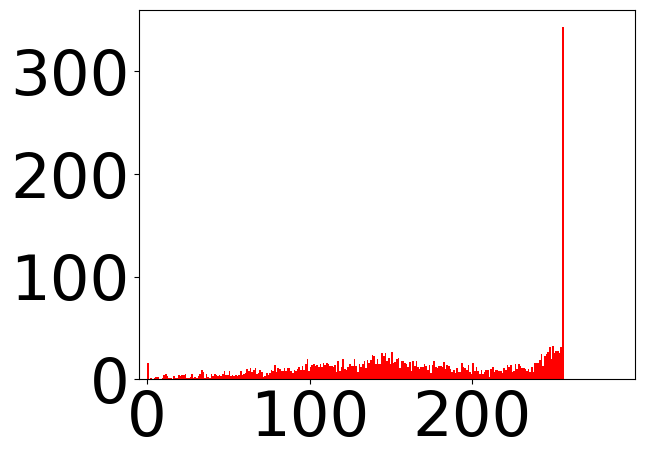}
    \includegraphics[width=\width]{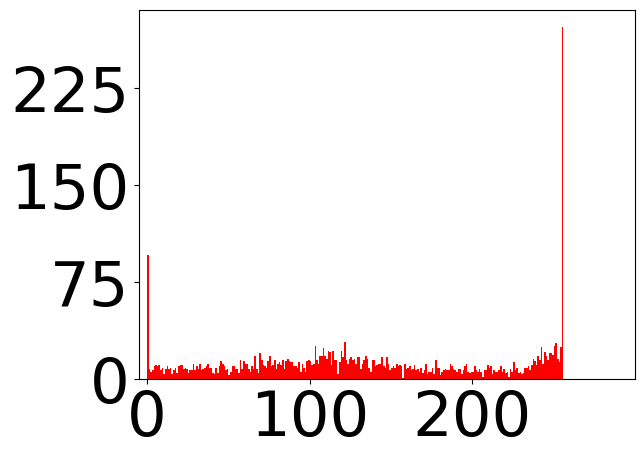}
    \includegraphics[width=\width]{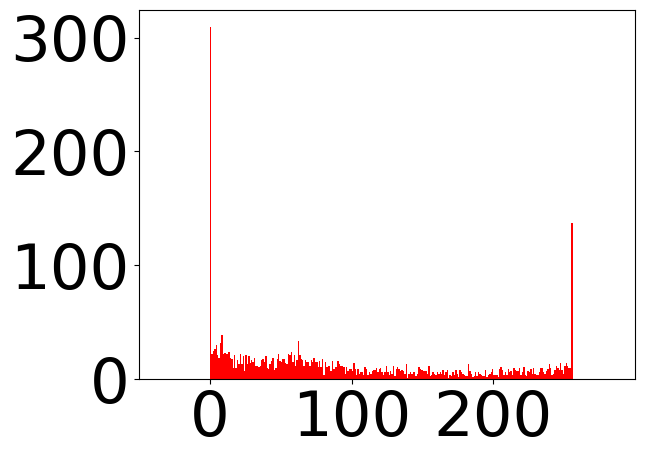}
    \includegraphics[width=\width]{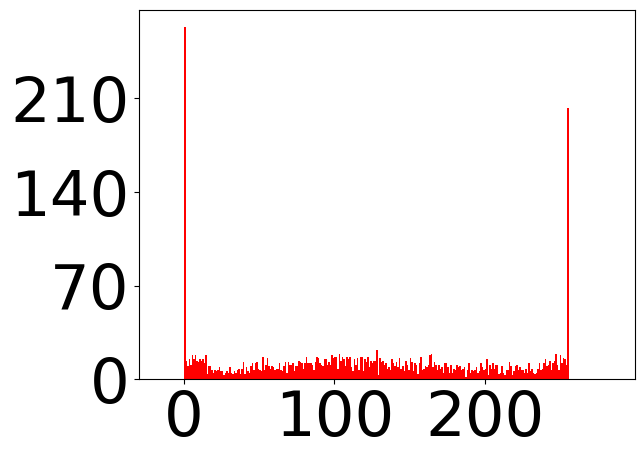}
    \includegraphics[width=\width]{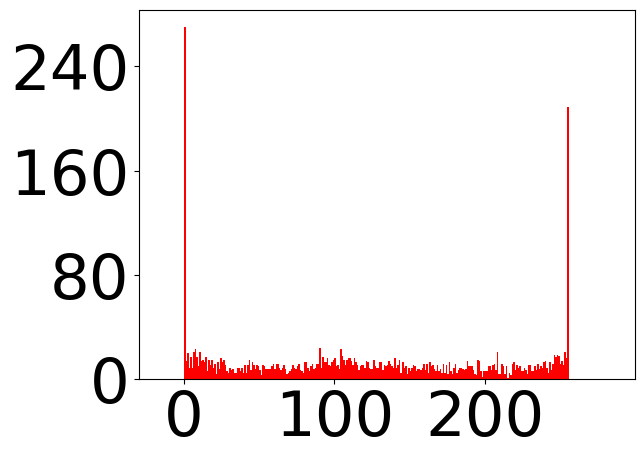}
    \includegraphics[width=\width]{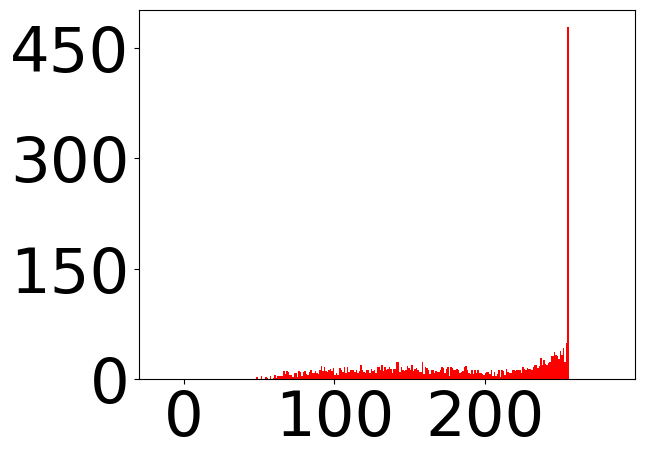}
    \includegraphics[width=\width]{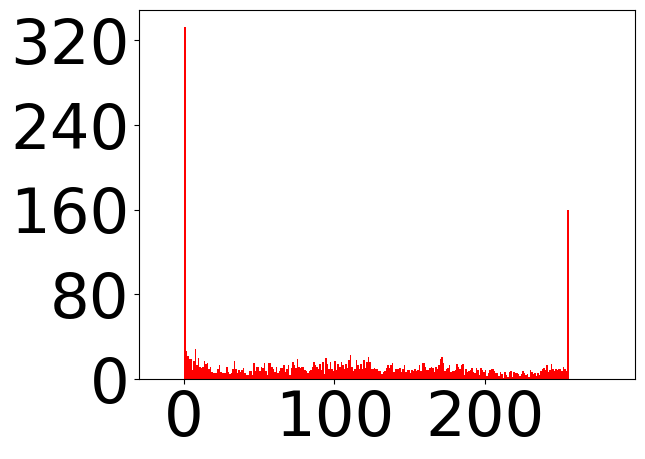}
    \\
    \includegraphics[width=\width]{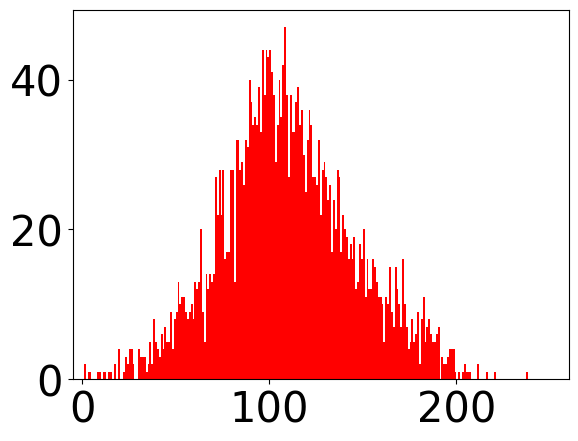}
    \includegraphics[width=\width]{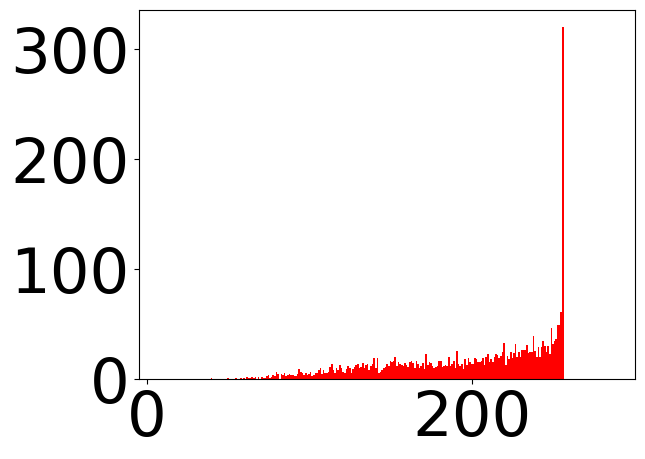}
    \includegraphics[width=\width]{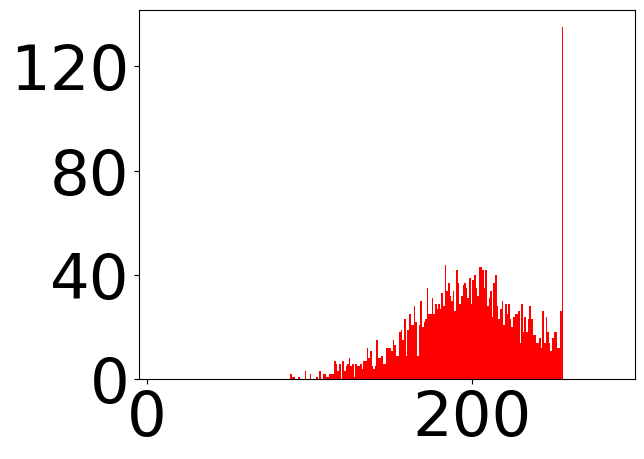}
    \includegraphics[width=\width]{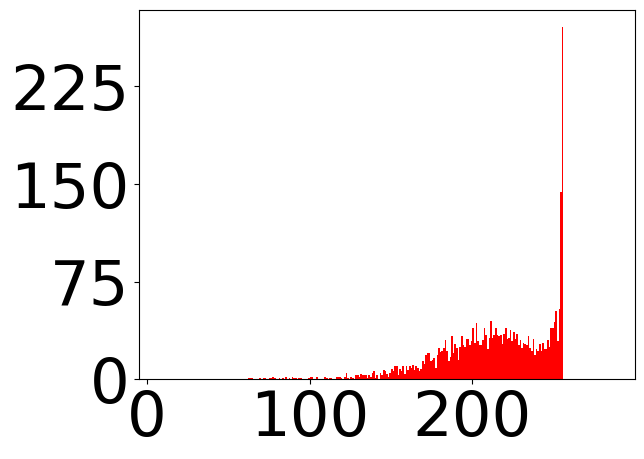}
    \includegraphics[width=\width]{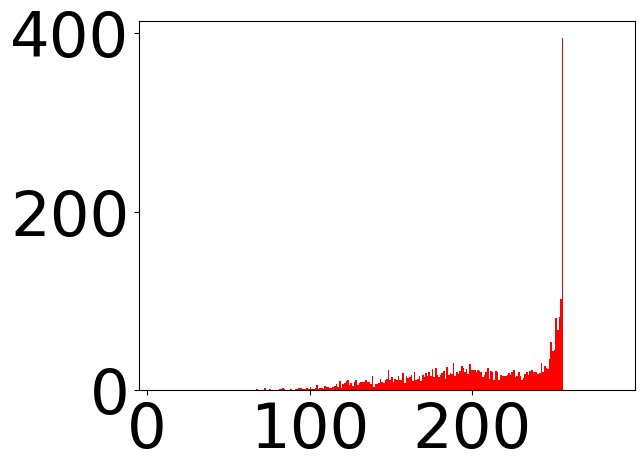}
    \includegraphics[width=\width]{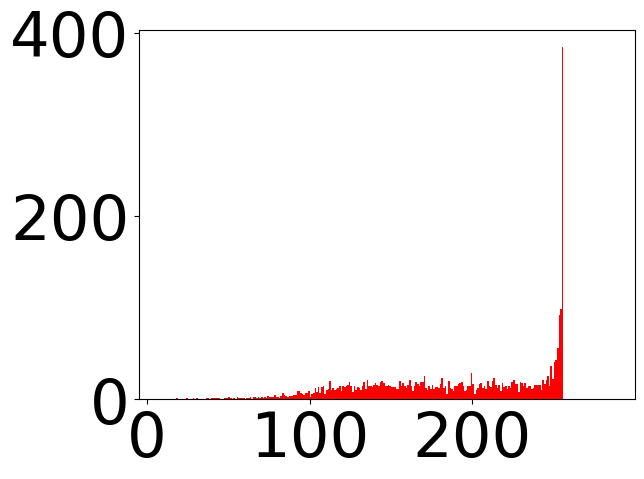}
    \includegraphics[width=\width]{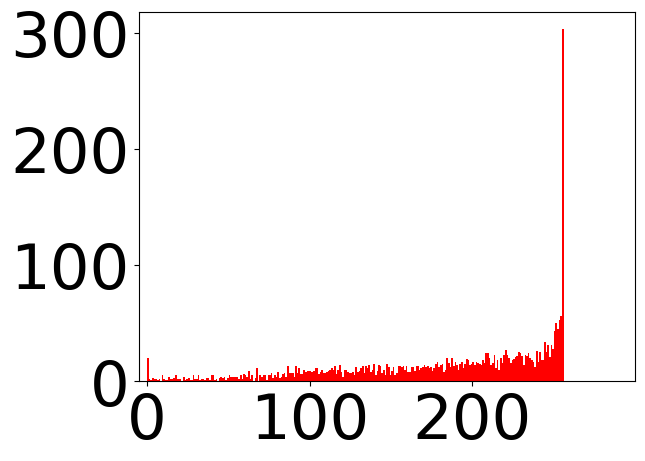}
    \includegraphics[width=\width]{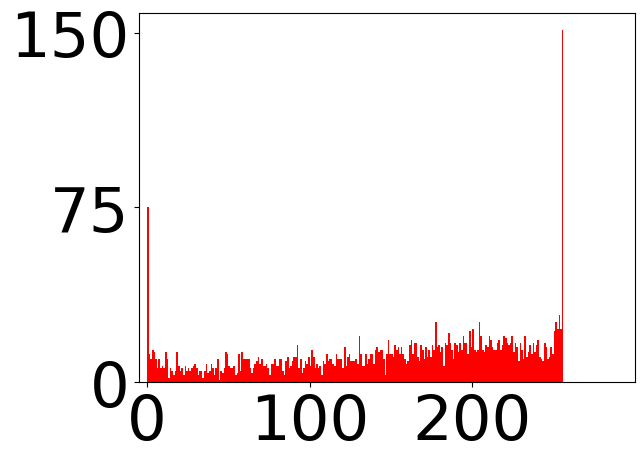}
    \includegraphics[width=\width]{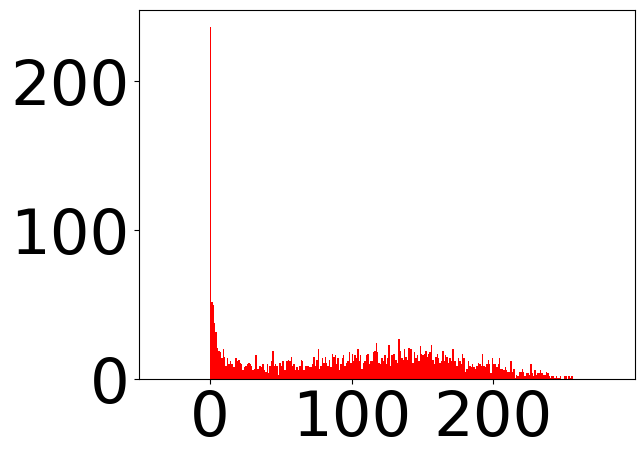}
    \includegraphics[width=\width]{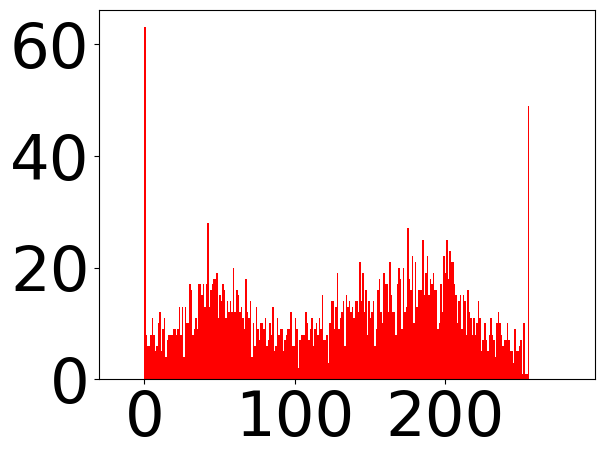}
    \includegraphics[width=\width]{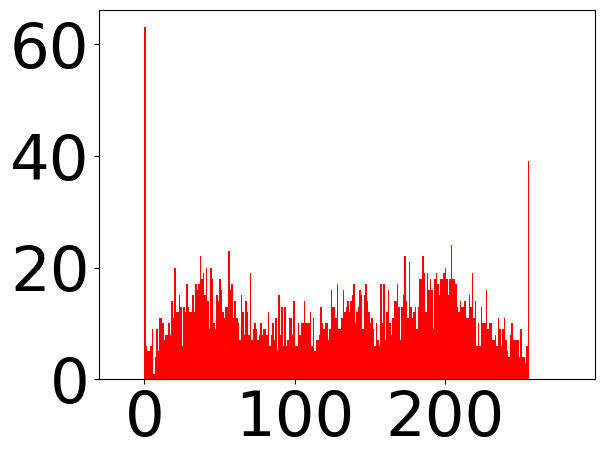}
    \includegraphics[width=\width]{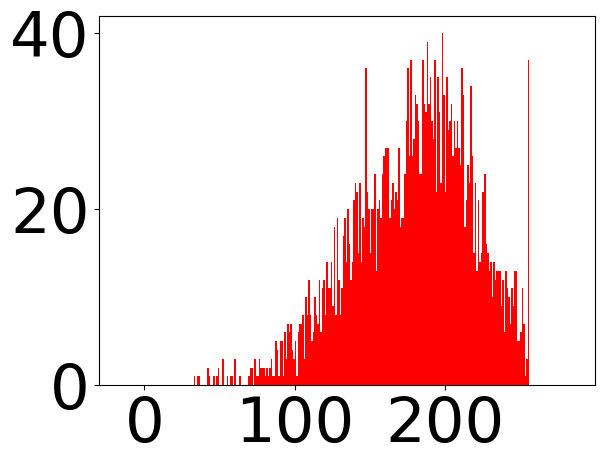}
    \includegraphics[width=\width]{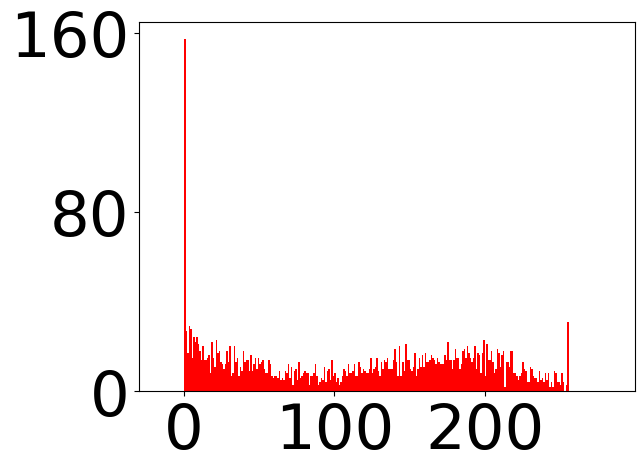}
    
    \caption{The histograms of the Figure 2 images of main paper in the same order. $1^{st}$ column is the histograms for the original images. Similarly, the later columns are the histograms for images after the NUI attack by Mask $1$ to Mask $12$, respectively.}
    \label{fig:hist}
\end{figure*}

\autoref{fig:hist} shows the effect of various masks on the image pixel value distribution using histograms. 
The $1^{st}$ column contains the histograms corresponding to the original images used in Figure 2 of main paper. Similarly, the later columns from left to right contain the histograms for images after the NUI attack by Mask $1$ to Mask $12$, respectively. 
The change in the distribution of the pixel values can be observed. We generate all the images using positive values of $k$, thus the number of pixels having higher pixel values has increased causing the histogram to be right-shifted. Masks that cause both brightness and darkness in the image generate histograms equally distributed throughout the axis.  
Following Figure 2 of main paper and Figure \ref{fig:hist} of Supplementary, we observed that though the histograms contain severely brighter pixels, the semantic meaning is intact and the histograms are similar for the majority of the images and thus generalize the NUI attack technique.

\begin{figure*}[t]
    \centering
    \newcommand\width{0.16\textwidth}
    \includegraphics[width=\width]{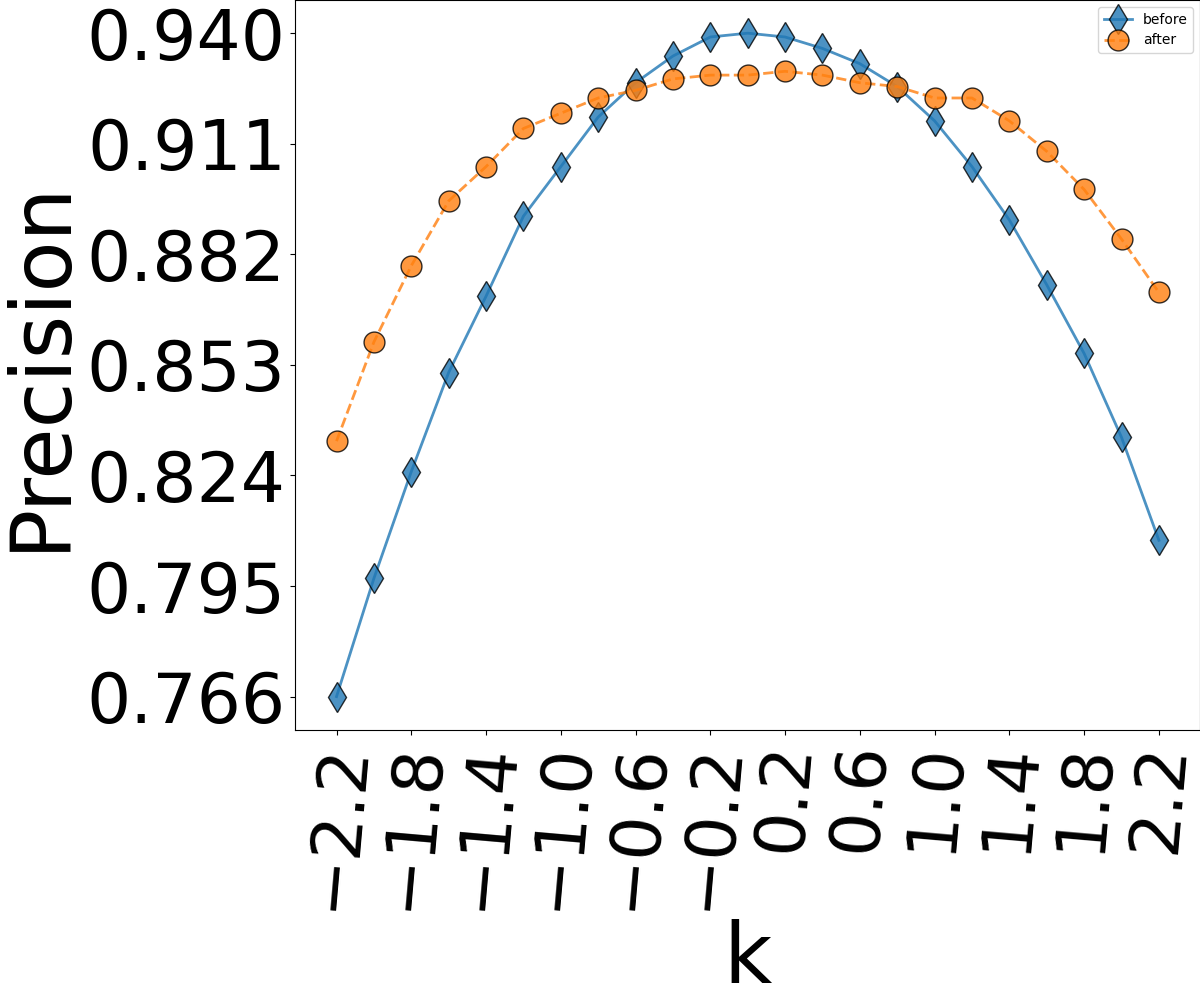}
    \includegraphics[width=\width]{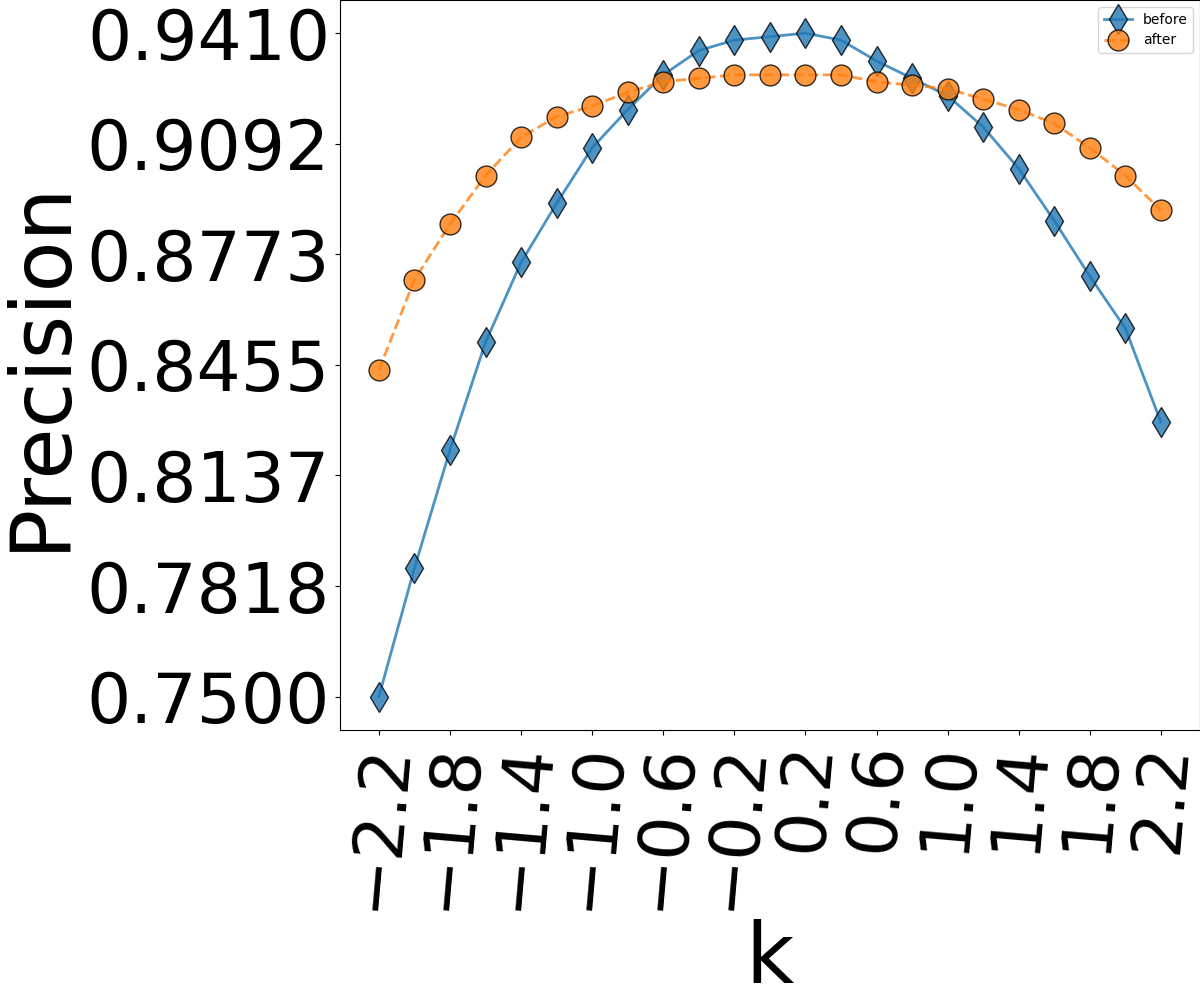}
    \includegraphics[width=\width]{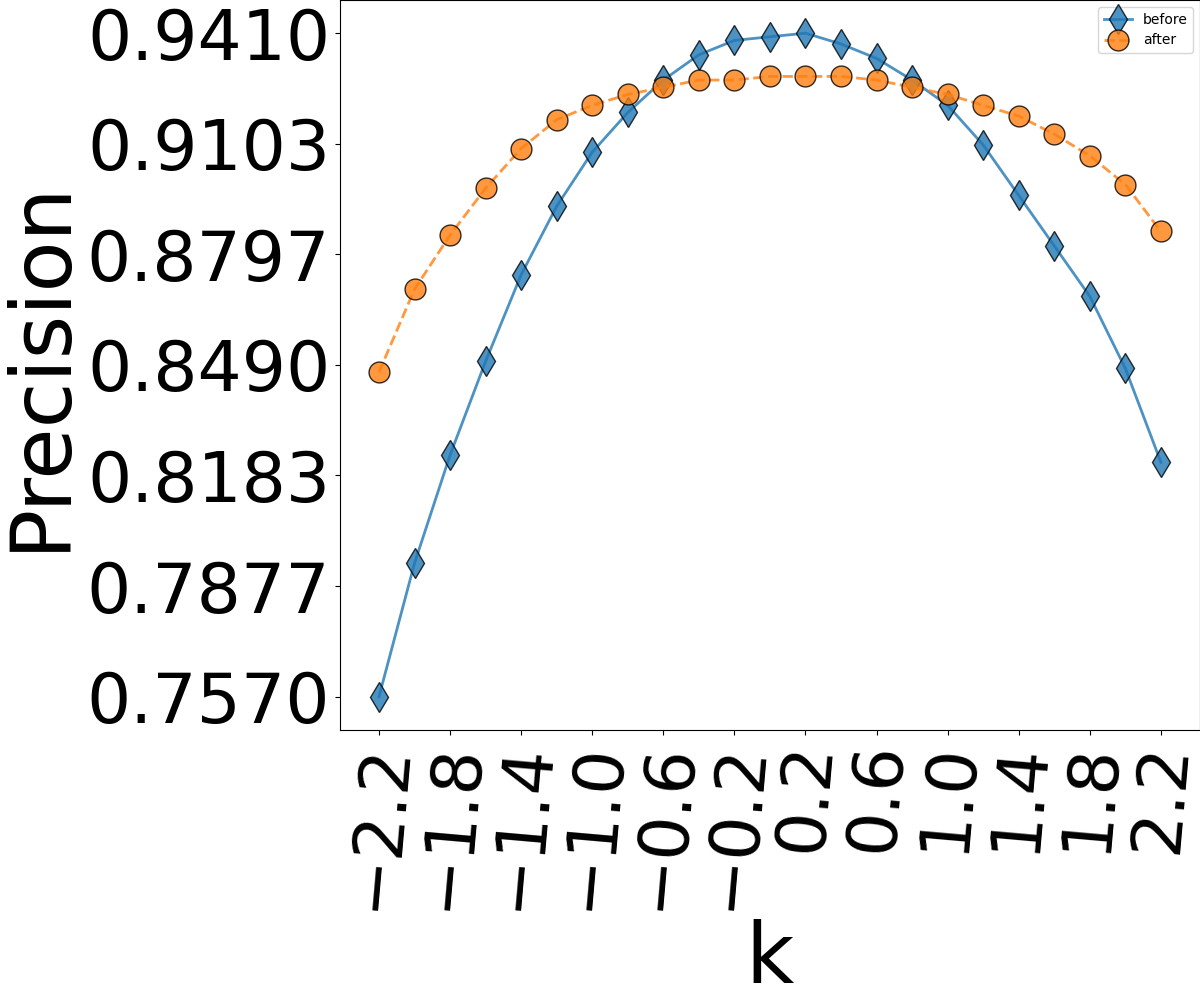}
    \includegraphics[width=\width]{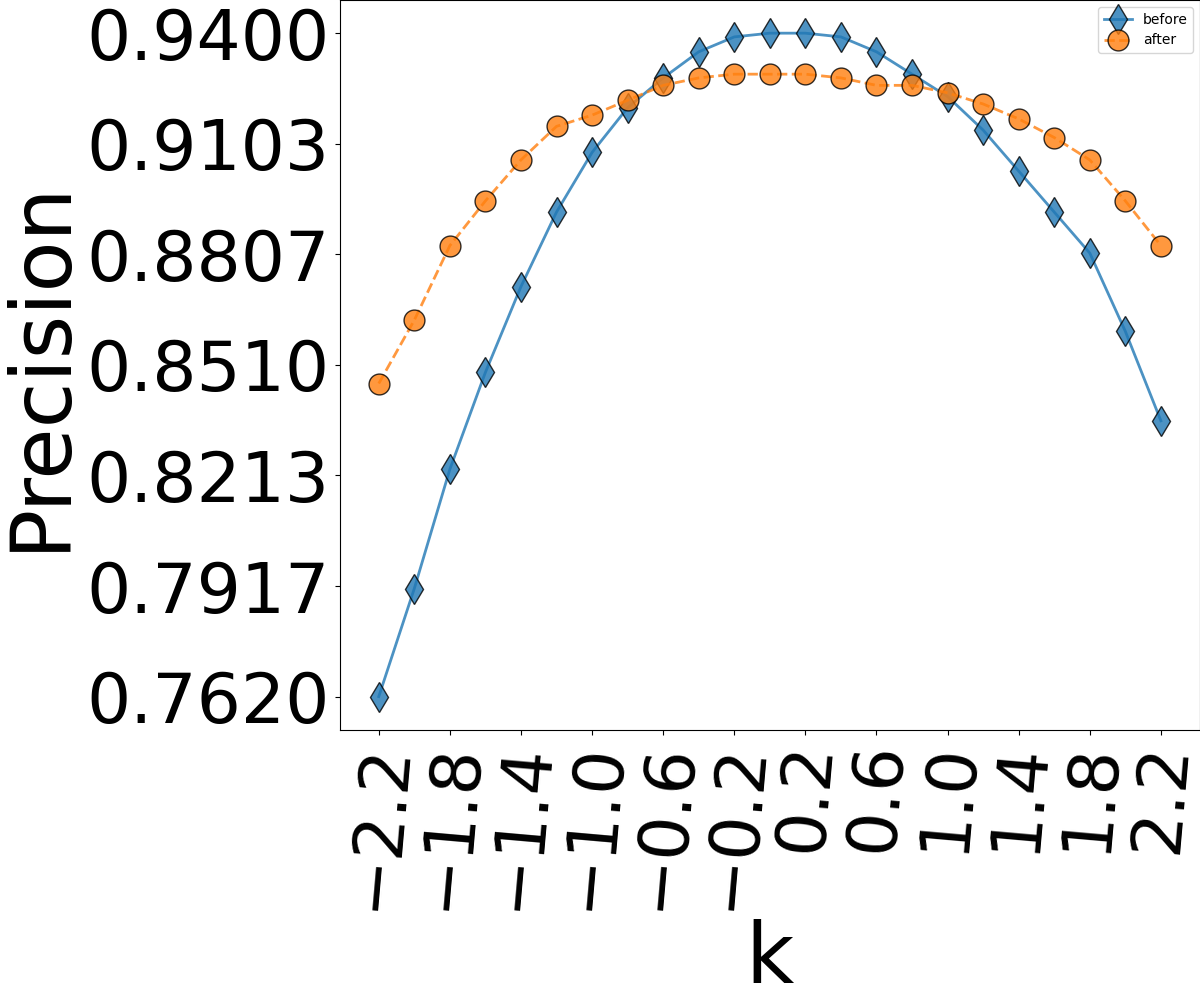}
    \includegraphics[width=\width]{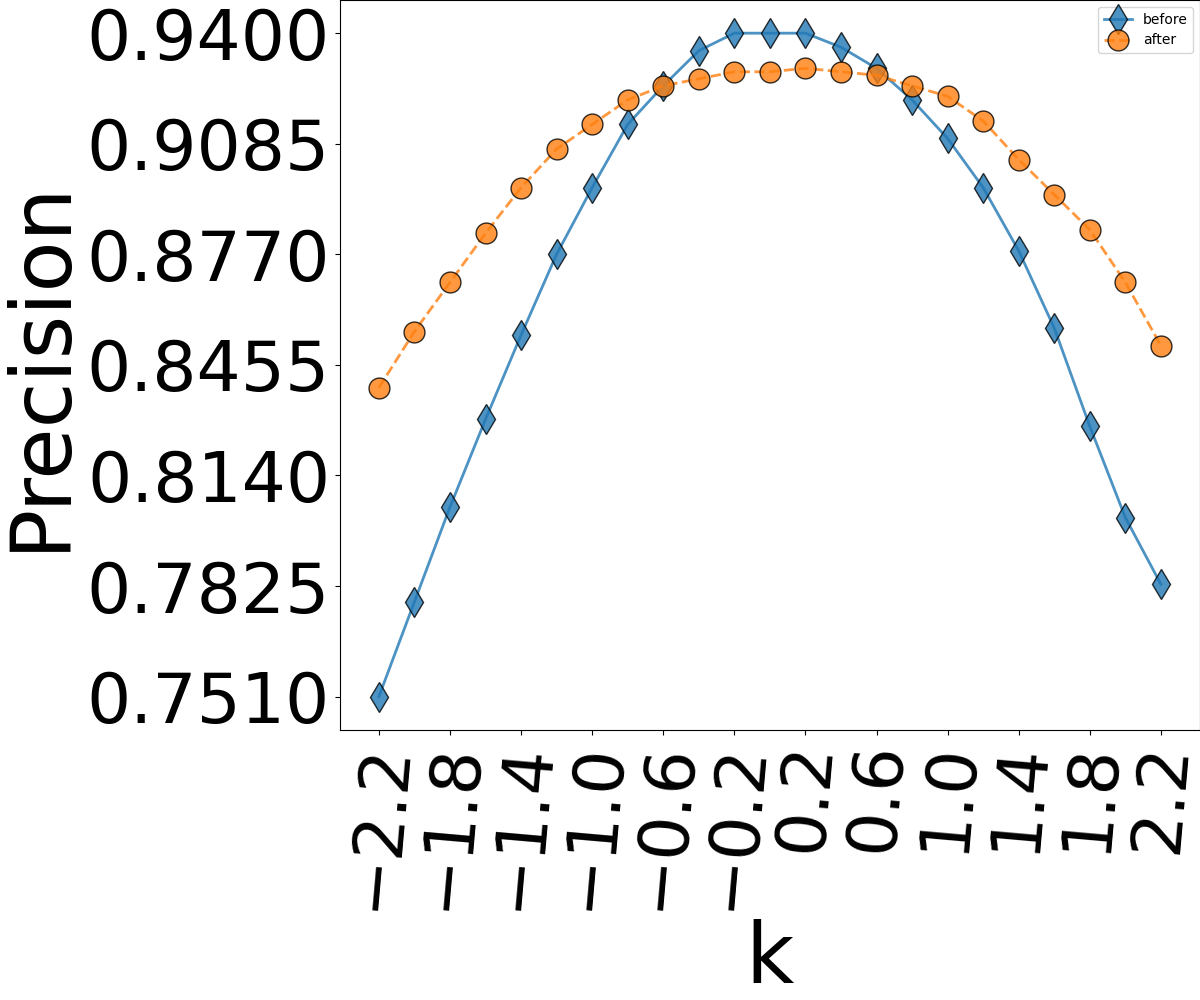}
    \includegraphics[width=\width]{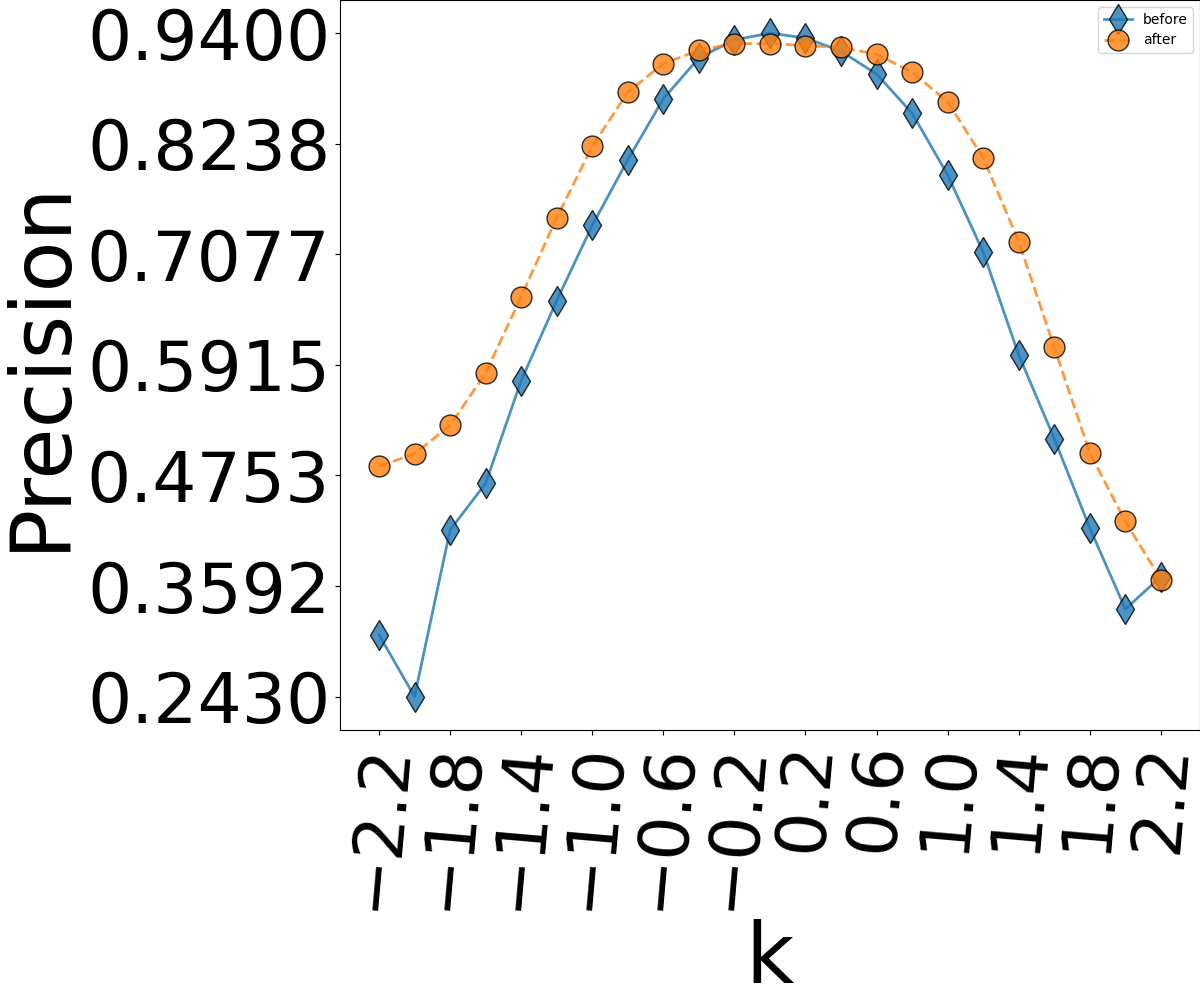}
    \includegraphics[width=\width]{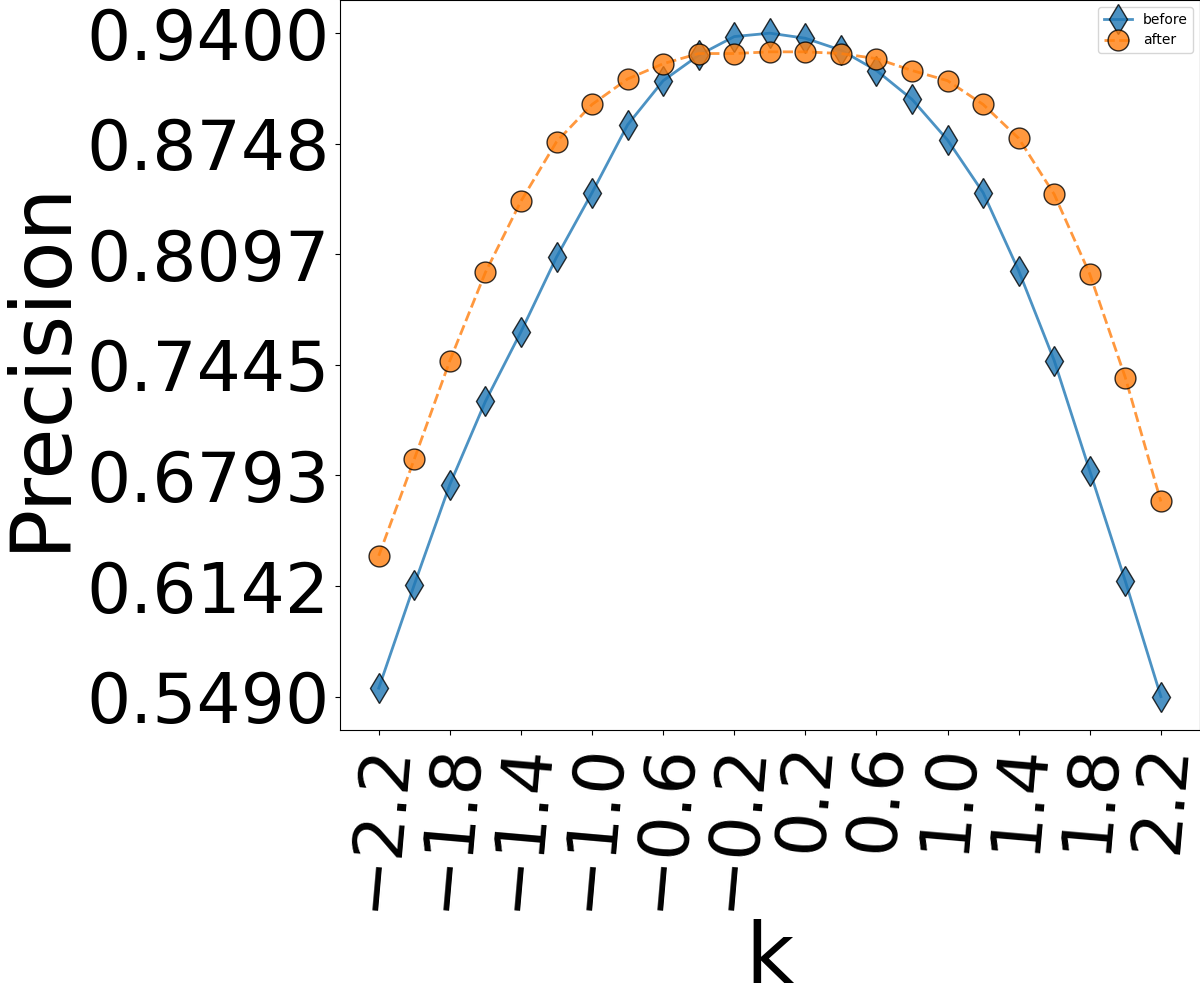}
    \includegraphics[width=\width]{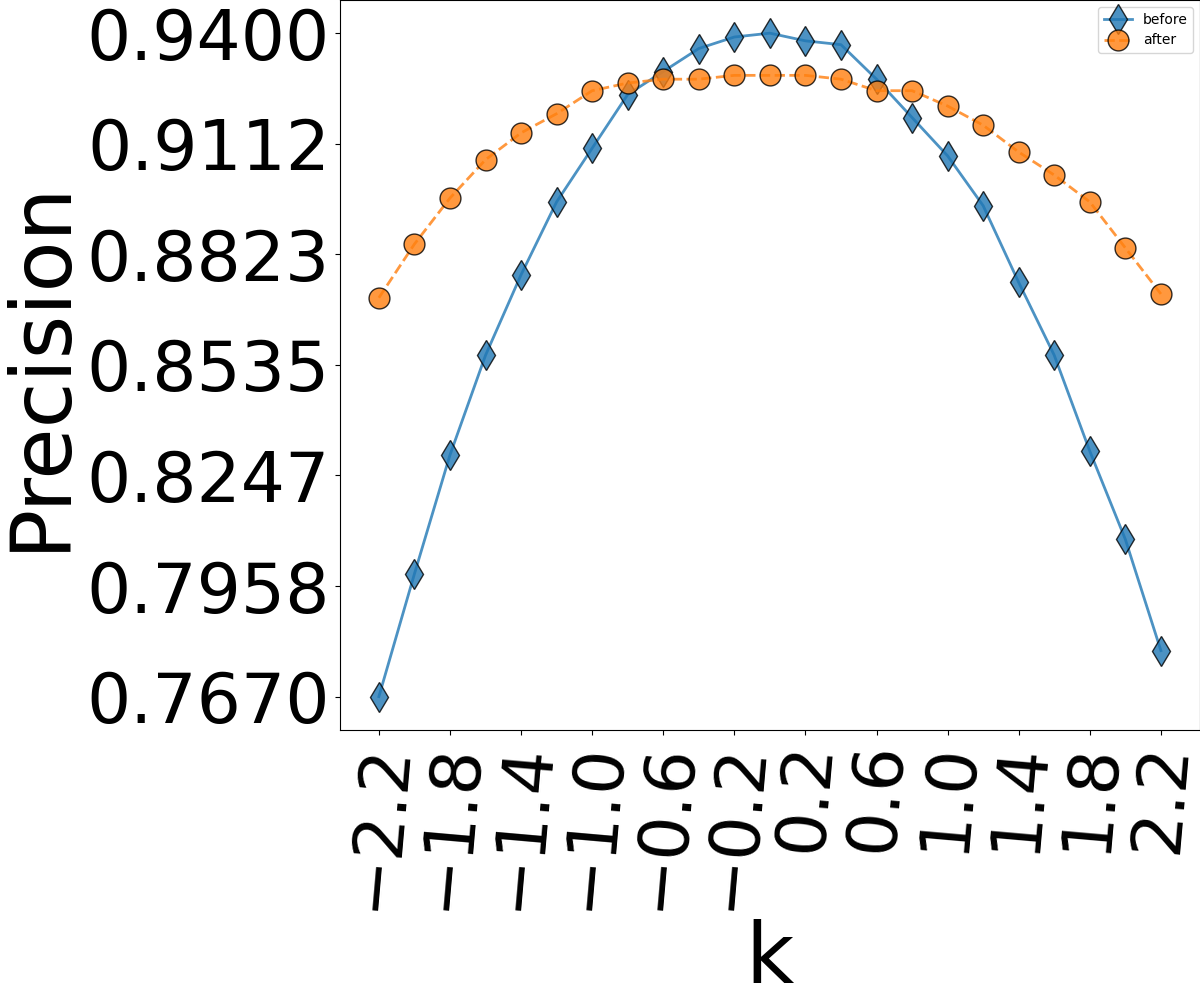}
    \includegraphics[width=\width]{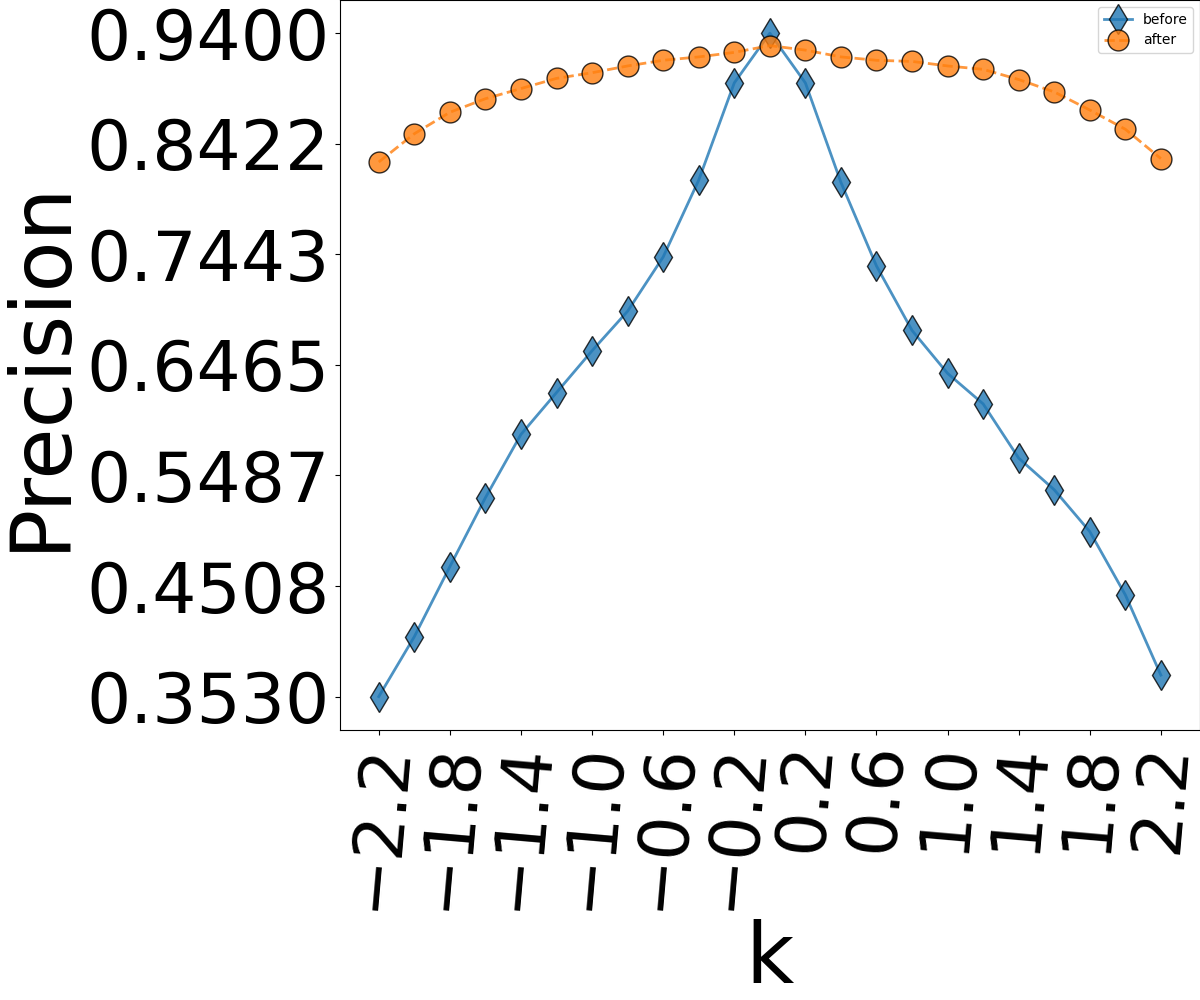}
    \includegraphics[width=\width]{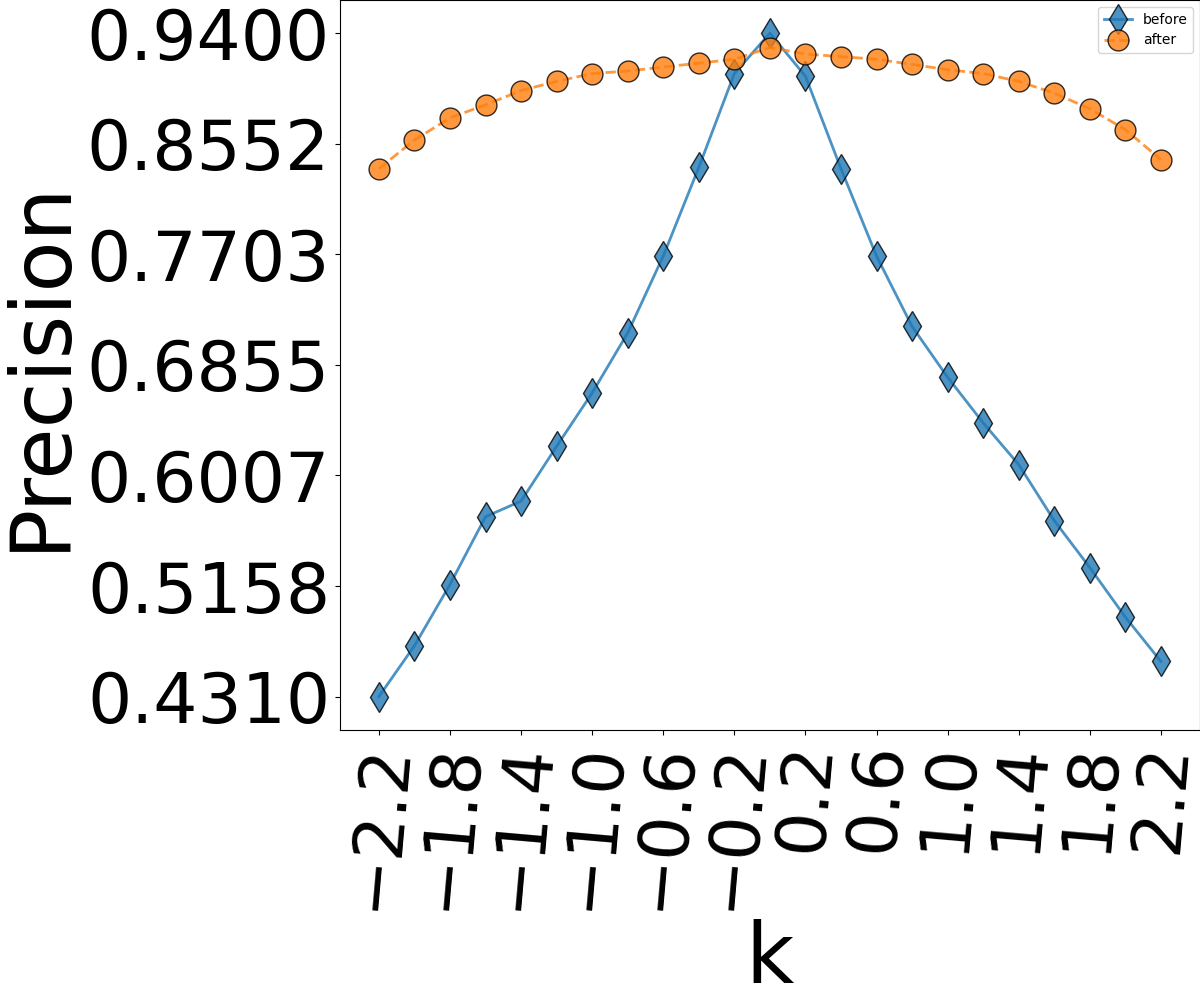}
    \includegraphics[width=\width]{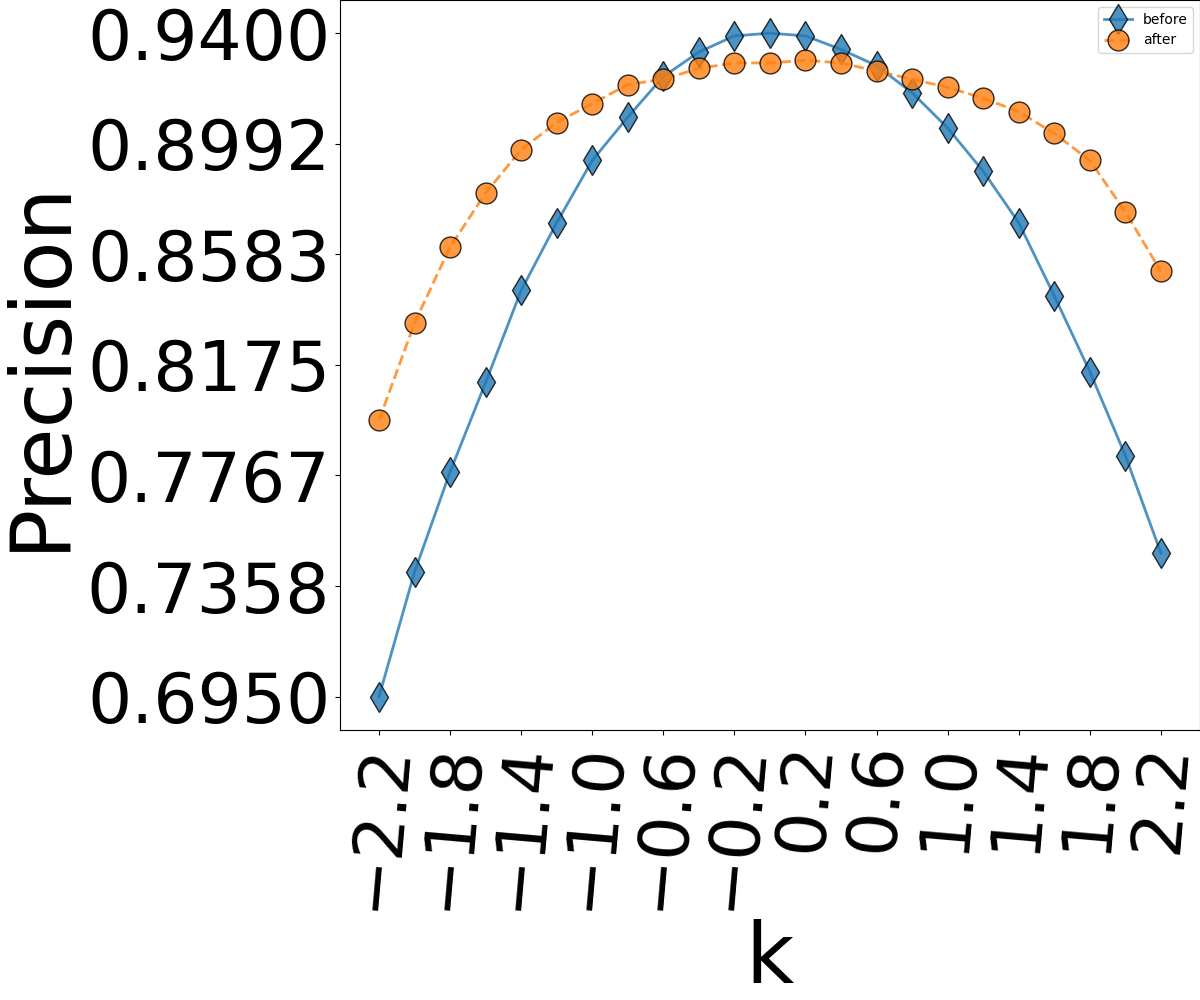}
    \includegraphics[width=\width]{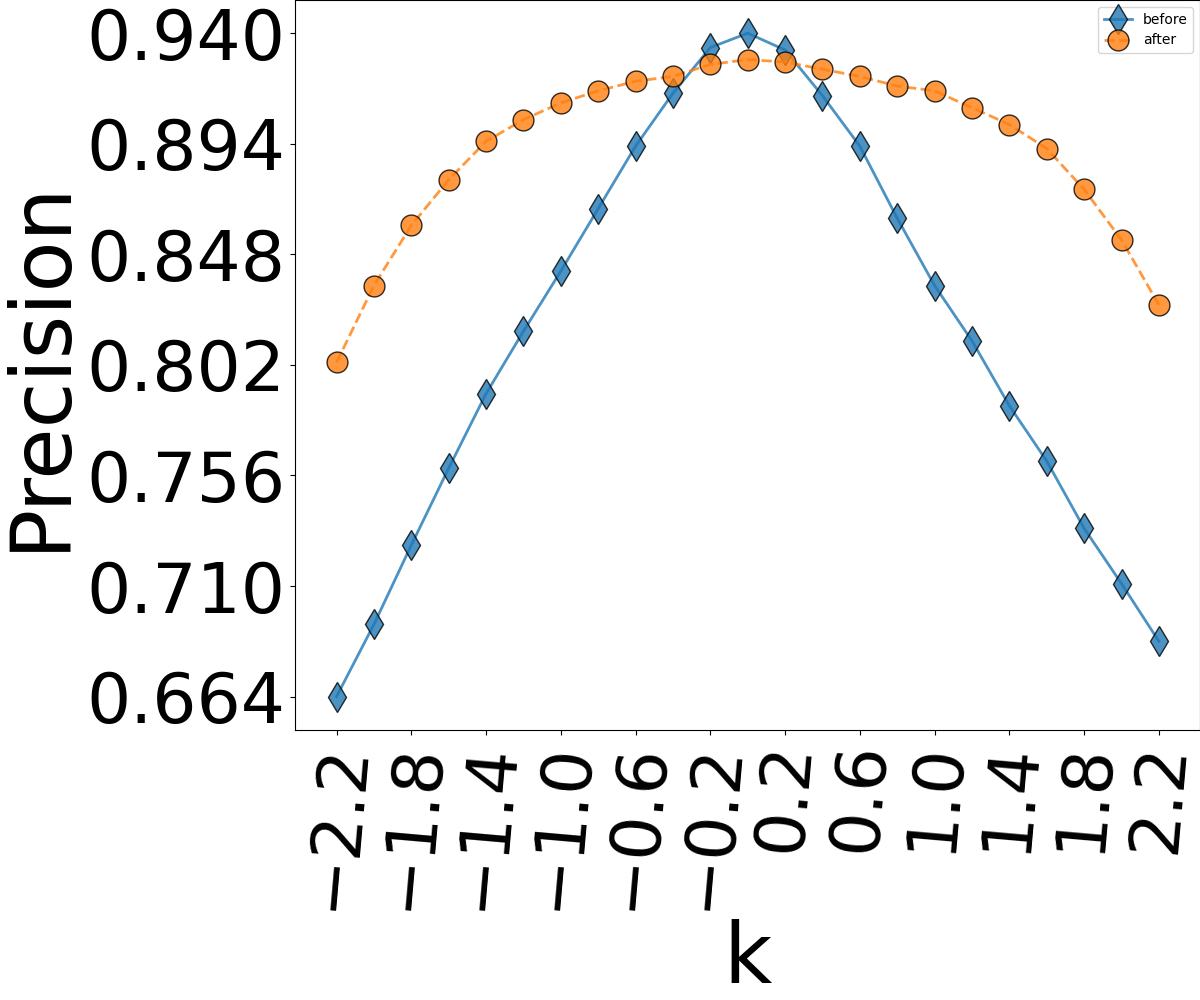}
    \\
    \includegraphics[width=\width]{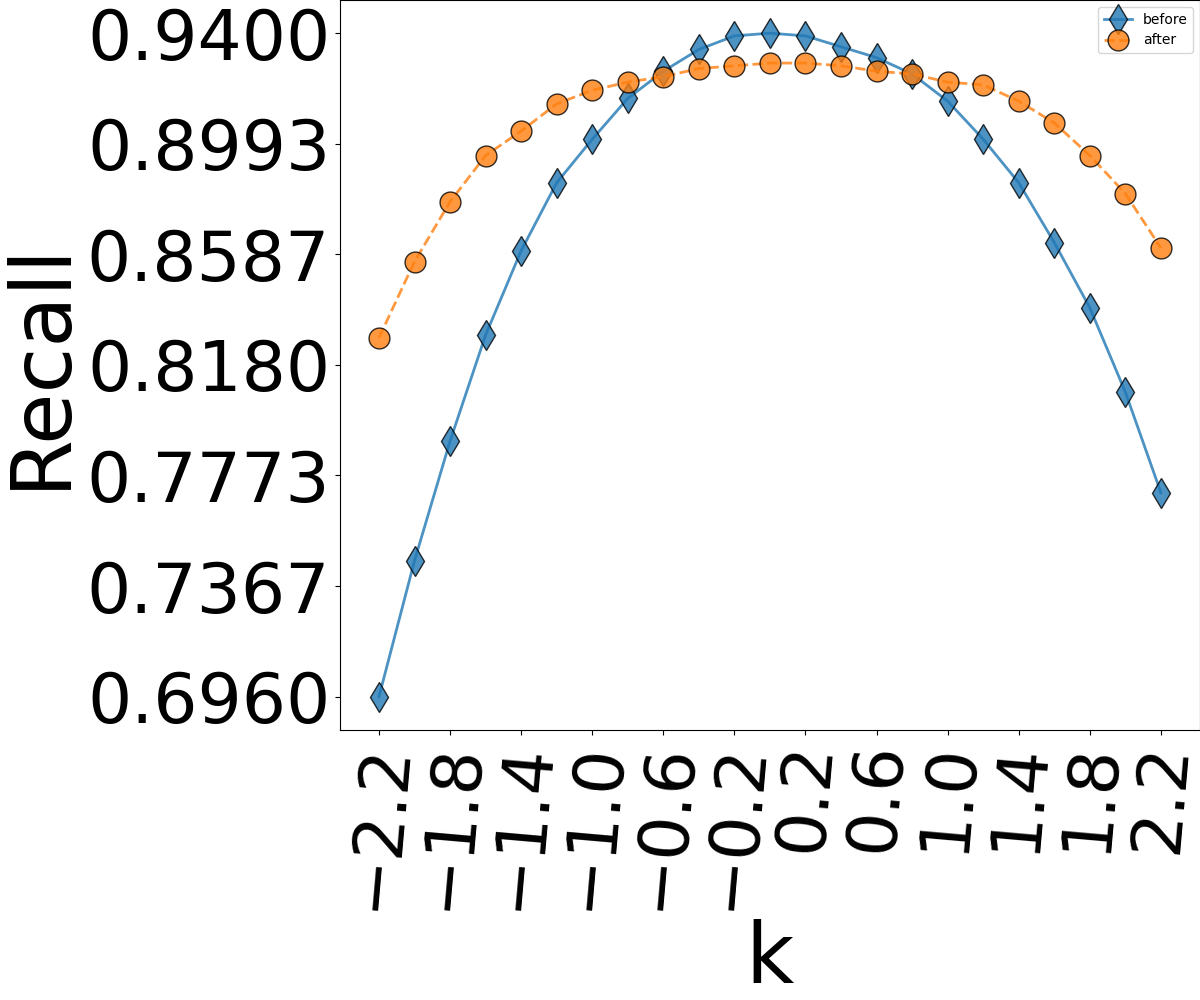}
    \includegraphics[width=\width]{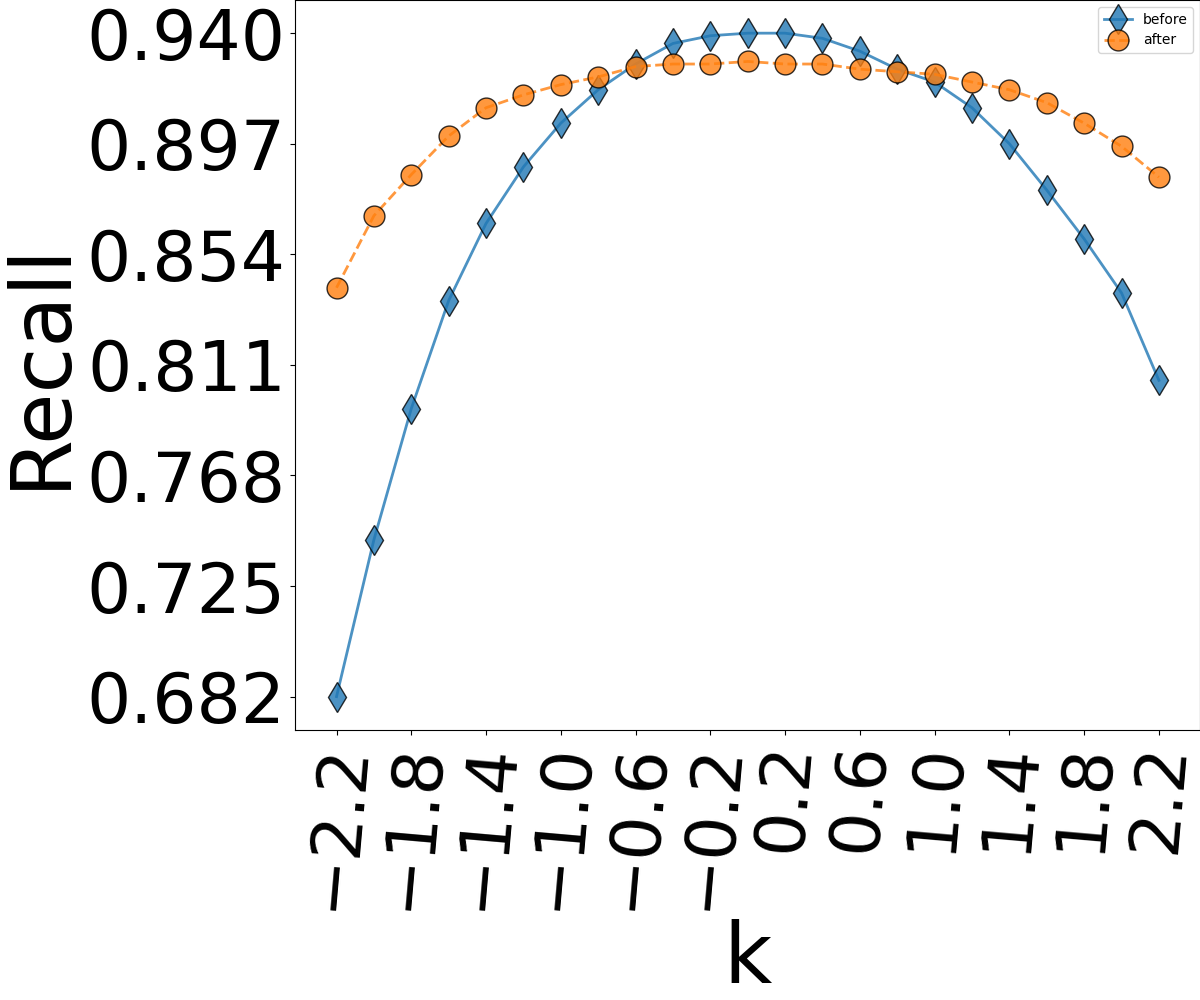}
    \includegraphics[width=\width]{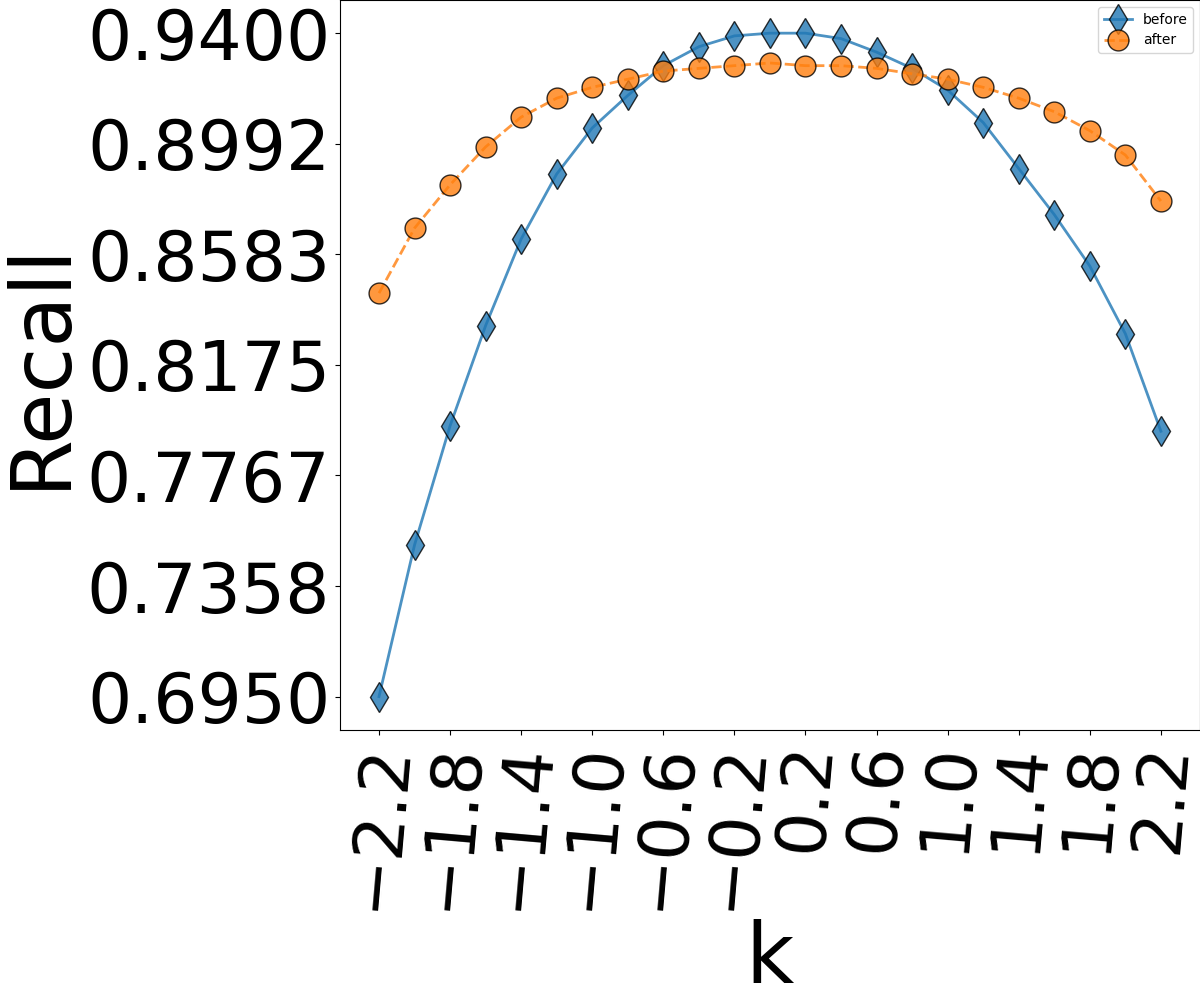}
    \includegraphics[width=\width]{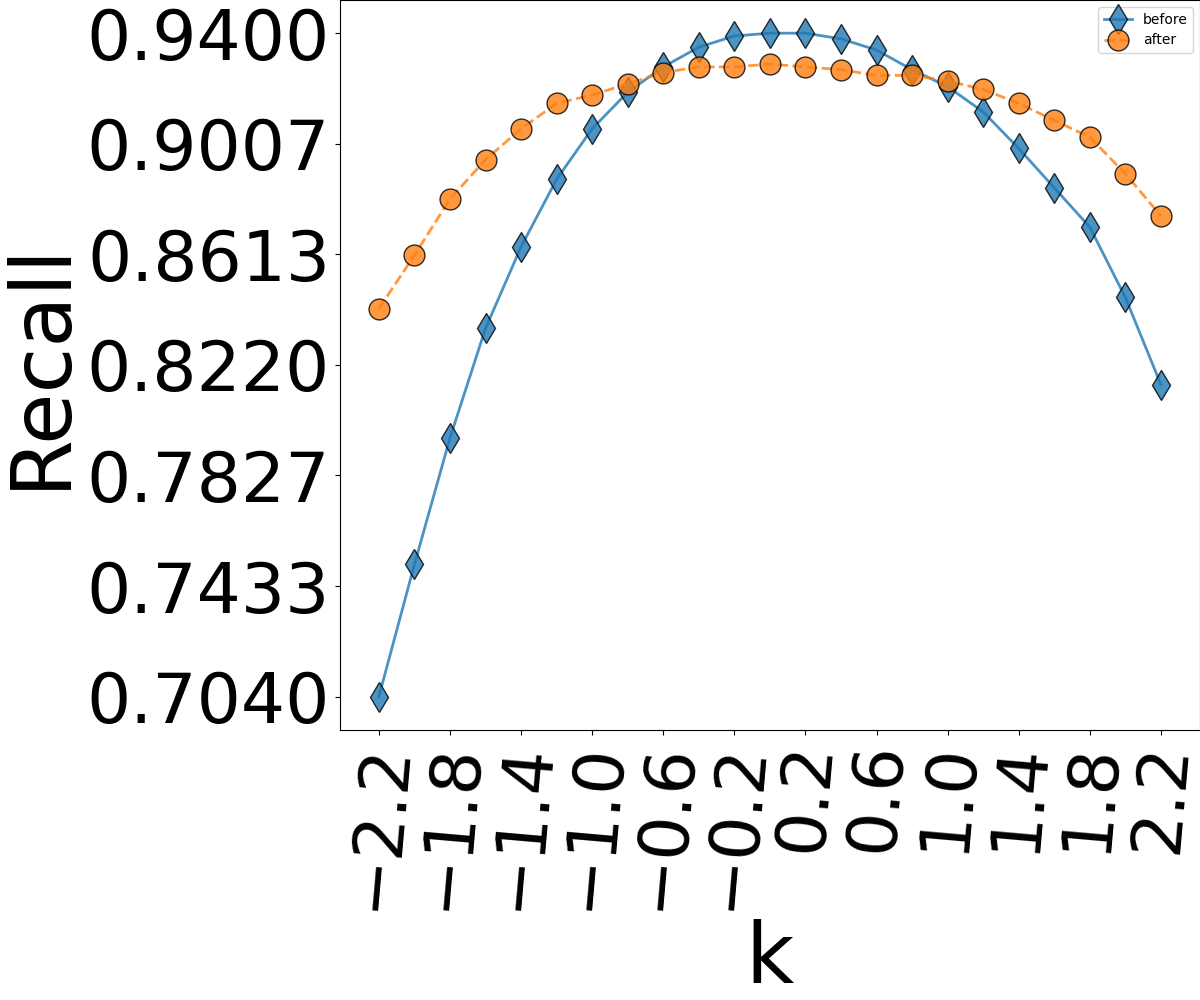}
    \includegraphics[width=\width]{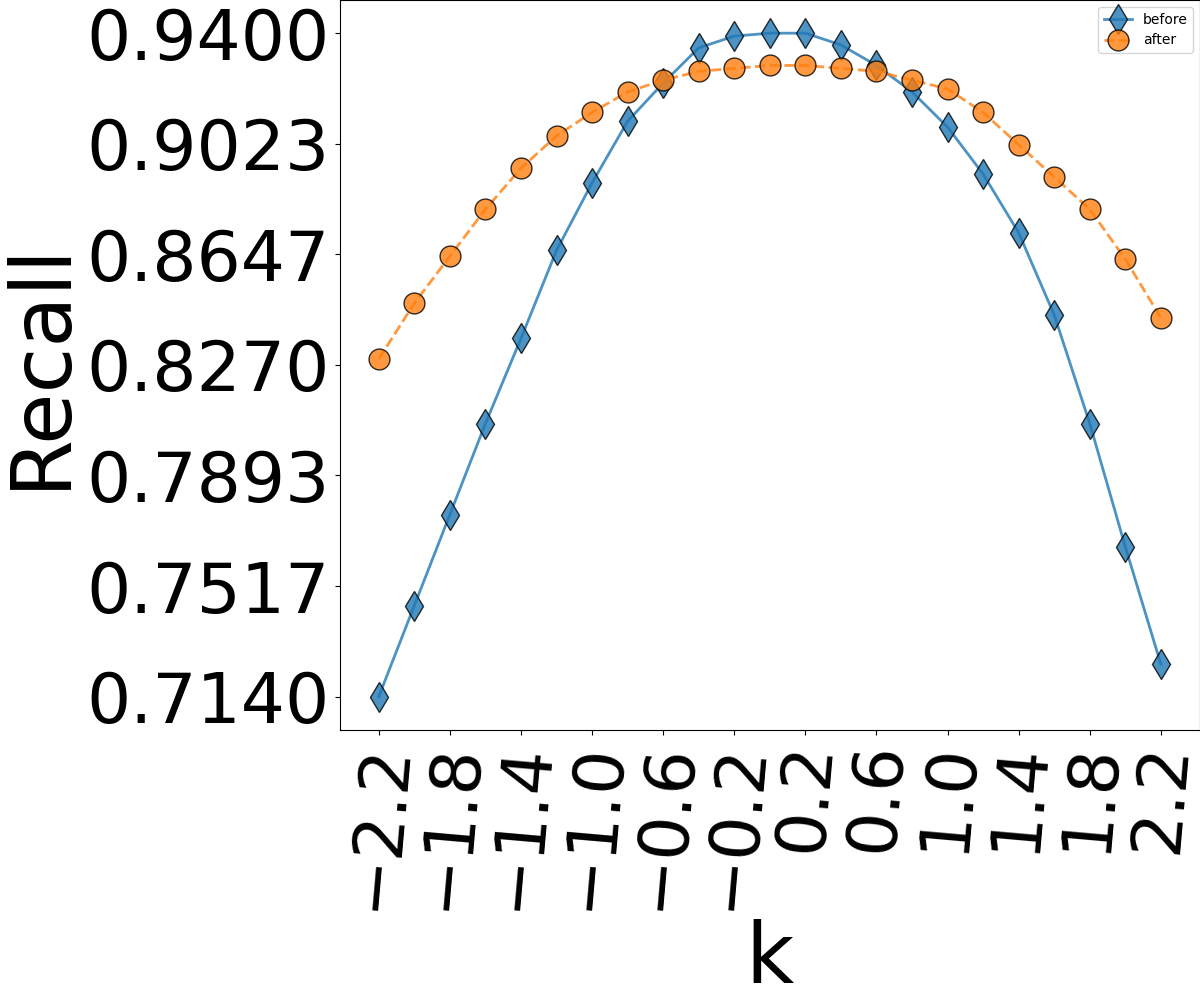}
    \includegraphics[width=\width]{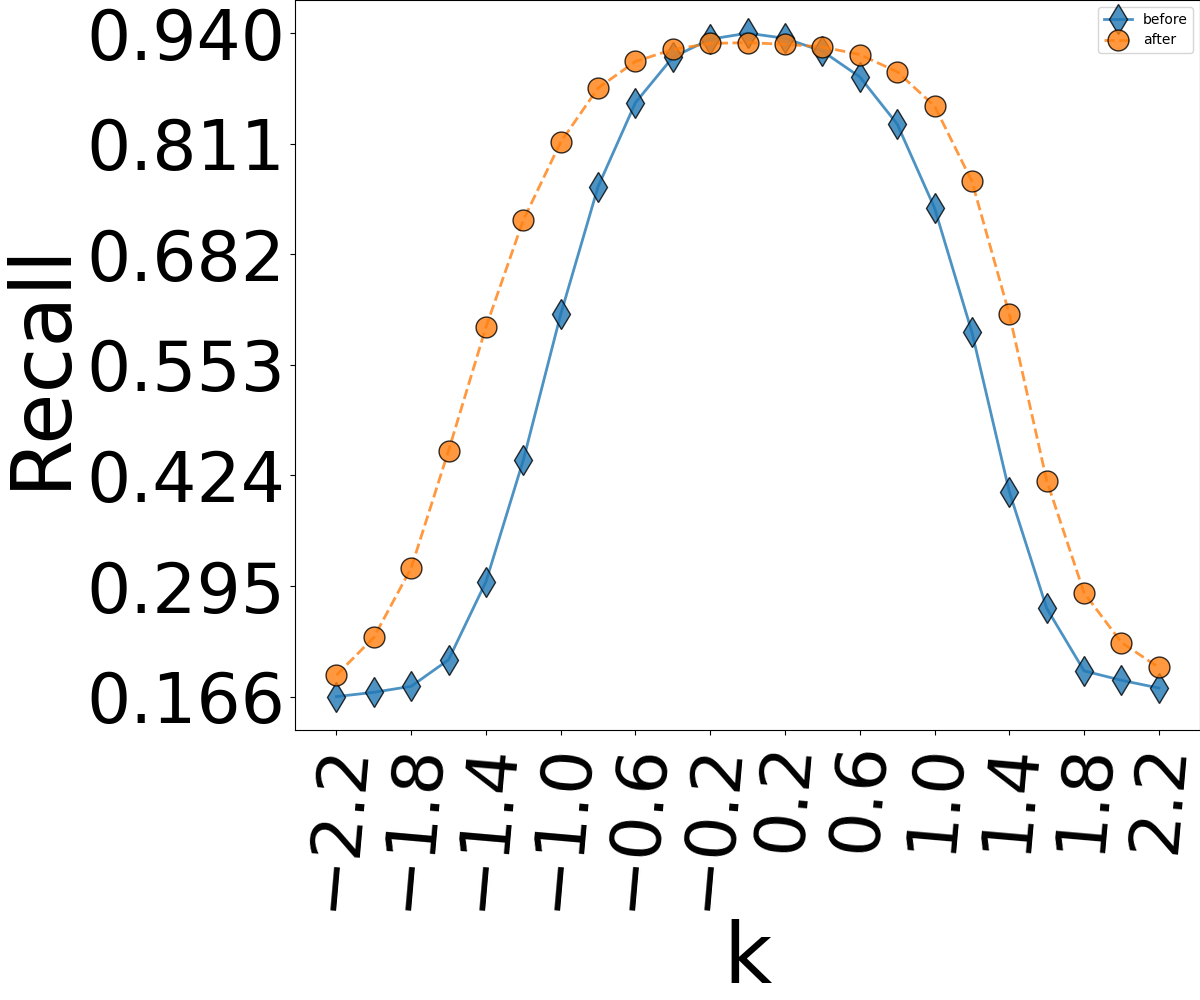}
    \includegraphics[width=\width]{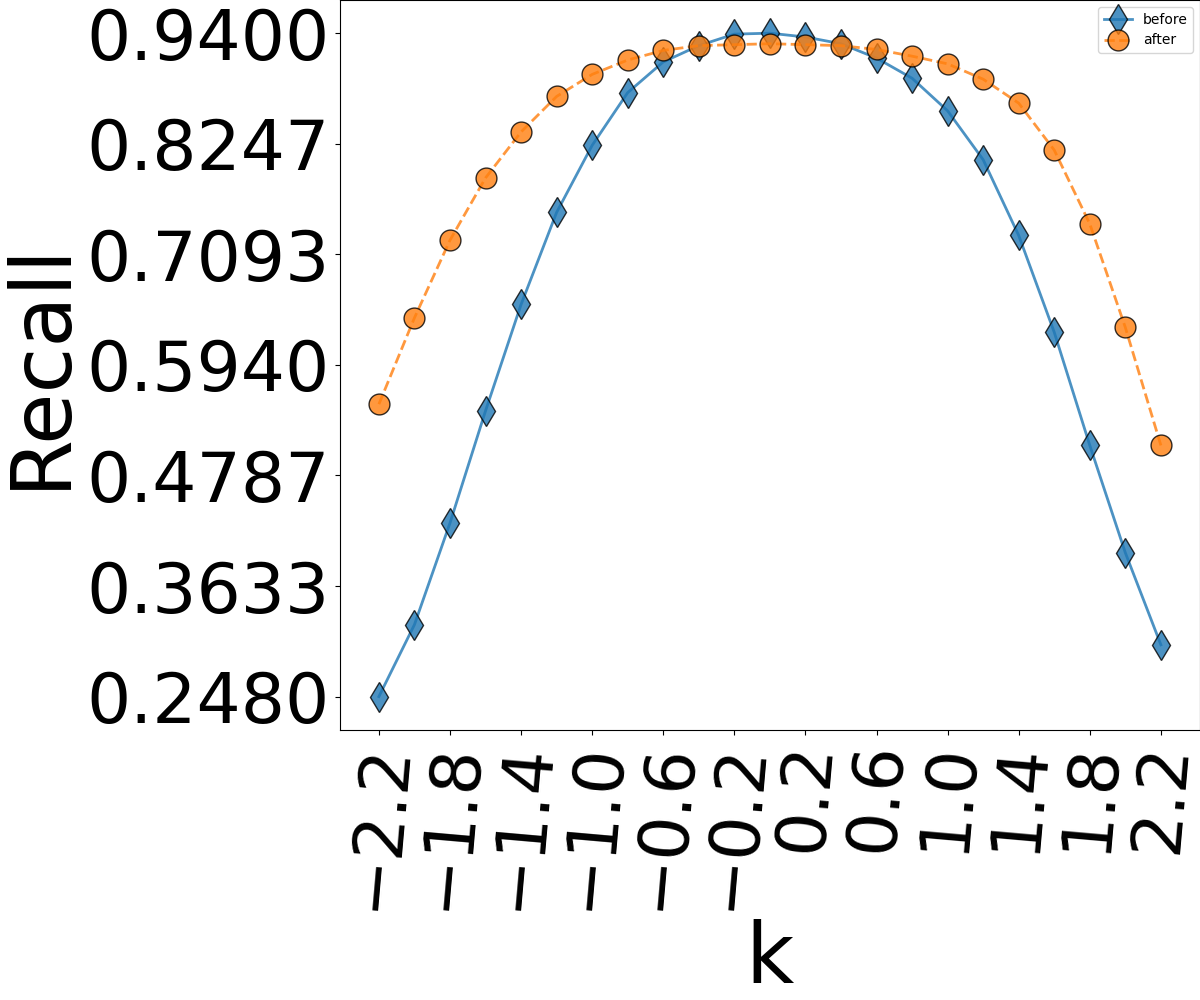}
    \includegraphics[width=\width]{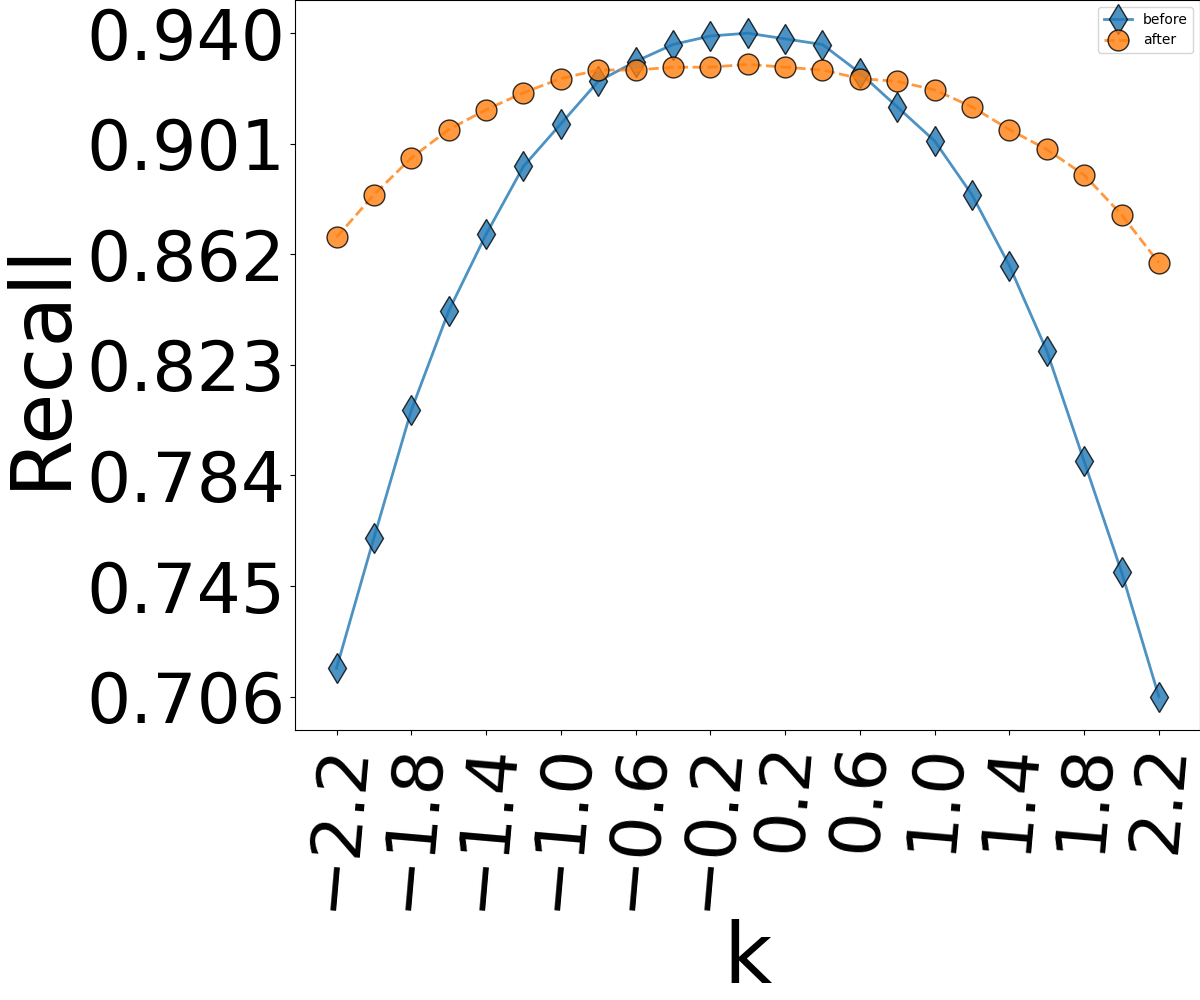}
    \includegraphics[width=\width]{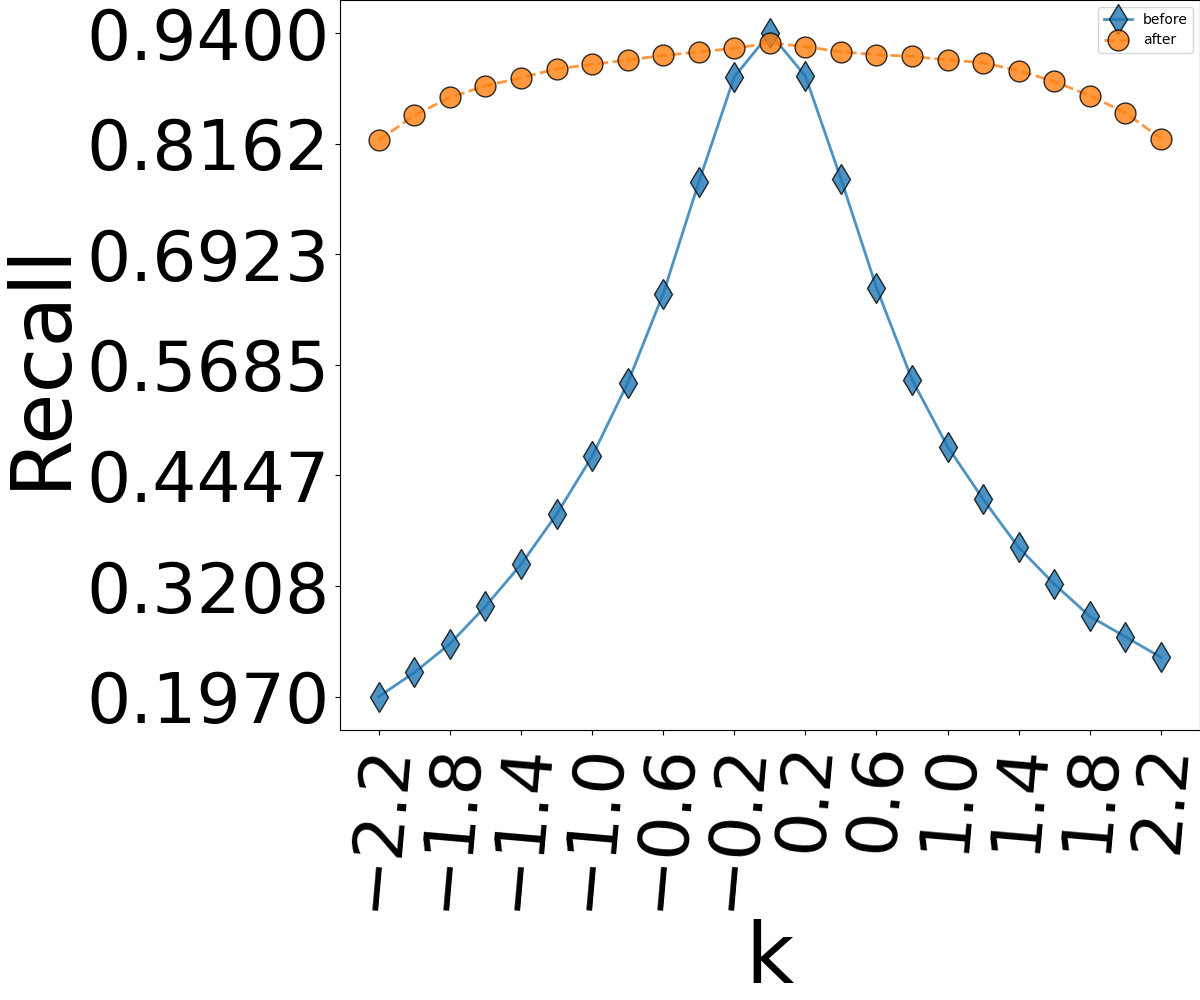}
    \includegraphics[width=\width]{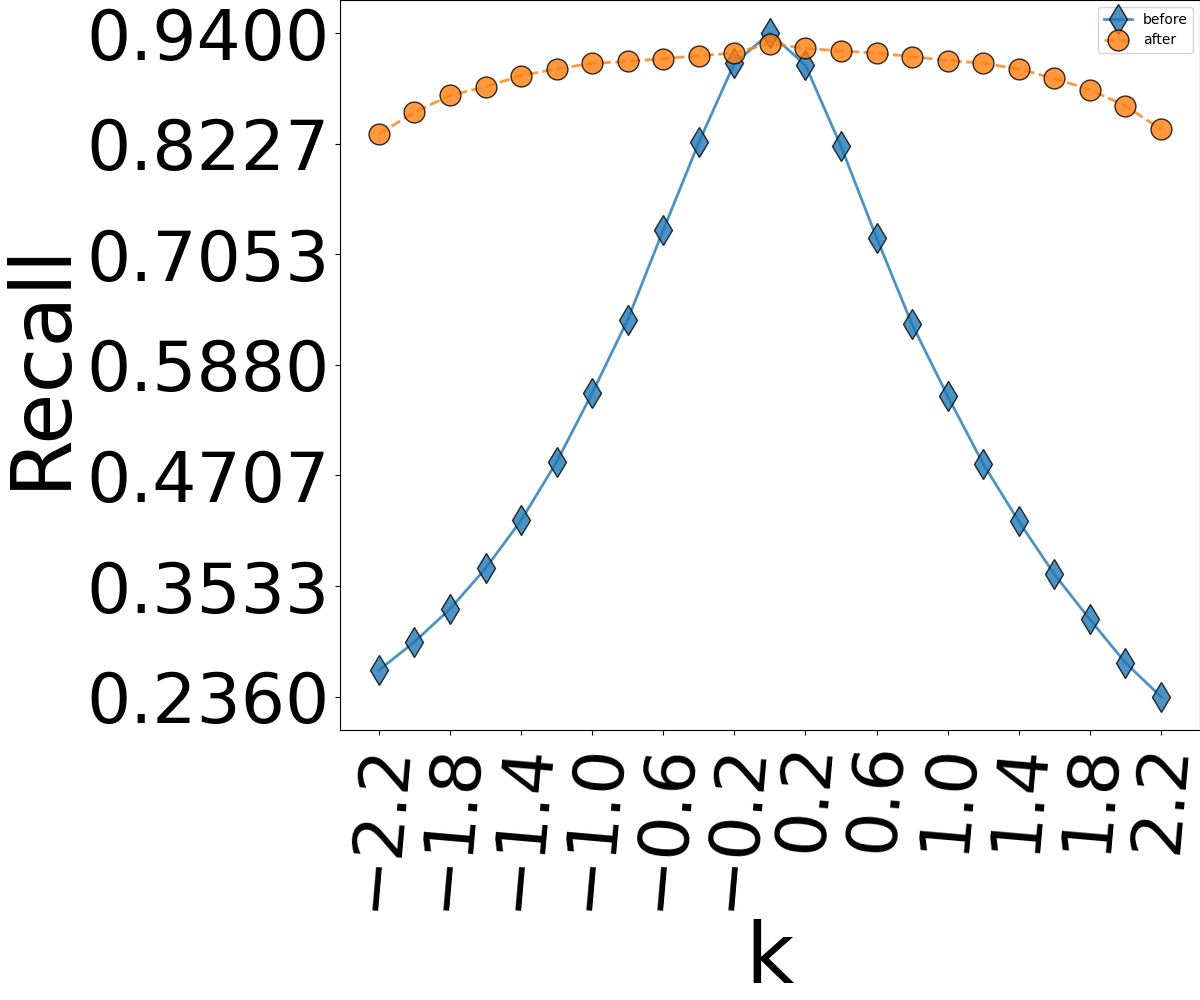}
    \includegraphics[width=\width]{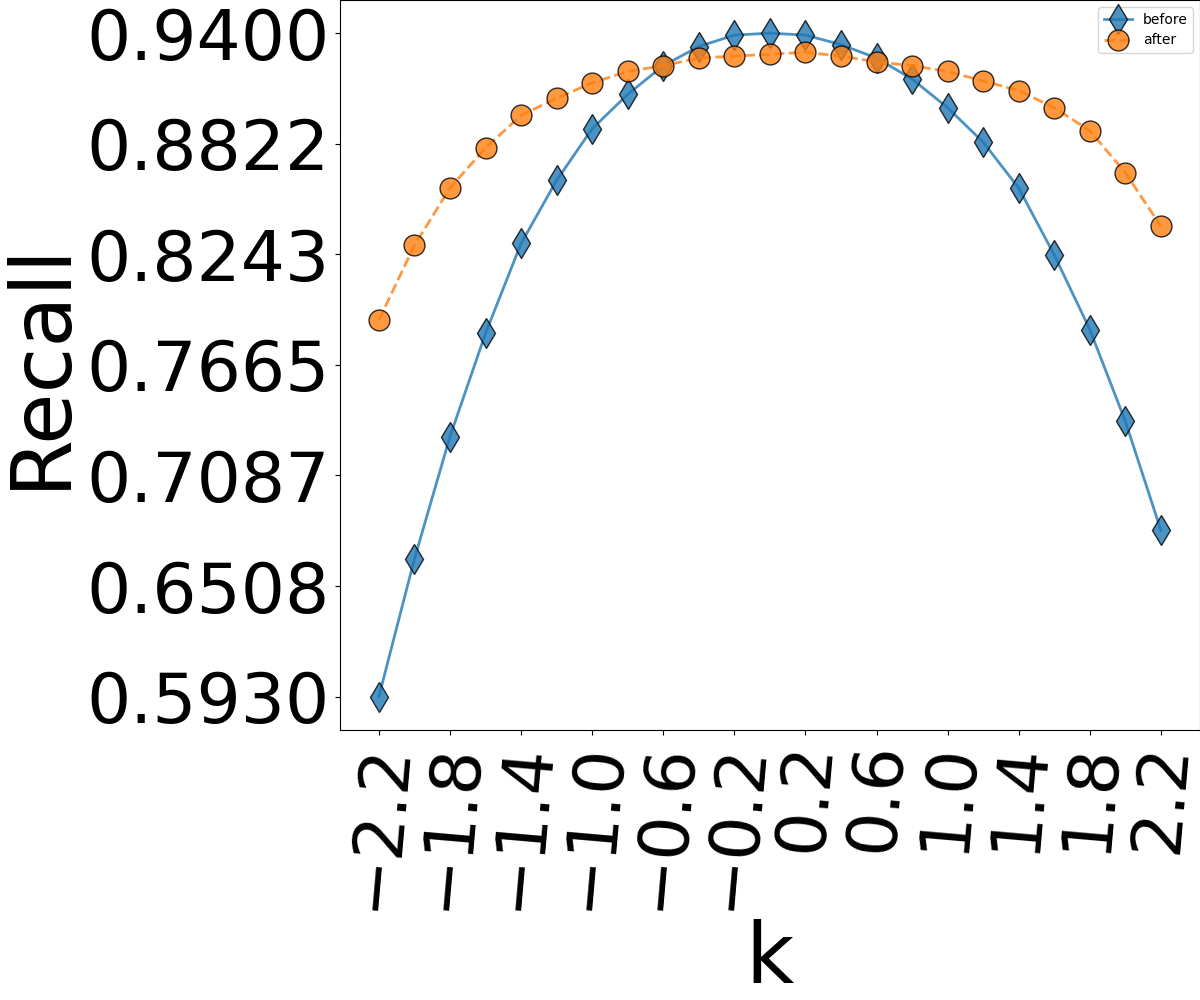}
    \includegraphics[width=\width]{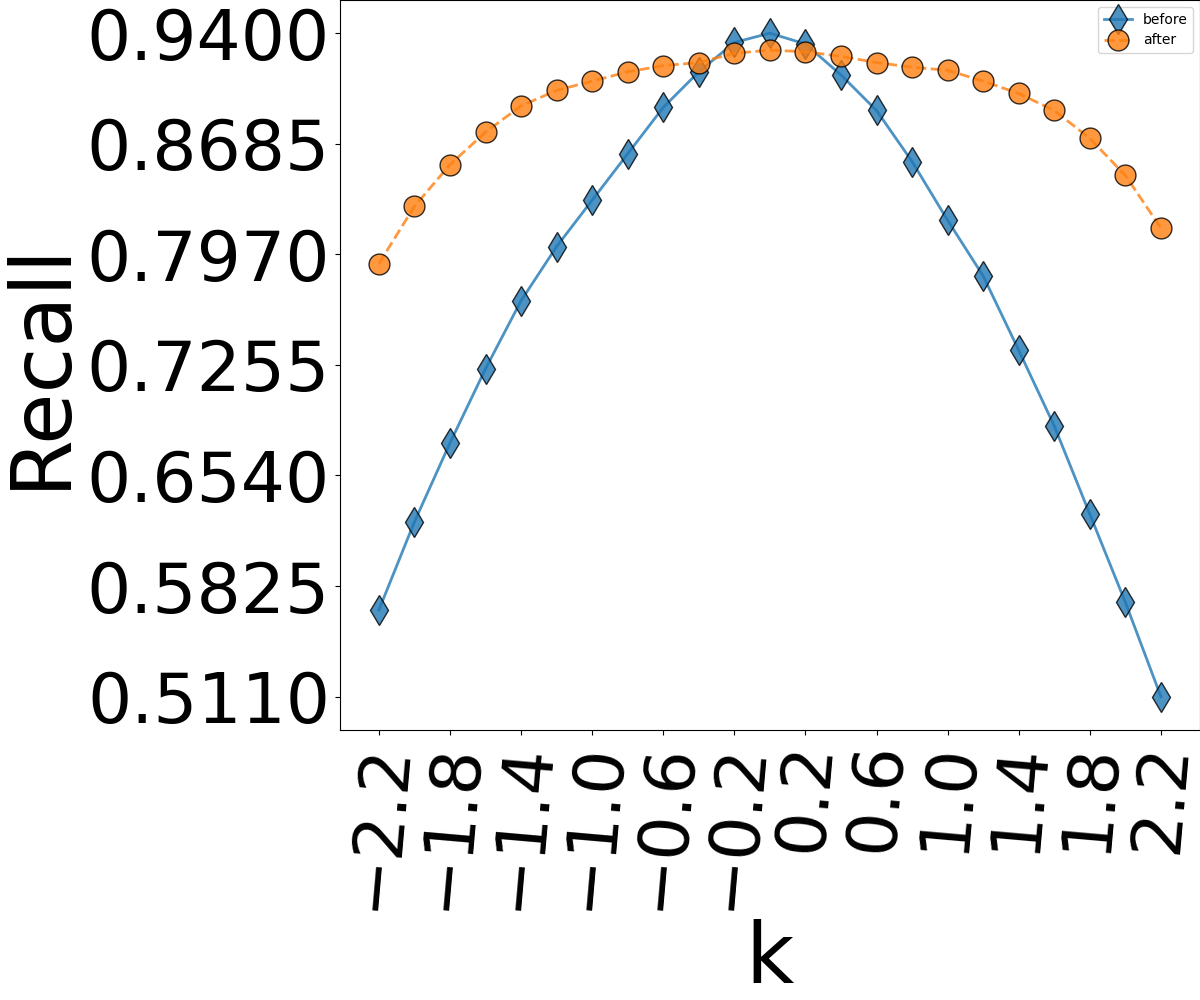}
    \\
    \includegraphics[width=\width]{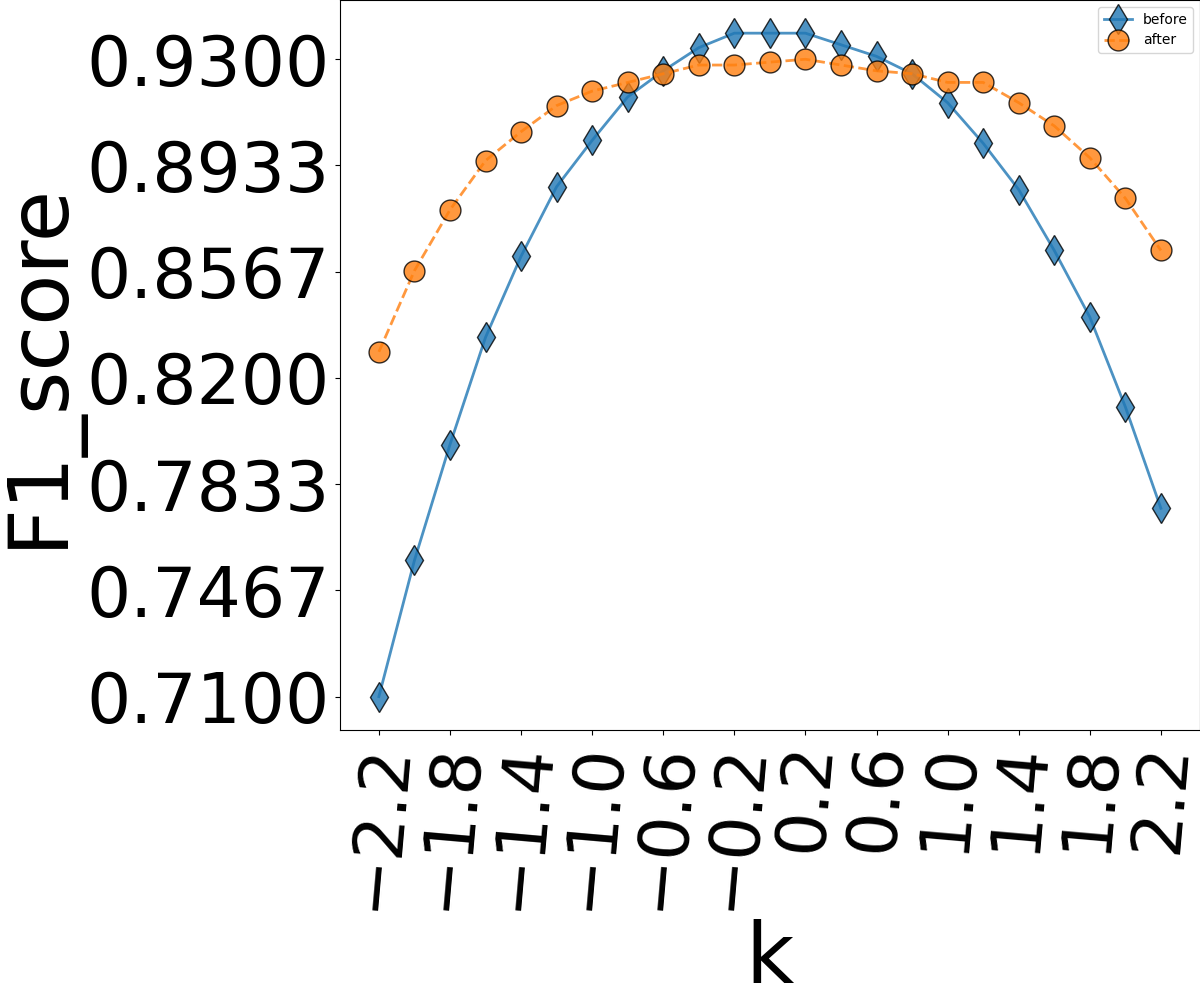}
    \includegraphics[width=\width]{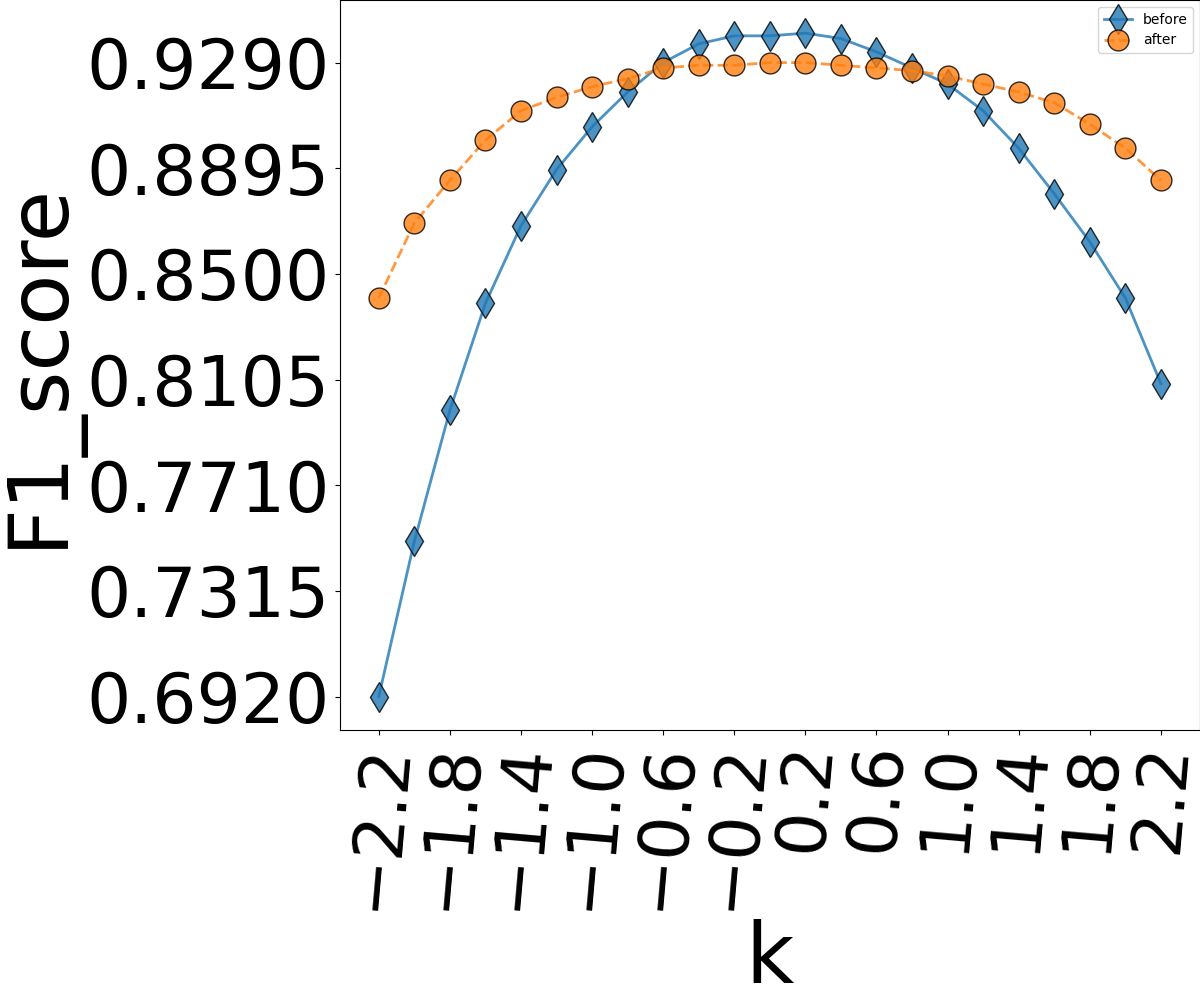}
    \includegraphics[width=\width]{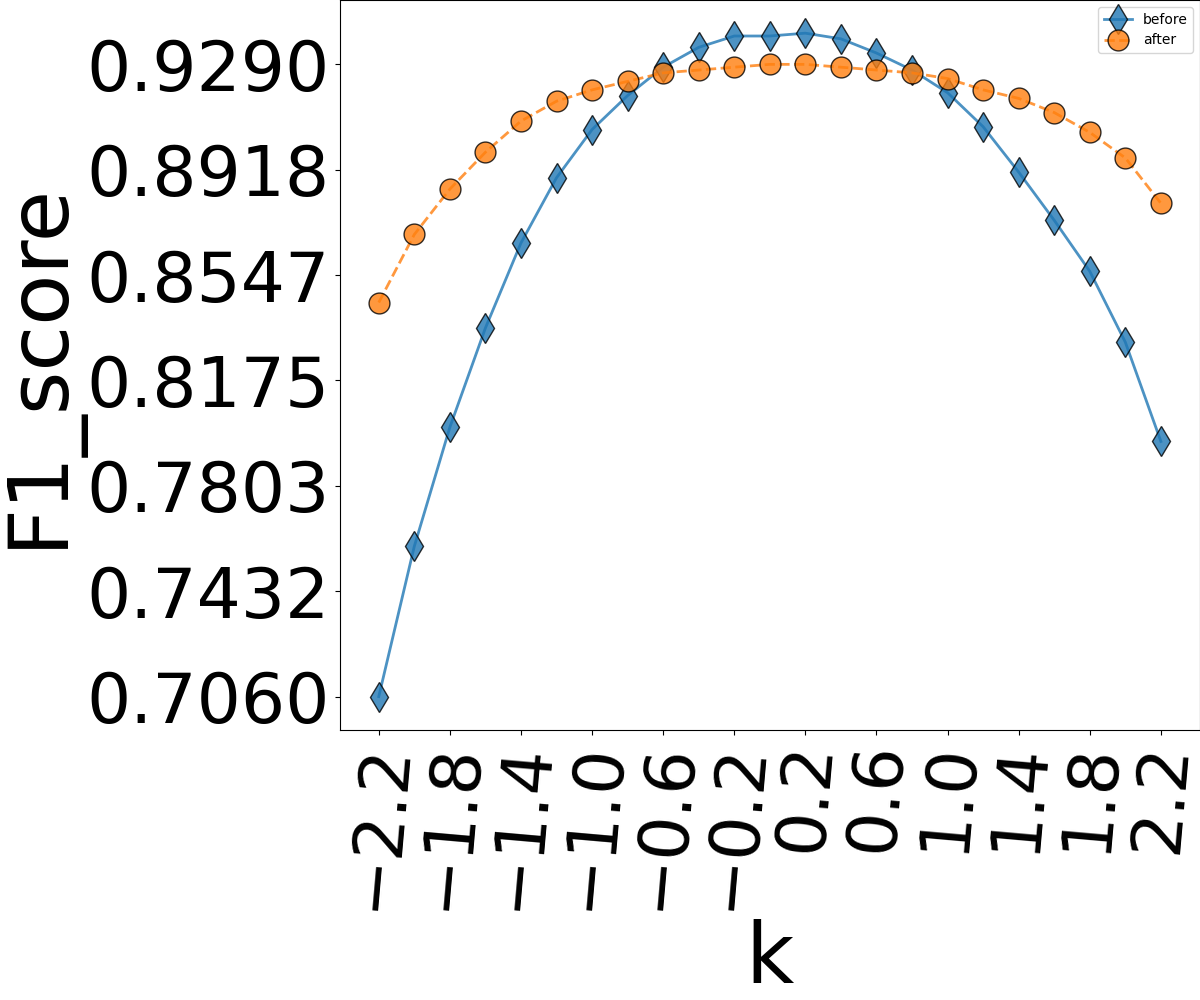}
    \includegraphics[width=\width]{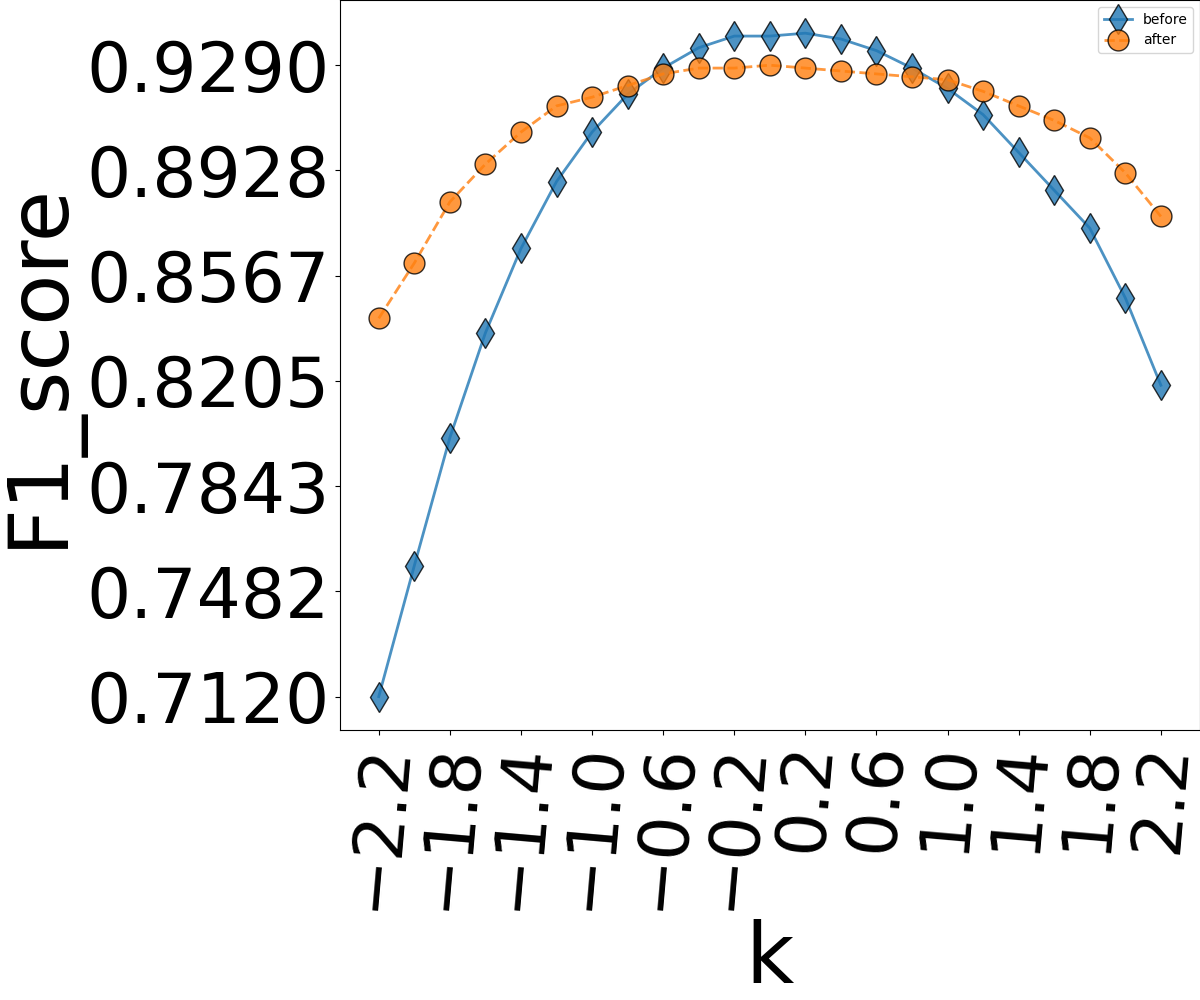}
    \includegraphics[width=\width]{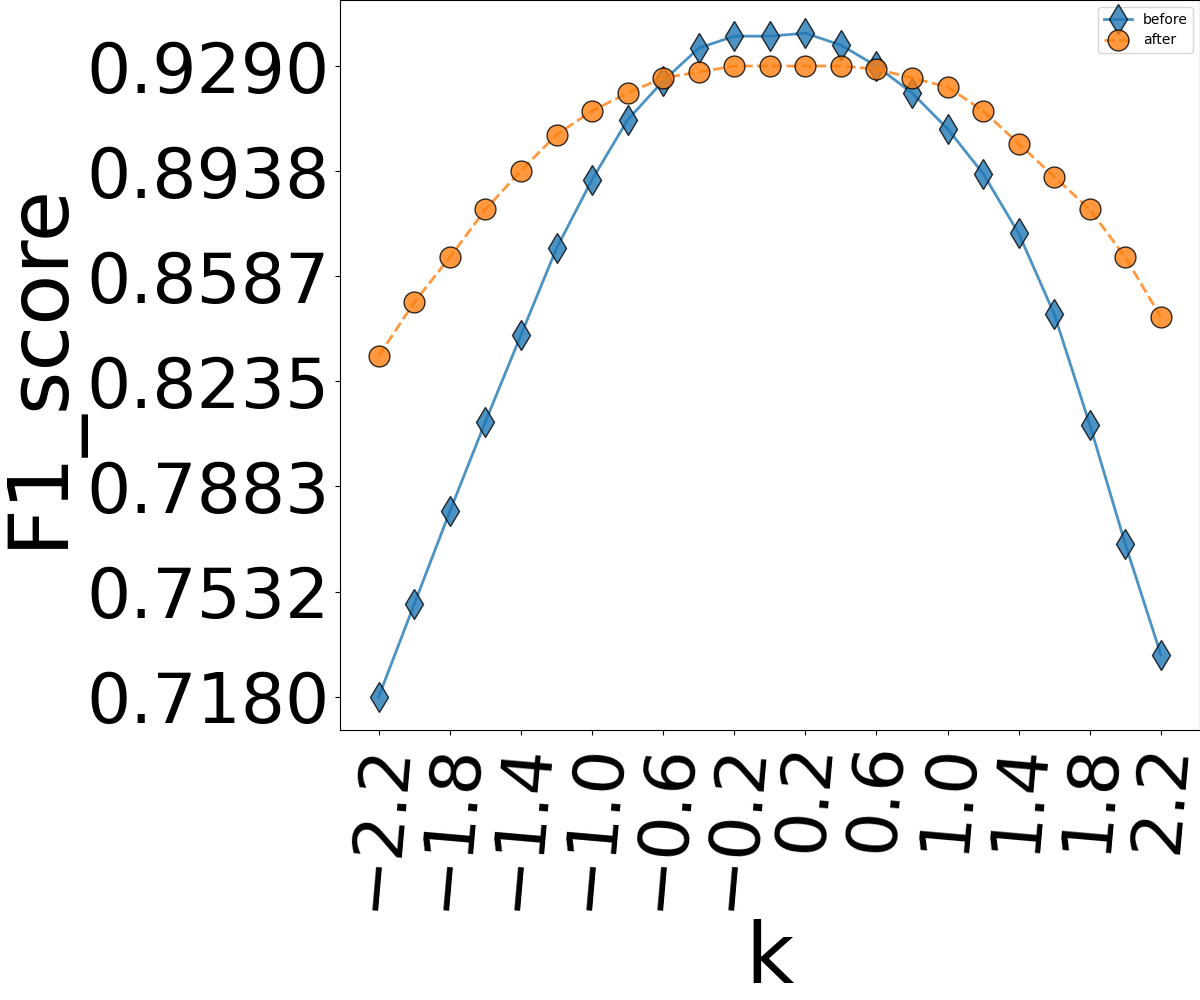}
    \includegraphics[width=\width]{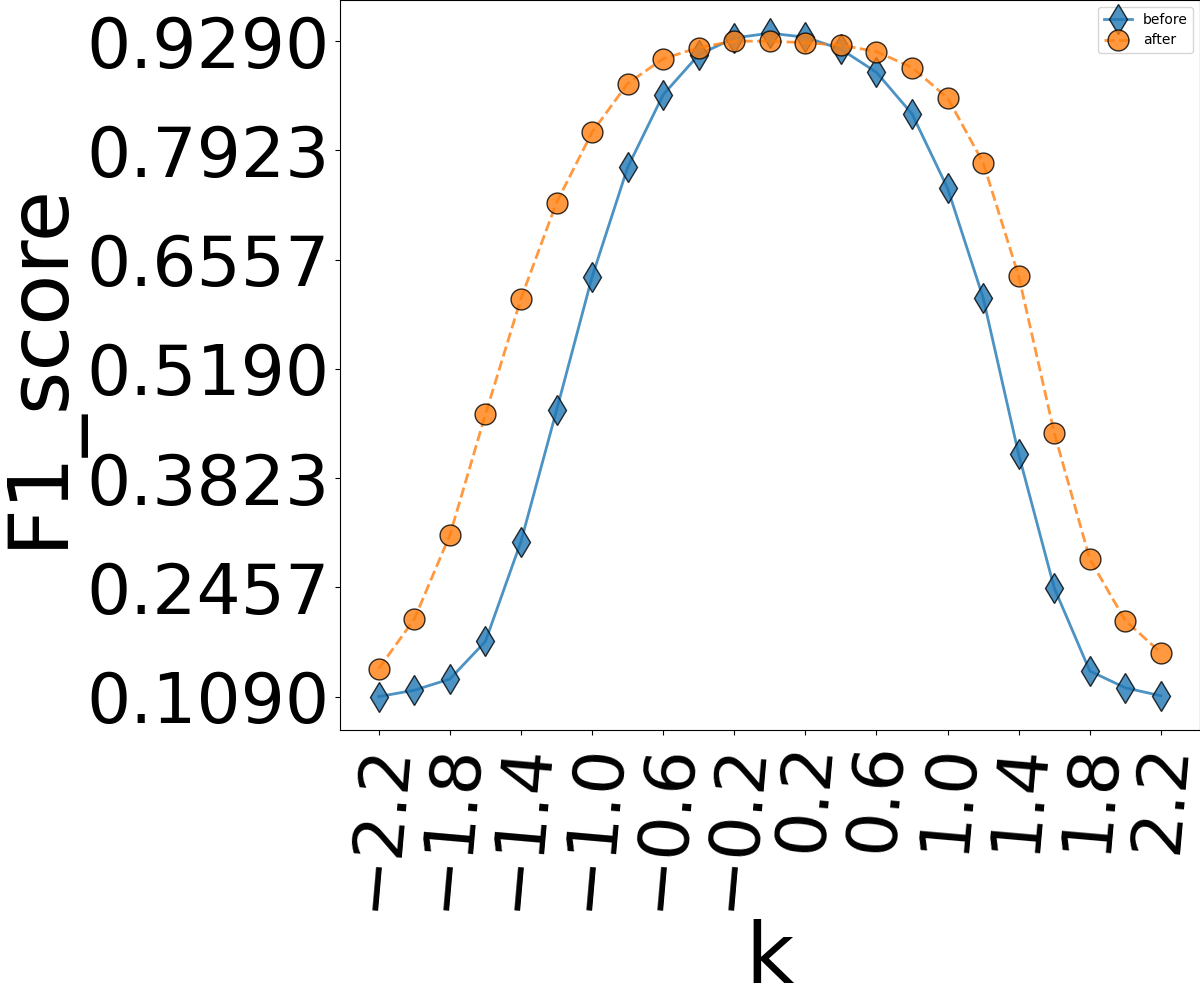}
    \includegraphics[width=\width]{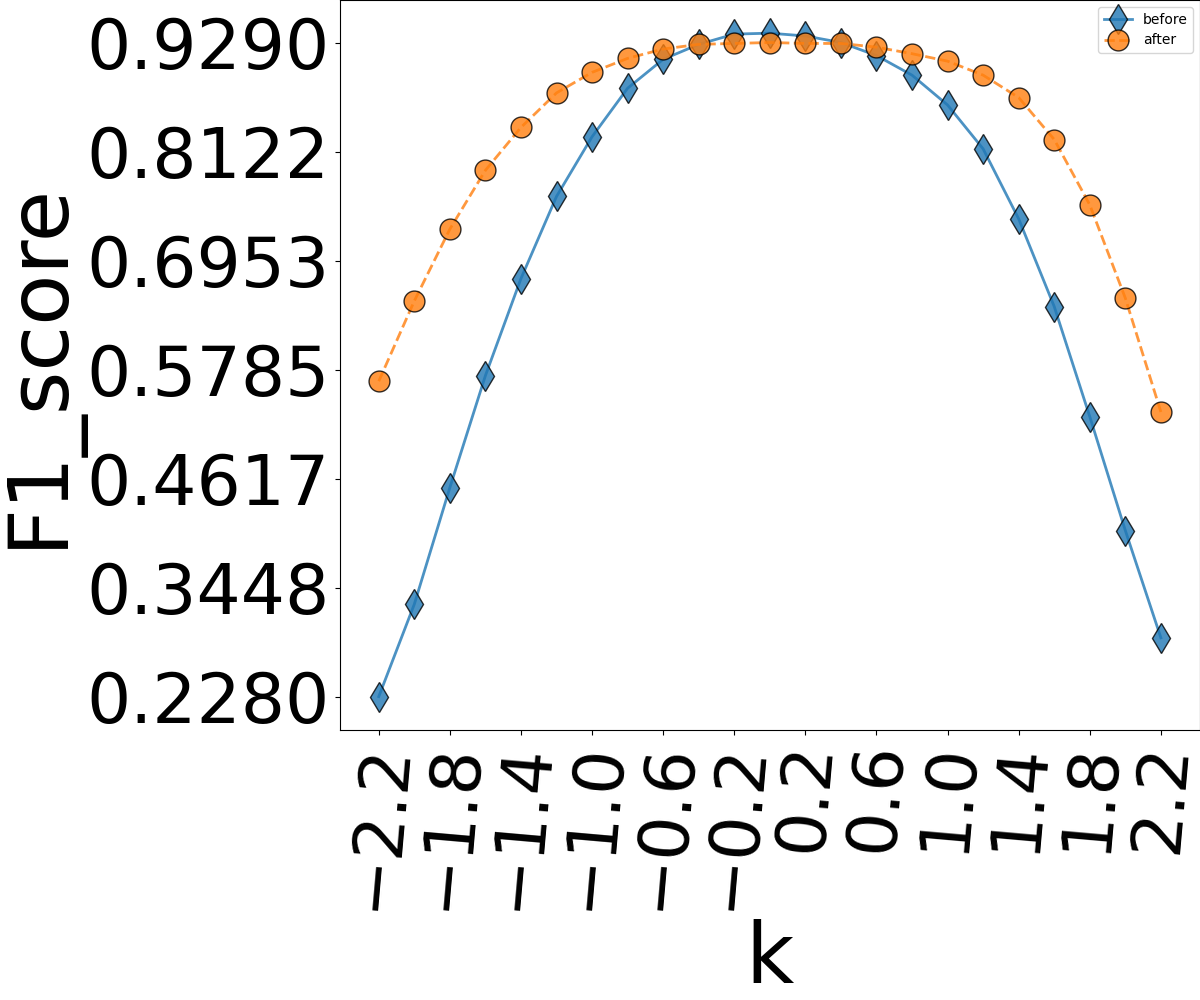}
    \includegraphics[width=\width]{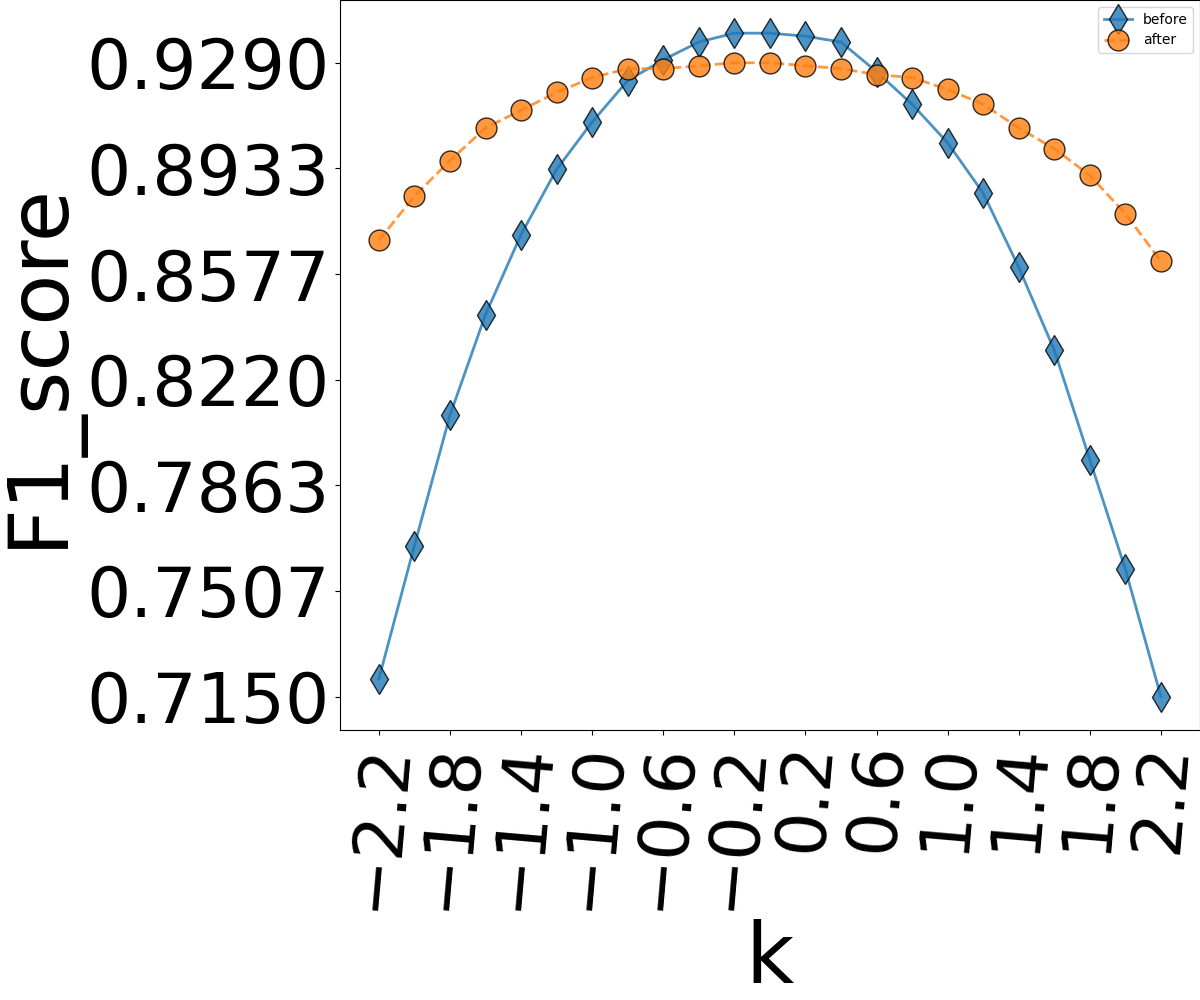}
    \includegraphics[width=\width]{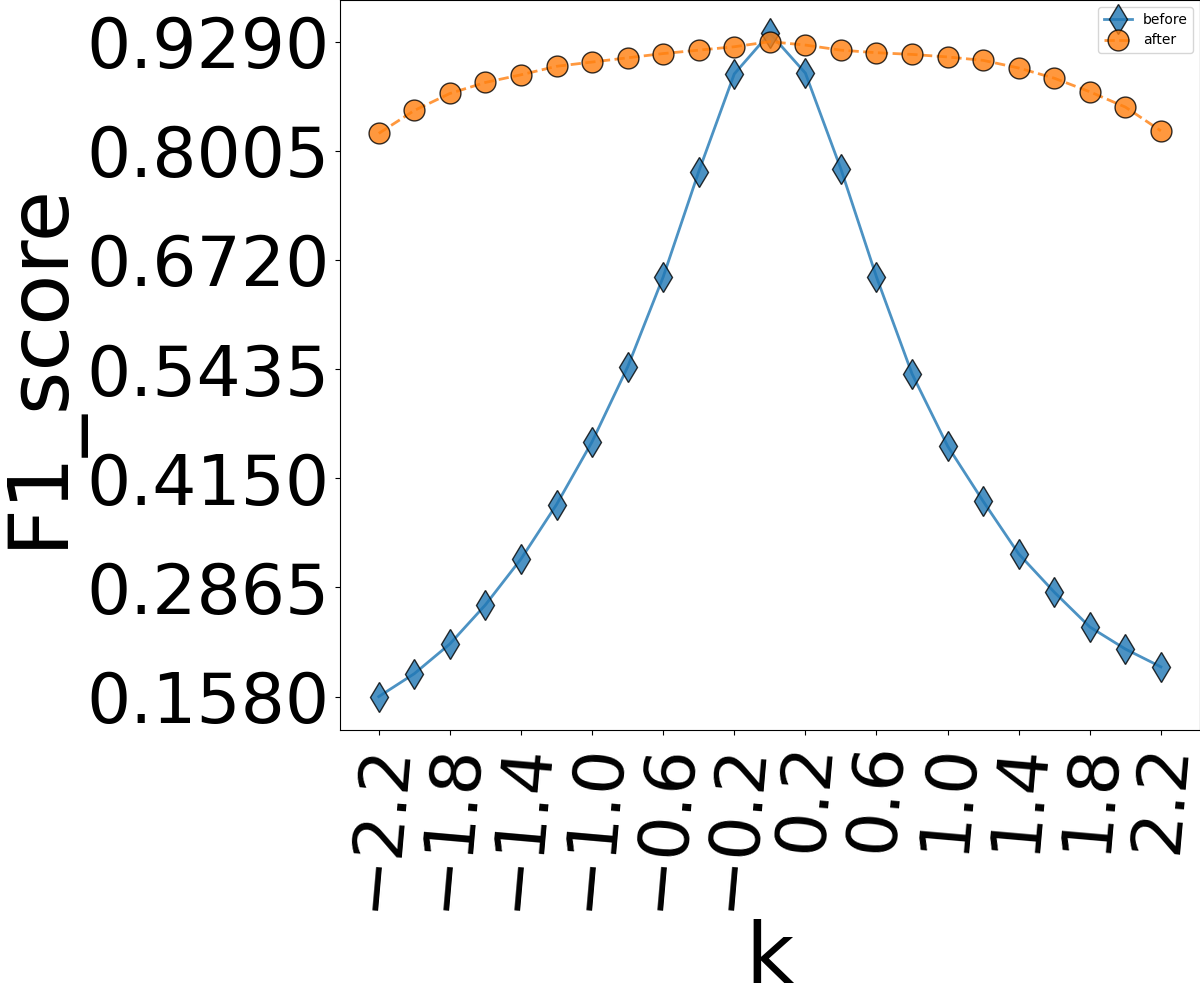}
    \includegraphics[width=\width]{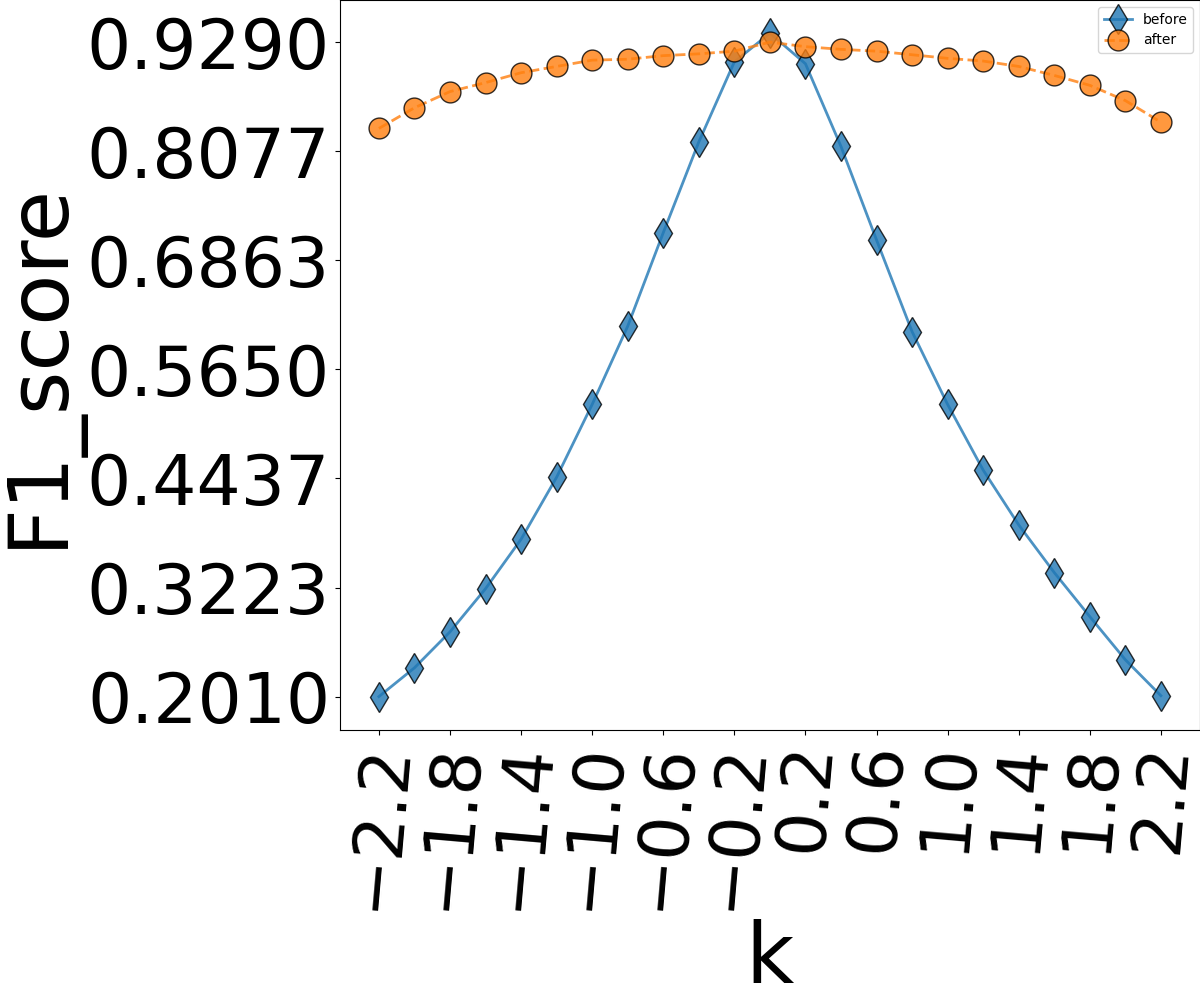}
    \includegraphics[width=\width]{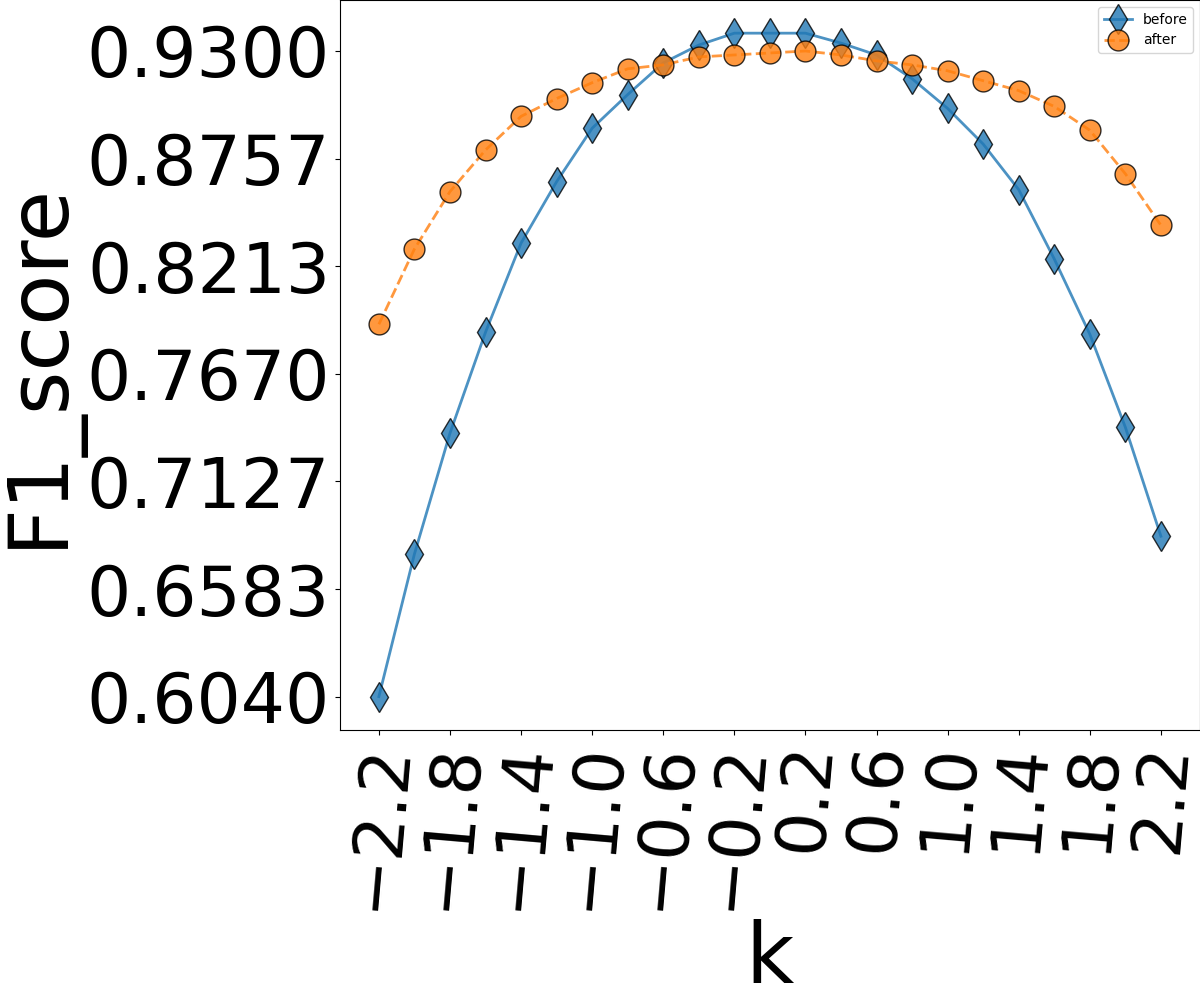}
    \includegraphics[width=\width]{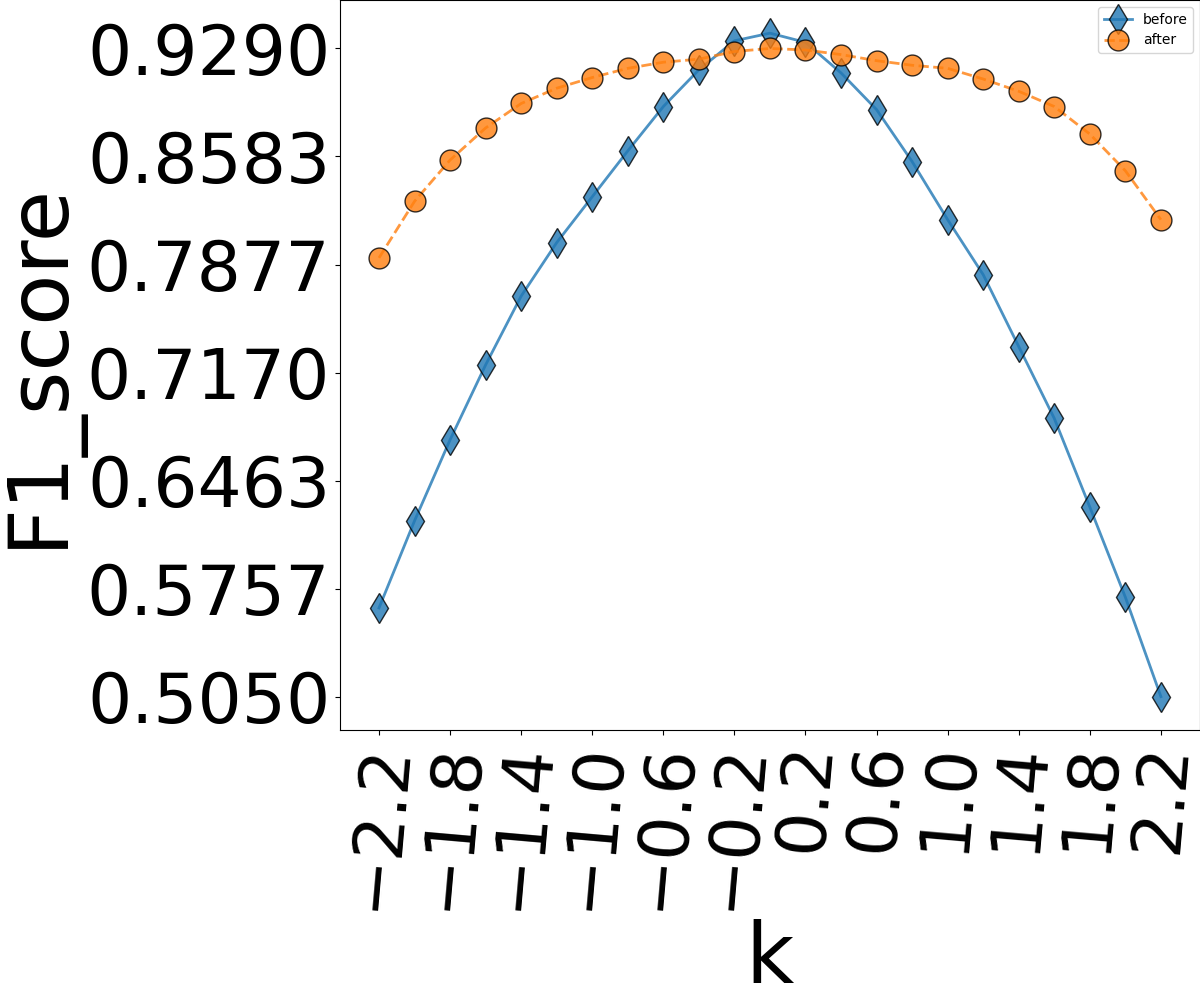}
    \\
 
    \caption{Figure shows the precision, recall and F1-score using InceptionV3 model on CIFAR10 dataset. The $1^{st}$ two rows contain Precision, the middle two rows contain recall and the last two rows contain F1-score, for $1^{st}$ to $12^{th}$ masks, respectively.}
    \label{fig:inc-prf}
\end{figure*}

\begin{figure*}[t]
    \centering
    \newcommand\width{0.16\textwidth}
    \includegraphics[width=\width]{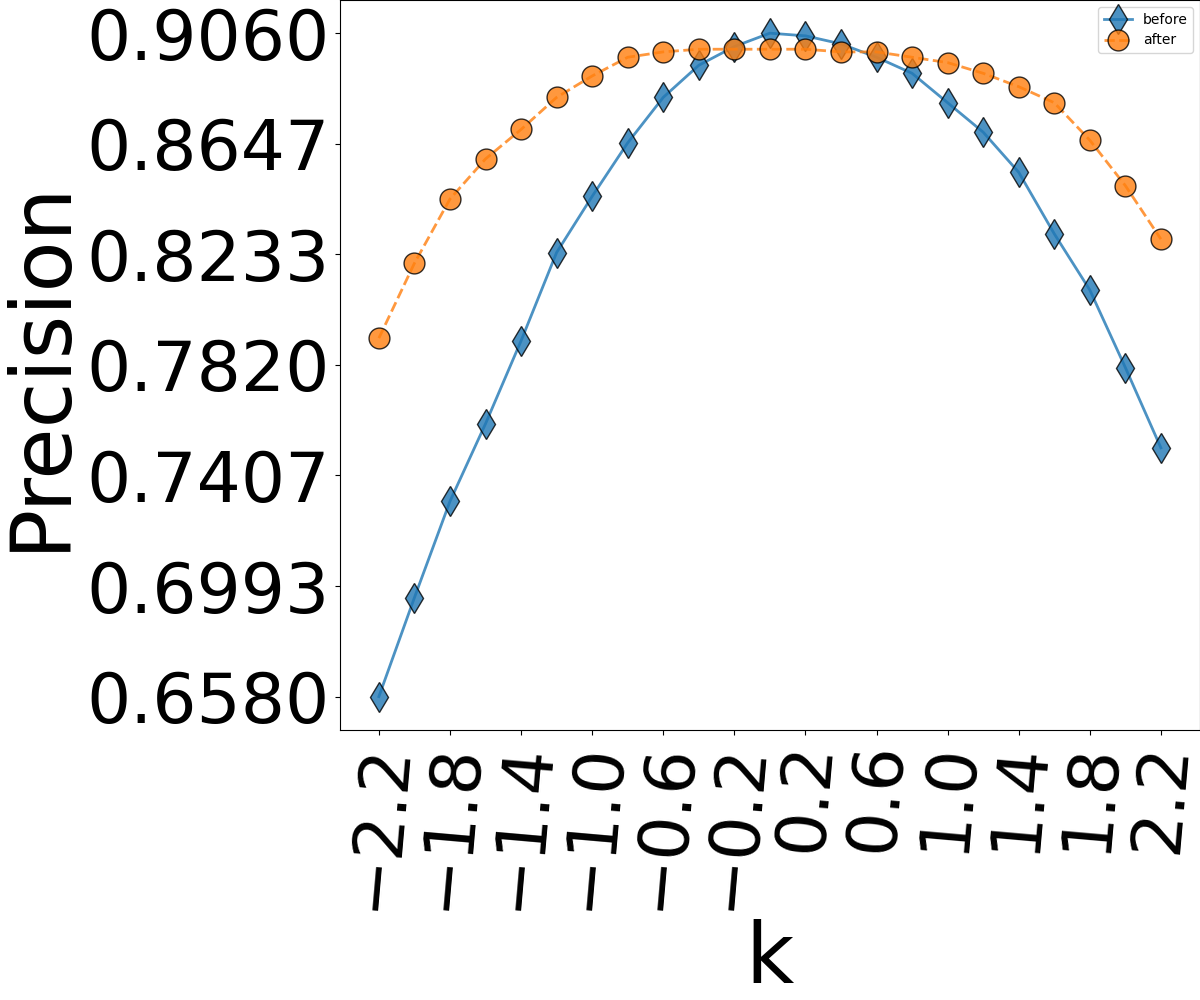}
    \includegraphics[width=\width]{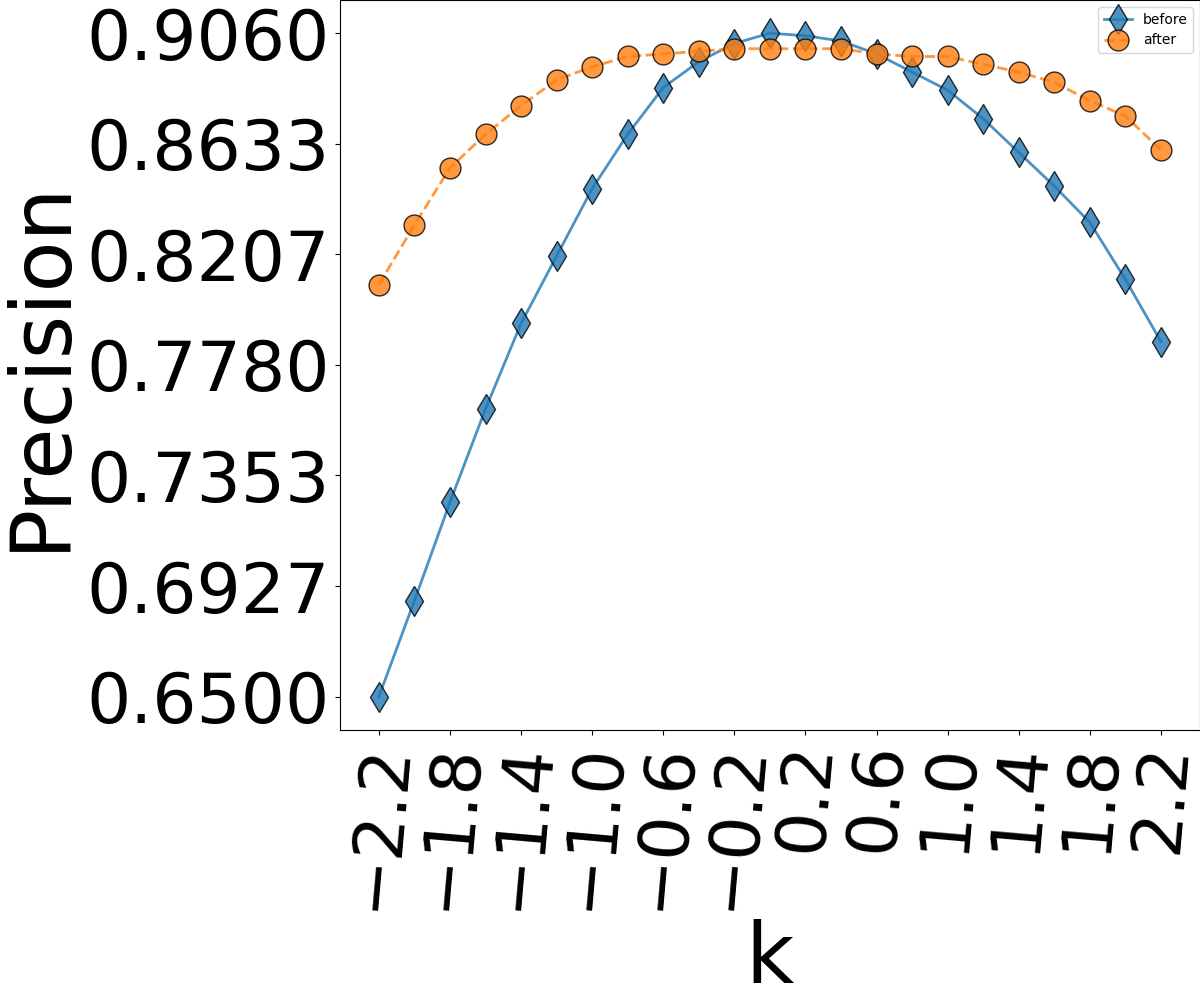}
    \includegraphics[width=\width]{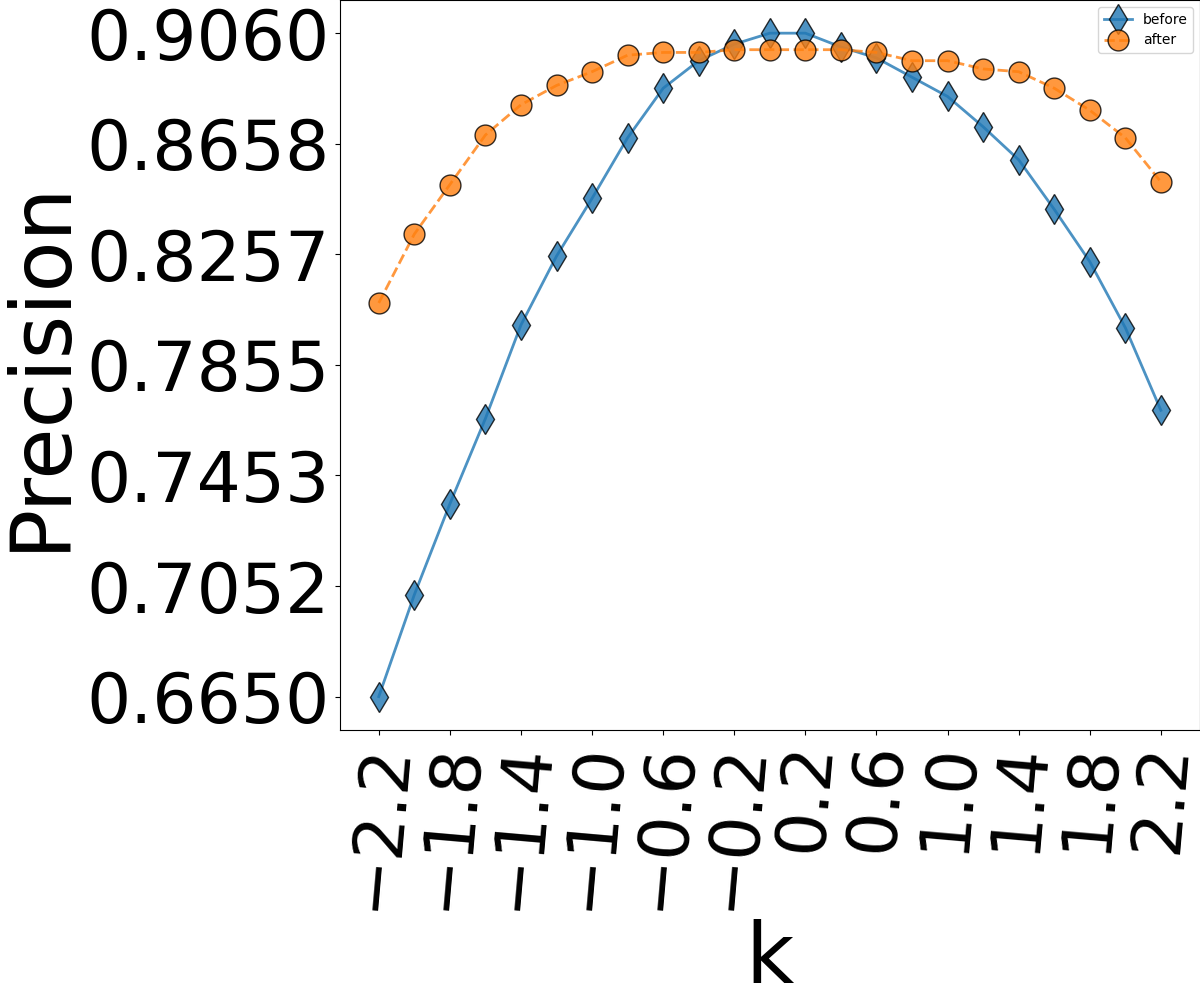}
    \includegraphics[width=\width]{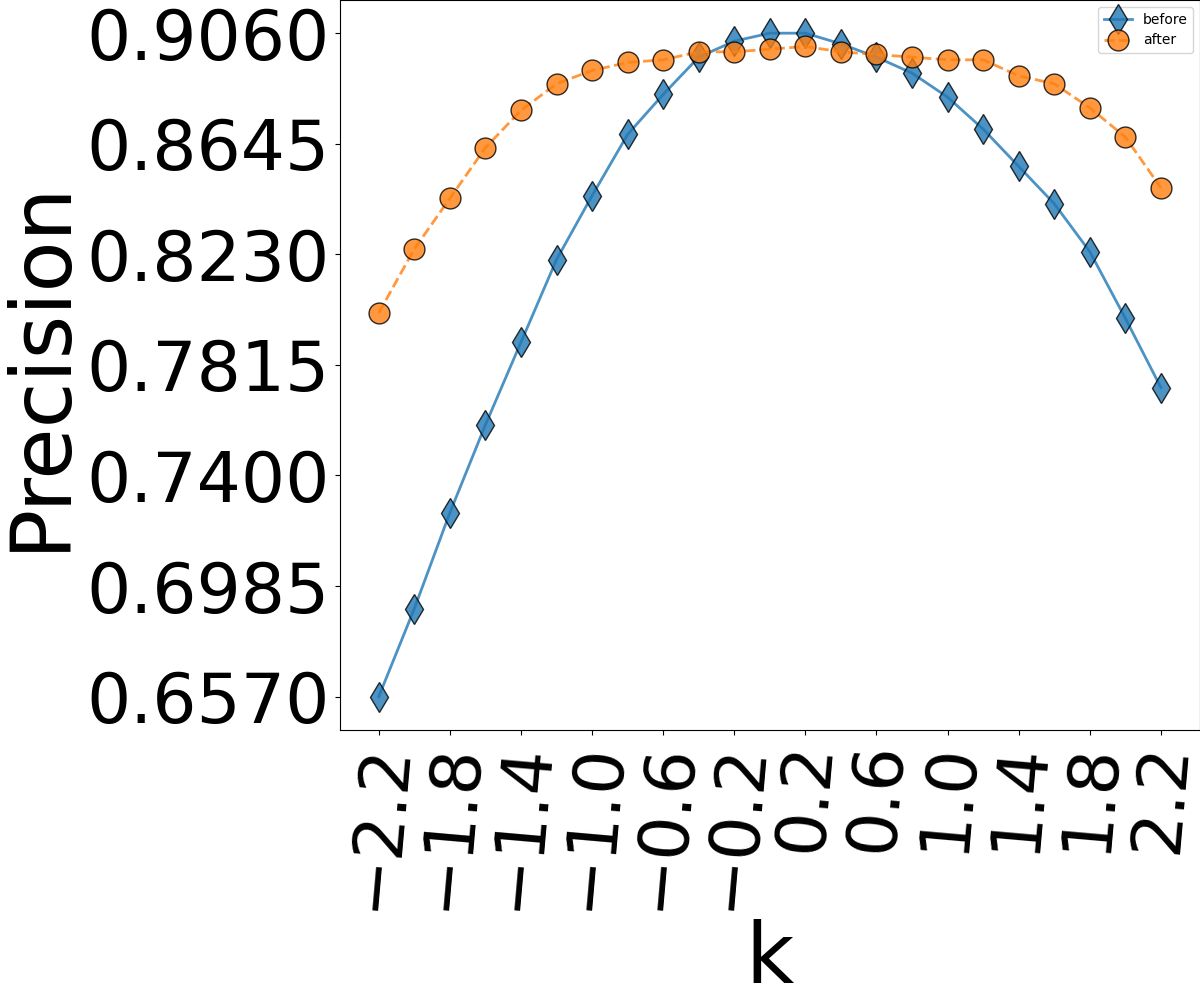}
    \includegraphics[width=\width]{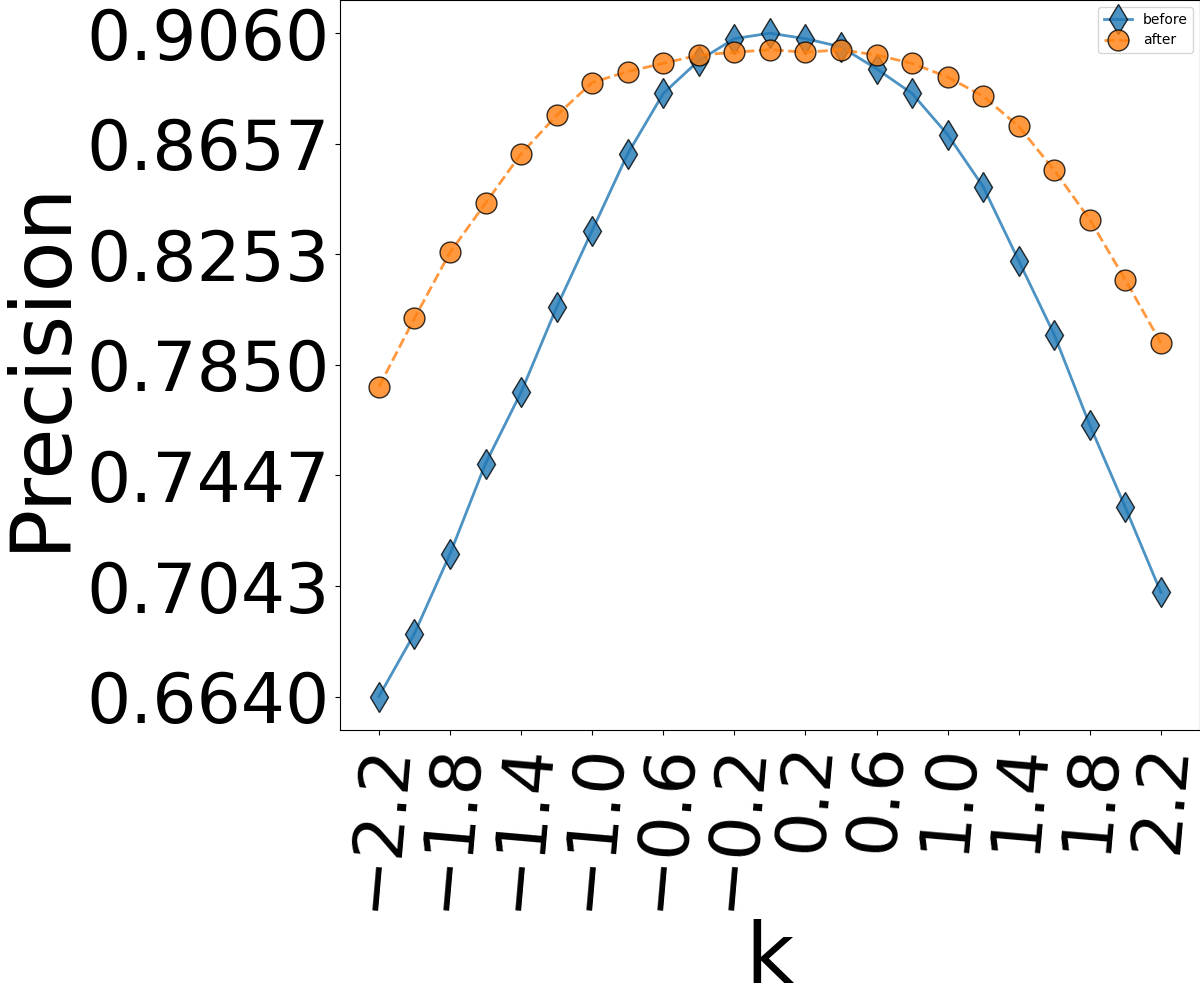}
    \includegraphics[width=\width]{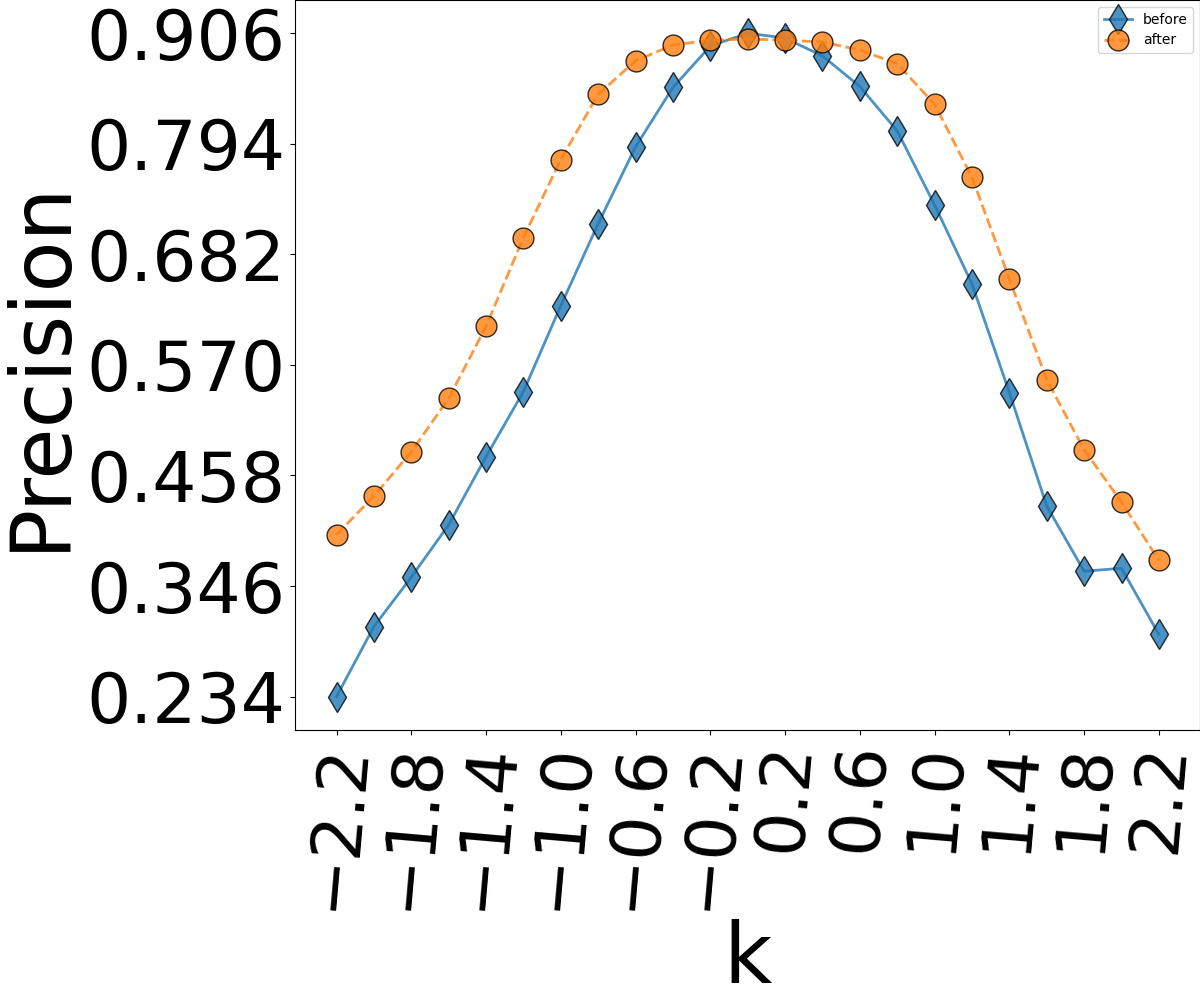}
    \includegraphics[width=\width]{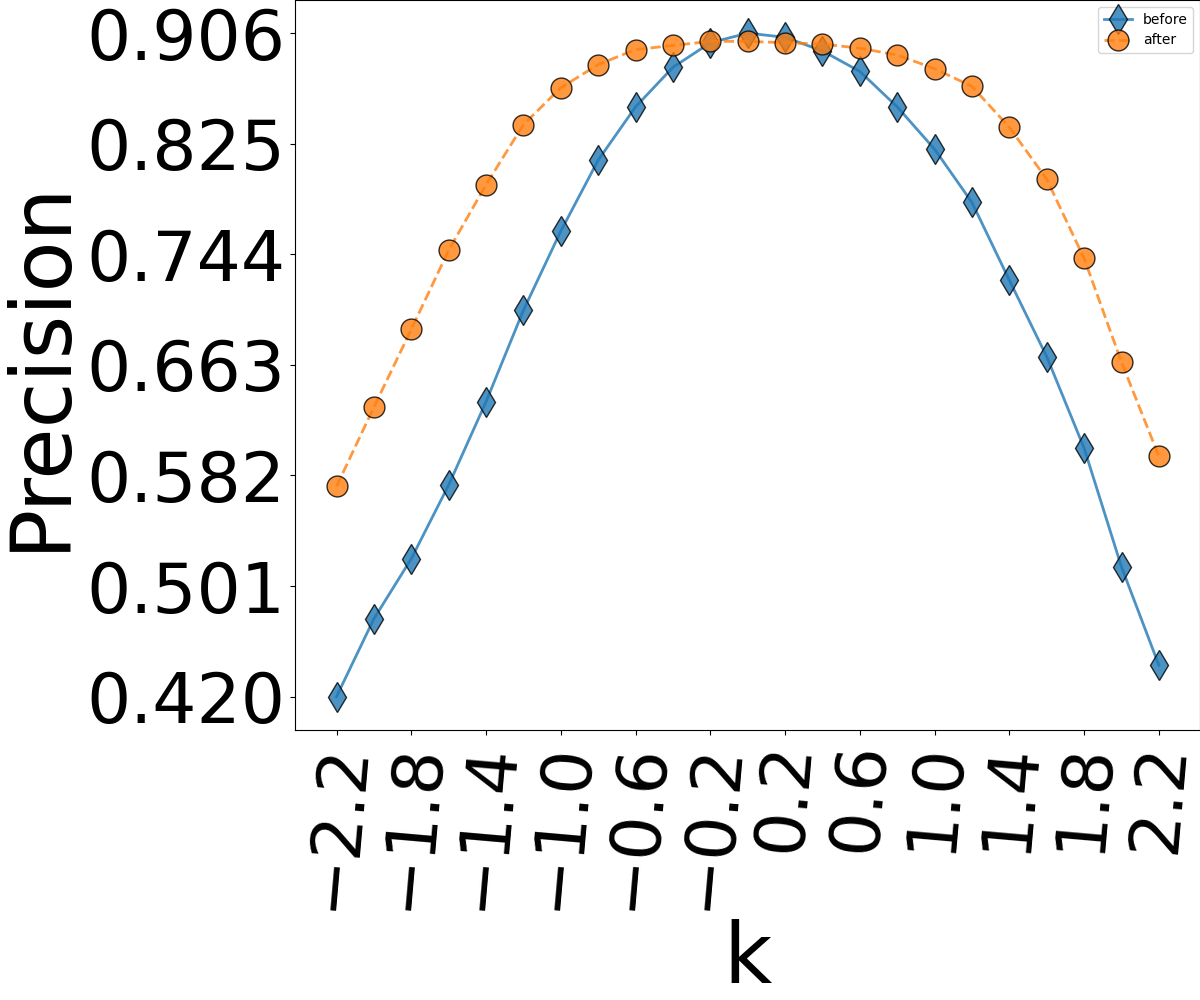}
    \includegraphics[width=\width]{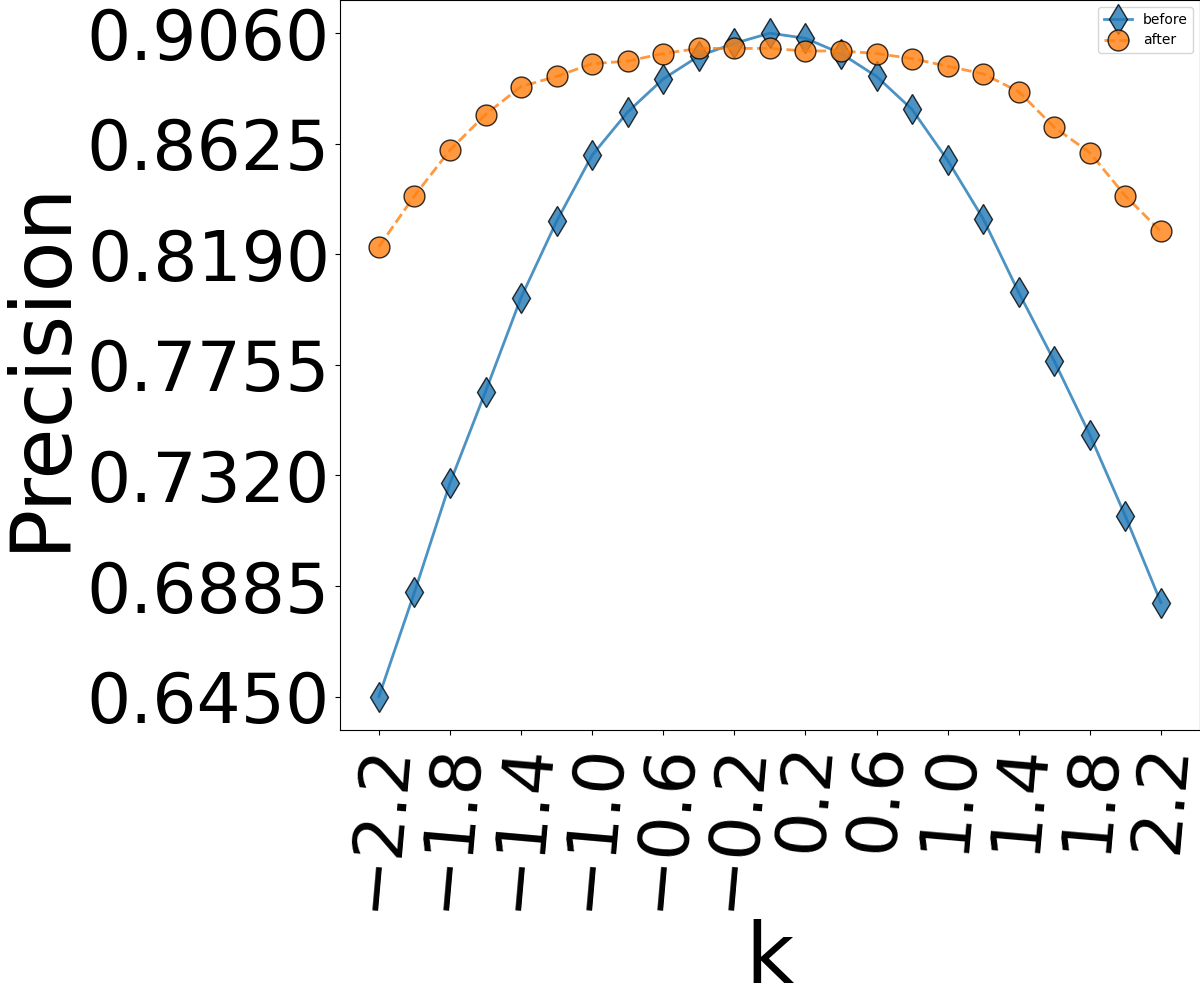}
    \includegraphics[width=\width]{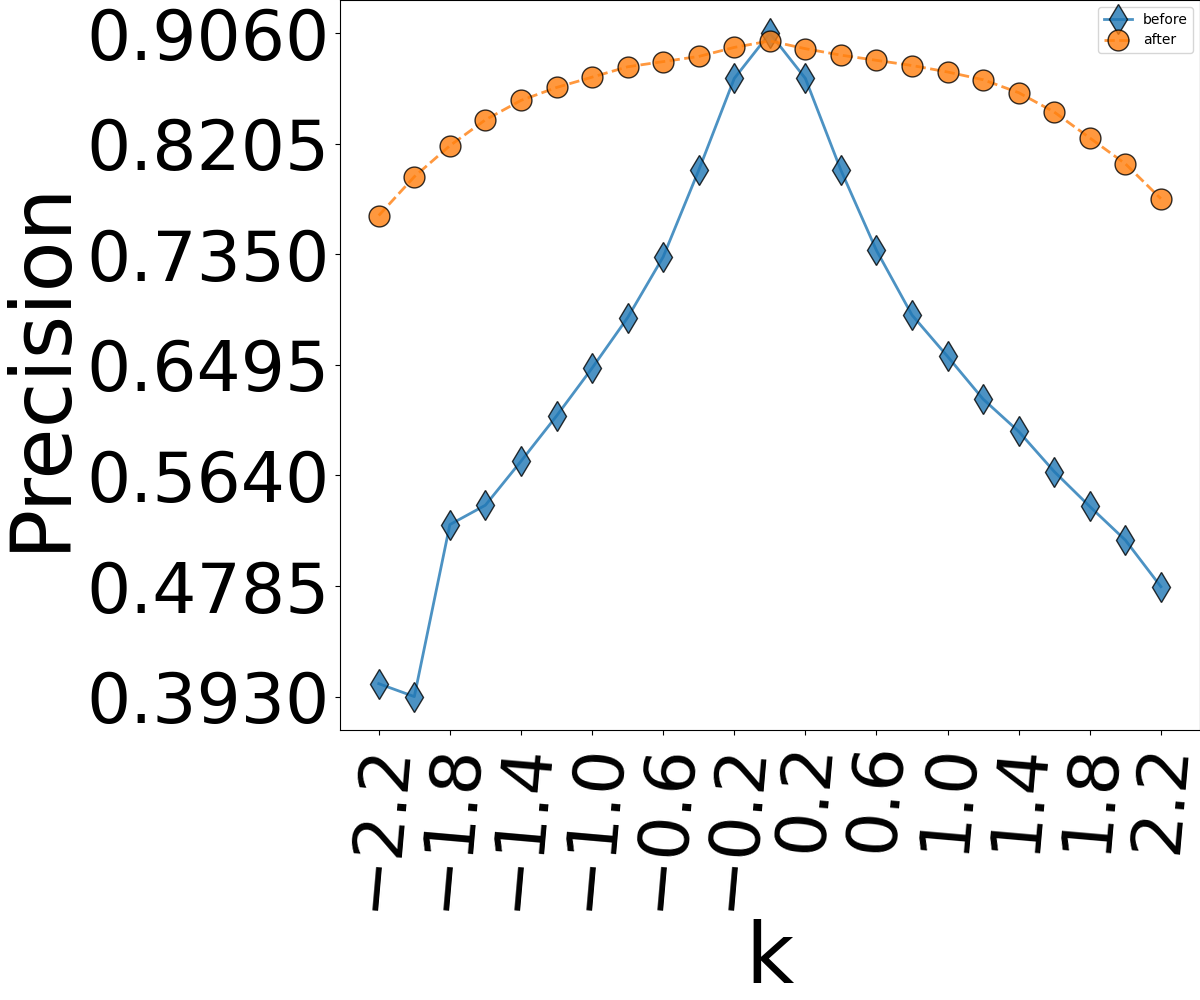}
    \includegraphics[width=\width]{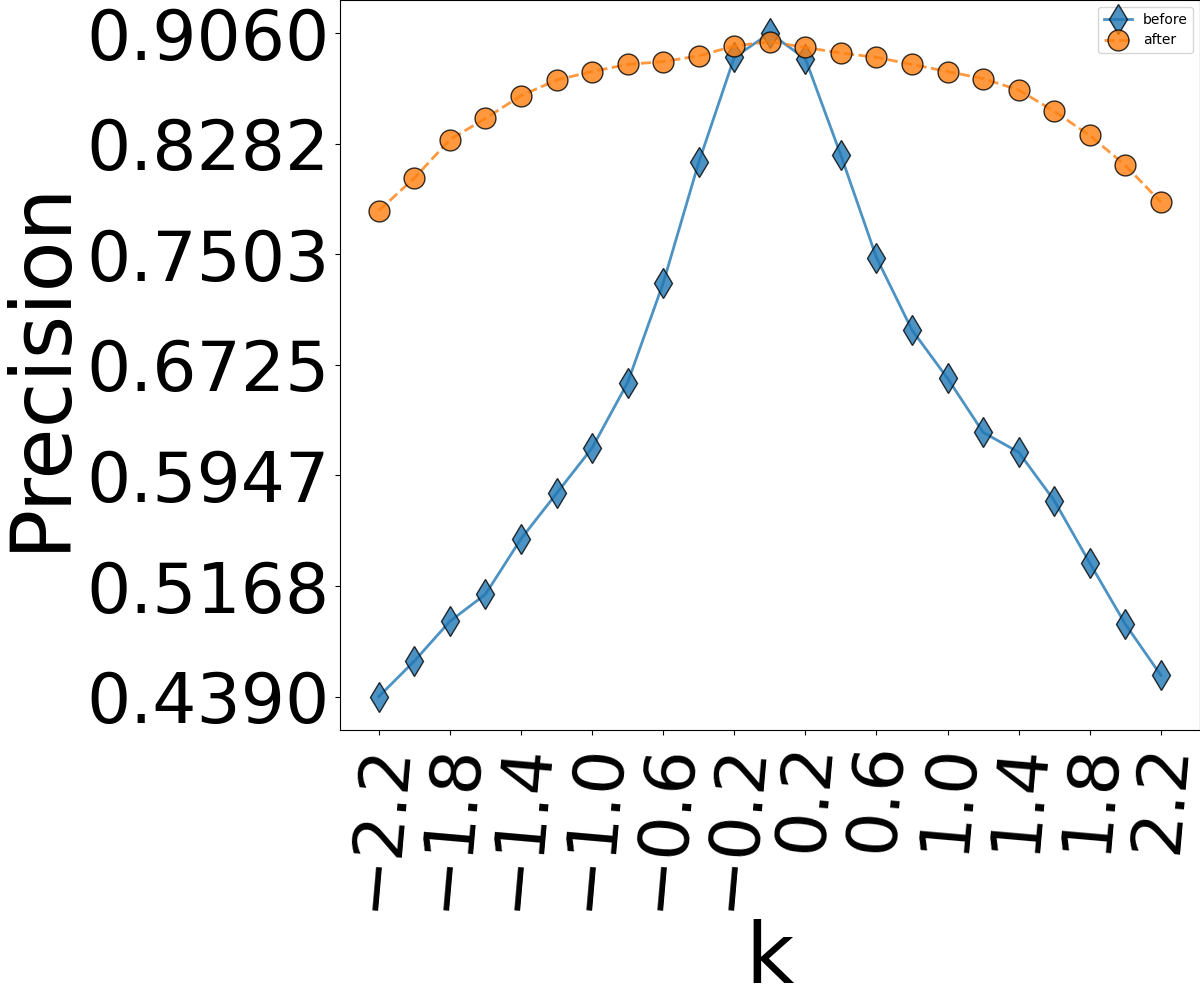}
    \includegraphics[width=\width]{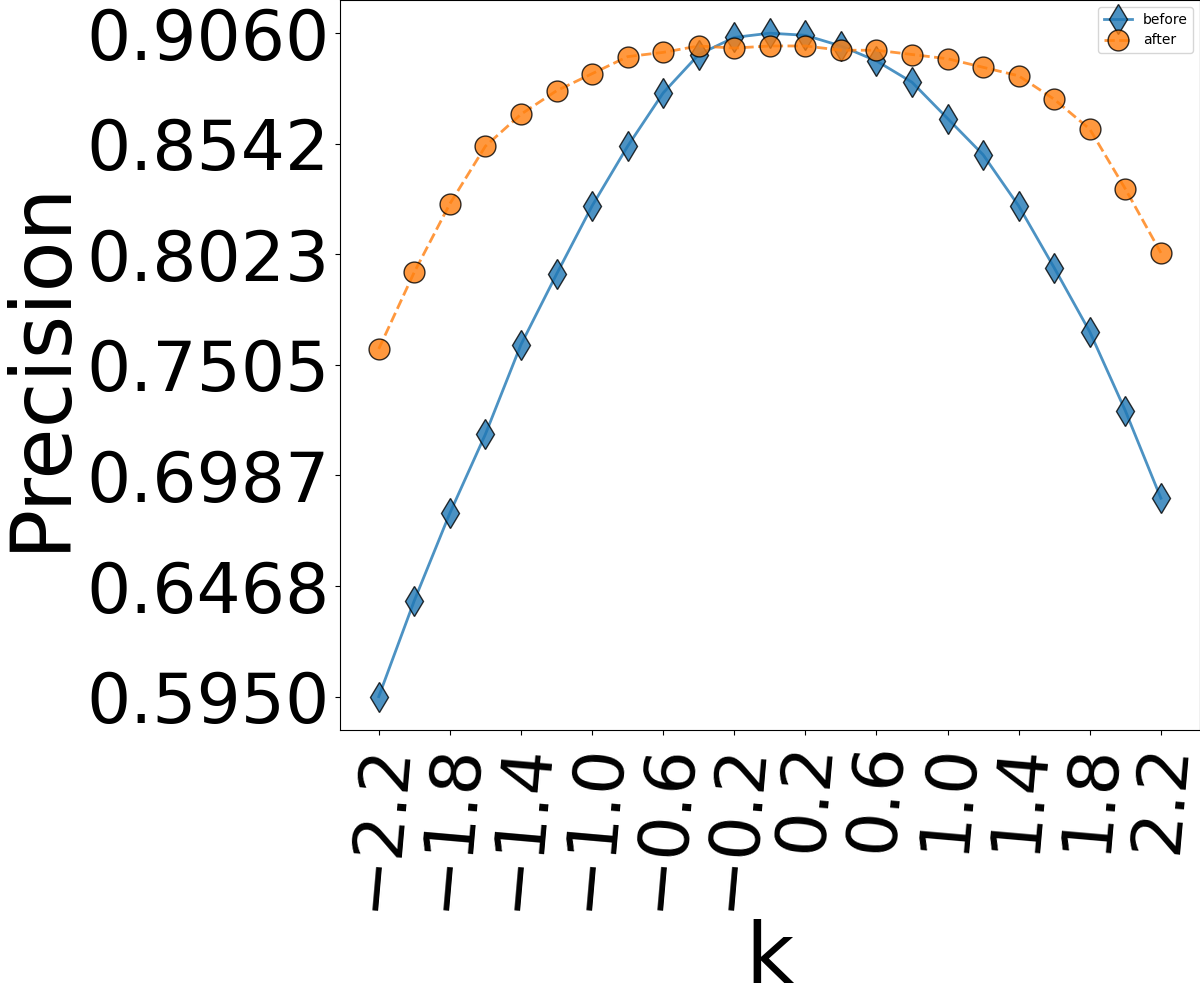}
    \includegraphics[width=\width]{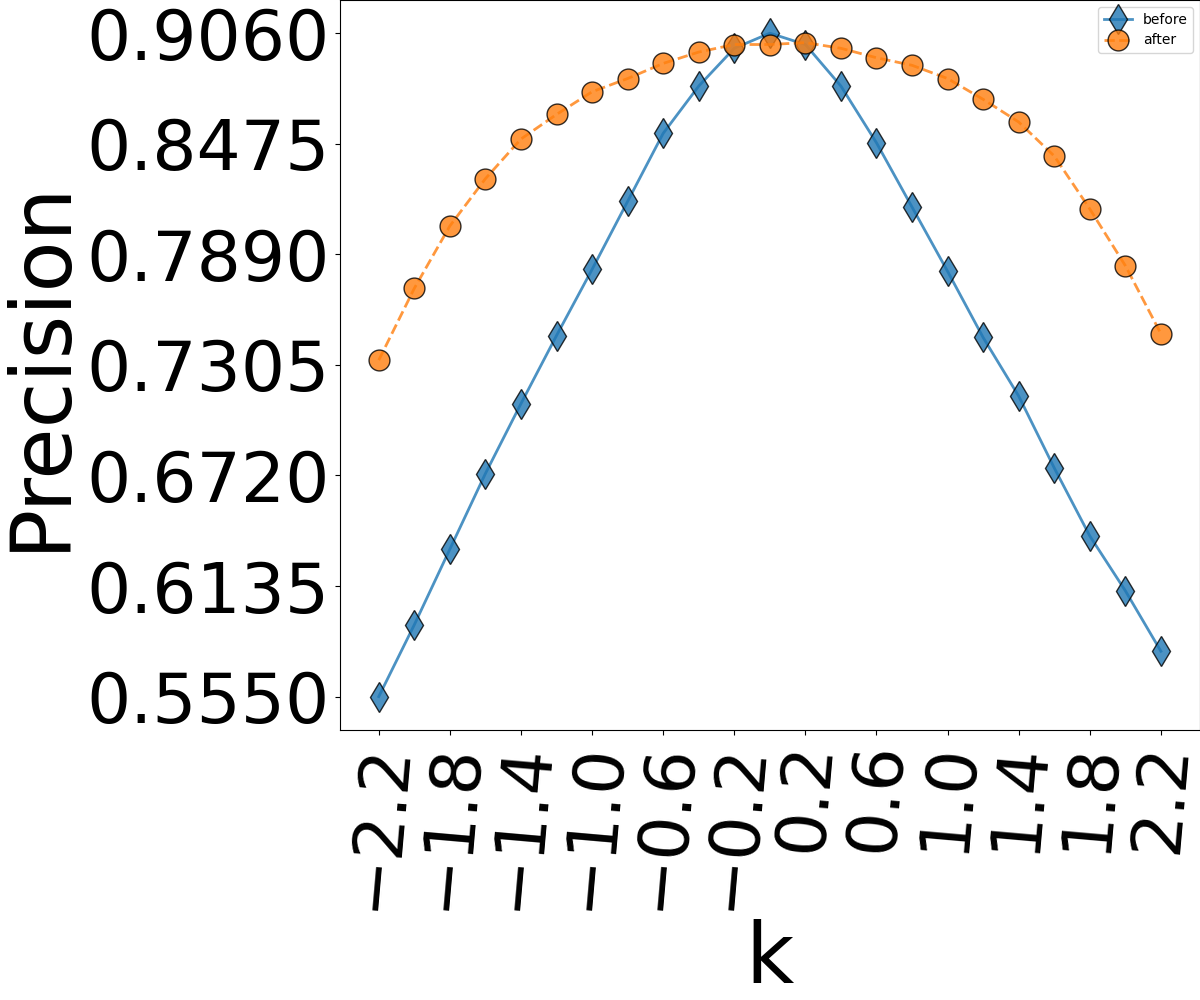}
    \\
    \includegraphics[width=\width]{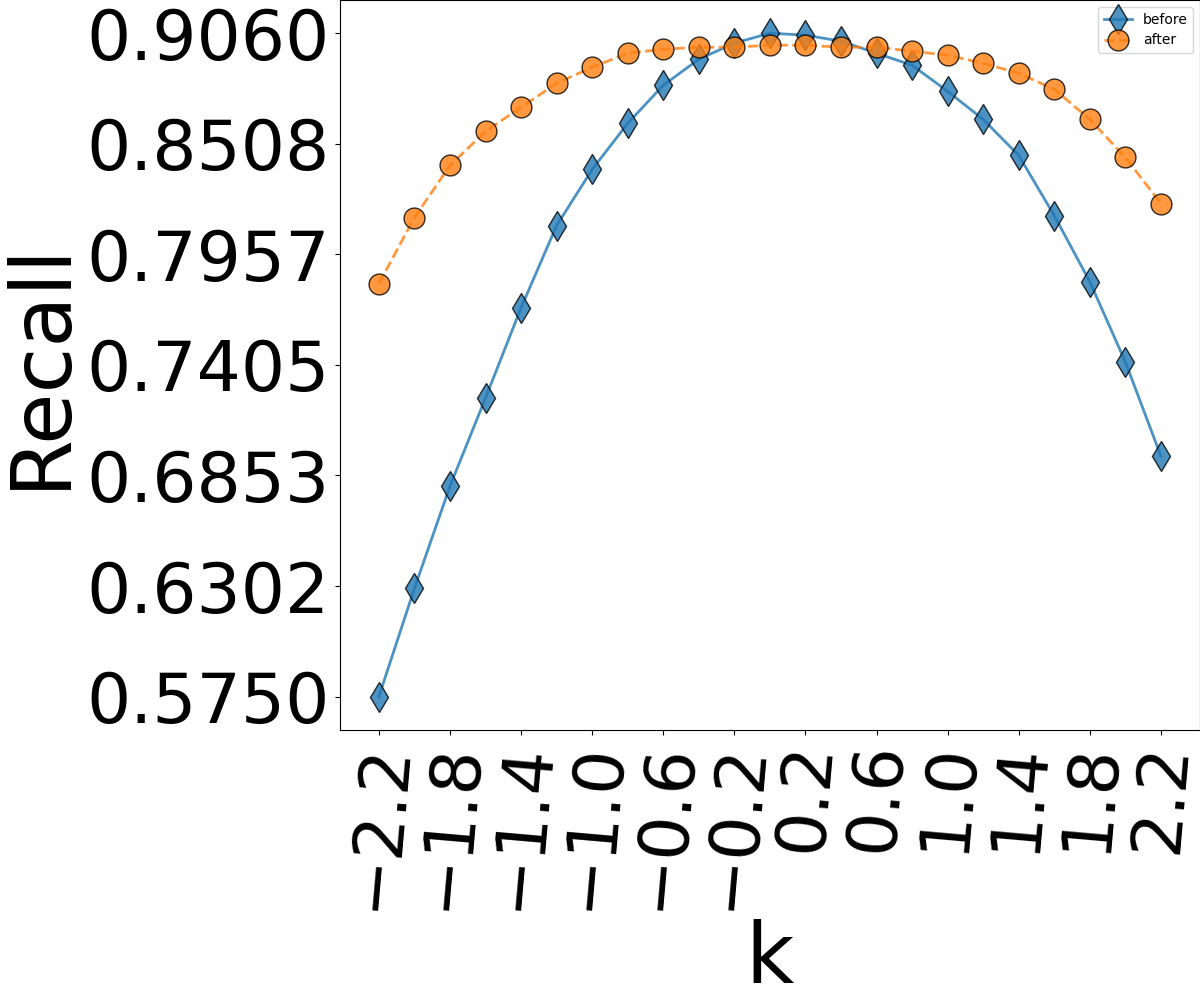}
    \includegraphics[width=\width]{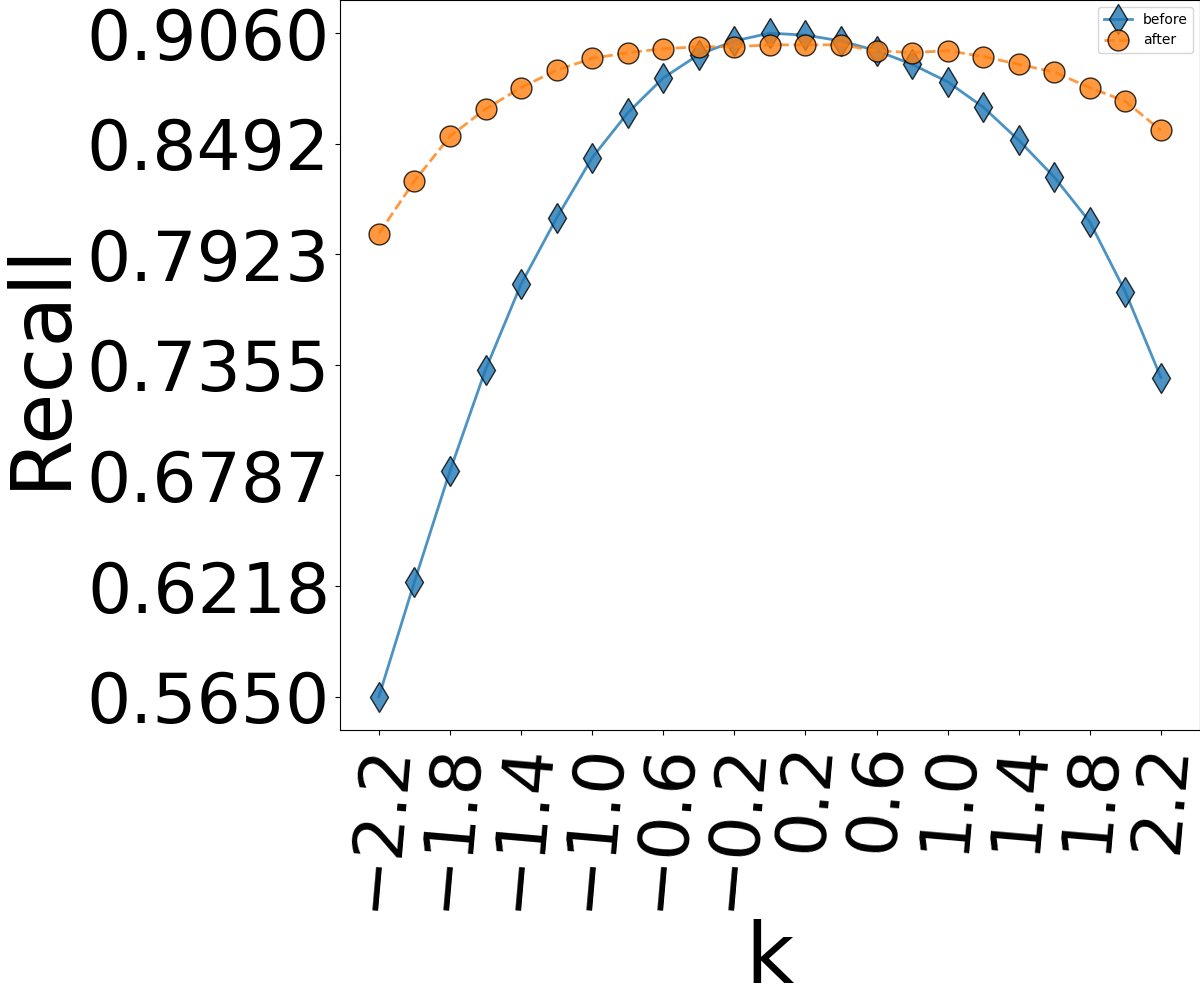}
    \includegraphics[width=\width]{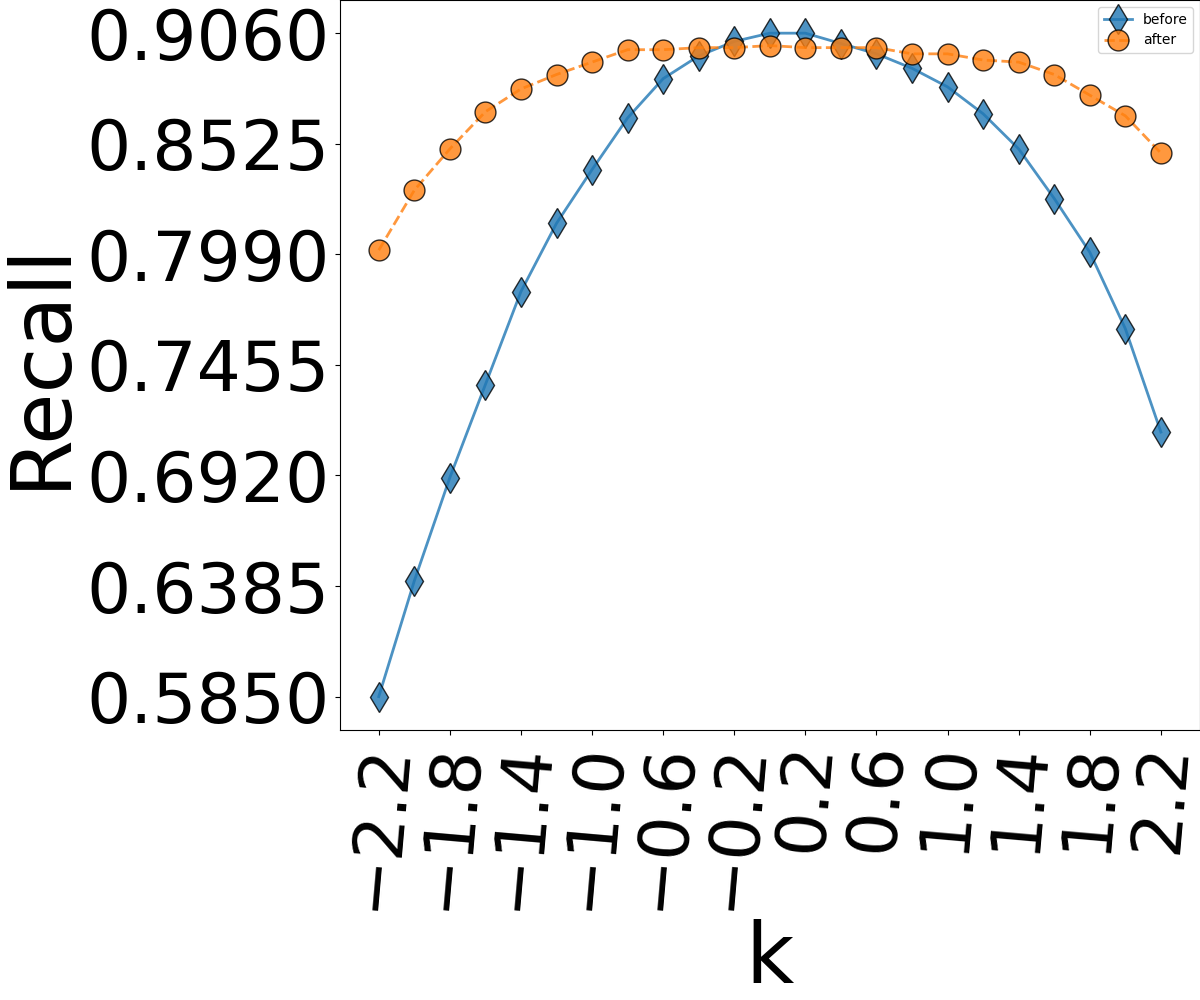}
    \includegraphics[width=\width]{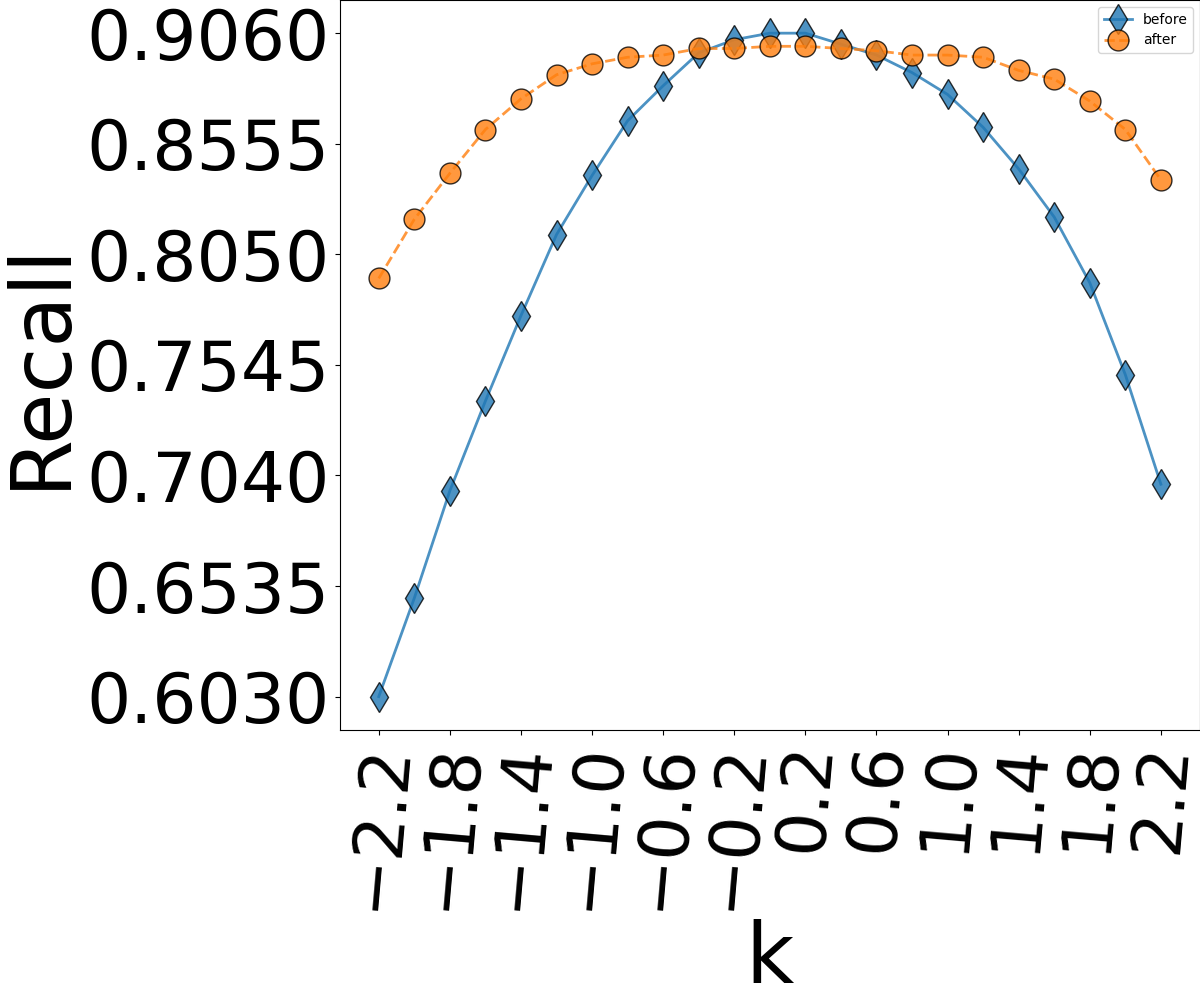}
    \includegraphics[width=\width]{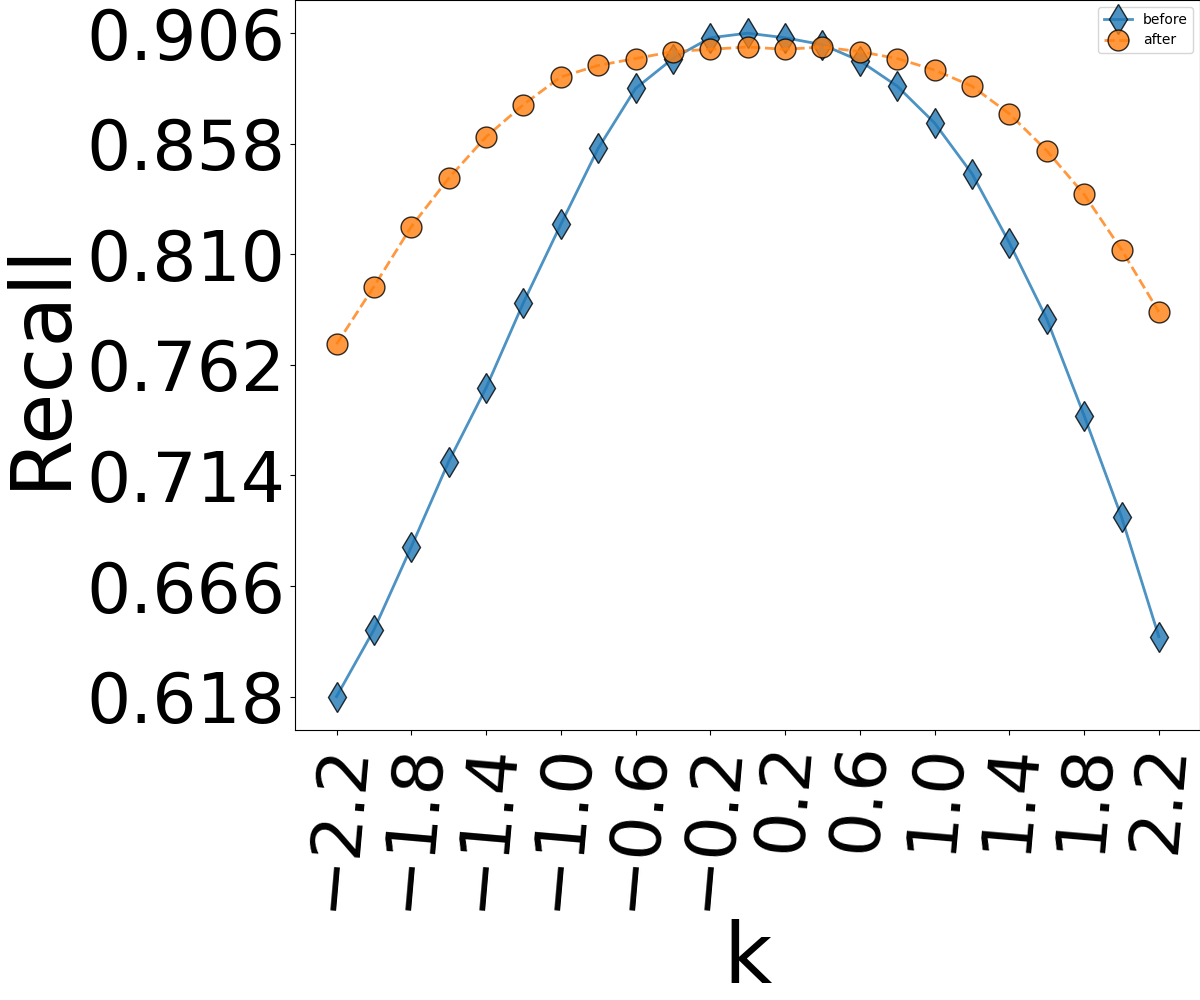}
    \includegraphics[width=\width]{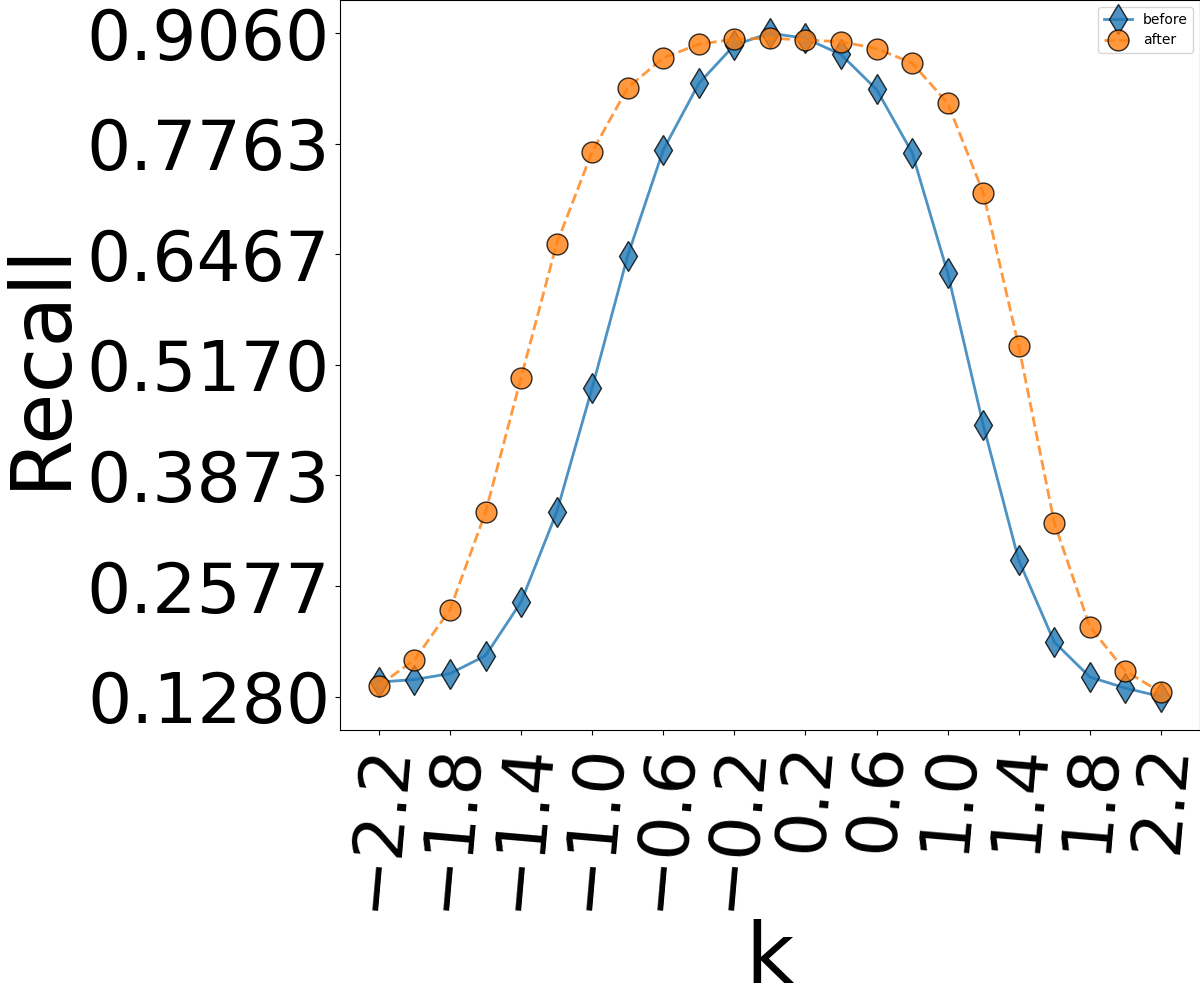}
    \includegraphics[width=\width]{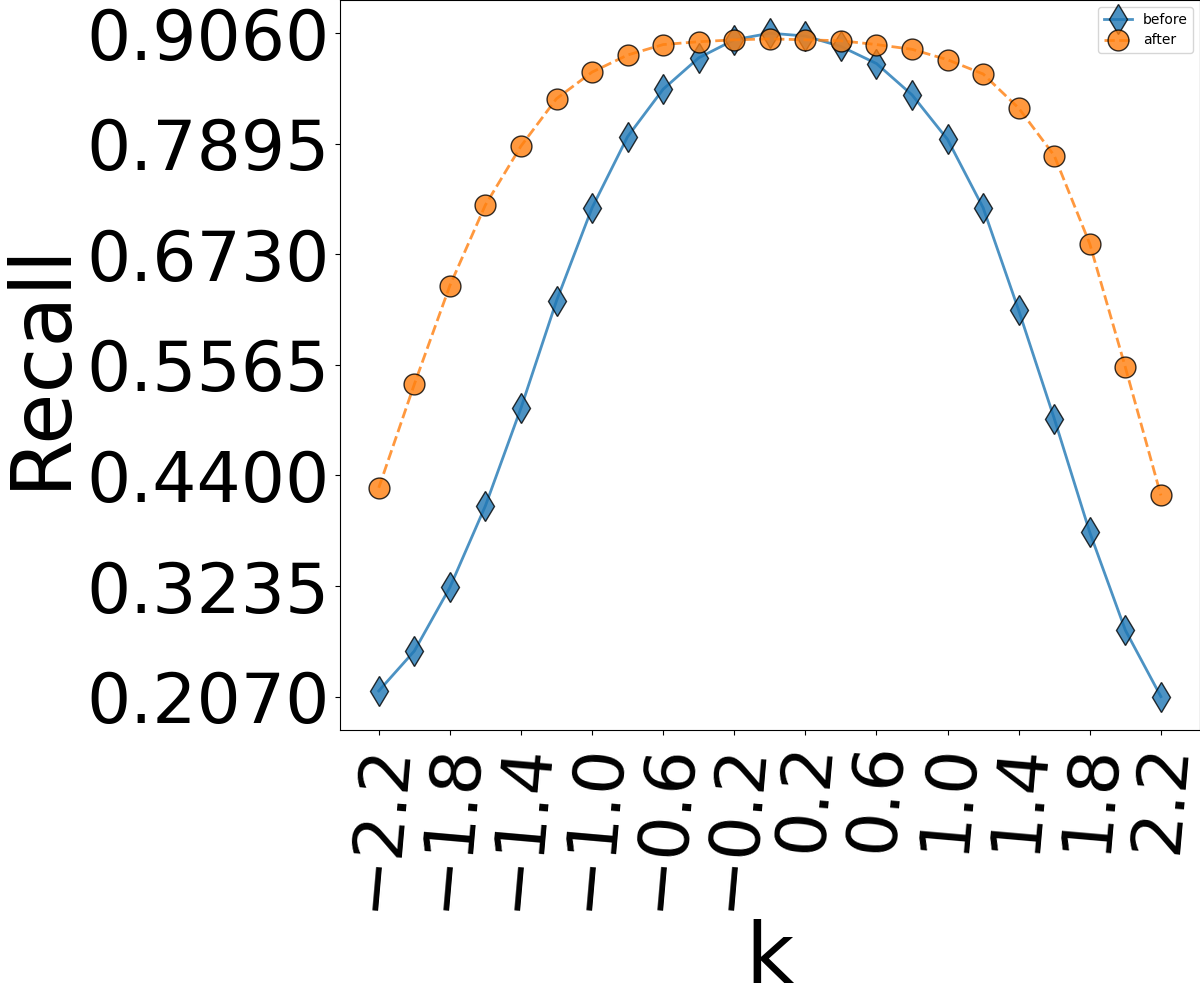}
    \includegraphics[width=\width]{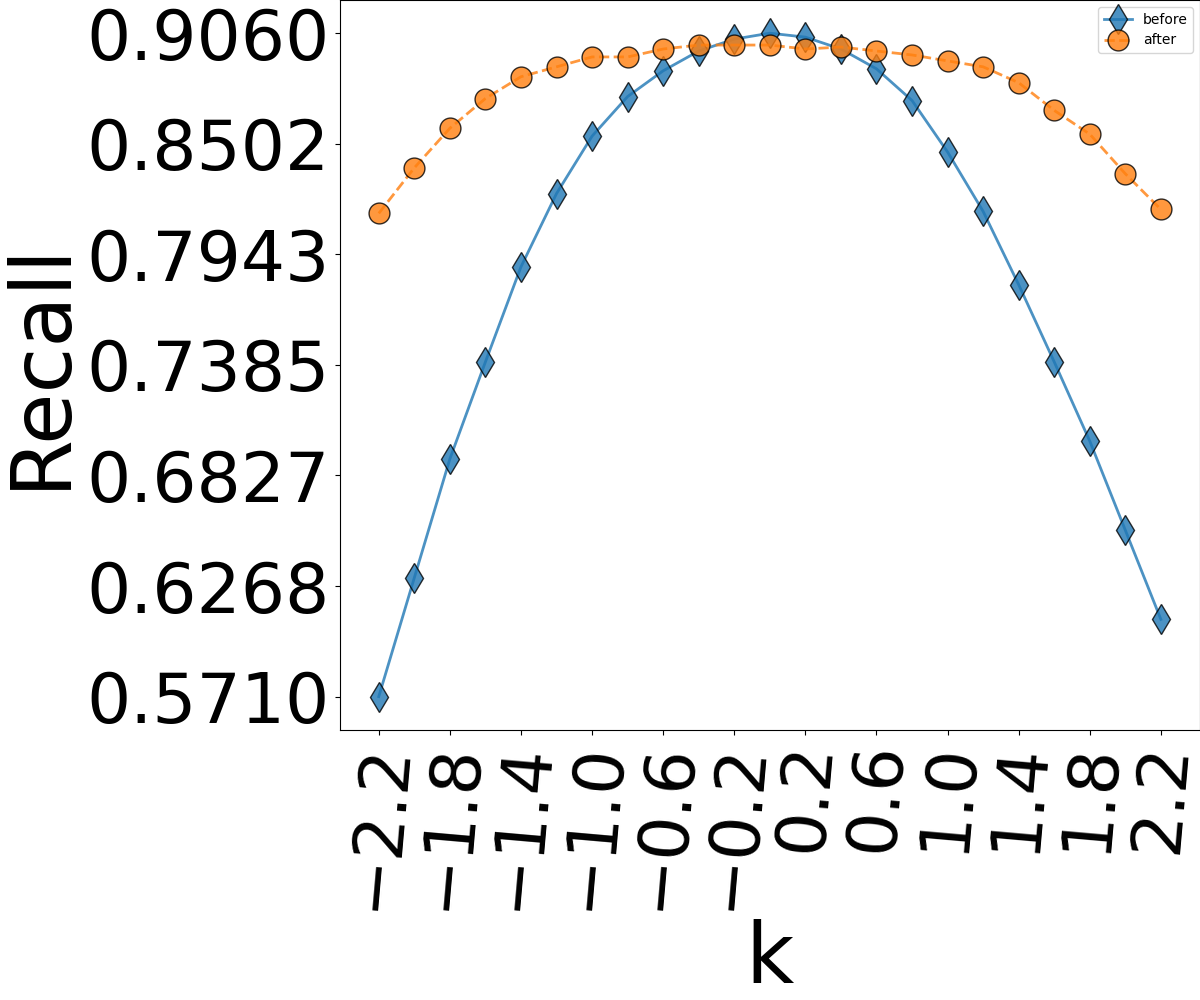}
    \includegraphics[width=\width]{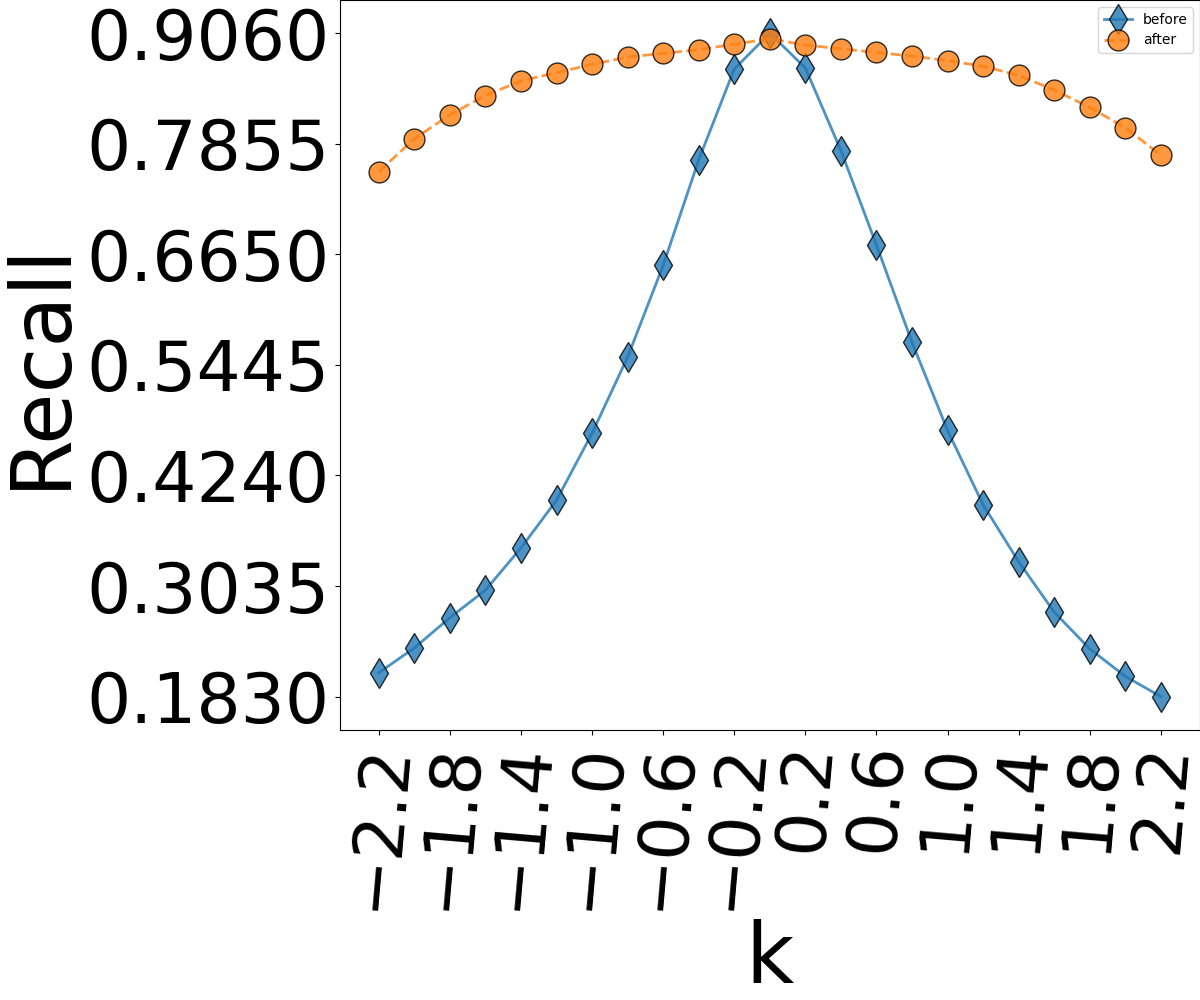}
    \includegraphics[width=\width]{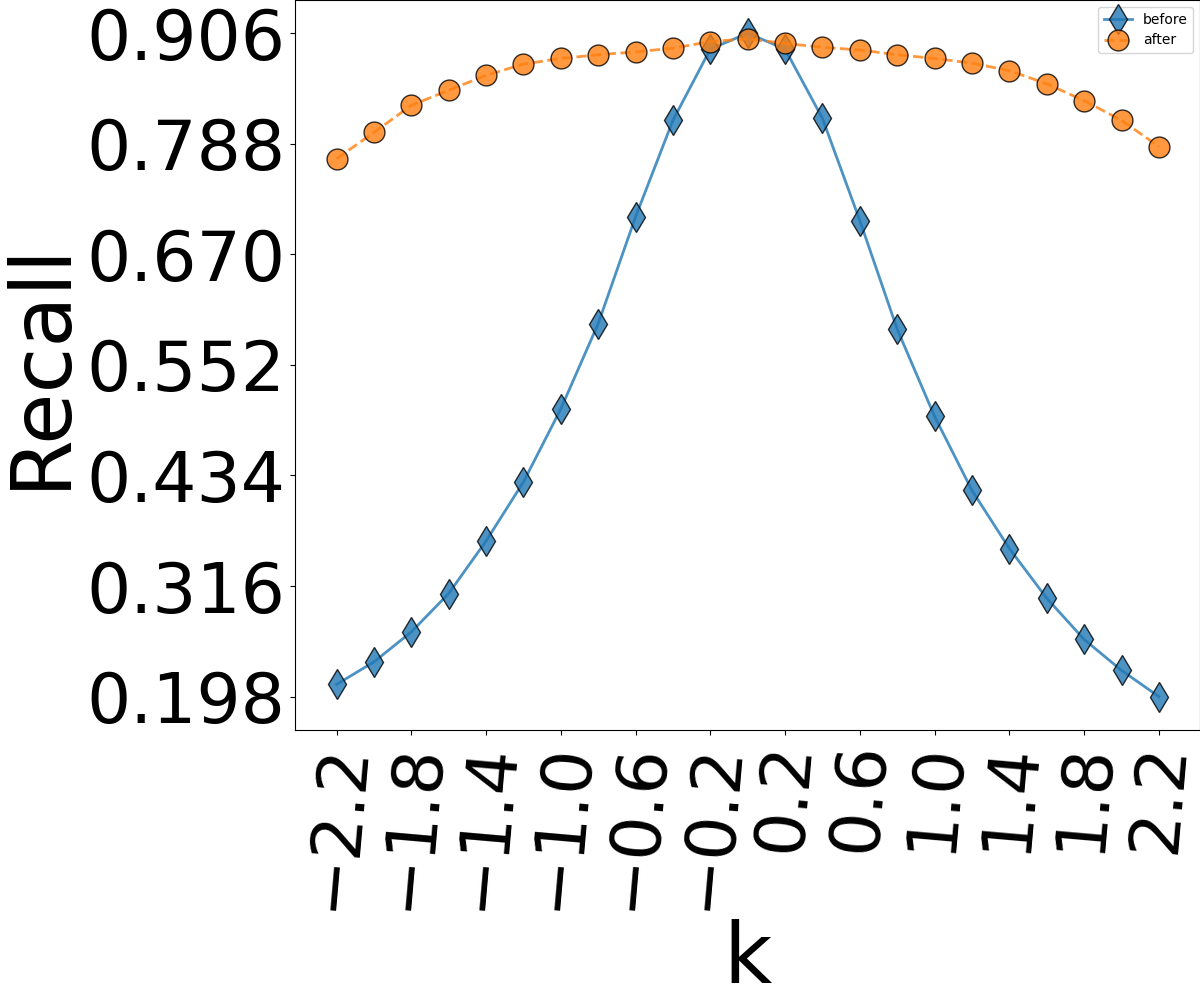}
    \includegraphics[width=\width]{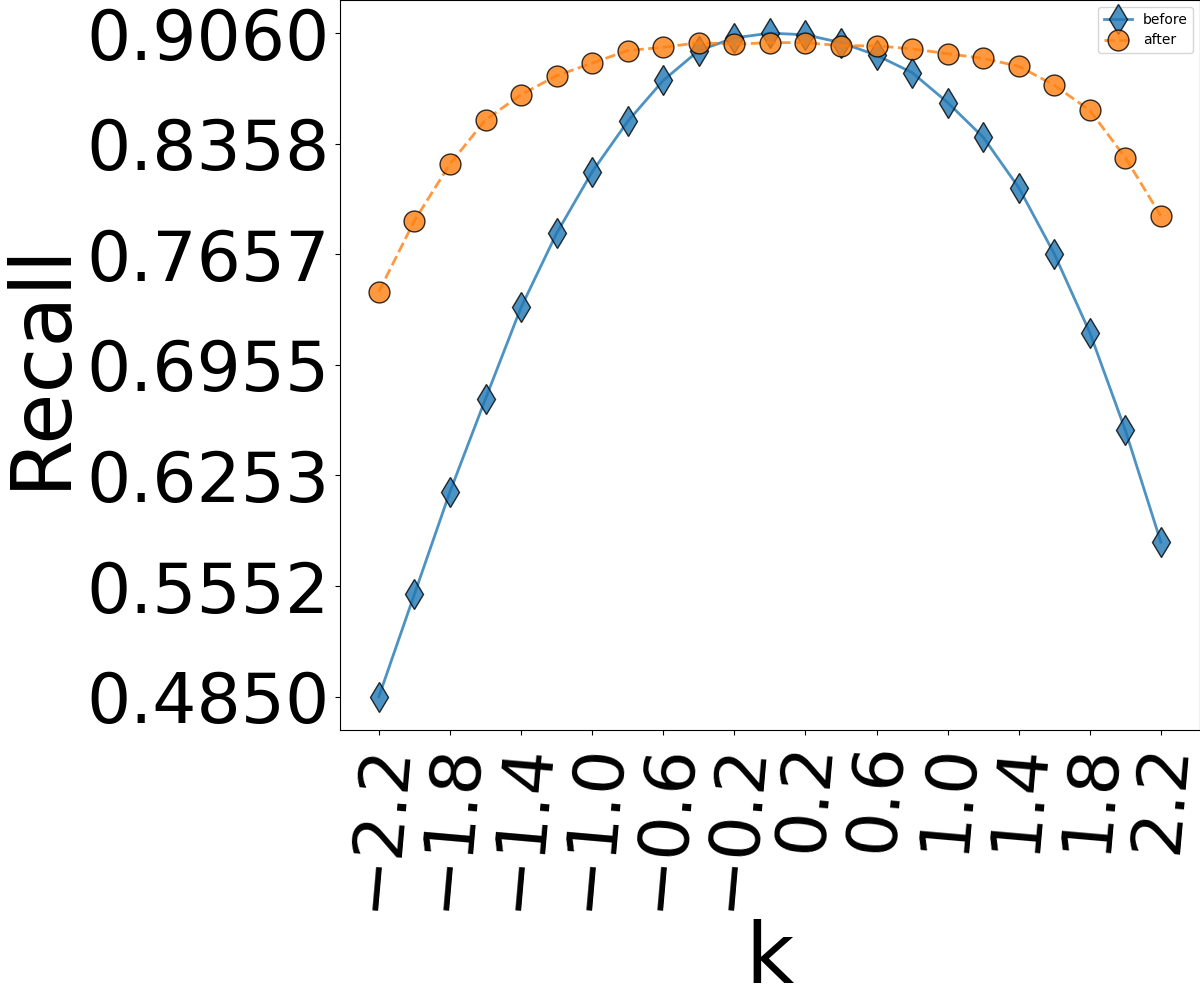}
    \includegraphics[width=\width]{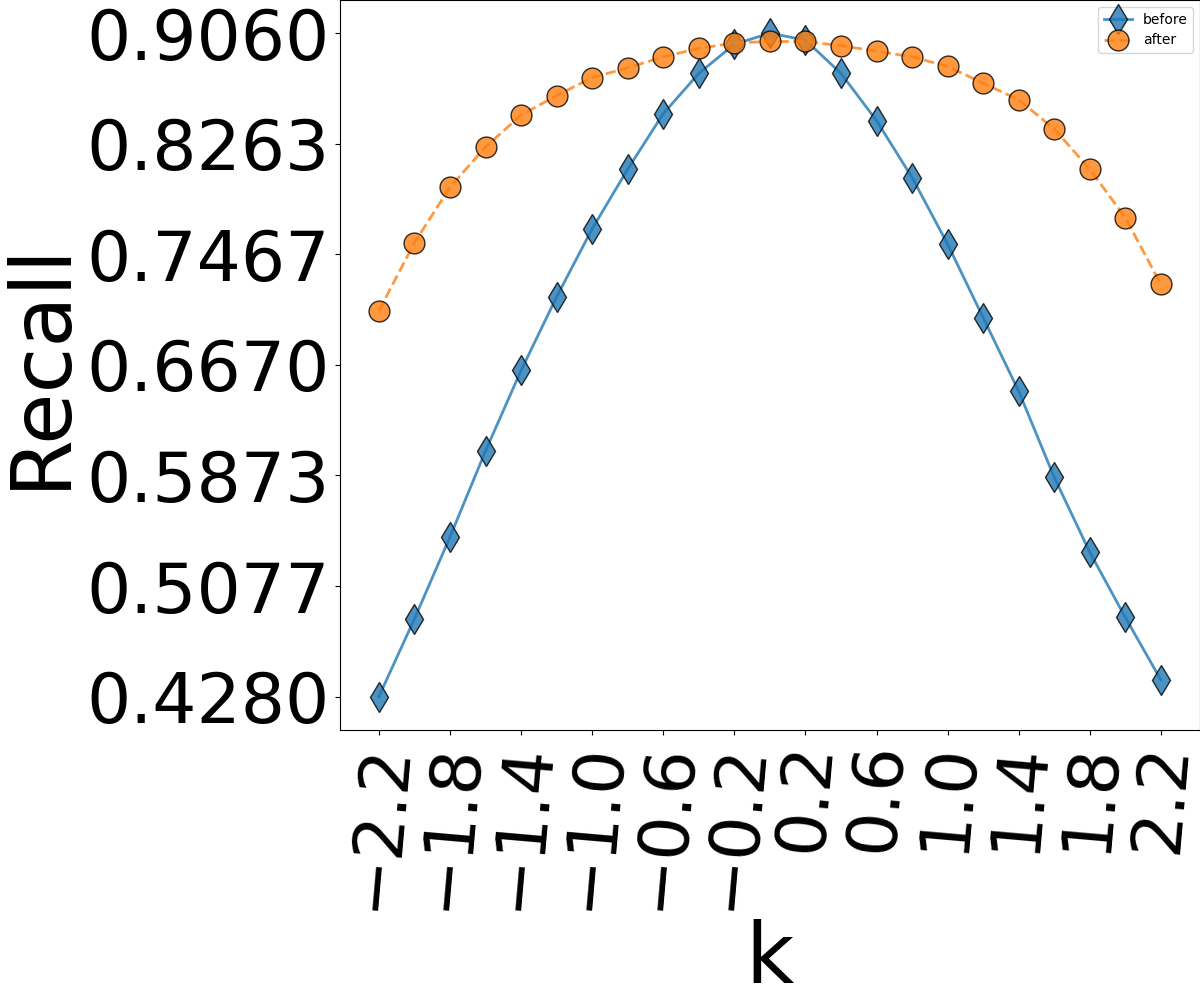}
    \\
    \includegraphics[width=\width]{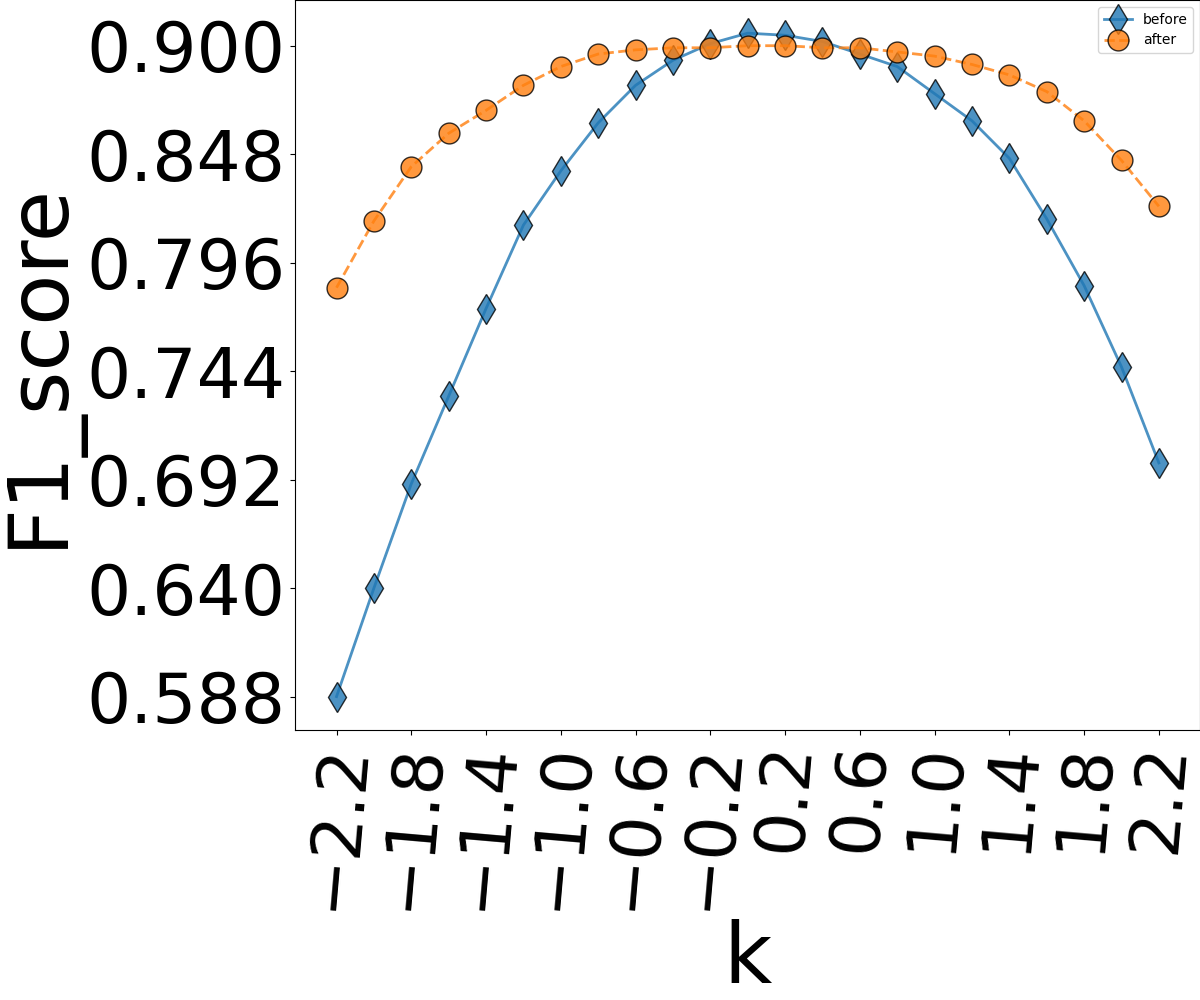}
    \includegraphics[width=\width]{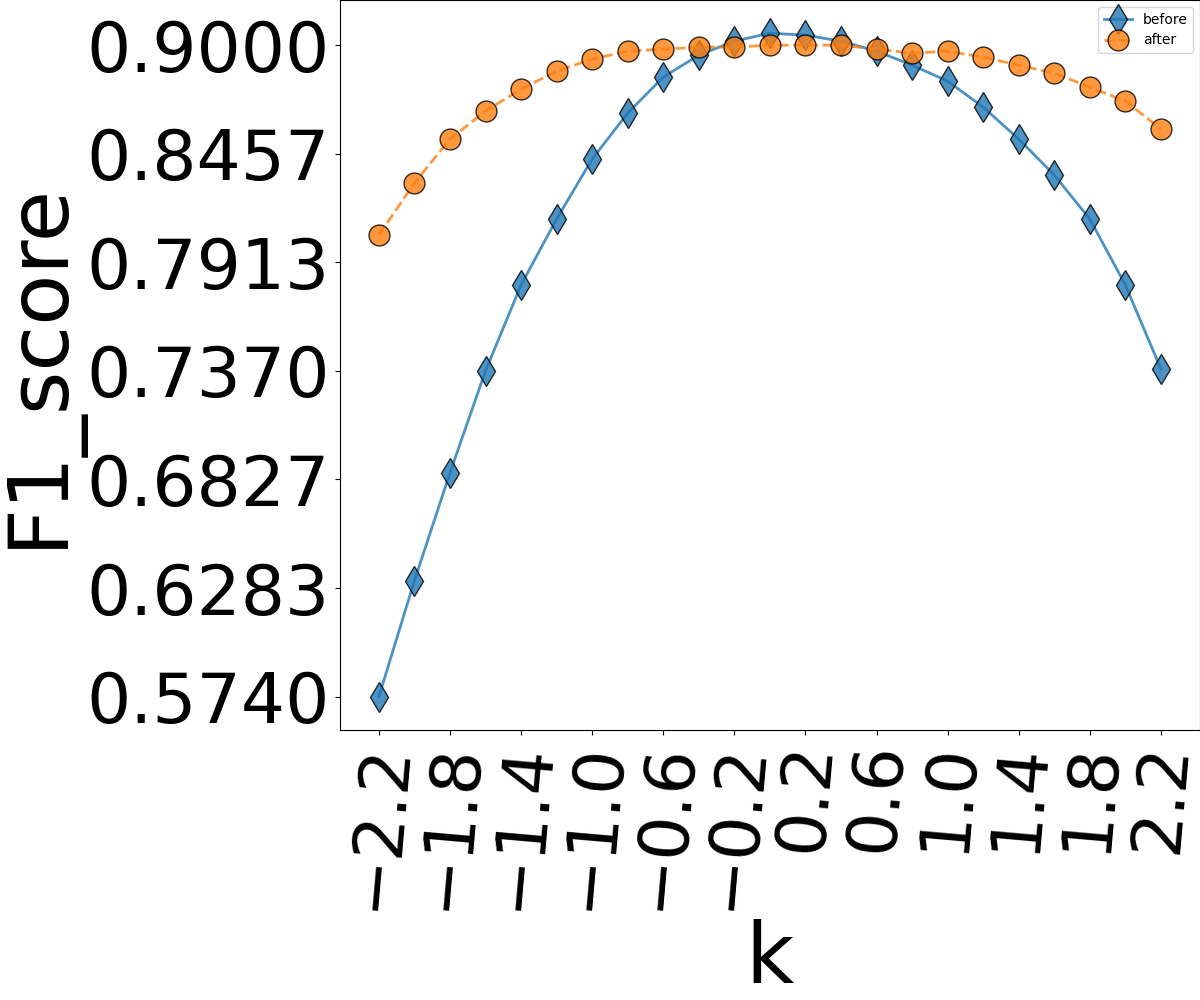}
    \includegraphics[width=\width]{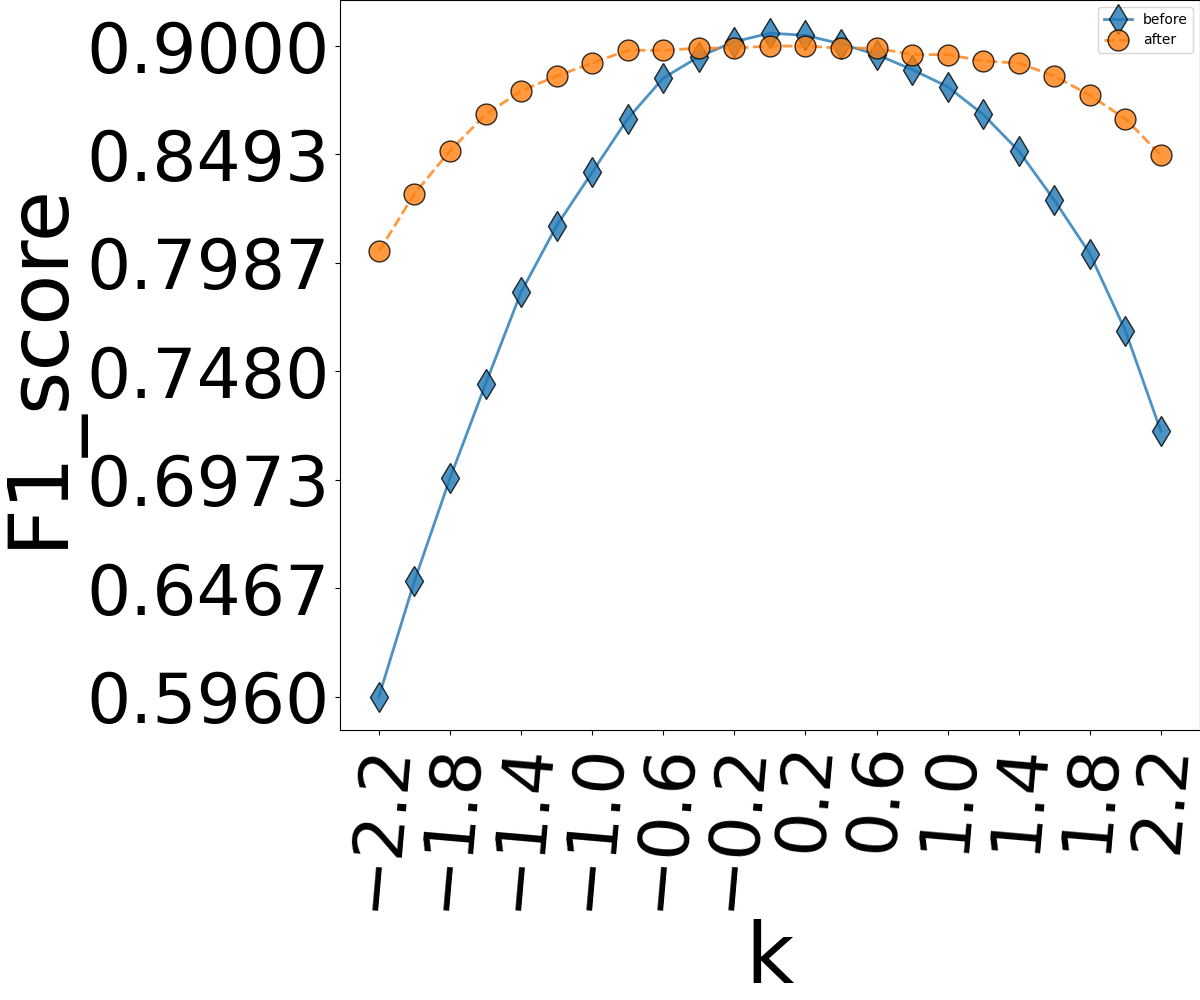}
    \includegraphics[width=\width]{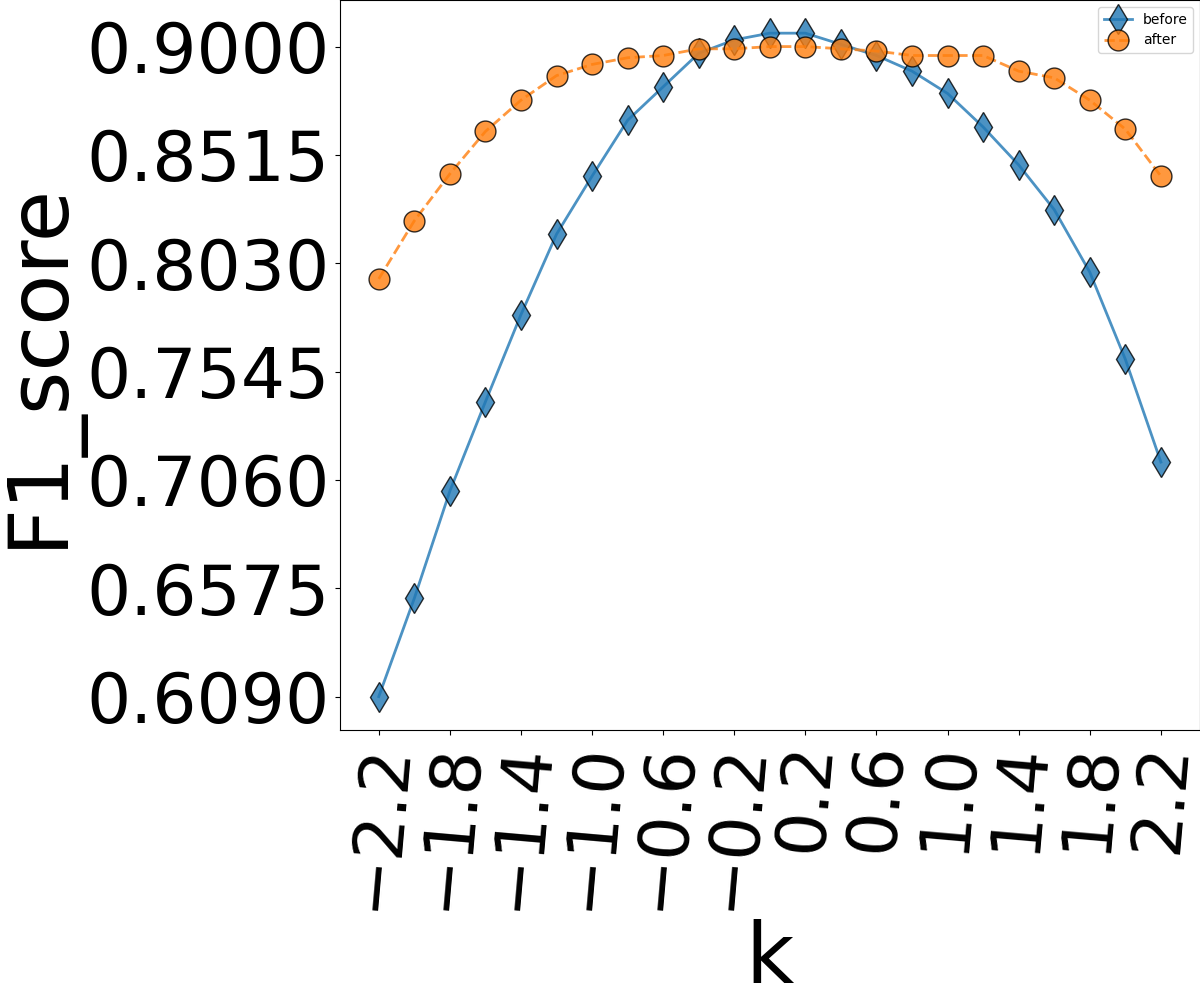}
    \includegraphics[width=\width]{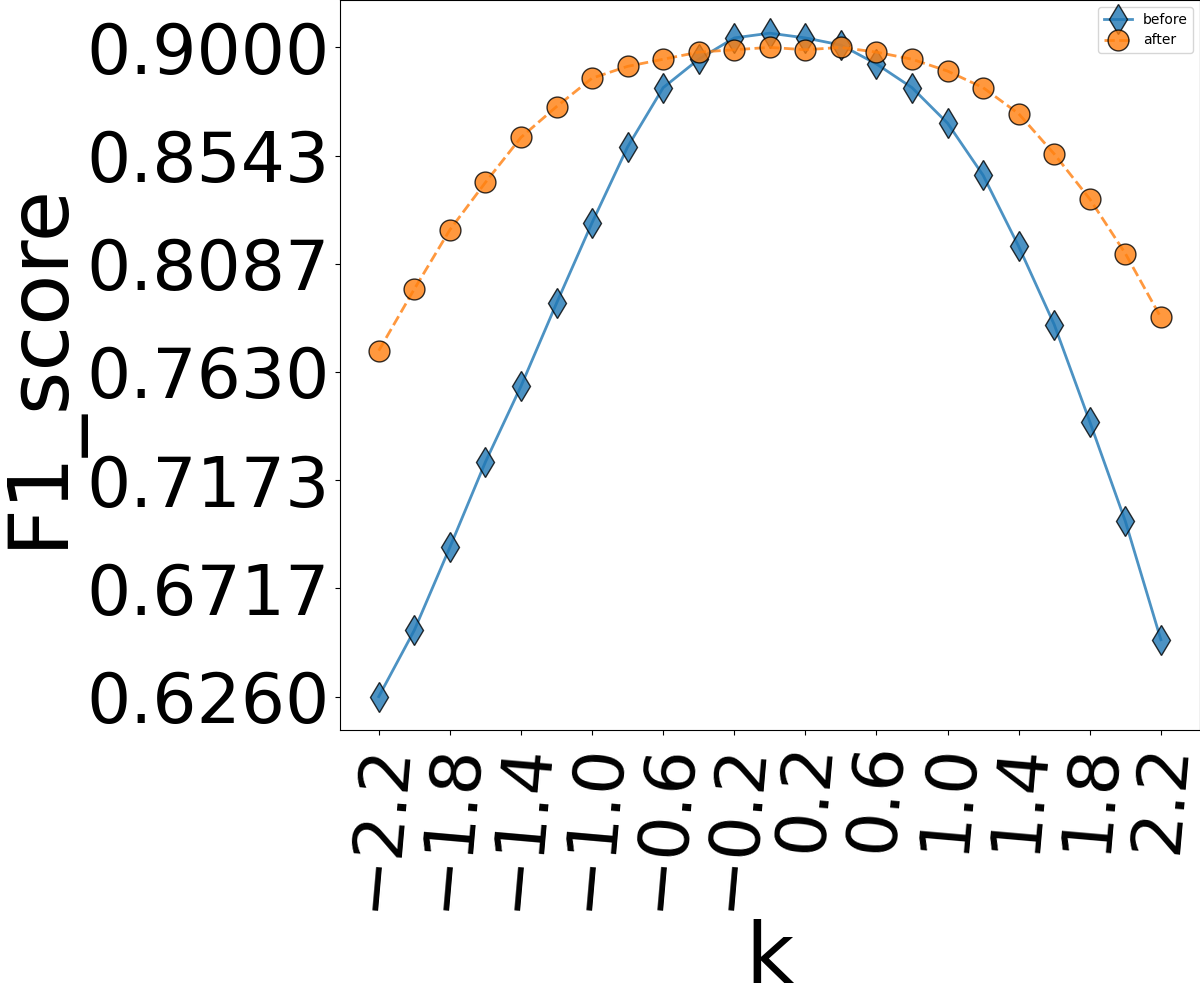}
    \includegraphics[width=\width]{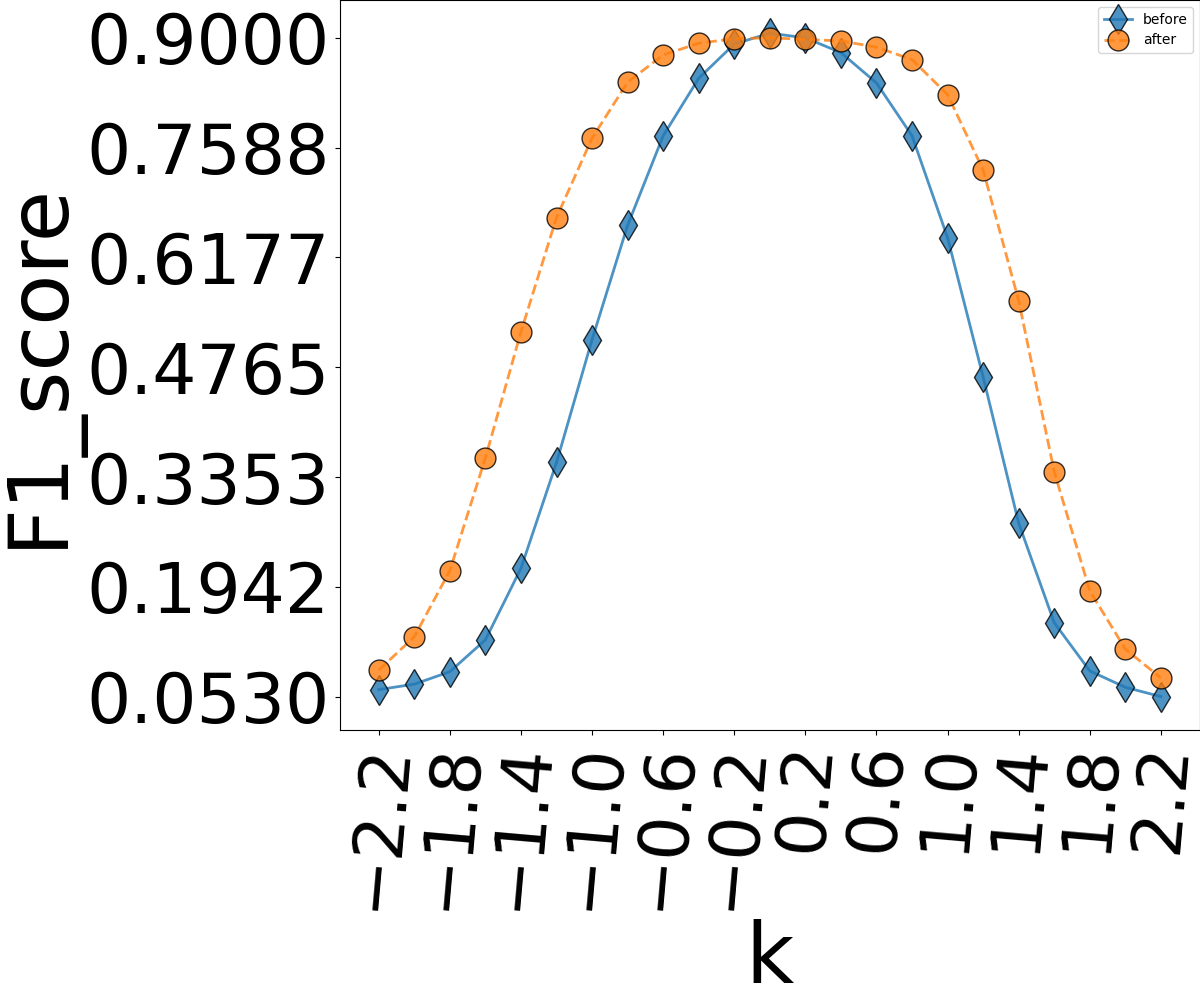}
    \includegraphics[width=\width]{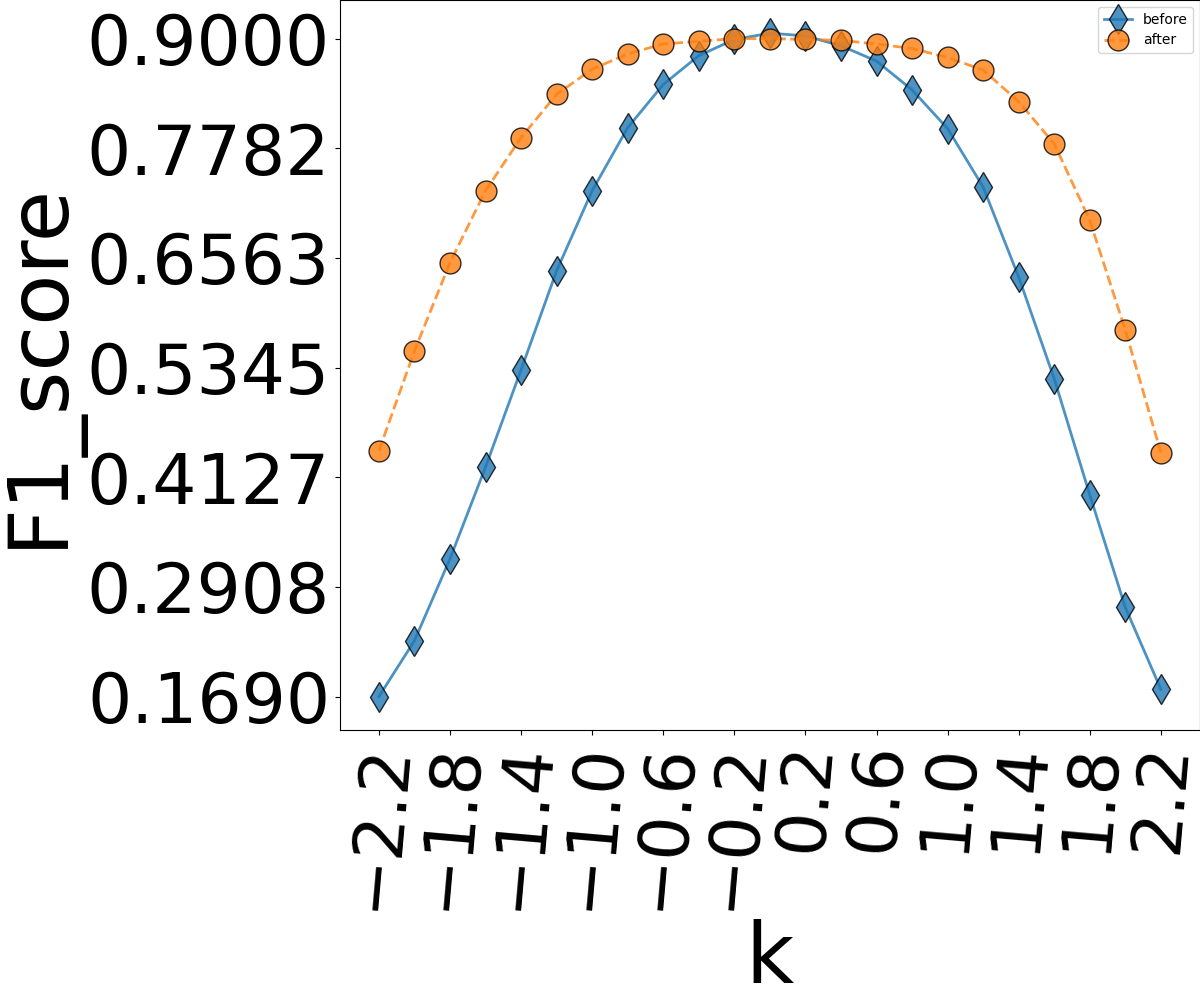}
    \includegraphics[width=\width]{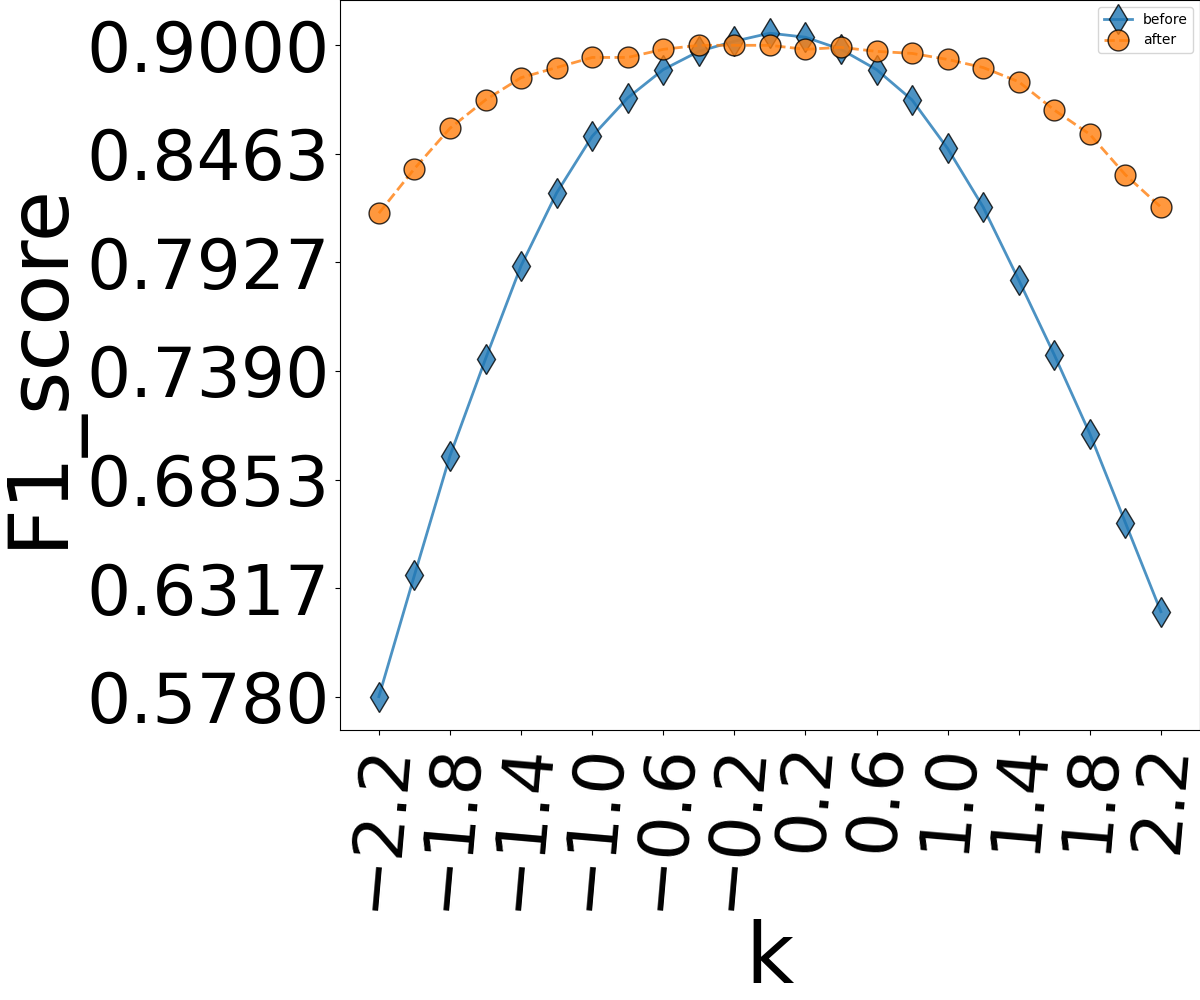}
    \includegraphics[width=\width]{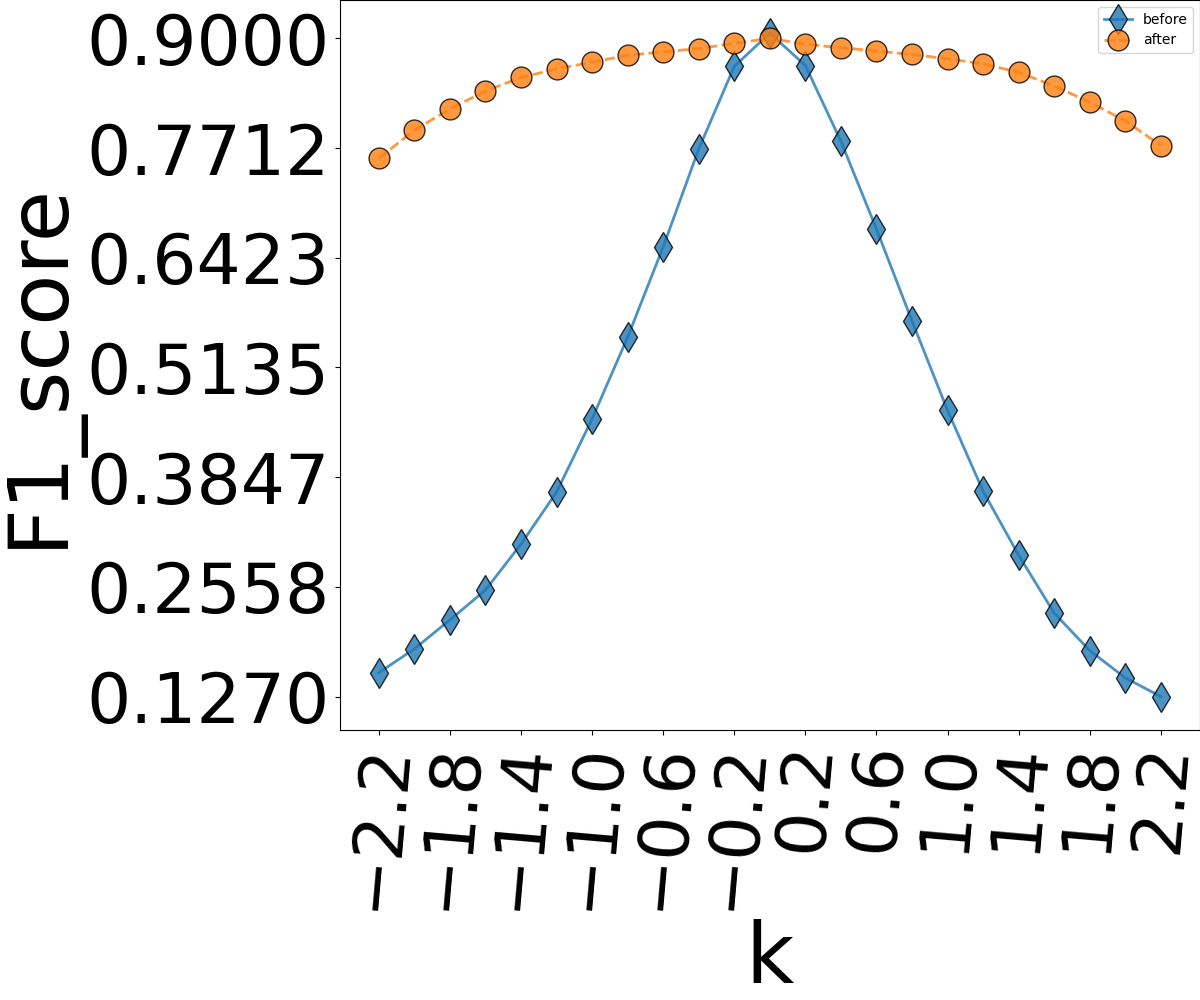}
    \includegraphics[width=\width]{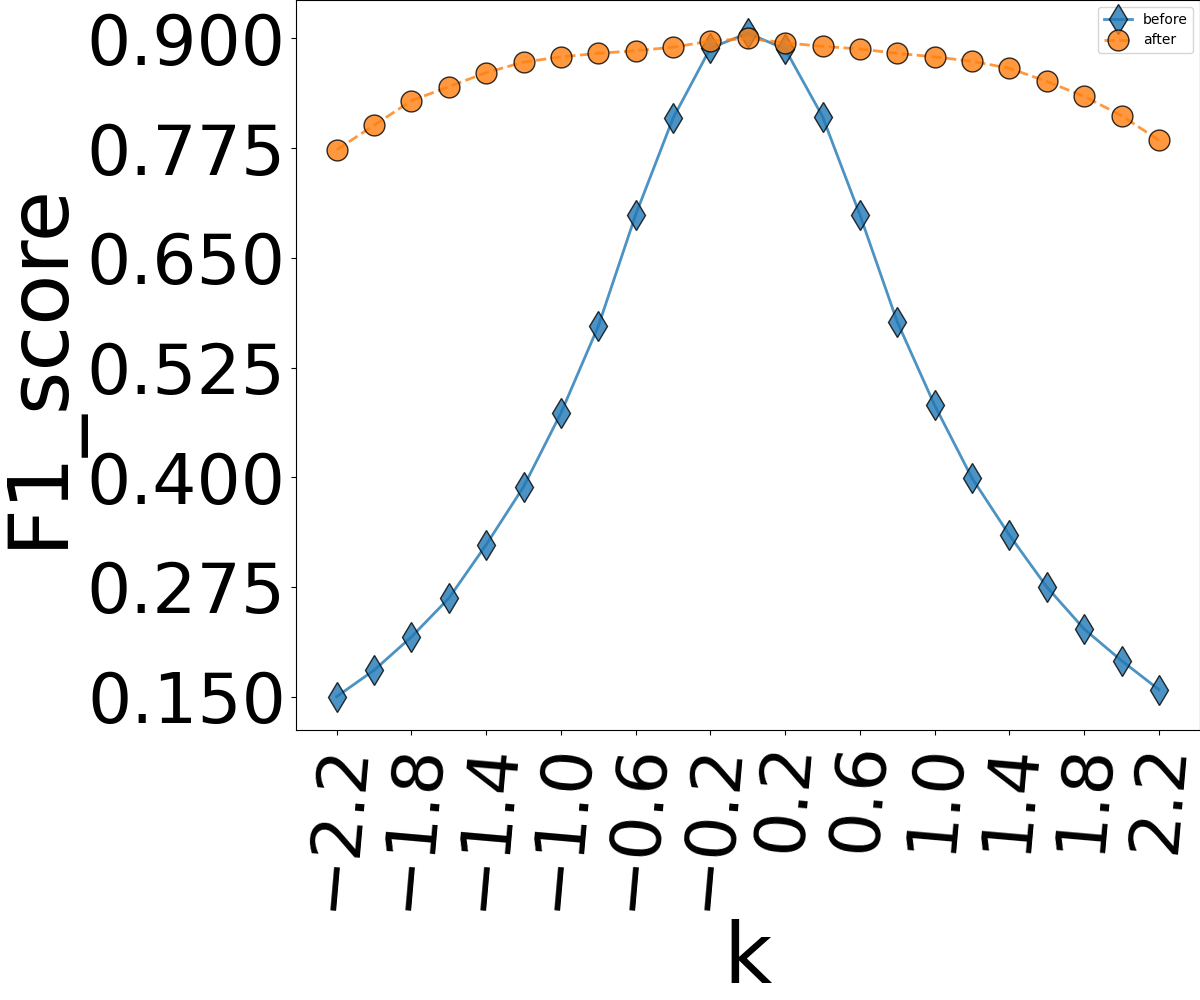}
    \includegraphics[width=\width]{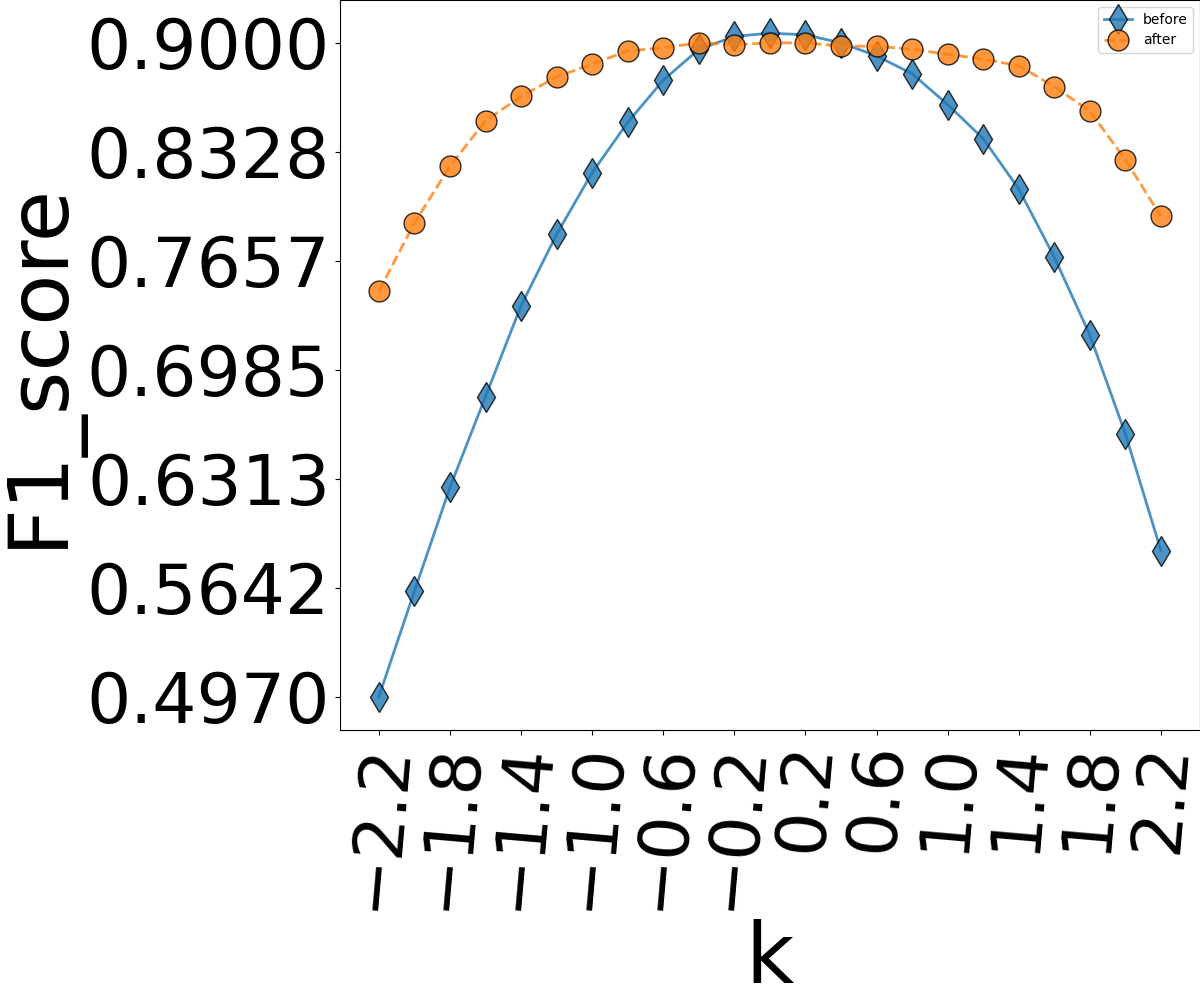}
    \includegraphics[width=\width]{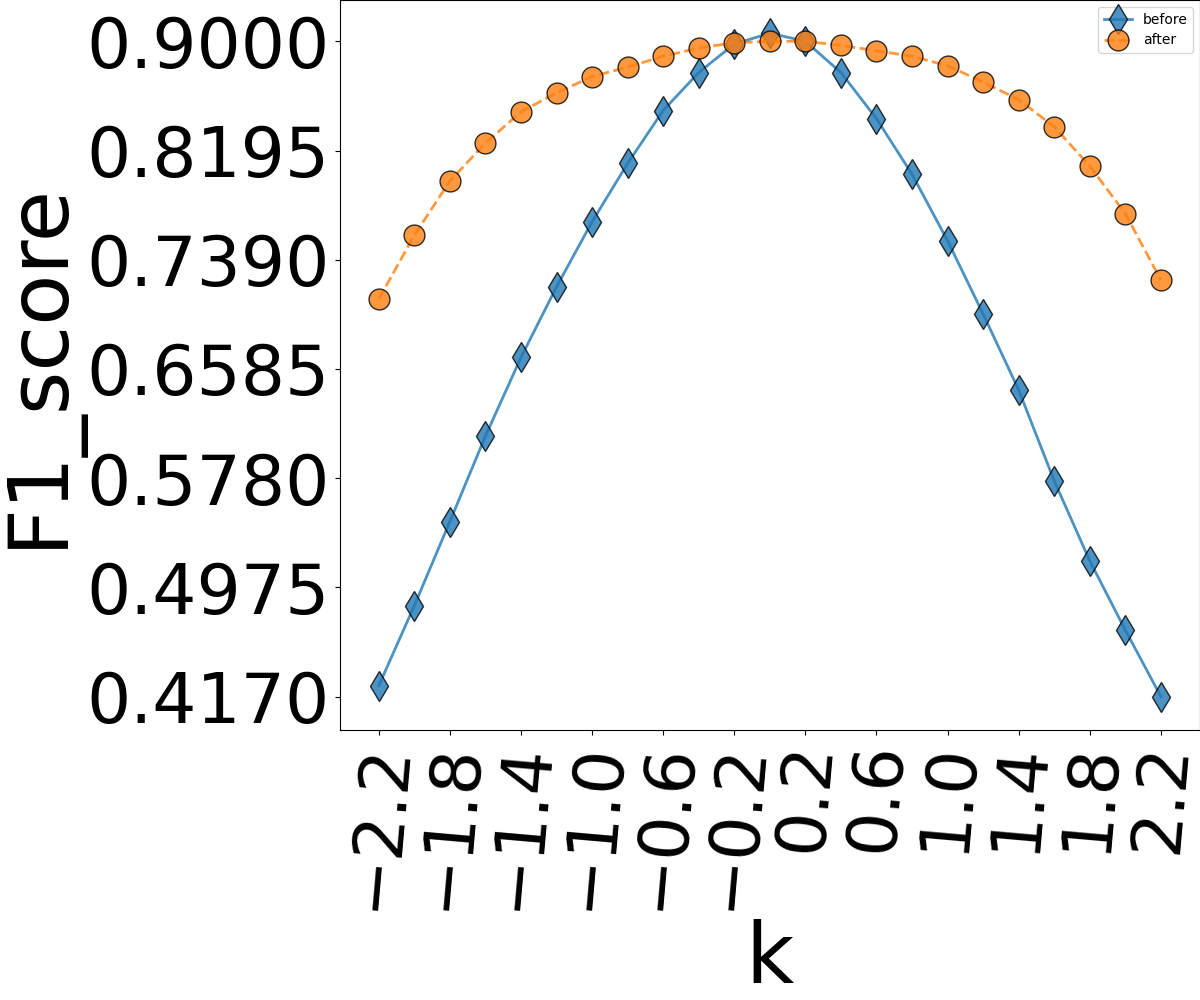}
    \\
 
    \caption{Figure shows the precision, recall and F1-score using MobilenetV3-small model on CIFAR10 dataset. The $1^{st}$ two rows contain Precision, the middle two rows contain recall and the last two rows contain F1-score, for $1^{st}$ to $12^{th}$ masks, respectively.}
    \label{fig:mob-prf}
\end{figure*}

\begin{figure*}[!b]
    \centering
    \newcommand\wide{2.93cm}
    \includegraphics[width=\wide]{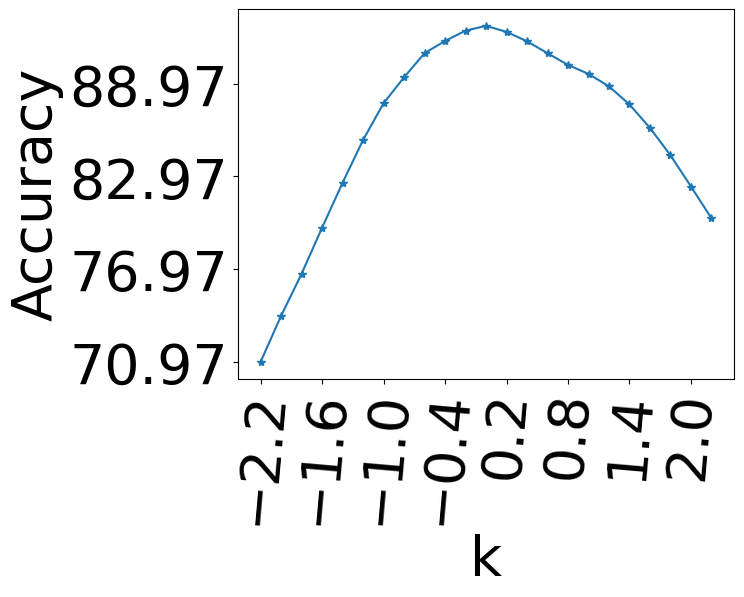}
    \includegraphics[width=\wide]{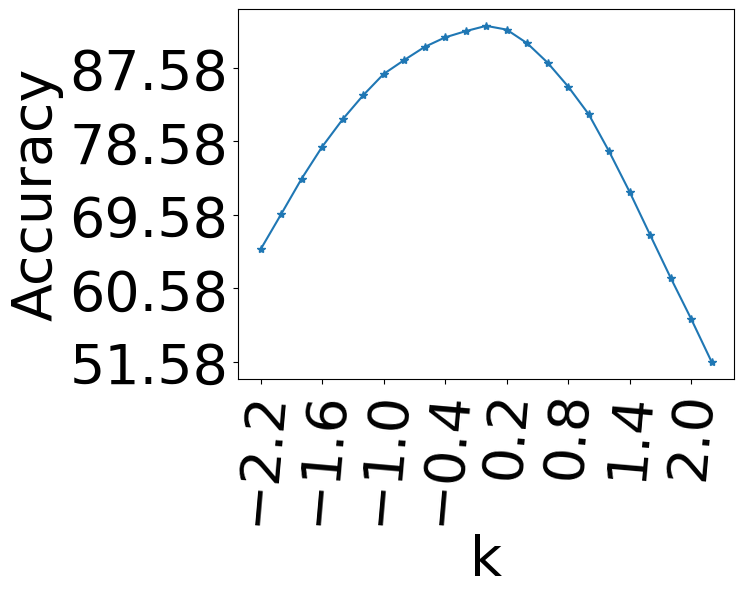}
    \includegraphics[width=\wide]{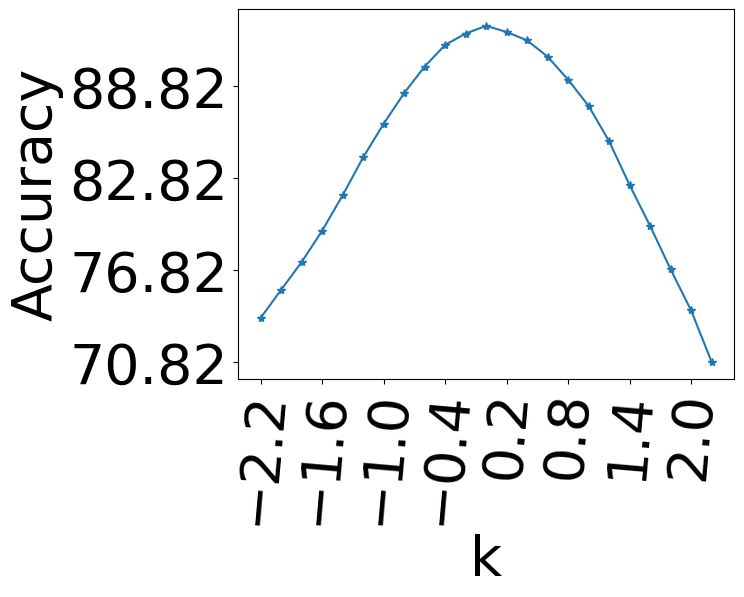}
    \includegraphics[width=\wide]{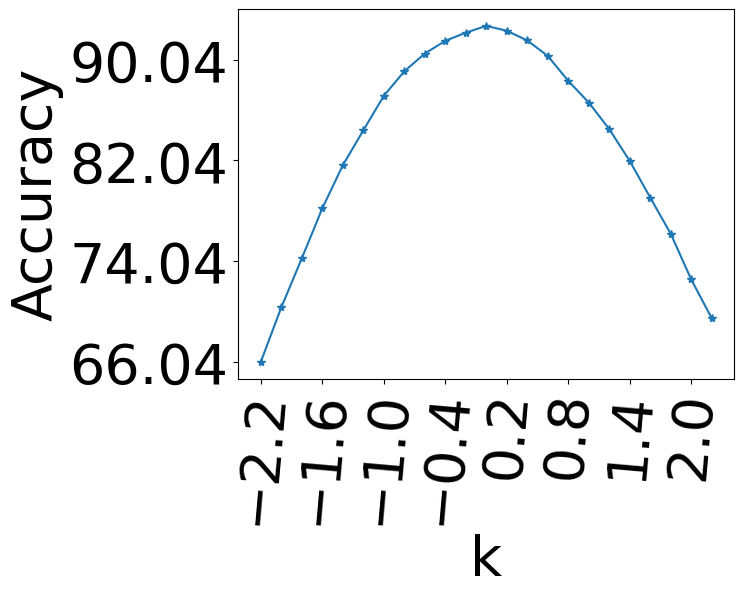}
    \includegraphics[width=\wide]{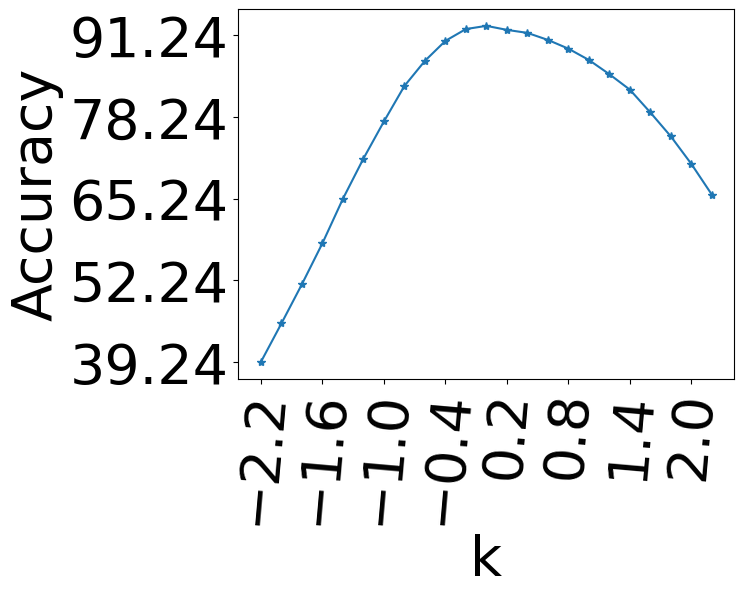}
    \includegraphics[width=\wide]{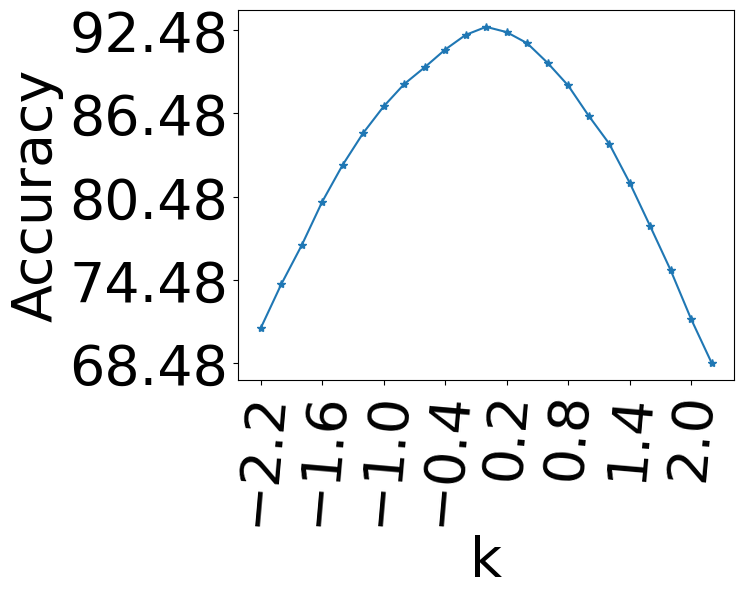}
    
    \caption{The effect of the NUI attack on the different colour channels. Left to right: The curve represents the accuracy when only $R$ channel, only $G$ channel, only $B$ channel, $RG$ channels, $RB$ channels, and $GB$ channels are perturbed, respectively.} 
    \label{fig:percha1}
\end{figure*}

\begin{figure*}[!b]
    \centering
    \newcommand\wide{2.48cm}
    \includegraphics[width=\wide]{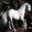}
    \includegraphics[width=\wide]{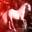}
    \includegraphics[width=\wide]{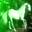}  
    \includegraphics[width=\wide]
    {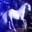}
    \includegraphics[width=\wide]{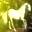}
    \includegraphics[width=\wide]{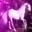}
    \includegraphics[width=\wide]{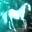}
    \caption{The effect of NUI attack on the different colour channels. The $1^{st}$ image is the original image, $2^{nd}$ to $4^{th}$ images are the result of perturbing only the red, green and blue channels, respectively, $5^{th}$ to $7^{th}$ images are the result of perturbing red and green channels, red and blue channels, and green and blue channels, respectively.}
    \label{fig:perchal2}
\end{figure*}

\subsection{Quantitative Analysis:}
Fig. \ref{fig:inc-prf} and Fig. \ref{fig:mob-prf} show the comparison of precision, recall and f1-score before and after the model trained on perturbed data. These results also support similar trend as observed using Accuracy reported in the main paper.

\subsection{Extension -- Effect of NUI Attack on Color Channels}
\label{sec:Color_channel}
We also test the effect of the proposed NUI attack on specific channels of $RGB$ images. For this experiment, the VGG16 model is used on the CIFAR10 dataset with a NUI attack using Mask $1$ on the test set. Six $RGB$ experimental settings are tested for different values of $k$, including perturbations applied to ${R, G, B, RG, RB, GB}$, where $R$, $G$ and $B$ represent the Red, Green and Blue channels, respectively. The results are illustrated in \autoref{fig:percha1}. The NUI attack shows a high impact on the combination of the Red and Blue channels as depicted in the $5^{th}$ plot. The effect on a sample image is shown after the NUI attack using Mask $1$ with $k=1.8$ in \autoref{fig:perchal2}. All the images are perceptible and preserve the semantic meaning.

\end{document}